%% file: main.tex
\documentclass[11pt]{guidelabs}
\usepackage{multirow}
\usepackage{float}
\usepackage{caption}
\usepackage{wrapfig}
\newcommand{\titlename}{Scaling Inherently Interpretable \\ Language Models }

\title{\titlename}

\usepackage{fontawesome5}
\newcommand{\heartmark}{\textcolor{GuideLabsTocPurple}{\faHeart}}

\newcommand{\heartfootnote}[1]{%
  \begingroup
  \renewcommand\thefootnote{\heartmark}%
  \footnote{#1}%
  \endgroup
}

\author{%
  {\LARGE Guide Labs Team}\\[6pt]
  Andreas Madsen \quad Aya Abdelsalam Ismail \quad Giang Nguyen \quad Isaac Plant \\
  Muawiz Chaudhary \quad Nathaniel Monson \quad Saqib Azim \quad Zhichen Guo \\[6pt]
  Julius Adebayo\heartfootnote{Steerling was a team effort; authors other than the last are sorted alphabetically. See author contributions \hyperref[sec:contributions]{here}.}
}
\date{} % keep blank

\GuideLabsSetTocEntryColor{GuideLabsTocPurple!50!black} % slightly bluer
\GuideLabsSetTocSecSkip{10pt}
\usepackage{svg}

\usepackage{placeins}

\usepackage[normalem]{ulem}
\PassOptionsToPackage{normalem}{ulem}

\usepackage{xspace}
\newcommand{\steerling}{Steerling\xspace}
\newcommand{\steerlingB}{Steerling-8B\xspace}

\newcommand{\causaldiff}{Causal Diffusion\xspace}
\newcommand{\atlas}{Atlas\xspace}
\newcommand{\blockdiff}{Block Diffusion\xspace}

\newcommand{\nconcepts}{$33{,}732$\xspace}            % exact -- use 1-2 times only
\newcommand{\nconceptsapprox}{over $33{,}000$\xspace} % rounded prose -- use everywhere

\newcommand{\citenote}[1]{{\color{orange}[cite]}}
\newcommand{\TBD}[1]{\textbf{??}}

\input{math_commands}

\GuideLabsAbstractLogoOff
\GuideLabsAbstractBrandTextOff

\begin{document}

\maketitle
\begin{abstract}

Interpretability is often treated as a tax on capability: language models are trained as opaque systems, then explained  after the fact, with methods whose reliability is difficult to establish. 
In this work, we challenge this premise. 
Rather than reverse-engineering a model, we make interpretability a constraint of the training  pipeline, optimized alongside the language modeling objective. 
Across three orders of magnitude of compute, on  both autoregressive and diffusion language models, interpretability scales with capability rather than against it.
Surprisingly, model representations become more disentangled and aligned with human-understandable concepts with scale. \\

We instantiate the training-time recipe with \steerlingB, a diffusion language model with a causal attention mask.  
For any group of generated tokens, \steerlingB attributes the output to relevant input tokens, human-understandable concepts, and training data. 
This enables closed-loop intervention: diagnose an output through its concept or feature attribution, retrieve similar training data, and correct the behavior through concept steering without retraining. 
\steerlingB remains competitive with open peer models trained on substantially 2–16×  more compute, suggesting a different scaling paradigm: interpretability can be designed into training, and it improves with scale.

%Interpretability does not have to be recovered after training: it can be built in from the start, and it gets better with scale.

\end{abstract}

\begin{figure}[!h]
  \centering
  \includegraphics[width=0.9\textwidth]{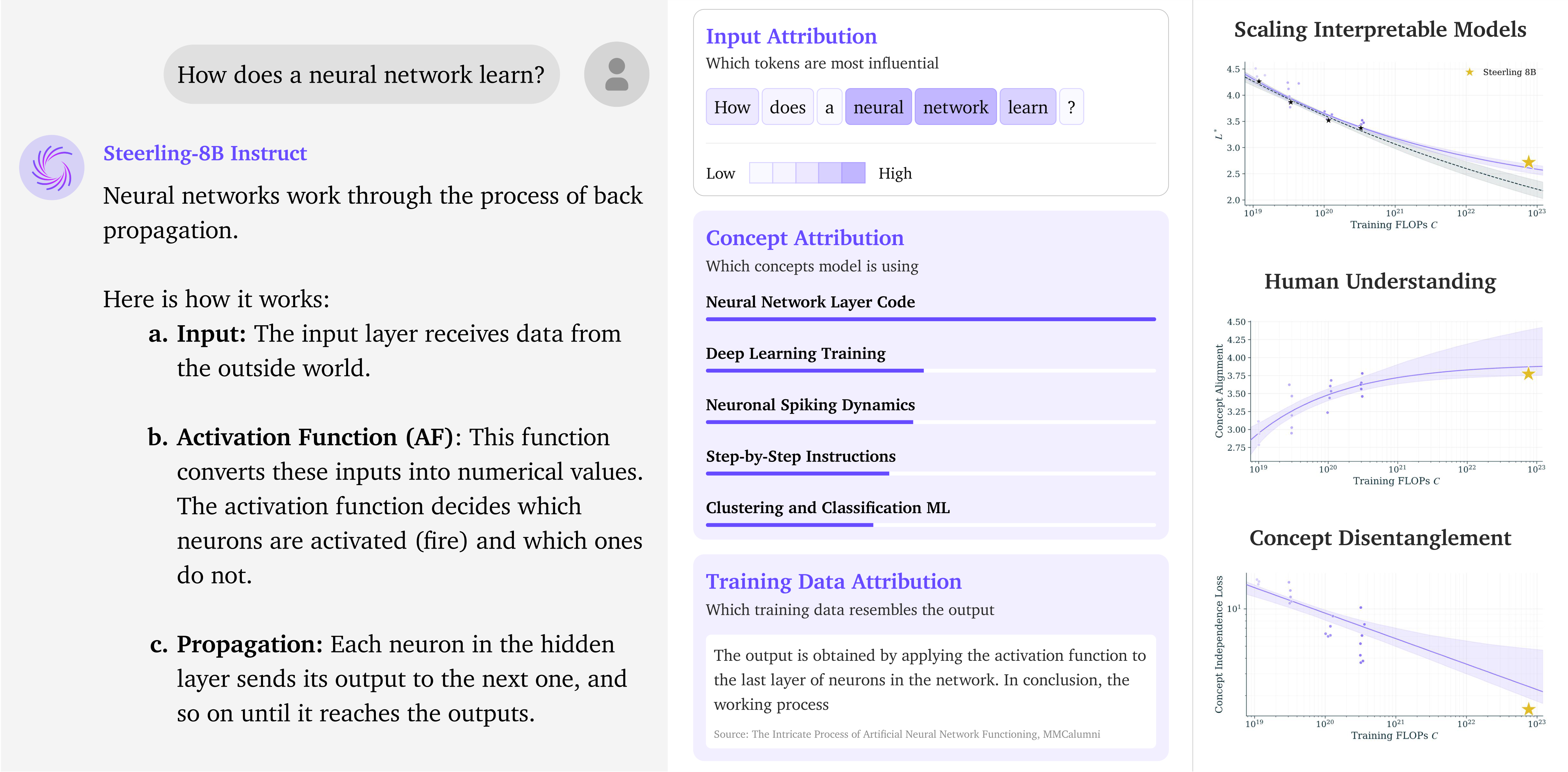}
%   \caption{For the response 
% shown, the model traces which prompt tokens mattered (\emph{Input 
% Attribution}), which internal concepts drove the output and by how much 
% (\emph{Concept Attribution}), and which training data the output 
% resembles, with sources (\emph{Training Data Attribution}).}
%   \label{fig:main}
\end{figure}

\clearpage

\pagenumbering{gobble}   % hide numbers (all ToC pages)
\tableofcontents
\clearpage% --- Main paper starts at page 1 ---
\pagenumbering{arabic}
\setcounter{page}{1}

\input{sections/introduction/main}

\input{sections/background/main}
\input{sections/recipe/main}

\input{sections/data/main}
\input{sections/architecture/main}
\input{sections/attribution/main}

\input{sections/interpretability_metrics/main}

\input{sections/scaling_laws/main}
\input{sections/steerling_pretraining/main}

\input{sections/steerling_midtraining/main}
\input{sections/steerling_posttraining/main}

\input{sections/related_works/main}
\input{sections/conclusion}
\input{sections/contributions}
% \clearpage
\bibliographystyle{plainnat}
\bibliography{refs}

\appendix
\input{sections/appendix/architecture}
\input{sections/appendix/capabilities}
\input{sections/appendix/data}

\input{sections/appendix/interpretability}

\input{sections/appendix/scaling_laws}
\input{sections/appendix/steerling_pretraining}

\input{sections/appendix/steerling_midtraining}

\end{document}

%% file: math_commands.tex
\usepackage{bm}

\def\1{\bm{1}}

\def\eps{{\epsilon}}

\DeclareMathAlphabet{\mathsfit}{\encodingdefault}{\sfdefault}{m}{sl}
\SetMathAlphabet{\mathsfit}{bold}{\encodingdefault}{\sfdefault}{bx}{n}

\newcommand{\E}{\mathbb{E}}

\newcommand{\R}{\mathbb{R}}

%% file: sections/introduction/main.tex
\section{Introduction}
Today's most capable AI systems are also the least understood.
This state of affairs is often treated as the price of progress: if a model is constrained toward human-meaningful structure, the assumption goes, it must have weaker performance.
The consequence is a now-pervasive workflow: we train the most capable model we can, then try to reverse-engineer it after the fact, as if training were a law of nature whose results we can only observe, never intervene on, lest we harm performance.
In this work, we put this premise to the test.
Across three orders of magnitude of compute, on both autoregressive and diffusion models, we show that building interpretability constraints into the training pipeline introduces a small, fixed scaling offset rather than a growing penalty.
More surprisingly, we find that training with interpretability constraints produces models whose representations become \emph{more} disentangled and aligned with human-understandable concepts with scale; coupling capability and understanding.

\textbf{Reverse engineering models and post-hoc interpretability.} A tremendous amount of scholarship has gone into the status-quo: train a model, then inspect it with classifier probes~\citep{alain2016understanding}, feature attributions~\citep{simonyan2013deep, sundararajan2017axiomatic}, sparse autoencoders~\citep{bricken2023monosemanticity}, perturbation tests~\citep{zeiler2014visualizing, lundberg2017unified}, or chain-of-thought~\citep{wei2022chain, nye2021show}.
These tools are useful, but they share a structural limitation: they explain a model that was never trained to make the explanation itself a valid interface into the model~\citep{meloux2025dead}. 
A probe reports that information is decodable, not that the model uses it.
Feature attribution often measures local sensitivity, not necessarily the effect of a human-relevant intervention. 
A sparse feature may reconstruct an activation without corresponding to a causal unit in the computation. 
A chain-of-thought may describe a plausible reason without being tied to the computation that produced the answer.
The problem is that the standard model training recipe does not create an interface whose variables, interventions, and semantic labels are coupled to the prediction computation. In \cref{sec:posthoc}, we examine these challenges in detail.

\textbf{Inherent interpretability.} Instead of asking how to explain an opaque model after training, we ask what conditions must hold for an explanation to be faithful, then build those conditions into the data, architecture, objective, and loss functions.  In \cref{sec:recipe}, we formalize this as \emph{inherent interpretability}: an attribution is
not an auxiliary visualization, but a trained interface satisfying key conditions.
 
\textbf{Three model understanding axes.} \steerlingB instantiates the inherent interpretability recipe  in the diffusion language modeling paradigm. For any output, the model traces its
prediction along three axes:
\begin{enumerate}
    \item to input tokens that affect the output under a trained absence baseline;
    \item to human-understandable concepts, in its representations, that contribute to the output; and
    \item to training data.
\end{enumerate}

\textbf{From explanation to control.} The same interfaces that provide model understanding also support intervention.
In the model, each concept is reprsentation as a direction that the model uses in its forward pass to produce an output. 
Consequently, amplifying or suppressing a concept is a simple edit. 

%, which makes systematic a closed loop that standard post-hoc workflows do not provide by construction: decompose an output into concept contributions, surface training data the model represents as similar, edit the relevant concept, and verify the resulting change, all without retraining.

\textbf{Atlas.} A central obstacle to our proposed training recipe is that no suitable concept library existed at the scale of modern pretraining corpora. To address this, we build Atlas, a concept annotation pipeline that starts from millions of documents, extracts hundreds of millions of free-form tags, canonicalizes them into over 33,000 concepts, and trains an annotator that labels arbitrary text at chunk level. 
In total, Atlas annotates over 1 Trillion tokens across web text, code, mathematics, and academic prose. We describe this system in  \cref{sec:data}.
 
\textbf{Architecture.} The architecture makes these concepts native to the model's computation.
\steerlingB uses a backbone with block-causal attention that preserves diffusion-style parallelism within blocks while retaining autoregressive-style KV caching across blocks.
Between the transformer backbone and the language-modeling head, we insert an additive concept bottleneck that makes the logit decomposition algebraically exact. The masking objective gives the model a trained representation of ``no information at this position,'' making feature-removal baselines in-distribution by construction. The overhead is small and decreases with scale: the concept module accounts for 4\% of parameters at 8B, and under the same parameterization, would fall below 1\% at frontier scales.  \cref{sec:architecture} describes the architecture and training objective.
 
\textbf{Scaling.} The main empirical question is whether this structure makes the model weaker.
We answer with IsoFLOP scaling sweeps across three orders of magnitude of compute and four model families: autoregressive, causal diffusion, autoregressive with concepts, and causal diffusion with concepts.
Adding the concept module shifts the compute-optimal scaling exponents by a small, fixed per-backbone offset; the cost of interpretability does not grow with scale. 
Simultaneously, all interpretability metrics improve with compute on both backbones: the model predicts concepts more accurately, separates known and unknown representations more cleanly, routes more of its prediction through concepts rather than the residual, and aligns its concept embeddings more closely with human-meaningful labels.
The validation loss of \steerlingB is predicted within 0.11 nats from small-scale fits using the joint Chinchilla form, and three of four interpretability metrics are predicted within tight bounds.
Under the metrics we measure, the model does not become harder to understand as it scales.
It becomes easier.
 \cref{sec:scaling-laws} presents the full analysis.
 
\looseness=-1\textbf{An interpretable model can be competitive with opaque peers trained on far more compute.} We train \steerlingB on 1.2 trillion tokens followed by 150 billion midtraining tokens on a code and math augmented mixture. 
Compared with open peer models at similar parameter scale, each trained on roughly 2--16$\times$ more compute, \steerlingB lands within approximately 10\% of their average benchmark performance, despite carrying interpretability constraints throughout training. A model can be both interpretable and competitive. Sections~\ref{sec:pretraining} and~\ref{sec:midtraining} describe the full training process.
 
\paragraph{Overview.}
The remainder of the paper is organized as follows.  \cref{sec:background} reviews transformers, diffusion language models, and concept bottleneck models.  \cref{sec:recipe} presents the interpretable training recipe and formalizes faithfulness and inherent interpretability.  \cref{sec:data} describes Atlas, the concept annotation pipeline.  \cref{sec:architecture} describes the causal-diffusion architecture and concept module.  \cref{sec:capabilities} presents attribution and steering capabilities.  \cref{sec:interp_metrics} defines the interpretability metrics.  \cref{sec:scaling-laws} presents the scaling-law analysis. Sections~\ref{sec:pretraining} and~\ref{sec:midtraining} describe \steerlingB pretraining and midtraining.  \cref{sec:related-work} discusses related work, and  \cref{sec:conclusion} concludes.

%% file: sections/background/main.tex
\clearpage
\section{Background}
\label{sec:background}
In this section, we present material that is core to our discussion in the rest of the paper. In addition, we present the notation that we use across the rest of the work. 

\subsection{Transformer notation}
\label{sec:bg-transformer}

A Transformer~\citep{vaswani2017attention} maps a sequence of input tokens to a sequence of hidden states per position. 
Each hidden state summarizes the token at that position together with its contextual information, and the model projects these hidden states into logits over the vocabulary. We summarize the notation in \cref{tab:notation}.

\input{sections/background/tables/notation}

\subsection{Autoregressive language models}
\label{sec:bg-ar}

Autoregressive language models~\citep{radford2018improving} are trained on the task of next-token prediction: given an input sequence, the model predicts the next token, one at a time, conditioned on all previous tokens. We use the following notation:

\begin{itemize}
    \item $p_\theta(x^i \mid \mathbf{x}^{<i})$ is the model's predicted distribution over the next token at position $i$, given all previous tokens $\mathbf{x}^{<i} = (x^1, \ldots, x^{i-1})$, where $\theta$ denotes the model parameters.
    \item The training loss is the negative log-likelihood of the next token, averaged over all positions in the sequence:
    \begin{equation}
    \mathcal{L}_{\text{AR}} = -\frac{1}{N} \sum_{i=1}^{N} \log p_\theta(x^i \mid \mathbf{x}^{<i}).
    \label{eq:ar-loss}
    \end{equation}
\end{itemize}

\subsection{Masked diffusion language models}
\label{sec:bg-mdm}

Diffusion language models~\citep{austin2021, ou2025, sahoo2024, shi2024} are trained by reversing a forward corruption process applied to the input sequence, rather than by predicting the next token.
A common variant is the masked diffusion language model (MDM), where the corruption process independently replaces tokens with a special \texttt{[MASK]} token. The model is trained to reconstruct the original tokens from the corrupted sequence. The amount of corruption is controlled by a noise level $t \in [0,1]$, where each token is masked with probability $t$. Thus, $t=0$ corresponds to no corruption, while $t=1$ corresponds to the fully masked sequence. We use the following notation:
\begin{itemize}
    \item $\mathbf{x}_t$ is the corrupted sequence at noise level $t$.
    \item $M(\mathbf{x}_t)$ is the set of positions in $\mathbf{x}_t$ that have been masked.
    \item $p_\theta(x^i \mid \mathbf{x}_t)$ is the model's predicted distribution over the original token at masked position $i$, given the corrupted sequence.
\end{itemize}

The training objective is a cross-entropy loss over the masked positions only, averaged over noise levels and sequences:
\begin{equation}
\mathcal{L}_{\text{MDM}} =
\mathbb{E}_{t, \mathbf{x}, \mathbf{x}_t}
\left[
\frac{1}{|M(\mathbf{x}_t)|}
\sum_{i \in M(\mathbf{x}_t)}
-\log p_\theta(x^i \mid \mathbf{x}_t)
\right].
\label{eq:mdm-loss}
\end{equation}

\subsection{Concept bottleneck models} 
\label{sec:bg-cbms}

Concept Bottleneck Models (CBMs)~\citep{cbm} add interpretability to a black-box neural network by inserting a layer of human-interpretable concepts between the input and the output. A sample $x$ is first mapped to concept activations $c = \phi(x) \in \mathbb{R}^n$, where each entry of $c$ corresponds to a supervised, human-interpretable concept. A second function $\psi$ then predicts the label from the concepts, $y = \psi(c)$. When $\psi$ is linear, as in the original formulation, the final prediction is a weighted sum of interpretable concepts:
\begin{equation}
x \xrightarrow{\;\phi\;} c \xrightarrow{\;\psi\;} y.
\label{eq:cbm}
\end{equation}

Training a CBM requires ground-truth concept labels $c$ alongside the target $y$, giving two losses. A concept loss matches $\phi(x)$ to $c$, and a prediction loss matches $\psi(\phi(x))$ to $y$.

% These can be combined in different ways. Joint training optimizes both losses end-to-end. Sequential training fits $\phi$ first, then trains $\psi$ on the predictions $\phi(x)$. Independent training fits $\psi$ on the ground-truth concepts $c$ rather than the predictions. Joint training is the most expressive, but it lets the prediction loss push unintended information into the concept activations, a failure mode known as concept leakage~\citep{mahinpei2021promises} which can be avoided through independent training.

\looseness=-1Concept Bottleneck Generative Models (CBGMs)~\citep{cbgm} extend this idea to generative modeling. Since a fixed set of supervised concepts cannot capture everything in the input, CBGMs add an unsupervised concept channel $u$ alongside the known concepts $c$, so that generation routes through both an interpretable known part and an unknown part that absorbs the remaining information:
\begin{equation}
x \xrightarrow{\;\phi\;} (c, u) \xrightarrow{\;\psi\;} y.
\label{eq:cbgm}
\end{equation}
CBGMs introduce an additional orthogonality loss, which encourages the unknown embeddings to be orthogonal to the known concept embeddings so that the unknown channel encodes information distinct from the supervised concepts.

%% file: sections/background/tables/notation.tex
\begin{table}[!ht]
\centering
\begin{tabular}{ll}
\toprule
Symbol & Meaning \\
\midrule
$\mathbf{x} = (x^1, \ldots, x^N)$ & Input token sequence of length $N$ \\
$x^i$ & Token at position $i$ \\
$V$ & Vocabulary \\
$T_{x^i} \in \mathbb{R}^d$ & Learned embedding of token $x^i$ \\
$d$ & Hidden dimension \\
$L$ & Number of transformer layers \\
$h \in \mathbb{R}^d$ & Transformer hidden state at a given position \\
$W \in \mathbb{R}^{|V| \times d}$ & Language modeling head \\
$W_y$ & Row of $W$ corresponding to token $y$ \\
$\ell_y = h^\top W_y$ & Logit for output token $y$ \\
$p_\theta$ & Model with parameters $\theta$ \\
$\mathbf{x}^{<i}$ & Sub-sequence of tokens before position $i$ \\
$\mathcal{L}_{\text{AR}}$ & Autoregressive training loss \\
\midrule
$\mathbf{x}_t$ & Corrupted sequence at noise level $t$ \\
$t \in [0, 1]$ & Noise level \\
$M(\mathbf{x}_t)$ & Set of masked positions in $\mathbf{x}_t$ \\
$\mathcal{L}_{\text{MDM}}$ & Masked diffusion training loss \\
\bottomrule
\end{tabular}
\caption{Transformer and language-model notation.}
\label{tab:notation}
\end{table}

%% file: sections/recipe/main.tex
\clearpage
\section{A recipe for building interpretable models}
\label{sec:recipe}

\looseness=-1In this section, we present a general recipe for building interpretable language models. The recipe follows the standard pipeline for training language models (data curation, architecture and loss design, optimization, and evaluation) but modifies each stage to introduce human-interpretability constraints. 

\begin{figure}[htbp!]
  \centering
  \includegraphics[trim={2cm 3cm 2cm 2cm}, clip, width=\textwidth]{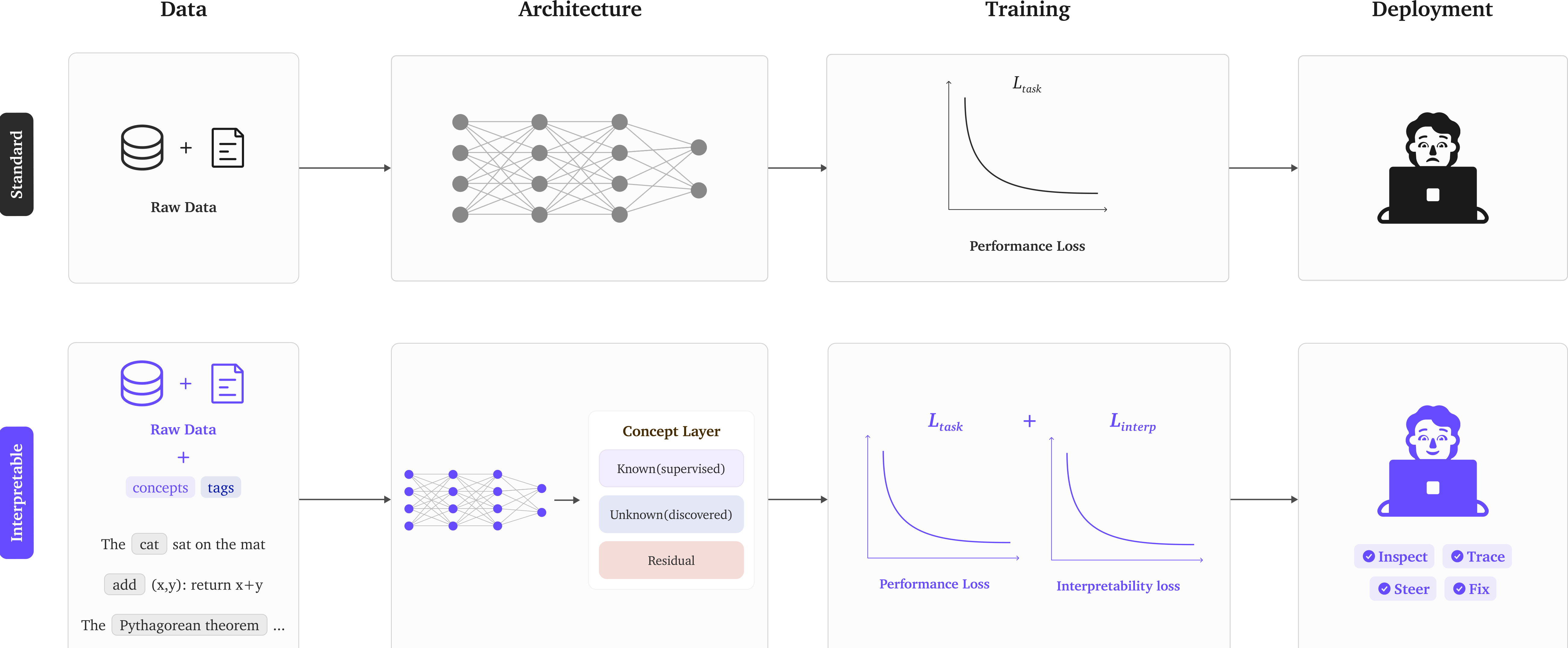}
  \caption{The recipe for training an interpretable model.}
  \label{fig:interpreable_vs_reg}
\end{figure}

\paragraph{Roadmap.}
We proceed in four parts:
\begin{itemize}
    \item \cref{sec:requirements} defines the interpretability requirements that we aim to satisfy in this work.
    \item \cref{sec:posthoc} shows why the standard recipe does not provide the required guarantees.
    \item \cref{sec:faithfulness-and-interp} formalizes explanation faithfulness and inherent interpretability.
    \item \cref{sec:interp-recipe} presents the interpretable recipe: pipeline modifications traced to specific conditions (summarized in Figure~\ref{fig:interpreable_vs_reg}).
\end{itemize}

\subsection{Interpretability requirements}
\label{sec:requirements}

Here we formalize our interpretability requirements: \emph{input attribution}, which measures the effect of removing each input token; \emph{concept attribution}, which decomposes the model's output into per-concept contributions; and \emph{training data (similarity) attribution}, which retrieves training examples the model considers similar to its output.

Interpretability describes the interface between a model and a human: the structures through which a person can inspect, evaluate, understand, and act on a model's behavior. We require three specific capabilities of this interface.

\begin{gltitledbox}{Informal Interpretability Requirements}
Given a model, $p_\theta$, for any output it produces, we want a human user to be able to answer three~\mbox{questions:}
\begin{enumerate}
    \item \textbf{``What in the input mattered?''}  This requirement asks which parts of the input most influence the output?

    \item \textbf{``What topics/ideas are responsible for that?''} This requirement asks what topics/ideas in the model's internal representation drove this output?
    \item \textbf{``Where is the output coming from?''} This requirement asks which training data resembles what the model just produced.
\end{enumerate}
\end{gltitledbox}

\looseness=-1 The literature refers to such artifacts as explanations, interpretations, or attributions. In this work, we use \emph{attribution} throughout: each requirement asks the model to attribute to a specific component.

We translate each question into a precise operation, which we term an \emph{attribution}:
\begin{enumerate}
    \item \textbf{``What in my input mattered?" $\rightarrow$ Input attribution}. We can measure the change in the model's output when each input is replaced by a trained absence baseline, and rank tokens by the magnitude of this change.
    \item \textbf{``What topics/ideas are responsible for that?" $\rightarrow$ Concept attribution}. We can decompose the model's internal representation into contributions from human-understandable topics, and trace each topic's contribution to the output.
    \item \textbf{``Where is this coming from?" $\rightarrow$ Training data attribution}. We can retrieve the training examples whose representations are most similar to the representation of the model's~\mbox{output.}
\end{enumerate}

To move forward, we will now concretize core definitions as we specify the components necessary to build an inherently interpretable language model.

\paragraph{Concepts.} In this work, and as is common in the literature, we term a human-understandable topic a \emph{concept}. More generally, a concept is a coherent unit of meaning that a human recognizes and that we seek to associate with the model's internal representations. Concepts span multiple levels of granularity: high-level themes like \texttt{sports}, \texttt{politics}, or \texttt{tourism}; mid-level topics like \texttt{gradient descent optimization}, \texttt{state elections}, or \texttt{Mediterranean cuisine}; and fine-grained units like \texttt{dropout regularization in recurrent networks} or \texttt{ranked-choice ballot counting}. They may also capture stylistic or functional attributes, \texttt{formal tone}, \texttt{sarcasm}, \texttt{step-by-step reasoning}, rather than topical content. The defining property is not the level of abstraction but human recognizability: a domain expert, given sufficient context, can identify whether the concept is present in a piece of text. Definition~\ref{def:concept} formalizes this intuition.

\begin{definition}[Concept]
\label{def:concept}
A \emph{concept} is a pair $c = (z_c, m_c)$ consisting of two components:
\begin{itemize}
    \item $z_c$, \textbf{what the model computes with}: an activation, embedding, or dimension that participates in producing the model's output, $\ell_y$, and can be modified to change the model's behavior (formalized in \cref{sec:inherent-interp}).
    \item $m_c$, \textbf{what the human sees}: a description that explains, in natural language, what $z_c$~\mbox{represents.}
\end{itemize}
In this work, $m_c = (l_c, d_c, \mathcal{T}_c)$, where $l_c$ is a short label, $d_c$ is a one-sentence description, and $\mathcal{T}_c$ is a set of characteristic words grounding the concept in observable language.
\end{definition}

We refer to $z_c$ as the concept's \emph{model variable} and $m_c$ as its \emph{semantic card} throughout the paper. The data pipeline (\cref{sec:data}) produces semantic cards before any model exists; the model variable $z_c$ is instantiated when the architecture described in \cref{sec:architecture} is trained.

With the definition of a concept in place, we now make each of our three attributions---input, concept, and training data---mathematically precise. We do this for a specific reason: in \cref{sec:posthoc}, we will show that post-hoc methods fail to deliver these attributions reliably, and in \cref{sec:faithfulness-and-interp}, we will define formal conditions under which they \emph{are} reliable. 

\paragraph{Notation.} Let $p_\theta$ be the model with parameters $\theta$, $\mathbf{x} = (x^1, \ldots, x^N)$ an input sequence of $N$ tokens, $y$ a predicted output token, and $\ell_y = h^\top W_y$ the output logit for token $y$, where $h \in \mathbb{R}^d$ is the hidden state and $W_y$ is the row of the language modeling head corresponding to $y$. We write $\mathcal{A}$ for a generic attribution functional, and use superscripts to distinguish the three types.
\paragraph{Input attribution.}
\emph{If an input token were absent, how much would the output change?} The input attribution functional maps the model, input, and output to a vector of per-token relevance scores:
\begin{equation*}\label{eq:feat-attr}
    \mathcal{A}^{\mathrm{input}}(p_\theta, \mathbf{x}, y) = \bigl(\alpha_1,\, \ldots,\, \alpha_N\bigr) \;\in\; \mathbb{R}^{N},
\end{equation*}
where
\begin{equation*}\label{eq:feat-deletion}
    \alpha_i = \ell_y(\mathbf{x}) - \ell_y(\mathbf{x}_{\setminus i})
\end{equation*}
measures the effect of replacing token, $x^i$, with a baseline representing absence. This definition has a prerequisite that we formalize later: it is meaningful only if $\mathbf{x}_{\setminus i}$ is in-distribution. If the model has never seen inputs with position $i$ absent, the output $\ell_y(\mathbf{x}_{\setminus i})$ reflects extrapolation. We return to this in \cref{sec:faithfulness}, where we define \emph{validity}.

\paragraph{Concept attribution.}
\emph{Which concepts contributed to the model's prediction, and by how much?} The concept attribution functional maps the model, input, and output to a vector of per-concept relevance scores:
\begin{equation*}\label{eq:concept-attr}
    \mathcal{A}^{\mathrm{concept}}(p_\theta, \mathbf{x}, y) = \bigl(\phi_1(y),\, \ldots,\, \phi_n(y),\, \rho(y)\bigr) \;\in\; \mathbb{R}^{n+1},
\end{equation*}
where $\phi_c(y)$ is the contribution of concept $c$ to the output logit $\ell_y$ and $\rho(y)$ is the residual; the portion of the prediction not attributed to any concept. The output logit decomposes as:
\begin{equation*}\label{eq:logit-decomp}
    \ell_y = \sum_c \phi_c(y) + \rho(y).
\end{equation*}
How the decomposition above is achieved, is an architectural choice described in \cref{sec:architecture}.

\paragraph{Training data (similarity) attribution.}
\emph{Which training examples does the model consider similar to this output?} The similarity attribution functional maps the model output and input to a set of training data, ranked by similarity, under the model's own encoder $r_\theta$:
\begin{equation*}\label{eq:retrieval}
    \mathcal{A}^{\mathrm{retrieval}}(p_\theta, \mathbf{x}) = \operatorname{TopK}_{D \in \mathcal{D}} \; \mathrm{sim}\bigl(r_\theta(\mathbf{x}),\, r_\theta(D)\bigr).
\end{equation*}

\paragraph{Similarity not causation.} Our training data attribution requirement does not seek causation. It only requires that the model represents these training data similarly to its output. We do not claim training influence, which would require approximating counterfactuals over retraining~\citep{koh2017understanding, bae2022if, grosse2023studying}.

\noindent Retrieval differs from the other two attributions in an important way: it is a \emph{representational provenance} functional, not a causal attribution. Concept attribution decomposes the output; input attribution measures counterfactual effects; retrieval reports similarity in the model's own geometry. Its criterion for being meaningful is that the similarity scores reflect the model's actual representation, not an externally imposed metric. This holds by construction when the retrieval encoder is the same encoder that feeds into prediction.

\subsection{Why post-hoc methods fail to satisfy the interpretability requirements}
\label{sec:posthoc}
Each of the three attributions defined above has post-hoc analogues: methods that attempt to compute input, concept, or training data attributions from an already-trained model, without modifying its training in any way. Each can in principle succeed. Each, in standard practice, does not \emph{guarantee} the required conditions because the model was not trained to support them. 

We begin with the root cause: standard training does not create an explanatory interface whose variables are coupled to prediction. We then examine post-hoc attempts at each attribution in turn: concept attribution via probes and sparse autoencoders, input attribution via gradient and perturbation methods, and training data attribution via influence functions. We close with the unifying diagnosis and a deeper obstacle, the Rashomon problem, which shows that even faithful post-hoc explanations can be arbitrary.

\paragraph{Root cause.} The standard model training pipeline optimizes for prediction. 
\begin{gltitledbox}{The core issue with post-hoc interpretability}
A standard model is trained for prediction. Its training objective
does not ask for representations that decompose into meaningful units, baselines
that make ``what if this input were absent?'' a well-defined question, or reasoning
traces that reflect the actual computation. Post-hoc interpretability methods assume
one or more of these structures exist and attempt to recover them after training.
But they were never created: the model's representations have no reason to be
interpretable, its responses to missing inputs have no reason to be meaningful,
and its verbal explanations have no reason to match its internal computations.
The failures below are not limitations of specific methods; they are consequences
of analyzing a model that was built without specifying formal interpretable conditions.
\end{gltitledbox}

\paragraph{Probes.} A common approach to post-hoc concept attribution is to train a linear classifier on 
the hidden states $h$ of an already-trained model and interpret high accuracy as evidence 
that the model ``represents'' a concept~\citep{alain2016understanding, belinkov2022probing}. 
The fundamental gap is between information \emph{presence} and information \emph{use}: 
a probe asks whether concept $c$ is decodable from $h$, not whether $p_\theta$ relies 
on $c$ for prediction. A perfectly accurate probe is compatible with the model ignoring 
the probed feature entirely~\citep{ravichander2021probing}. Several lines of evidence 
sharpen this concern:
\begin{itemize}
    \item Linear and structural probes recover information with high accuracy from 
          randomized contextualized embeddings~\citep{conneau2018you, hewitt2019structural}.
    \item Syntactic probes do not generalize across 
          domains~\citep{maudslay2021syntactic}.
    \item Probes extract features merely encoded in token embeddings but unused 
          during inference~\citep{ravichander2021probing}.
    \item Capacity-controlled, information-theoretic, and amnesic probing mitigate 
          some false discoveries but do not close the presence-versus-use gap~\citep{voita2020information, elazar2021amnesic}.
\end{itemize}
Even when probes identify directions useful for steering, one cannot assert that the 
model uses those directions for prediction. Each new concept requires training a separate 
probe, with no guarantee of coherence across probes.

\paragraph{Sparse autoencoders.}
Sparse autoencoders (SAEs) decompose a frozen model's hidden states $h$ into 
sparse dictionaries of features~\citep{bricken2023towards}, 
and have surfaced striking results: features corresponding to the Golden Gate 
Bridge, emotional states, and safety-relevant 
behaviors~\citep{templeton2024scaling}. However, the gap between discovering 
interesting features and guaranteeing that those features are the variables 
through which $p_\theta$ computes its predictions is substantial. Because the SAE is trained on a frozen model, its features are 
post-hoc descriptions with no structural relationship to the prediction 
pathway. More concerning, recent work shows that SAEs trained with different 
random seeds produce different feature sets, and that random SAE baselines 
match fully-trained SAEs on sparse probing and causal editing 
metrics~\citep{korznikov2026sanity}. The explanations may seem convincing, 
but they are not uniquely grounded in what the model learned.

\paragraph{Gradient-based input attribution.}
Gradient-based attribution methods compute $\partial \ell_y / \partial x^i$ and present the result as a measure of feature importance. Several variants abound including saliency maps~\citep{simonyan2013deep}, Grad-CAM~\citep{selvaraju2016grad}, and integrated gradients~\citep{sundararajan2017axiomatic}, and many others.
 
Gradients measure infinitesimal sensitivity; however, the human typically cares about the effect of removing or substantially changing a feature, a finite but larger perturbation. For gradients to be a useful proxy, the model must be approximately linear over the perturbation budget the human cares about or be off-manifold robust~\citep{dombrowski2019explanations, srinivas2022efficient, srinivas2023models}. However, unless the training procedure is regularized to produce models that are locally linear, there is no incentive for the model to satisfy this requirement. Handed a model, the user might not have a way to assess how locally linear it is, and therefore no way to know whether the gradient is a valid proxy for the perturbation they care about.
 
\paragraph{Perturbation-based input attribution.}
Perturbation-based methods like SHAP~\citep{lundberg2017unified} and occlusion~\citep{zeiler2014visualizing} replace input tokens with a reference value and measure the output change. Common reference values (zero vectors, random tokens, arbitrary masks) are out of distribution: the observed effect reflects the model's extrapolation behavior, not the feature's actual role~\citep{hooker2019benchmark,kumar2020problems, jain2022missingness}. The resolution is to choose a null state that is in-distribution by construction for the model at hand.
 
\paragraph{Unfaithful chain of thought.}
Chain-of-thought reasoning~\citep{nye2021show, wei2022chain} produces natural-language explanations alongside outputs. The model's stated reasoning need not reflect its actual computation: a model can produce a correct-sounding chain that arrives at the answer for entirely different internal reasons, or a plausible chain that is confabulated~\citep{lanham2023measuring, madsen2024self, turpin2024language, korbak2025chain, chen2025reasoning}. The explanation channel (generated text) and the computation channel (hidden states $\to$ output) are not structurally coupled by training.

\paragraph{A deeper problem: explanatory multiplicity.}
Even if post-hoc methods could deliver faithful attributions for a given model, 
the result would be contingent on which model happened to emerge from training. 
Models with identical performance can differ arbitrarily in their internal 
mechanisms, and explanatory multiplicity is decoupled from predictive 
multiplicity: two models with indistinguishable accuracy can produce 
contradicting attributions for the same input~\citep{d2022underspecification,brunet2022implications}. 
This instability propagates to any downstream action, e.g. steering, editing, 
counterfactual recommendations, derived from the 
explanation~\citep{pawelczyk2020counterfactual}. The recipe in \cref{sec:interp-recipe} 
addresses this by anchoring the explanatory interface: the concept library, the 
absence baseline, and the additive decomposition are shared across all models 
the training procedure can produce. Different runs yield different parameters, 
but the interface through which attributions are computed is fixed. 
\cref{sec:related-work} discusses the underspecification 
and Rashomon literature in detail.

\begin{gltitledbox}{The common thread}
Every gap above traces to the same structural absence: the model was not trained
to make the attribution valid. The training was indifferent to whether probed
features would be causally relevant, perturbation baselines would be
in-distribution, gradients would be informative, or stated reasoning would
reflect computation. This observation is not
new~\citep{meloux2025dead}:
post-hoc attributions are not identifiable estimators for the quantities they
claim to measure. What is needed is to make the attributions identifiable, and one way to do that is to build the attributions into the training process itself.
\end{gltitledbox}

\subsection{Faithfulness and inherent interpretability}
\label{sec:faithfulness-and-interp}
 
The gaps identified in \cref{sec:posthoc} share a common structure: the attribution's claims either disagree with what happens when one intervenes on the model, or the intervention itself is invalid. We now formalize these two failure modes as conditions on attributions (\cref{sec:faithfulness}), then define the class of models that satisfy them by construction (\cref{sec:inherent-interp}).
 
%% ---------------------------------------------------------------
\subsubsection{Faithfulness}
\label{sec:faithfulness}

\textbf{Prior work on faithfulness.} The concept of attribution faithfulness has 
been widely discussed in the literature~\citep{hooker2019benchmark, deyoung2020eraser, lyu2024towards, atanasova2023faithfulness}. 
Most treatments define faithfulness informally as ``the explanation/attribution reflects the 
model's actual reasoning''~\citep{jacovi2020towards} or operationalize it through 
specific tests, perturbation checks, sufficiency, 
comprehensiveness~\citep{deyoung2020eraser}, without unifying these into 
conditions that a training procedure can be designed to satisfy. We distill the 
literature into two concrete conditions, agreement and validity, that are 
checkable, traceable to specific design choices, and help to 
diagnose every post-hoc failure identified in \cref{sec:posthoc}.
 
\begin{definition}[Faithfulness]
\label{def:faithfulness}
An attribution functional, $\mathcal{A}$, is \emph{faithful} with respect to model, $p_\theta$, intervention family, $\mathcal{I}$, and tolerance $\xi$, if:
\begin{enumerate}
    \item \textbf{Agreement.} The output relevance scores, from the attribution, predict the observed effects of interventions in $\mathcal{I}$ within tolerance $\xi$.
    \item \textbf{Validity.} The interventions in $\mathcal{I}$ belong to a \emph{training-supported intervention family}: the model has been trained on, or explicitly regularized for, the perturbed states used to define the attribution.
\end{enumerate}
\end{definition}
 
\textbf{Two types of interventions.} We distinguish input interventions $\mathcal{I}_x$, such as replacing token, $x^i$, with the mask token, from latent interventions $\mathcal{I}_z$, such as modifying a concept variable $z_c$. For masking, validity means the model saw masked contexts during training. For concept interventions, validity means edits are bounded to observed activation ranges or sampled concept values.
 
\textbf{Mapping to \cref{sec:posthoc}.} Each post-hoc issue, previously identified, is a violation of one or both conditions. Probes violate agreement: information presence does not imply information use. SAEs violate agreement: reconstruction geometry does not imply causal role. Gradient methods violate agreement: local sensitivity does not match the human's perturbation budget. Perturbation methods violate validity: the baseline is out of distribution. Chain-of-thought violates agreement: the text channel is not coupled to the computation channel.
 
%% ---------------------------------------------------------------
\subsubsection{Inherent interpretability}
\label{sec:inherent-interp}
We now have all the relevant definitions and terms to state what we mean by
inherent interpretability in this work.

\begin{definition}[Inherent interpretability]
\label{def:inherent-interp}
 A model, $p_\theta$, is \emph{inherently interpretable} with respect to an attribution functional, $\mathcal{A}$, an intervention family, $\mathcal{I}$, a semantic map $\mathcal{M}: c \mapsto m_c$ assigning each concept its semantic card (Definition~\ref{def:concept}), a tolerance $\xi$, and a coverage threshold $\zeta$ if its training (or finetuning) procedure produces models for which:

\medskip
\noindent\textbf{Causal faithfulness:}
\begin{enumerate}
    \item \textbf{Nativeness.} The attributed variables are part of the computation that produces the model's output, $\ell_y$.

\item \textbf{Agreement.} The attribution predicts the effect of an
    intervention before it is applied. For an intervention in $\mathcal{I}$,
    let $\Delta \ell_y$ be the \emph{observed} change: the difference in the
    output logit between a forward pass with the intervention applied (e.g.,
    a token replaced by mask, or a concept variable
    modified) and the unmodified forward pass. Let $\hat{\Delta} \ell_y$ be
    the \emph{predicted} change: the change implied by the attribution's
    relevance scores alone, computed without running the intervened forward
    pass. Agreement requires
    \begin{equation*}
        |\Delta \ell_y - \hat{\Delta} \ell_y| \leq \xi.
    \end{equation*}
    \item \textbf{Validity.} Interventions in $\mathcal{I}$ belong to a training-supported intervention family.
\end{enumerate}
 
\noindent\textbf{Semantic faithfulness:}
\begin{enumerate}
    \setcounter{enumi}{3}
    \item \textbf{Interpretation.} The semantic map $\mathcal{M}$ gives human-valid descriptions of the attributed variables: each card $m_c$ accurately describes what its variable $z_c$ encodes.
    
        \item \textbf{Coverage.} The attributed variables together account for at
    least a fraction $\zeta$ of the model's prediction, with the unattributed
    residual $\rho(y)$ explicitly reported.

\end{enumerate}
\end{definition}

\textbf{Inherent Interpretability does not mean total mechanistic transparency.} Inherent interpretability is a targeted guarantee for specified attribution queries, not a global claim about every neuron or attention head. When we say inherent interpretability in this work, we do not claim that the entire model is human-understandable.

\begin{gltitledbox}{Causal vs.\ semantic faithfulness}
Conditions 1--3 establish \emph{causal faithfulness}: the attribution correctly describes the model-side variable $z_c$ and its role in producing $\ell_y$. Conditions 4--5 establish \emph{semantic faithfulness}: the human description $m_c$ validly represents $z_c$, and the interpretable component is useful for the output.\\[4pt]
Concept leakage~\citep{mahinpei2021promises} is a failure of semantic faithfulness: the attribution may correctly predict the causal effect of $z_c$ (conditions 1--3 hold) while $m_c$ incompletely describes what $z_c$ encodes (condition 4 fails). For example, if $z_c$ encodes ``sports'' plus hidden information about ``gender,'' the numeric attribution $\phi_c(y)$ correctly predicts the effect of modifying $z_c$, but the semantic card $m_c = \text{``sports''}$ is incomplete.
\end{gltitledbox}

\paragraph{Examples of inherent interpretability.} The definition is agnostic to the form of explanation. We illustrate with examples in the literature:

\begin{enumerate}
    \item \textbf{Masking $\to$ inherently interpretable w.r.t.\ $\mathcal{A}^{\mathrm{input}}$.} Training with a masking objective places the absence baseline in-distribution, satisfying validity. \citet{madsen2024interpretability} make this move for finetuning; we extend it to pretraining.
    \item \textbf{Concept bottleneck $\to$ inherently interpretable w.r.t.\ $\mathcal{A}^{\mathrm{concept}}$.} An additive bottleneck gives nativeness and exact agreement ($\xi = 0$). \citet{cbm} introduced this for classification. And \citet{cbgm} extend that approach to generative models.

\end{enumerate}

%% 3.4 THE INTERPRETABLE MODEL-TRAINING RECIPE
\subsection{The interpretable model-training recipe}
\label{sec:interp-recipe}
 
We now have three attribution functionals (\cref{sec:requirements}), a 
demonstration that post-hoc methods do not guarantee them (\cref{sec:posthoc}), 
and a formal definition of inherent interpretability specifying what is needed 
(Definition~\ref{def:inherent-interp}). The recipe modifies the standard 
training pipeline at exactly the points needed to satisfy the five conditions. Every 
modification traces to a specific condition; Figure~\ref{fig:recipes} summarizes 
the full pipeline.

\input{sections/recipe/recipe_table}
 
\textbf{Why pretraining.} Interpretability constraints are most effective when 
present during representation formation. Finetuning restructures existing 
representations; pretraining shapes them from the start. Our recipe targets 
pretraining, though the contracts apply to finetuning with weaker guarantees.
% [Cite: references for pretraining vs finetuning evidence]

\paragraph{The standard recipe.}
A standard training pipeline trains a model $p_\theta$ on input-output pairs 
$\{(\mathbf{x}_i, y_i)\}$ by minimizing a task loss, 
and evaluates on downstream metrics. This recipe is effective for prediction but 
does not require the model to expose explanations. 
\paragraph{The interpretable recipe.}
The interpretable recipe modifies the standard pipeline at five points (right 
column of Figure~\ref{fig:recipes} and bottom row of  Figure~\ref{fig:interpreable_vs_reg}). Each modification is motivated by a 
specific condition of Definition~\ref{def:inherent-interp}.

\paragraph{Concept annotations (Interpretation).}
The training data is augmented with concept annotations: 
$\mathcal{D} = \{(\mathbf{x}_i, y_i, \mathbf{c}_i)\}$. Without supervised 
targets, the model has no signal to organize its representations 
around human-meaningful categories; it will default to whatever geometry 
minimizes the task loss (\cref{sec:posthoc}). Concept annotations provide 
the semantic targets that ground the interpretation condition: each concept 
variable, $z_c$, has a corresponding semantic card, $m_c$, whose meaning is 
established before training. The annotations also enable a control interface, 
since concepts with prior semantics can be monitored, amplified, or 
suppressed. \emph{Our instantiation:} \cref{sec:data} describes the 
Atlas pipeline that produces concept annotations.

\paragraph{Bottleneck architecture (Nativeness, Agreement).}
A model architecture that routes the model's predictions through concepts makes it so that each concept's contribution to the output can be directly computed from the forward pass. This is a structural requirement: the 
concept variables, $z_c$, must lie on the computational path from the 
model's internal state to the output.
The specific mechanism, additive bottleneck, multiplicative gating, or 
another decomposition, is an implementation choice.  \cref{sec:architecture} describes an additive concept bottleneck with a 
linear output head.

\paragraph{Trained absence baseline (Validity).}
Input attribution requires replacing input features with a baseline 
representing absence and measuring the change in output. For this to be 
meaningful, the model must have a learned representation of ``no information 
at this position''. Without 
this, the perturbed input $\mathbf{x}_{\setminus i}$ is out of distribution, 
and the observed output change reflects extrapolation rather than the 
feature's actual role. 
The contract is that the training objective must 
include the perturbed states used to define $\mathcal{A}^{\mathrm{input}}$. 
Standard autoregressive next-token objective does not typically incorporate this type of masking.
\emph{Our instantiation:} \cref{sec:architecture} describes a masked  diffusion objective ($\mathcal{L}_{\mathrm{MDM}}$) that trains the model on  corrupted sequences, making the mask token an in-distribution absence baseline.

\paragraph{Interpretability Losses (Interpretation, Coverage).}
The architecture provides the slot for concepts; the losses ensure the slot is used. Two failure modes must be prevented. First, \emph{concept leakage}: 
the concept variable $z_c$ may encode information beyond what the semantic 
card $m_c$ describes, degrading the interpretation condition. Concept losses that align $z_c$ activations with the annotations $\mathbf{c}_i$ counteract  this by maintaining semantic alignment between the model-side variable and 
its human description. 
Second, \emph{residual domination}: the residual  $\rho(y)$ may absorb most of the predictive capacity, leaving the concept  decomposition algebraically exact but practically vacuous. Losses that  penalize the residual, enforce reconstruction of the unexplained hidden  state, and encourage independence between concept components address this. 
The general form is $\mathcal{L} = \mathcal{L}_{\mathrm{task}} + \sum_j 
\lambda_j \mathcal{L}_{\mathrm{interp},j}$, where the weights $\lambda_j$ 
may be annealed during training to balance task performance and 
interpretability. 
\emph{Our instantiation:} \cref{sec:architecture} describes the specific loss components and annealing schedule.

\paragraph{Evaluation.}
Task metrics (perplexity, accuracy) remain necessary to ensure 
interpretability is not obtained by discarding predictive information. In 
addition, the model should be evaluated against the five conditions of 
Definition~\ref{def:inherent-interp}:

\begin{enumerate}
    \item \textbf{Causal faithfulness (Nativeness).} Are the attributed 
variables on the computational path? Agreement: do 
interventions on $z_c$ change $\ell_y$ as predicted by $\phi_c(y)$? Validity: are the interventions in-distribution?
    
    \item \textbf{Semantic faithfulness.} Interpretation: do humans 
independently recognize the concepts from the model's characteristic 
evidence? Coverage: what fraction of prediction mass is 
carried by concepts versus the residual?

\end{enumerate}

\emph{Our instantiation:} \cref{sec:data-human-eval} describes the human recoverability study 
(Condition~4) and the quantitative evaluation (Conditions~2, 3, 5).

\looseness=-1\paragraph{Summary.}
Every modification traces to a condition of Definition~\ref{def:inherent-interp}; Table~\ref{table:interp_contract} summarizes the consequence of removing each one. The recipe itself is not new: its elements are implicit in work on concept bottlenecks~\citep{cbm}, interpretable-by-design architectures~\citep{rudin2019stop}, masking-based attribution~\citep{madsen2024interpretability}, and recent theoretical treatments that derive interpretability constraints from explicit premises~\citep{barbiero2026standard}. Our contribution is to make the recipe explicit, trace each element to a formal condition, and instantiate it for language models at pretraining scale.

\begin{table}[htbp!]
\centering
\small
\begin{tabular}{lll}
\toprule
\textbf{Remove\ldots} & \textbf{Condition broken} & \textbf{Consequence} \\
\midrule
Concept annotations & Interpretation (4) & No semantic targets \\
Bottleneck architecture & Nativeness (1), Agreement (2) & $\phi_c(y)$ not computable \\
Trained absence baseline & Validity (3) & Input-attribution baselines OOD \\
Interpretability losses & Coverage (5), Interpretation (4) & Residual absorbs capacity \\
\bottomrule
\end{tabular}
\caption{Removing any single recipe modification breaks a specific condition of Definition~\ref{def:inherent-interp}.}
\label{table:interp_contract}
\end{table}

This completes the recipe. The interpretability requirements (\cref{sec:requirements}) defined what we want; the post-hoc analysis (\cref{sec:posthoc}) showed why the standard pipeline cannot deliver it; the definitions of faithfulness and inherent interpretability (\cref{sec:faithfulness-and-interp}) formalized what is needed; and the recipe above specified how to modify each stage of the pipeline to satisfy those conditions. The remainder of the paper instantiates this recipe: \cref{sec:data} instantiates the data contract, building the concept library; \cref{sec:architecture} instantiates the architecture, objective, loss, and optimization contracts, building the model; and \cref{sec:scaling-laws} shows that these contracts preserve compute-optimal scaling while the interpretability properties themselves improve with compute. Finally, \cref{sec:pretraining,sec:midtraining} carry the recipe to full scale, pretraining and mid-training \steerlingB{}.

%% file: sections/recipe/recipe_table.tex
\definecolor{modelcolor}{RGB}{0,120,0}
\definecolor{datacolor}{RGB}{0,100,200}
\definecolor{conceptcolor}{RGB}{200,0,0}
\definecolor{residualcolor}{RGB}{140,100,0}
\definecolor{losscolor}{RGB}{140,0,140}
\definecolor{attrcolor}{RGB}{0,140,140}
\definecolor{evalcolor}{RGB}{200,120,0}
\definecolor{stagecolor}{RGB}{80,80,80}
\definecolor{warncolor}{RGB}{180,80,0}
\definecolor{okcolor}{RGB}{0,130,60}

\definecolor{rowlight}{RGB}{245,245,250}
\definecolor{rowdark}{RGB}{232,233,240}

\begin{figure*}[!ht]
\centering
\small
\setlength{\tabcolsep}{7pt}
\setlength{\extrarowheight}{8pt}
\newcommand{\stage}[1]{{\color{stagecolor}\textbf{#1}}}
\begin{tabular}{
  >{\raggedright\arraybackslash}m{1.5cm}
  >{\centering\arraybackslash}m{5.0cm}
  >{\centering\arraybackslash}m{8.4cm}
}
\toprule
& \textbf{Standard Training Recipe} & \textbf{Interpretable Training Recipe} \\
\midrule

\rowcolor{rowlight}
\stage{Data}
&
$\mathcal{D} = \Bigl\{\overbrace{(\mathbf{x}_i, y_i)}^{{\color{datacolor}\textbf{\scriptsize Input--Output}}}\Bigr\}_{i=1}^{n}$
&
$\mathcal{D} = \Bigl\{\overbrace{(\mathbf{x}_i, y_i)}^{{\color{datacolor}\textbf{\scriptsize Input--Output}}},\;\overbrace{\mathbf{c}_i}^{{\color{conceptcolor}\textbf{\scriptsize Explanations}}}\Bigr\}_{i=1}^{n}$
\\[14pt]

\rowcolor{rowdark}
\stage{Arch.}
&
$\ell_y = \underbrace{\;h\;}_{\color{modelcolor}\textbf{\scriptsize Opaque}}^{\!\top} W_y$
&
$\ell_y = \underbrace{\textstyle\sum_c \phi_c(y)}_{\color{conceptcolor}\textbf{\scriptsize Bottleneck}} + \underbrace{\rho(y)\vphantom{\textstyle\sum_c}}_{\color{residualcolor}\textbf{\scriptsize Residual}}$\newline
{\color{okcolor}\scriptsize$\boldsymbol{\to}$ \textbf{concept \& retrieval attribution}}
\\[14pt]

\rowcolor{rowlight}
\stage{Objective}
&
$\mathcal{L}_{\mathrm{AR}}$ \;{\scriptsize(\S2.2)}
&
Objective with \textbf{trained absence baseline}\newline
{\color{okcolor}\scriptsize$\boldsymbol{\to}$ \textbf{input attribution}}
\\[14pt]

\rowcolor{rowdark}
\stage{Loss}
&
$\textstyle\sum_{i}\ell\bigl((\mathbf{x}_i, y_i);\,\theta\bigr)$
&
$\underbrace{\textstyle\sum_{i}\ell_{\mathrm{task}}}_{\color{losscolor}\textbf{\scriptsize Task}} + \underbrace{\textstyle\sum_{j}\lambda_j\,\ell_{\mathrm{interp},j}(\mathbf{c}_i)}_{\color{losscolor}\textbf{\scriptsize Interpretability}}$
\\[14pt]

\rowcolor{rowlight}
\stage{Evaluate}
&
$\underbrace{\mathrm{score}}_{\color{evalcolor}\textbf{\scriptsize Task Metric}}$
&
$\underbrace{\mathrm{score}_{\mathrm{task}}}_{\color{evalcolor}\textbf{\scriptsize Task}}\;,\quad\underbrace{\mathrm{score}_{\mathrm{interp}}}_{\color{conceptcolor}\textbf{\scriptsize Interpretability}}$
\\[14pt]

\rowcolor{rowdark}
\stage{Attribute}
&
$\underbrace{\hat{\mathcal{A}}^{\mathrm{post\text{-}hoc}}(p_\theta, \mathbf{x}, y)}_{\color{warncolor}\textbf{\scriptsize Approximate\;(\S3.2)}}$
&
$\underbrace{\mathcal{A}^{\mathrm{input}}}_{\color{datacolor}\textbf{\scriptsize In-dist.}}\;,\quad\underbrace{\mathcal{A}^{\mathrm{concept}}}_{\color{conceptcolor}\textbf{\scriptsize Exact}}\;,\quad\underbrace{\mathcal{A}^{\mathrm{retrieval}}}_{\color{attrcolor}\textbf{\scriptsize Native}}$
\\[10pt]

\bottomrule
\end{tabular}

\caption{The standard and interpretable recipes, compared stage by stage (\cref{sec:interp-recipe}).
The interpretable recipe augments each stage to satisfy the conditions of
Definition~\ref{def:inherent-interp}.
Each attribution functional (bottom row, right) traces to a specific upstream
modification ({\color{okcolor}green arrows}):
the bottleneck architecture enables exact concept attribution and native
retrieval in the model's own concept-aligned geometry;
the trained absence baseline in the objective makes feature attribution
in-distribution.
The standard recipe supports only post-hoc attribution
({\color{warncolor}orange}), which does not guarantee the required conditions
(\cref{sec:posthoc}).}
\label{fig:recipes}
\end{figure*}

%% file: sections/data/main.tex
\section{Data}
\label{sec:data}
In this section, we describe the process of building large-scale concept-annotated pretraining, midtraining, and post-training corpora. As discussed in \cref{sec:recipe}, to satisfy the concept interpretability constraints, we need supervision: data annotated with concepts at a level of granularity to capture meaning. Towards this end, we present \textbf{Atlas}, an automated system for annotating language modeling corpora with human-understandable concepts at a fine-grained level. Using Atlas, we annotated a 1.5 trillion-token corpus spanning web text, scientific writing, code, and synthetic data with \nconceptsapprox concepts across science, technology, philosophy, medicine, law, etc.

\textbf{Overview.} First, we discuss why existing concept libraries are inadequate and state the desiderata that a suitable library must satisfy (\cref{sec:data-desiderata}). We then describe the Atlas pipeline: a three-stage process that moves from sampled documents to high-recall tags---short free-form words or phrases associated with local spans of text---(\cref{sec:data-stage1}), from tags to a canonical concept library (\cref{sec:data-stage2}), and from the library to a trained concept annotator that can label arbitrary text (\cref{sec:data-stage3}); each stage is validated under a common evaluation framework introduced at the start of \cref{sec:data-atlas}. We next describe the hierarchical taxonomy imposed on the library (\cref{sec:data-taxonomy}) and a human interpretability study validating that our concepts are human-meaningful rather than merely LLM-fluent (\cref{sec:data-human-eval}). Finally, we describe how the trained annotator is applied to the full training corpus (\cref{sec:data-processing}) and how the resulting annotations and embeddings are indexed for test-time training data attribution (\cref{sec:training-data-indexing}).

\subsection{Motivation: No human-interpretable concept library exists at scale}
\label{sec:data-desiderata}

Before building our own library, we surveyed existing approaches for large-scale concept extraction. Broadly, existing concept libraries fall into three categories.

\begin{itemize}
    \item \textbf{Word-based concept dictionaries}. \citet{luo2024pace} construct a 40,000-item concept dictionary by selecting the most frequent words from the Brown Corpus~\citep{francis1967computational} and prompting GPT-4~\citep{gpt4} to generate sentences illustrating each word. Single words, however, cannot capture the higher-level abstractions, multi-sentence topics, or domain-specific ideas that appear throughout pretraining datasets.
    
    \item \textbf{Activation-derived, unsupervised concepts}. These approaches extract concepts from open-weight model activations, for example, directions discovered via sparse autoencoders (SAEs) \citep{bricken2023towards}. Activation-derived concepts are not constrained to be human-meaningful;  a direction in activation space may be statistically salient without corresponding to anything a person would recognize as a coherent idea. There is also no guarantee that these directions will correspond to topics that are expressed in the corpus that the interpretable language model will be trained on~\citep{leask2025saes, korznikov2026sanity}.
    
    \item \textbf{Narrow domain libraries.} Some concept sets target specific tasks such as sentiment analysis or toxicity detection. For instance, \citet{cblm} define concepts using ChatGPT for SST2, Yelp Polarity, and AGNews, specifying categories like world, sport, business, and technology news for AGNews. While these libraries offer high-quality labels, they are far too narrow to supervise models trained on diverse corpora spanning web text, code, mathematics, and scientific writing.
\end{itemize}

\textbf{Desiderata for an ideal concept library.} We posit that a concept library suitable for supervising language model pretraining at scale must satisfy five properties. The library must be --
\begin{enumerate}
    \item \emph{Multi-scale}: cover both high-level themes (e.g., ``machine learning'') and fine-grained units (e.g., ``gradient clipping for recurrent networks'').
    \item \emph{Localizable}: be applicable to spans within a document, enabling sub-document-level supervision and control.
    \item \emph{Stylistically expressive}: capture attributes like document tone, formality, or even communicative intent, not restricted to semantic categories.
    \item \emph{Representative}: possess the breadth necessary to cover the true distribution of large-scale pretraining corpora across web text, code, mathematics, and scientific writing.
    \item \emph{Human-meaningful}: comprise concepts that people---given expertise appropriate to the topic---can understand and would want to use when assessing a language model's behavior.
\end{enumerate}
 
The last desideratum deserves emphasis. It is tempting to define ``human-meaningful'' as ``labeled by a human'' or ``labeled by an LLM in a way that sounds reasonable''. Neither of these is sufficient. It is possible for an LLM to produce a fluent label for even semantically incoherent document groupings. 
A human can rate a label as ``reasonable'' without checking whether they would have arrived at the same label independently. In \cref{sec:data-human-eval}, we describe a direct test that enables us to assess the correspondence between human-generated labels and LLM-generated labels.

\begin{figure}[t]
  \centering
  \includegraphics[trim={2cm 4cm 2cm 4cm}, clip, width=\textwidth]{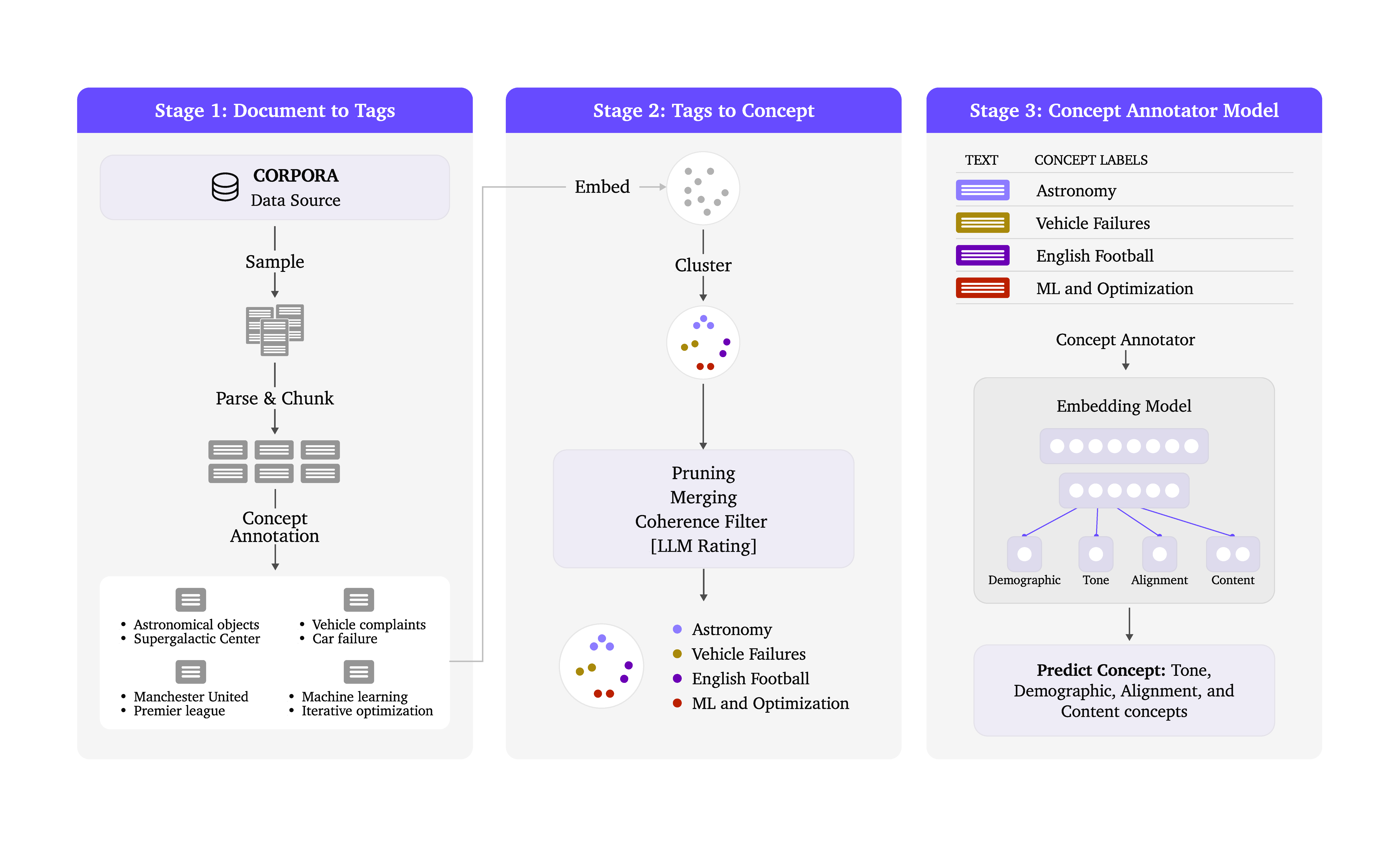}
  \caption{Overview of the \atlas three-stage annotation pipeline: documents are chunked and tagged (Stage~1), tags are clustered and canonicalized into a concept library (Stage~2), and a concept annotator model is trained for scalable text annotation (Stage~3).}
  \label{fig:data_pipeline}
\end{figure}

\subsection{Atlas: Documents to concepts}
\label{sec:data-atlas}

Atlas is a three-stage pipeline. Stage~1 samples documents, chunks them into local semantic units, and extracts a high-recall pool of free-form tags. Stage~2 clusters these tags, filters low-quality tag clusters, labels them, and deduplicates them into a canonical library of human-understandable concepts. Stage~3 trains a concept annotator model that can assign concepts from the library to arbitrary text, enabling annotation of the full training corpus. We describe each stage in turn.

\textbf{Evaluation overview.} The evaluation of Atlas must answer two questions. The first is an engineering question: does the pipeline produce high-quality annotations at each stage? The second is a scientific question: are the resulting concepts genuinely human-interpretable, or merely LLM-fluent? For the engineering question, we use a single framework throughout the pipeline and report results alongside each stage below. A sampled annotation---a raw tag from Stage~1, a canonical concept from Stage~2, or a predicted label from the Stage~3 annotator---is scored against its text chunk on a 1--5 relevance scale, by an LLM judge at scale and by human annotators on smaller matched samples. A score of 2 or higher counts as successful: the tag or concept is at least minimally present in the chunk. This threshold is intentionally permissive, reflecting the high-recall goal of the pipeline and the fact that minor tags can capture fine-grained contextual details without being the dominant topic. We therefore report both success rates and full score distributions, visualized as histograms of per-chunk and per-tag (or per-concept) average scores. On the matched samples, human ratings track the LLM judge but run roughly 0.4 points higher, so the LLM-judged scores reported below are, if anything, conservative.

The scientific question cannot be settled by an LLM judge: an LLM can rate an LLM-generated label as coherent without any guarantee that a human would recognize the same concept. We address it separately with a two-phase human study in \cref{sec:data-human-eval}, testing whether the concept names are human-interpretable rather than merely fluent.

\subsubsection{Stage~1: Corpus sampling, chunking, and documents to tags}
\label{sec:data-stage1}

Stage~1 constructs the high-recall tag pool from which the concept library is later derived. We begin with a representative sample of our pretraining and midtraining mixtures, split documents into local semantic chunks, prompt an annotator model to assign structured free-form tags to each chunk, and validate the resulting tags. At this stage, tags are not canonical concepts: they are short words or phrases that may be redundant, overlapping, or overly specific. This is deliberate. Stage~1 is optimized for recall; Stage~2 handles clustering, filtering, labeling, and deduplication.

Throughout this section, we distinguish between \emph{tags} and \emph{concepts}. 

\important{A \emph{tag} is a short free-form string assigned to a specific chunk of text during Stage~1 annotation.

A \emph{concept} is a canonical human-interpretable idea produced by clustering, filtering, labeling, and deduplicating many related tags. Each concept has a name, description, and supporting evidence from its associated tags.}

Tag assignment is an empirical output of a high-recall annotation process: a tag either was or was not assigned to a chunk, but unassigned tags are not treated as negatives. Concept assignment is a calibrated library-level prediction problem: once tags have been canonicalized into concepts, we evaluate whether a named concept is genuinely present in a chunk and whether humans recognize the concept from its supporting evidence. Concepts are necessarily less fine-grained than raw tags, and much of the calibration work in Atlas concerns whether a named concept should be assigned to particular chunks of text.

\textbf{Source mixture.} We sample from five major document categories: \emph{web text}, using a deduplicated version of DCLM from Zyda-2 \citep{li2024datacomp, tokpanov2024zyda}; \emph{general academic knowledge}, including peS2o \citep{pes2o}, arXiv \citep{weber2024redpajama}, and Wikipedia and Wikibooks (from Dolma 1.7 \citep{soldaini2024dolma}); \emph{mathematics}, including Dolmino-math, GSM8K \citep{cobbe2021training}, OpenWebMath \citep{paster2024openwebmath}, and Algebraic Stack \citep{azerbayev2023llemma}; \emph{code}, using StarCoder \citep{starcoder}; and \emph{question--answer exchanges}, using FLAN v2 \citep{longpre2023flan}. In total, the sample contains 6.6 million documents balanced across these categories. This mixture was chosen not as a benchmark distribution, but as a substrate for concept extraction: it spans multiple writing styles, knowledge domains, and levels of conceptual density.

\textbf{Chunking.} We annotate at the chunk level rather than the document level. Long documents often contain several unrelated ideas, and whole-document annotation tends to collapse them into coarse summaries. A chunk is a short, semantically coherent span, created by detecting sentence boundaries with the BlingFire sentence splitter \citep{microsoft_blingfire_2019} and concatenating consecutive sentences until a domain-specific token threshold is reached. We use a threshold of 150 tokens for web text and general documents and 256 tokens for mathematics and code, where a single mathematical or algorithmic idea often requires more context. Sentences exceeding the threshold are retained as single chunks; extremely long sentences above 50,000 tokens are dropped as malformed or uninformative for local annotation. This process yields 44 million chunks from the 6.6 million sampled documents. \important{Chunking at this granularity is essential: it makes tags local to the semantic units they describe, rather than global summaries of entire documents.}

\textbf{Domain-aware annotation schemas.} Different domains express conceptual structure differently. A Wikipedia article, a mathematical proof, a Stack Exchange question, and a block of Python code call for different annotation fields. We therefore designed domain-specific structured tag schemas, each containing roughly 4 to 6 fields tailored to the content type. Each field elicits a hierarchy of tags from broad to narrow, and together the fields produce on average 10--15 tags per text chunk.

\input{sections/data/chunk_tag_example_webtext}

For each domain, the schema separates several complementary views of a chunk: what the chunk is about, what role it plays, and how it is expressed. For web text, fields include topic, communicative purpose, tone or style, and secondary entities or events; \cref{tab:chunk-tag-example-webtext} shows an example web text document split into chunks with its domain-specific structured tags. For mathematics, fields capture the mathematical object or topic, problem type, solution or proof technique, notation conventions, and difficulty. For code, fields capture programming language, design pattern, algorithmic structure, software-engineering role, and documentation style. The schemas are deliberately overcomplete: a single chunk may receive several overlapping tags, because Stage~1 is optimized for recall before Stage~2 clusters the tag space into canonical concepts.

\textbf{Annotator model selection.} We evaluated several open-weight models for structured annotation, including Phi-family models \citep{abdin2024phi}, \texttt{Qwen2.5-7B} \citep{qwen2025qwen25technicalreport}, and \texttt{Mistral-Small-3.1-24B-Instruct} \citep{mistralsmall24b}. The choice of annotator model is constrained by a practical requirement: reliability at scale matters more than raw capability. Even a 1\% schema deviation rate across 44 million chunks produces nearly half a million unusable annotations.

\begin{figure}[t]
  \centering
  \input{sections/data/topk_frequent_tags}
  \caption{The 50 most frequent raw tags produced by Stage~1. These tags illustrate the high-recall, redundant, non-canonical tag space before Stage~2 clustering and deduplication.}
  \label{fig:top50-frequent-tags}
\end{figure}
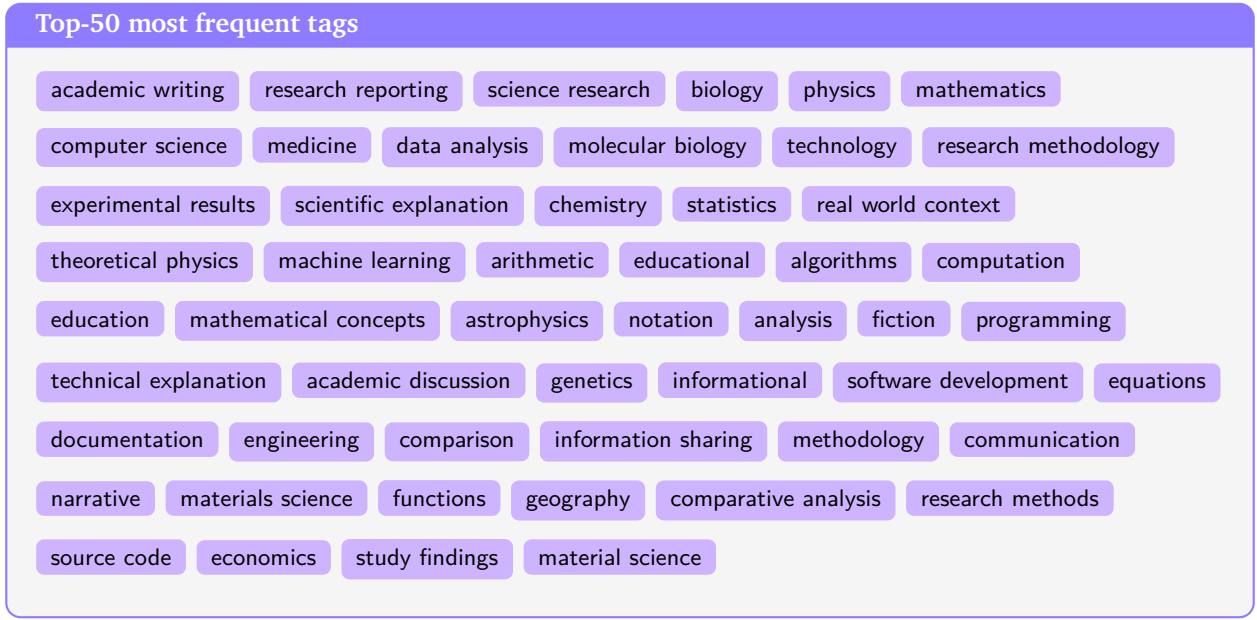

We found that Phi-family models could output consistently structured annotations but suffered from repetition collapse and degenerate loops, making them unsuitable for long-running annotation jobs. \texttt{Qwen2.5-7B} performed better semantically but less predictably syntactically, often producing the wrong number of fields or responses that were difficult to parse. \texttt{Mistral-Small-3.1-24B-Instruct} consistently adhered to the structured format, avoided repetition collapse, and maintained stable behavior across millions of prompts. Mistral was the smallest model that satisfied our formatting and reliability constraints.

\begin{figure}[htbp]
    \centering
    \includegraphics[width=\textwidth]{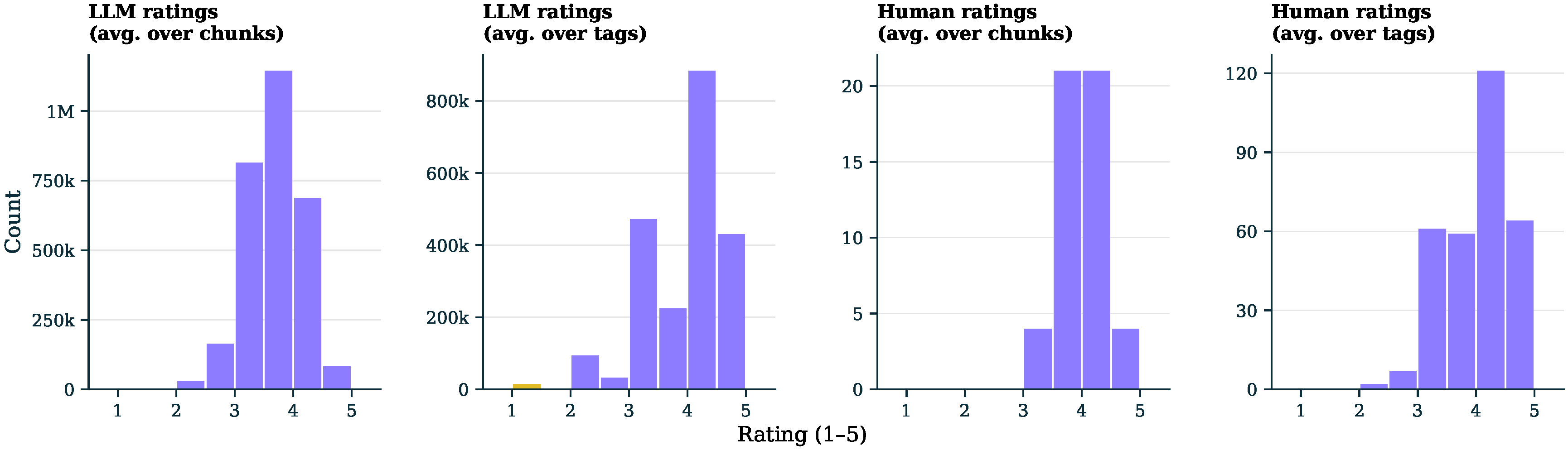}
    \caption{Stage~1 tag validation scores. Distributions of average tag-relevance ratings on a 1--5 scale: per-chunk averages (left column; each chunk's tag ratings averaged) and per-tag averages (right column; each unique tag's ratings averaged over the chunks it appears in), as rated by the LLM judge over 2.93M sampled chunks (top row) and by three human annotators on a 50-chunk sample (bottom row).}
    \label{fig:chunk_tag_eval}
\end{figure}

\textbf{Output.} Stage~1 produced nearly 44 million annotated chunks, from which we extracted approximately 500 million tags. After tag deduplication, 14 million unique tags remained, of which approximately 1 million appeared more than 15 times. The most frequent raw tags, shown in \cref{fig:top50-frequent-tags}, provide a sanity check on this high-recall regime: the tag pool captures broad recurring content and stylistic patterns, but remains redundant and non-canonical before Stage~2 clustering. This raw tag space is intentionally redundant and noisy. \important{At this stage of the pipeline, missing a relevant concept is more damaging than producing overlapping or synonymous tags: synonymy and over-specificity can be removed by clustering, while absent concepts cannot be recovered downstream.} The purpose of Stage~1 is therefore to cast a wide semantic net. Stage~2 consolidates this tag space into a large concept library.

Using the evaluation framework described at the start of this section, an LLM judge scored 21.3 million (tag, chunk) pairs over a sample of 2.93 million chunks (2.15 million unique tags), with human annotators rating a 50-chunk subsample; the judging prompt is given in \cref{app:data}. Individual tag ratings average 3.62, and 97.5\% score at least 2. \important{Across all domains, nearly every tag is at least minimally present in its chunk: per-chunk average scores center near 3.6, with two-thirds of chunks falling between 3.2 and 4.0 and 99.9\% averaging at least 2 (\cref{fig:chunk_tag_eval}).}

\subsubsection{Stage~2: Tags to concepts}
\label{sec:data-stage2}

Stage~2 transforms nearly 14 million noisy, free-form tags into a coherent library of \nconceptsapprox human-understandable concepts. The process embeds tags into a semantically meaningful space, clusters them, filters low-coherence clusters, labels each surviving cluster, and merges near-duplicate concepts through graph-based deduplication. We describe each step in turn.

\textbf{Tag normalization and embedding.} Raw tags produced by an LLM from Stage~1 are highly variable: minor formatting differences---hyphens, slashes, whitespace, punctuation artifacts---can split semantically identical tags into separate strings. We normalize all tags into a standardized form before embedding, \eg \texttt{astronomical-objects} $\rightarrow$ \texttt{astronomical objects}, \texttt{climate-change-adaptation} $\rightarrow$ \texttt{climate change adaptation}. Each unique tag is then embedded into an $n$-dimensional ($n=768$) space using the \texttt{all-mpnet-base-v2} sentence embedding model \citep{allmpnetbasev2}. This choice is empirical: in a comparison against \texttt{Qwen3-Embedding-0.6B} \citep{zhang2025qwen3}, the former produced slightly higher-quality clusters, as measured by standard quantitative clustering metrics---Silhouette score \citep{rousseeuw1987silhouettes}, Davies--Bouldin index \citep{davies1979cluster}---together with the LLM-judge cluster coherence evaluation described below. In total, this step produces nearly 14 million vectors, one per unique tag.

\textbf{Clustering.} With all tags embedded in a common semantic embedding space, we cluster them into groups representing candidate concepts; for instance, \textit{missing pet}, \textit{missing pet incident}, \textit{missing dog} should fall in a single cluster. We use ${k}$-means, implemented in the FAISS library \citep{johnson2019billion} for GPU-accelerated efficiency over tens of millions of vectors.

\textbf{The number of underlying semantic clusters is unknown}, so we sweep a wide range of cluster counts: $k \in \{100, 500, 1\text{k}, 10\text{k}, 20\text{k}, 30\text{k}, 50\text{k}, 80\text{k}, 100\text{k}\}$. Small $k$ produces large, diverse clusters that conflate unrelated concepts; large $k$ produces tight clusters but risks fragmenting genuinely related concepts or introducing clusters based on noise. We evaluate each $k$ with the Silhouette score, which rewards clusters that are internally cohesive and well separated, the Davies--Bouldin index, which penalizes clusters whose internal scatter is large relative to their separation, alongside the LLM-based cluster coherence evaluation described next.

\begin{figure}[htbp]
  \centering
  \includegraphics[width=0.9\textwidth]{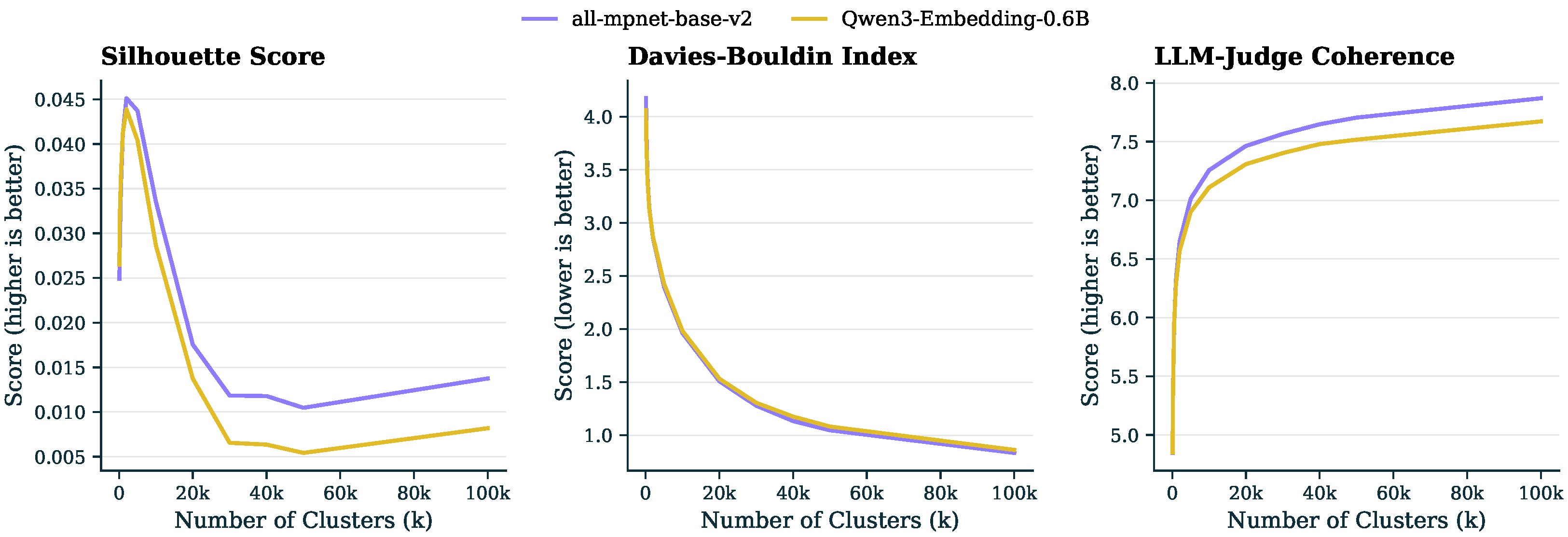}
  \caption{\looseness=1 Comparison of \texttt{all-mpnet-base-v2} and \texttt{Qwen3-Embedding-0.6B} embeddings for clustering $\sim$14M LLM-generated tag annotations via the $k$-means clustering algorithm. We evaluate Silhouette score (left plot), Davies--Bouldin index (middle plot), and coherence score (right plot) on randomly sampled cluster members. As $k$ increases, \texttt{all-mpnet-base-v2} consistently yields more coherent clusters than \texttt{Qwen3-Embedding-0.6B}.}
  \label{fig:tag_clustering_metrics}
\end{figure}

\textbf{LLM-based cluster coherence evaluation.} The two standard metrics capture broad trends across large changes in $k$ but do not reliably distinguish nearby values, so we complement them with a direct semantic coherence evaluation of every cluster. For each cluster, we sample three strata of tags---core (closest to the centroid), random (uniform), and edge (farthest from the centroid)---group them into sets of ten and query an LLM, \texttt{Mistral-Small-3.1-24B-Instruct}, to score the semantic coherence of each set on a 1--10 scale, yielding a coherence score per stratum. \important{Coherence improves steadily with $k$ and plateaus around $k=40{,}000$--$80{,}000$ as shown in \cref{fig:tag_clustering_metrics}; beyond this range, clustering computational cost rises while semantic gains diminish, as larger $k$ begins to split coherent groups across multiple clusters. We therefore set the initial number of clusters to $k=80{,}000$.}

\textbf{Quality filtering.} Not every cluster corresponds to a coherent concept; some arise from incidental lexical overlap rather than shared meaning. We retain only clusters that clear strict per-stratum coherence thresholds---core $\geq 9$, random $\geq 8$, and edge $\geq 7$. Of the 80,000 clusters, approximately 17,000 fail these criteria and are removed, leaving roughly 63,000 high-quality clusters.

\textbf{Cluster labeling.} Each surviving cluster is converted into a human-interpretable concept. We sample 50--100 representative tags per cluster, weighted by tag frequency, and prompt an LLM, \texttt{Mistral-Small-3.1-24B-Instruct}, to generate a concise label (1--6 words) and a one-sentence rich description; the prompt is given in \cref{app:data}. These labels provide a clean, human-friendly interpretation of what are otherwise dense numerical clusters of tags. \important{The LLM \emph{labels} structure that emerged from clustering millions of human-comprehensible tags; it does not fabricate concepts from nothing.} The tags were produced by annotating real text, the clusters capture statistical regularities in those tags, and the LLM assigns a name to each regularity. Whether the resulting names are genuinely human-interpretable is an empirical question we address in \cref{sec:data-human-eval}.

\textbf{Graph-based cluster deduplication.} The labeled concepts, now numbering $62{,}000$, still contain substantial redundancy: multiple clusters may express the same underlying idea, differing only in granularity or phrasing. We address this redundancy by iterative graph-based concept merging.

\begin{figure}[htbp!]
\centering
\includegraphics[width=\textwidth]{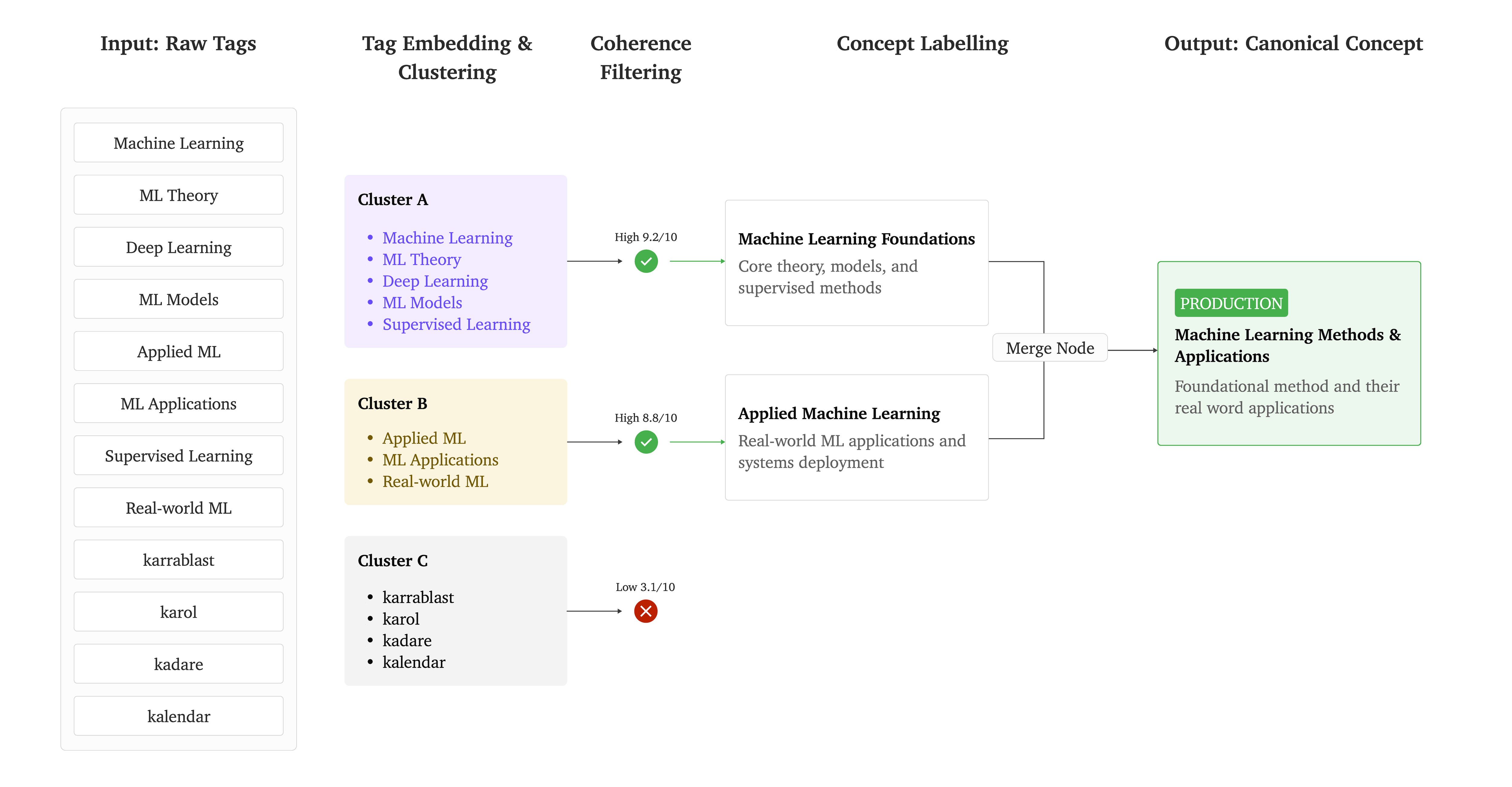}
\caption{\looseness=-1 Stage-2 transformation of noisy LLM tags into a canonical concept:
raw tags are embedded and clustered, incoherent clusters are filtered out, each
surviving cluster is labeled into a single concept (name with italic description),
and semantically adjacent concepts are merged by cosine similarity into one
canonical entry.}
\label{fig:tag_concept_pipeline}
\end{figure}

We embed each concept's label and description with \texttt{Qwen3-Embedding-0.6B} into a semantic $n$-dimensional vector space, then connect each concept to its $m$-nearest neighbors, retaining edges with cosine similarity above a certain threshold $\tau$. This produces an undirected similarity graph whose nodes are concepts and whose edges connect near-duplicates. We partition the graph with Louvain-community detection \citep{blondel2008fast} to identify groups of related concepts, treat each community as a candidate merge set, and prompt an LLM to regenerate a single unified label and description for each set. A single pass reduces the library to approximately 39,000 concepts. We then re-embed all concepts and repeat---rebuilding the graph and re-running community detection at each iteration---halting after two to three iterations once an iteration merges only a negligible number of concepts. This leaves \nconceptsapprox concepts. We chose $m=20$, $\tau=0.95$ for first iteration, reduced $\tau=0.9$ for second pass, and then $\tau=0.85$ for third pass based on empirical evidence.
 
\begin{figure}[t]
    \centering
    \includegraphics[trim={2cm 4cm 2cm 4cm}, clip, width=\textwidth]{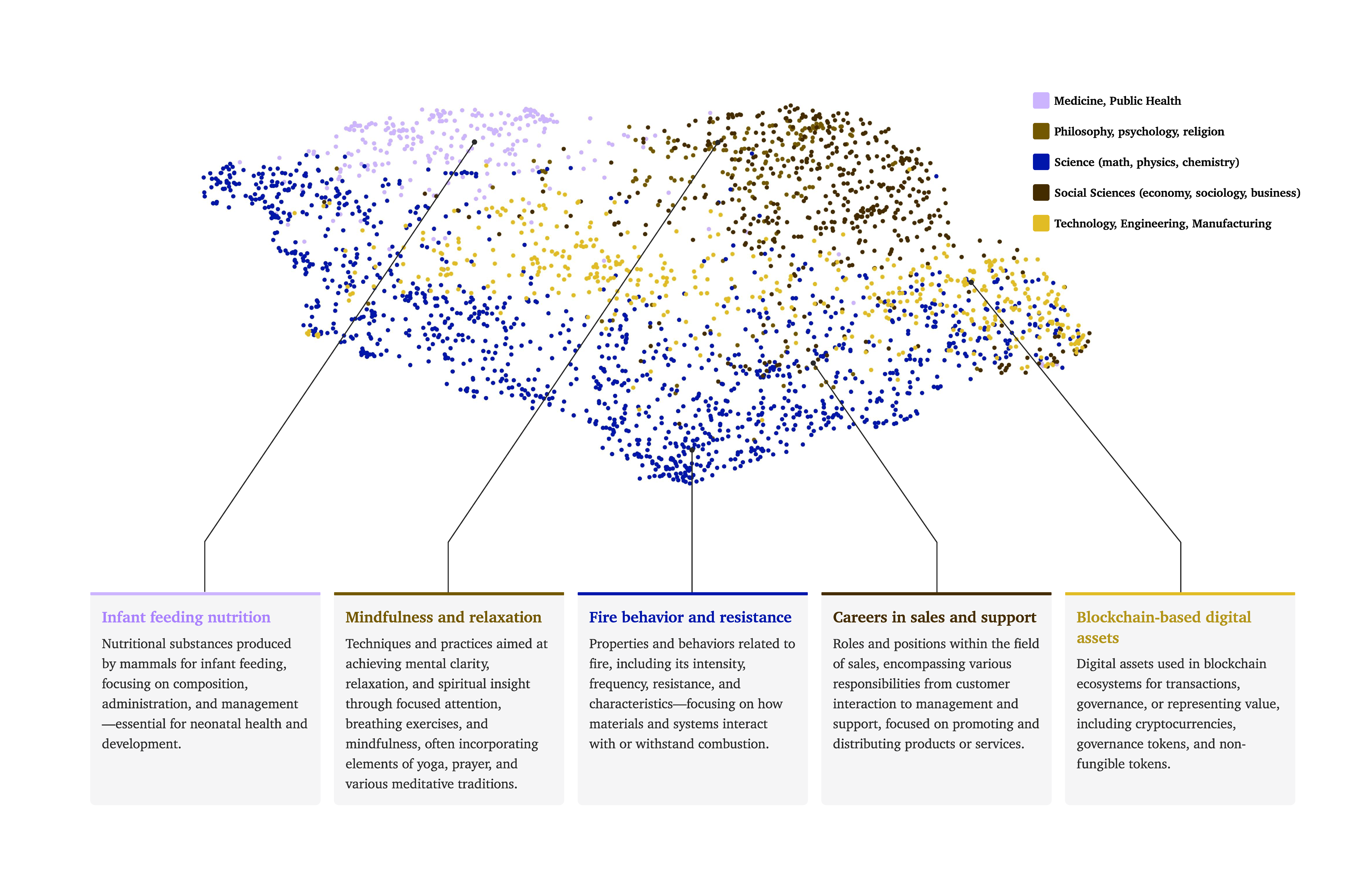}
    \caption{\looseness=-1 UMAP projection of a subsample of concept embeddings across five taxonomy groups. The cards showcase representative concepts from distinct regions of the embedding space to illustrate local semantic clusters.}
    \label{fig:concept_umap}
\end{figure}

\textbf{Output.} \important{The final canonical concept library contains \nconcepts concepts spanning science, technology, philosophy, medicine, law, and other domains.} Some conceptual overlap and hierarchy are unavoidable---real-world knowledge is not cleanly partitioned---but the deduplicated library strikes a practical balance between granularity and clarity. We visualize the concept embeddings in \cref{fig:concept_umap} using UMAP \citep{mcinnes2018umap}, a scalable nonlinear dimension-reduction method for visualization and manifold learning.

\begin{figure}[htbp]
     \centering
     \includegraphics[width=\textwidth]{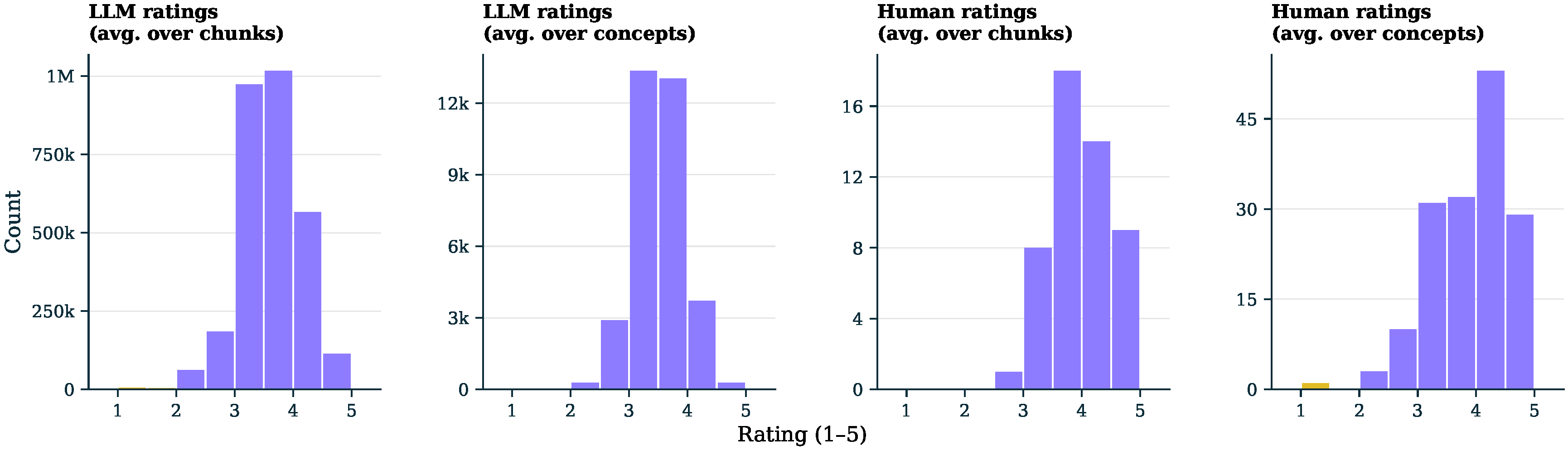}
     \caption{Stage~2 concept validation scores. Distributions of average concept-relevance ratings on a 1--5 scale: per-chunk averages (left column) and per-concept averages (right column; each concept's ratings averaged over the chunks it is assigned to), as rated by the LLM judge over 2.93M chunks (top row) and by human annotators on a 50-chunk sample (bottom row).}
     \label{fig:concept_relevance_ratings}
\end{figure}

\textbf{Validation.} Using the same framework, an LLM judge scored the concepts associated with each of the same 2.93 million chunks via their Stage~1 tags---12.2 million (concept, chunk) pairs covering nearly the entire library---with human annotators again rating a 50-chunk subsample (\cref{fig:concept_relevance_ratings}). Individual ratings average 3.50, with 98.0\% scoring at least 2; per-chunk averages center near 3.5, with two-thirds of chunks between 3.2 and 4.0 and 99.7\% averaging at least 2. \important{Per-concept average scores are notably uniform across the library (mean 3.51, standard deviation 0.41): concept quality is consistent rather than driven by a head of frequent, well-represented concepts.}

\subsubsection{Stage~3: Concept annotator model}
\label{sec:data-stage3}

\looseness=-1The outcome of Stages~1 and 2 is a large canonical library of content concepts, defined over the sampled corpus. Stage~3 trains a scalable annotator that maps an arbitrary input chunk to a set of target labels, allowing us to annotate the full 1.5 trillion-token pretraining corpus. The annotator predicts over a combined target space consisting of the Atlas-derived content concepts together with several lower-cardinality auxiliary domains: tone, demographic-reference, and alignment-relevant labels.

\textbf{Training targets.}
The Stage~3 training targets come from two sources. The first source is the content-concept library produced by Stages~1 and 2. For each chunk in the Stage~1 annotated sample, raw tags are mapped through the Stage~2 clustering and deduplication pipeline to the final content-concept IDs. Thus, if a tag assigned to a chunk belongs to a cluster that is ultimately canonicalized as concept \(c\), then \(c\) is inherited as a positive content label for that chunk. These inherited labels provide positive-only supervision: concepts not assigned to a chunk are treated as unlabeled rather than verified negatives.

Rather than train directly on all 44 million Stage~1 annotated chunks, we construct a long-tail-enriched training reservoir. The reservoir is selected to improve coverage of rare content concepts by amassing a minimum number of examples per concept. This procedure yields roughly 2.9 million chunks. The resulting reservoir is still highly imbalanced---frequent concepts can appear tens of thousands of times---but it gives substantially better coverage of the tail than uniform sampling from the full Stage~1 sample. We train the annotator on approximately 2.3 million reservoir chunks and reserve approximately 597{,}000 chunks for held-out evaluation.

The second source of supervision consists of fixed candidate vocabularies for auxiliary lower-cardinality domains. These domains are not produced by the Stage~2 clustering procedure. Instead, we define finite candidate lists for tone, demographic references, and alignment-relevant categories, and annotate chunks using the same LLM-based annotation infrastructure with separate lists for each domain. Given a chunk and a candidate list, the annotator is asked which categories are relevant to the text. Selected categories are used as positive labels for the corresponding auxiliary heads. The final prediction vocabulary contains \nconcepts labels in total, dominated by approximately \(33{,}606\) content concepts, together with roughly \(80\) tone labels, \(38\) demographic-reference labels, and \(8\) alignment-relevant labels.

\textbf{Design constraints.} A few key requirements shaped the annotator model design. 
\begin{itemize}
    \item First, we want a single model that can handle both the high-cardinality Atlas-derived content concept library and the lower-cardinality auxiliary domains described above, rather than maintaining separate classifiers with separate encoders, thresholds, and inference logic.
    \item Second, in Stage~2, we found that a simple $k$-nearest neighbor classifier over the concept embeddings, performs surprisingly well on content-label prediction, especially for frequent, well-represented concepts. We want a flexible architecture and loss function family that retains this simplicity, but incorporates learned signals for rare classes.
    \item Third, our training data contains only positive labels for each concept domain since we did not do negative annotation: each chunk is annotated with the concepts it \emph{has}, never with the concepts it explicitly does not have. This positive-unlabeled (PU) supervision regime requires careful treatment of loss functions and evaluation metrics, so we will restrict ourselves to settings that make it flexible to incorporate such design constraints \citep{denis1998pac, de1999positive, denis2005learning, kiryo2017positive}.
\end{itemize}
 
\textbf{Architecture.} The annotator uses the \texttt{Qwen3-Embedding-0.6B} model as a shared encoder, producing a pooled embedding for each input chunk. This embedding feeds into a lightweight MLP trunk (LayerNorm, ReLU, Dropout, projection to 1024 dimensions) shared by the auxiliary prediction heads. The content head follows a different path: it computes dot products between the encoder output and a matrix of content-concept embeddings, one per Atlas-derived content concept. This design preserves the KNN-like signal of direct embedding similarity while allowing learned improvements. The tone, demographic-reference, and alignment-relevant heads use simpler linear classifiers, which suffice for their lower-cardinality fixed vocabularies.
 
\textbf{Loss.} We combine two loss functions. The first is masked binary cross-entropy, applied only to positions where targets are non-zero, which handles the positive-only supervision. The second is a non-negative PU loss, a PU-compatible objective that penalizes overconfident predictions on unlabeled classes and stabilizes learning on long-tail distributions. Rare concepts are further supported by a rarity-weighted sampling scheme that boosts underrepresented labels during training.
 
\textbf{Evaluation.} Standard classification metrics---precision, recall, PR-AUC---are problematic under PU learning, because any unlabeled example predicted as positive is counted as a false positive even if the prediction is correct. This systematically deflates precision. We track these metrics for monitoring relative improvement across training runs, but we do not rely on them for absolute quality judgments. Instead, we rely on the LLM and human evaluation framework described at the start of the section.

\begin{figure}[t]
    \centering
    \includegraphics[width=0.8\textwidth]{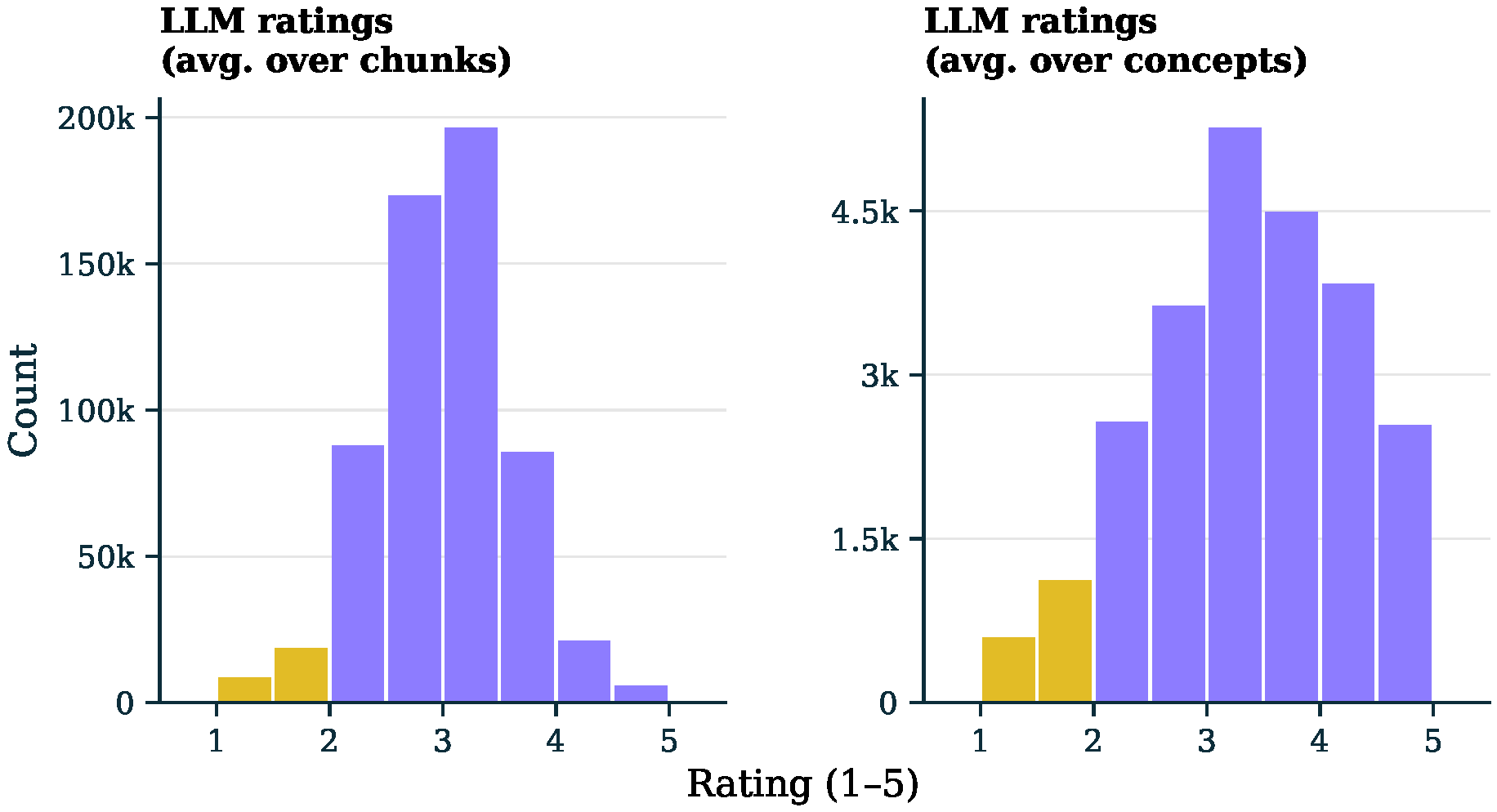}
    \caption{Stage~3 annotator validation on held-out chunks. Distributions of average predicted-concept relevance ratings from the LLM judge: per-chunk averages (left) and per-concept averages (right).}
    \label{fig:predicted_concept_relevance_ratings}
\end{figure}

The annotator is trained on 2.3 million of the sampled chunks; we evaluate its predictions on the remaining 597{,}000 held-out chunks, scoring 4.93 million predicted (concept, chunk) pairs with the LLM judge (\cref{fig:predicted_concept_relevance_ratings}). Predicted-concept ratings average 2.94, with per-chunk averages clustering between 2.5 and 3.5 (median 3.0); 95.5\% of per-chunk averages are at least 2, and per-concept averages center near 3.3. \important{The annotator thus assigns acceptable concept sets to the overwhelming majority of held-out chunks, though its predictions score lower and are more dispersed than original tag and concept ratings---plausibly reflecting the positive-unlabeled training regime and the per-domain prediction caps.}

\subsection{Concept taxonomy}
\label{sec:data-taxonomy}

\begin{figure}[!ht]
  \centering
  \includegraphics[width=\textwidth]{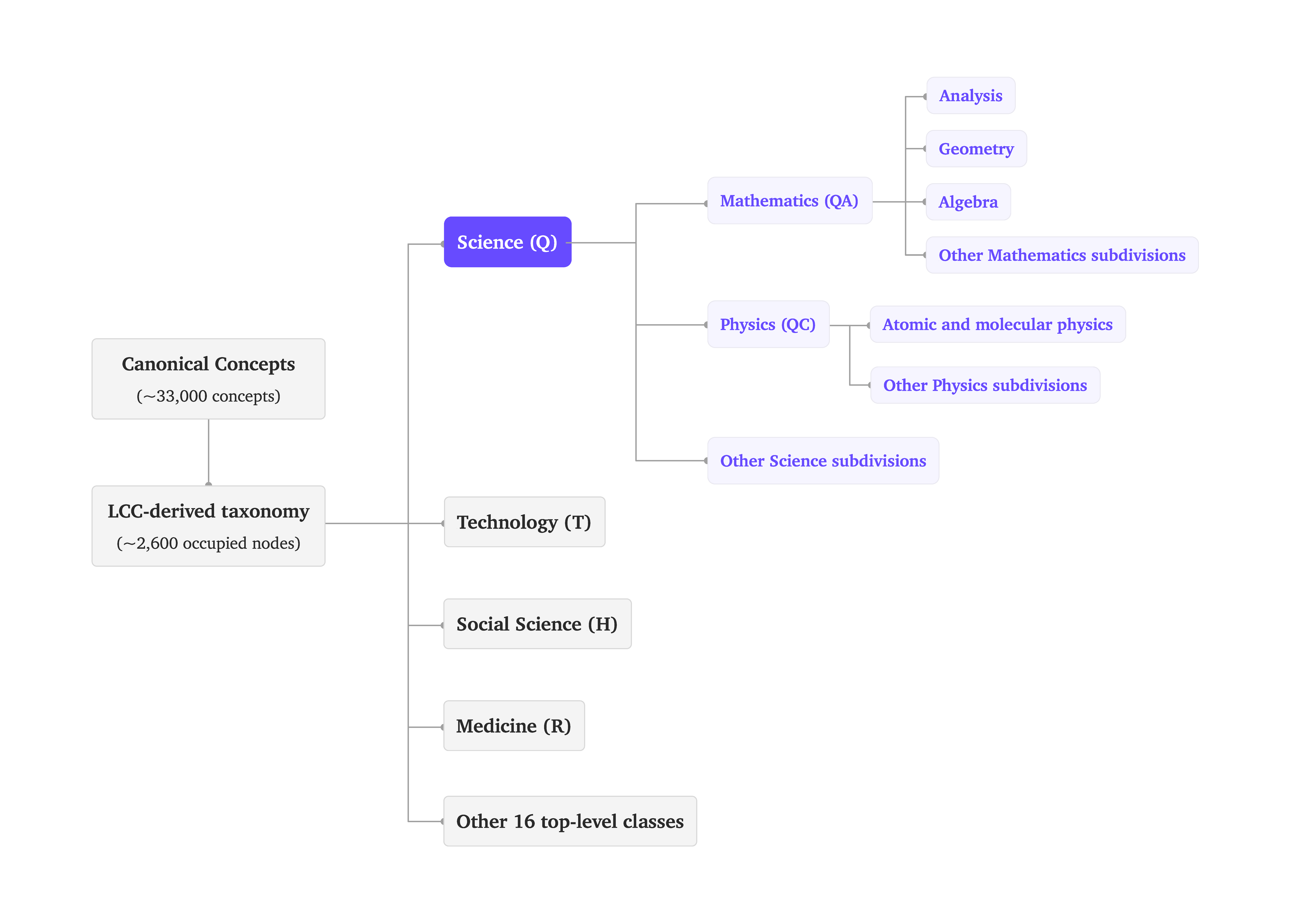}
  \caption{Schematic of the LCC-derived taxonomy used to organize the Atlas concept library. The \nconcepts canonical concepts are mapped onto approximately 2,600 occupied taxonomy nodes. The figure shows the root structure and expands the Science (Q) branch to illustrate how top-level classes decompose into more specific areas such as Mathematics (QA), Physics (QC), and their subdivisions; other branches are collapsed for readability.}
  \label{fig:taxonomy_tree_illustration}
\end{figure}

We organized the \nconcepts canonical concepts into a hierarchical taxonomy derived from the Library of Congress Classification (LCC), an established bibliographic classification system maintained by the Library of Congress \citep{loc_lcc}. We used the \texttt{agentlans/library-classification-systems} dataset \citep{tseng2024libraryclassificationsystems}, which provides a machine-readable outline of LCC entries with parent--child links. The resulting taxonomy maps the concepts onto roughly 2,600 occupied nodes within the full 6,517-node LCC outline, with populated paths reaching depth 9 (\cref{fig:taxonomy_tree_illustration}).
 
\textbf{Concept distribution across taxonomy.} Science (Q) dominates with 38\% of concepts, followed by Technology (T) at 15\%, Social Sciences (H) at 15\%, and Medicine (R) at 8\%. All 20 root branches are represented to varying degrees. Within the Science branch, mathematics and physics subcategories are particularly prominent: Mathematics (QA) subdivisions such as Analysis, Geometry, and Algebra account for over $3{,}700$ concepts combined, while Physics (QC) areas such as Atomic and Molecular Physics contribute over 660 concepts.

\begin{figure}[t]
  \centering
  \includegraphics[width=\textwidth]{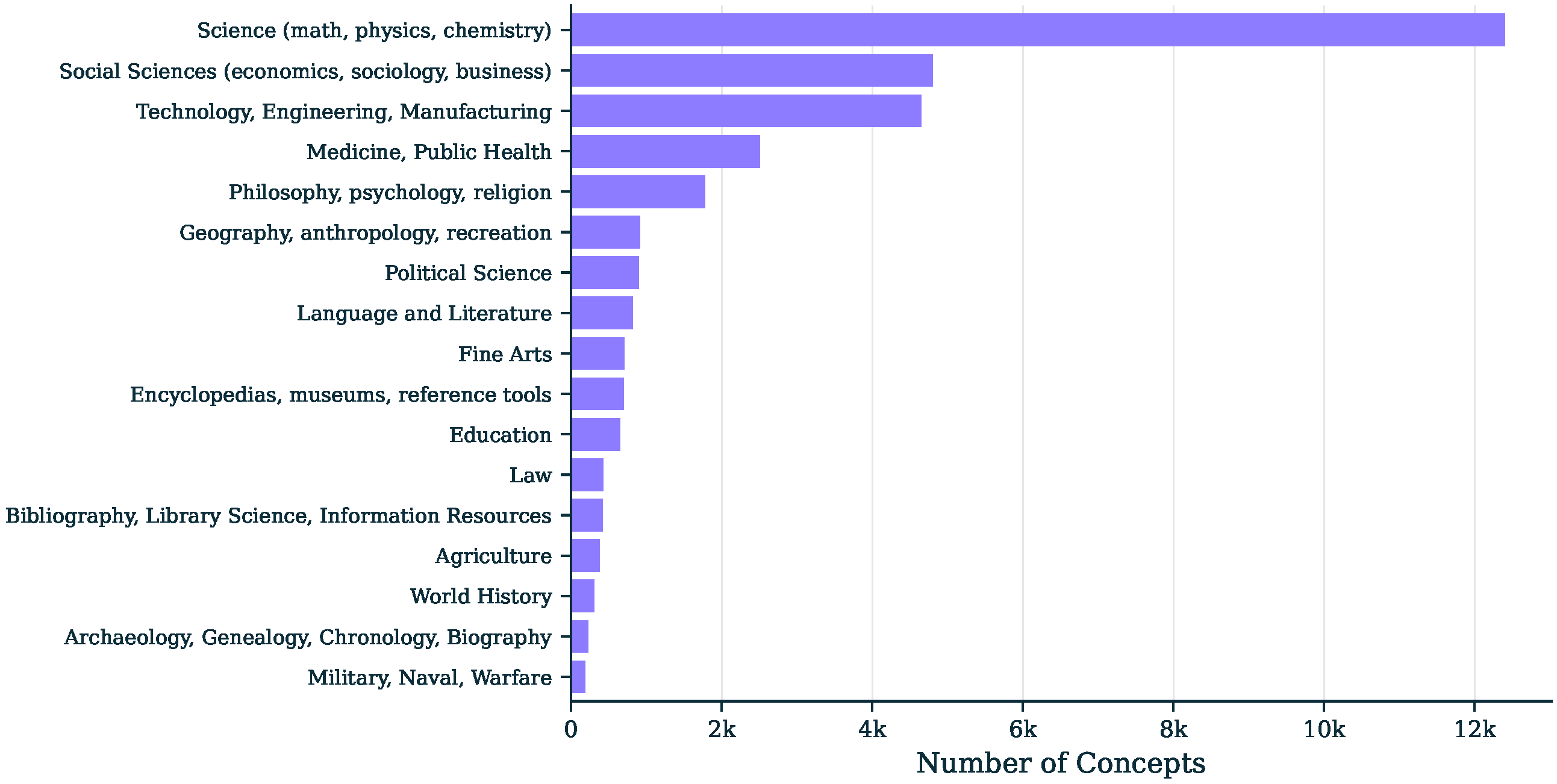}
  \caption{Distribution of \nconceptsapprox concepts across the top-level LCC classes and notable subclasses. Science (Q) accounts for the largest share, followed by Technology (T), Social Sciences (H), and Medicine (R); all root classes are represented to varying degrees. The distribution mirrors the composition of the pretraining corpus rather than a curatorial choice.}
  \label{fig:taxonomy_concept_distribution}
\end{figure}

The distribution in \cref{fig:taxonomy_concept_distribution} reflects the composition of the underlying pretraining corpus rather than an editorial choice about which domains matter. The taxonomy provides a structured framework for analyzing concept coverage across knowledge domains and, as we describe below, enables stratified evaluation of concept quality.

\subsection{Human interpretability of the concept library}
 \label{sec:data-human-eval}
 
The validations above show that Atlas is internally consistent: tags are relevant to chunks, clusters are semantically coherent, and the trained annotator assigns concepts to held-out text that an LLM judge scores as present. This does not, by itself, show that the final concept names are human-interpretable. Since the Stage~2 concepts are labeled by an LLM, there is a specific failure mode we must rule out: the LLM might assign fluent names to statistical patterns that do not correspond to concepts humans recognize. For example, a cluster whose tags co-occur for lexical rather than semantic reasons can still receive a confident, plausible-sounding name---and an LLM judge shown the same evidence may rate that name as coherent---without any person being able to independently arrive at, or even recognize, the underlying concept.

We therefore evaluate two claims. First, the evidence associated with a concept must itself contain recoverable semantic structure. Second, the Atlas label---generated by an LLM in Stage~2---must name that structure at least as well as labels generated independently by humans.

\textbf{Lifted-word evidence.} For each concept $c$, we construct a list of \emph{lifted words}: lemmatized words appearing in the text chunks assigned concept $c$, ranked by how strongly they are associated with $c$ relative to their background frequency across all chunks,
 \[
     \operatorname{lift}(w,c) = \frac{P(w \mid c)}{P(w)} ,
 \]
 with a minimum-support filter to remove idiosyncratic rare words. Lifted words provide a readable, corpus-level form of concept evidence. They also instantiate the more general evaluation problem used elsewhere in the paper: given only the words or tokens statistically associated with a concept, can a labeler recover a human-meaningful name?
 
\textbf{Two-phase human study.}

We ran a two-phase human study. In Phase~1, human annotators saw only the lifted words for a concept. Each annotator wrote a name or short phrase for the lifted-word list and rated, on a $1$--$5$ coherence scale---from $1$ (the words form no recognizable concept) to $5$ (the words clearly correspond to a single recognizable concept)---whether the words formed a recognizable concept at all. This generation task is deliberately stricter than asking humans to approve a provided label: annotators can mark the evidence as noisy rather than being forced to accept a fluent name.

In Phase~2, annotators performed blind comparative scoring. For each concept, we assembled a candidate set containing the Atlas label generated in Stage~2, two human labels written for the same concept by other annotators in Phase~1, a \emph{taxonomy distractor} from a nearby but distinct concept in the taxonomy, and an \emph{embedding distractor} from a different concept with a nearby label embedding. In a small number of cases where a second human label was unavailable, we substituted a deliberately generic filler label as a floor control: a label expected to fit poorly, confirming that raters used the low end of the scale. Annotators rated how well each candidate name fit the same lifted-word list, again on a $1$--$5$ scale. Candidate order was randomized, annotators were blind to label provenance, and no annotator scored a label they had written.

\begin{figure}[t]
\centering
\includegraphics[width=0.8\textwidth]{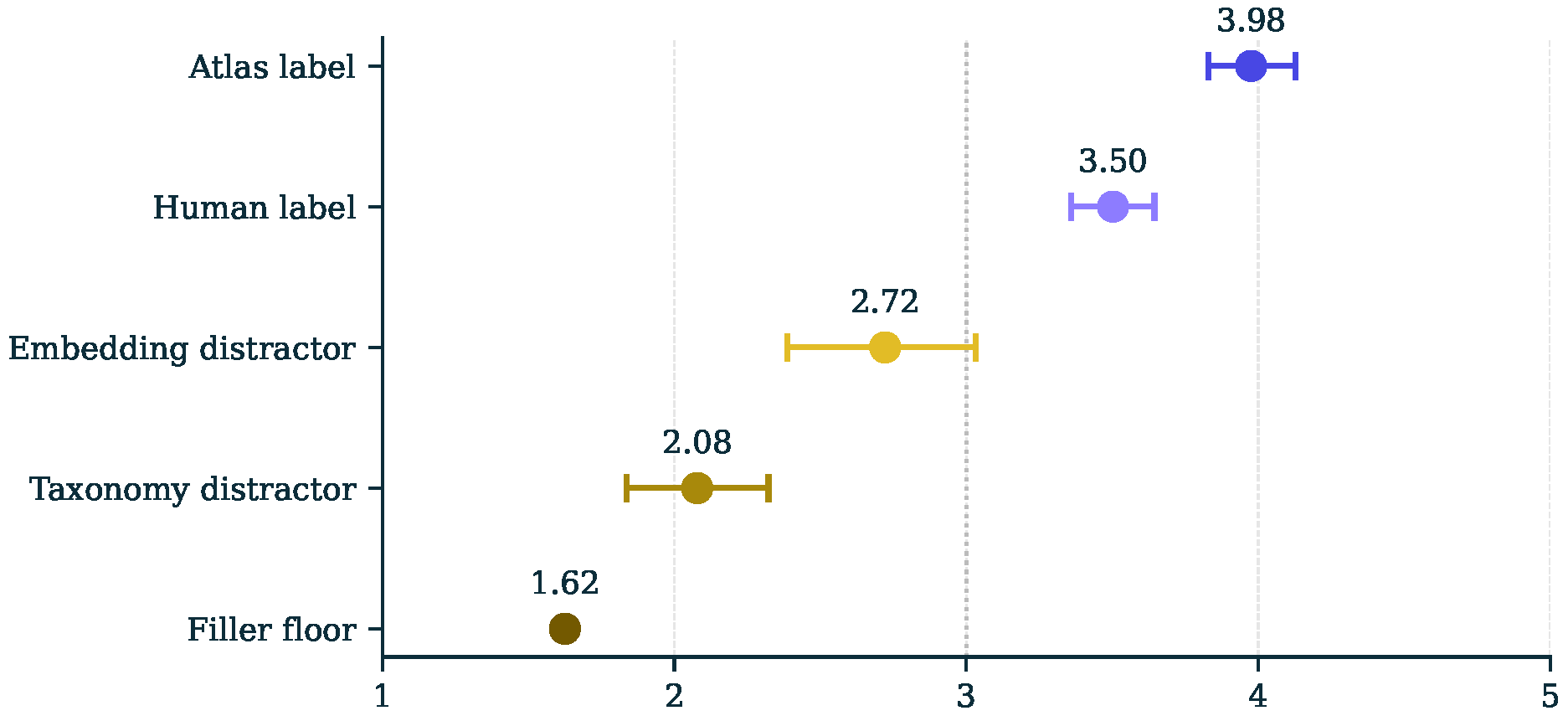}
\caption{Phase~2 human-evaluation fit scores by candidate type. Human annotators rated, on a $1$--$5$ scale, how well each candidate name fit the same lifted-word evidence for a concept. Points show mean fit scores and horizontal bars show $95\%$ confidence intervals; the dotted vertical line marks the neutral midpoint of the scale. Atlas labels score comparably to or above independently generated human labels, and both are clearly separated from embedding, taxonomy, and filler distractors, indicating that raters were not merely assigning high scores to any plausible label.}
\label{fig:human-eval-phase2-fit}
\end{figure}

\textbf{Do lifted words contain recognizable structure?}
Yes, although not in every case. We sampled $100$ concepts stratified by top-level taxonomy branch: ten concepts from each of the nine largest branches and ten from the aggregated remainder. Phase~1 collected $303$ named-concept responses from $9$ human annotators, with a median of $3$ annotators per concept. Human-written names averaged $4.1$ words.
 
The mean Phase~1 coherence score was $3.52$. Annotators judged $55\% $ of responses to form a recognizable concept ($\geq 4$), rated $27\%$ as borderline ($=3$), and flagged $17\%$ as incoherent or noisy ($\leq 2$; on this coherence scale, unlike the relevance scale used in the pipeline validations, low scores indicate unrecognizable or noisy evidence). \important{Thus the premise is empirically non-trivial: many lifted-word lists contain semantic structure that humans can independently recognize, while a minority remain ambiguous or noisy.}
 
\textbf{Do Atlas labels name that structure?}
Yes. Phase~2 collected $1{,}025$ individual candidate-name ratings: $205$ scoring records over $34$ concepts from $8$ annotators. In blind scoring, annotators sharply separated real labels from distractors, validating the task itself (\cref{tab:human-eval-phase2}). \important{Atlas labels and human-generated labels both scored far above taxonomy and embedding distractors, showing that raters were not merely assigning high scores to any plausible phrase.}

\begin{table}[t]
\centering
\small
\begin{tabular}{lccc}
\toprule
candidate type & mean fit ($1$--$5$) & sd & $n$ \\
\midrule
LLM (Atlas) label  & $3.98$ & $1.04$ & $205$ \\
human label           & $3.50$ & $1.17$ & $402$ \\
embedding distractor  & $2.72$ & $1.30$ & $205$ \\
taxonomy distractor   & $2.08$ & $1.11$ & $205$ \\
filler (floor)        & $1.62$ & $1.06$ & \phantom{00}$8$ \\
\bottomrule
\end{tabular}
\caption{Phase~2 blind fit scores by candidate type. Human annotators rated how well each candidate name fit the same lifted-word evidence. Counts sum to $1{,}025$ individual ratings.}
\label{tab:human-eval-phase2}
\end{table}

The strongest comparison is between the Atlas label and independently generated human labels. Atlas labels scored higher on average: $3.98$ versus $3.50$. They also received a top-two rating ($\geq 4$) $79\%$ of the time, compared with $63\%$ for human labels. In paired comparisons, the Atlas label outscores a human label with probability $0.62$ (cluster-bootstrap $95\%$ CI $[0.58,0.66]$). Additional ordinal mixed-model and robustness analyses are reported in \cref{app:human-eval-details}.

\textbf{Interpretation.}
These results address the central question of validity: humans can infer meaningful concepts from lifted-word evidence alone. \important{When asked to judge labels blindly, humans rate the Atlas labels (LLM-generated) at least as highly as independently human-generated labels, and far above nearby distractors.} This does not certify every concept in the library individually, but it does show that Atlas is not merely producing LLM-fluent names for arbitrary clusters. On a stratified pilot, the labels are human-recognizable, preferred to strong distractors, and competitive with or better than human-written names.

\textbf{Connection to known concept alignment.} The same principle underlies the known concept alignment metric used in model evaluation (\cref{sec:interp_metrics}). Known concept alignment asks whether the word/token evidence associated with a concept from the library---for example, lifted words or the tokens most boosted by a steering vector---matches that concept's name. The human study validates this question as meaningful and supports using an LLM judge as a scalable proxy: \important{humans independently recognize the semantic structure in the evidence and endorse the Atlas labels under blind comparison.} The validation chain is therefore: human-recognizable word/token evidence $\rightarrow$ human-endorsed Atlas labels $\rightarrow$ scalable LLM-judged concept alignment.

\subsection{Data Processing}
\label{sec:data-processing}

The Atlas pipeline described in \cref{sec:data-stage1}--\cref{sec:data-stage3} produces a concept library and a trained concept annotator model. This subsection describes how we apply the annotator to the full training corpus, producing the tokenized, concept-annotated training data that the \steerling model consumes. The process has three stages: distributed concept annotation with embedding extraction, tokenization with document-structure markup, and retrieval index construction (\cref{sec:training-data-indexing}).

\subsubsection{Distributed Concept Annotation}

\looseness=-1The Stage~3 annotator (\cref{sec:data-stage3}) must be applied to every chunk in the training corpus---approximately 11 billion text chunks spanning 1.5 trillion tokens. We distribute this workload across GPU nodes using SLURM array jobs, where each task processes and annotates a single unit of stored data (in parquet format) independently.
 
\paragraph{Input format.} Each input parquet file contains rows of chunked text with fields -- \texttt{chunk\_text} (the main content), \texttt{data\_src}, \texttt{doc\_id}, \texttt{chunk\_id}, \texttt{doc\_end\_flag} (marking the final chunk of a document), and \texttt{metadata}. The chunking follows the same procedure described in \cref{sec:data-stage1}: sentence boundaries are detected with BlingFire sentence splitter \citep{microsoft_blingfire_2019}, and consecutive sentences are concatenated until a domain-specific token threshold is reached (150 tokens for web text, 256 tokens for mathematics and code).
 
\looseness=-1\paragraph{Forward pass.} For each chunk, the annotator's \texttt{Qwen3-Embedding-0.6B} encoder produces a pooled 1024-dimensional embedding, which is then passed to the multi-head classifier. The classifier predicts concept activations across all four domains---content, tone, demographic, and alignment---using domain-specific thresholds and per-domain caps on the maximum number of predicted concepts. Each domain's predictions are local indices into that domain's concept vocabulary; these are mapped to global concept IDs via fixed offsets, producing a unified list of concept annotations per chunk.
 
\paragraph{Embedding extraction.} During the same forward pass, we cache the 1024-dimensional encoder embeddings for every text chunk, storing them in Zarr arrays at FP16 precision alongside that chunk's metadata (\texttt{data\_src}, \texttt{doc\_id}, \texttt{chunk\_id}). These embeddings serve two purposes: they form the basis of the retrieval index constructed in \cref{sec:training-data-indexing}, and they are available for downstream analysis without requiring additional encoder inference. At the scale of the full training corpus, this amounts to approximately 11 billion 1024-dimensional vectors.
 
\looseness=-1\paragraph{Output format.} Each input parquet file produces a corresponding annotated parquet file with the original columns preserved and new columns added: \texttt{concepts} (list of global concept IDs), and per-domain arrays (\texttt{content\_global\_ids}, \texttt{tone\_global\_ids}, \texttt{demographic\_global\_ids}, \texttt{alignment\_global\_ids}). These annotated parquet files are the input to the tokenization stage below.

\subsubsection{Tokenization and document structure}

\begin{figure}[t]
  \centering
  \includegraphics[trim={2cm 4.2cm 2cm 6.7cm}, clip, width=0.95\textwidth]{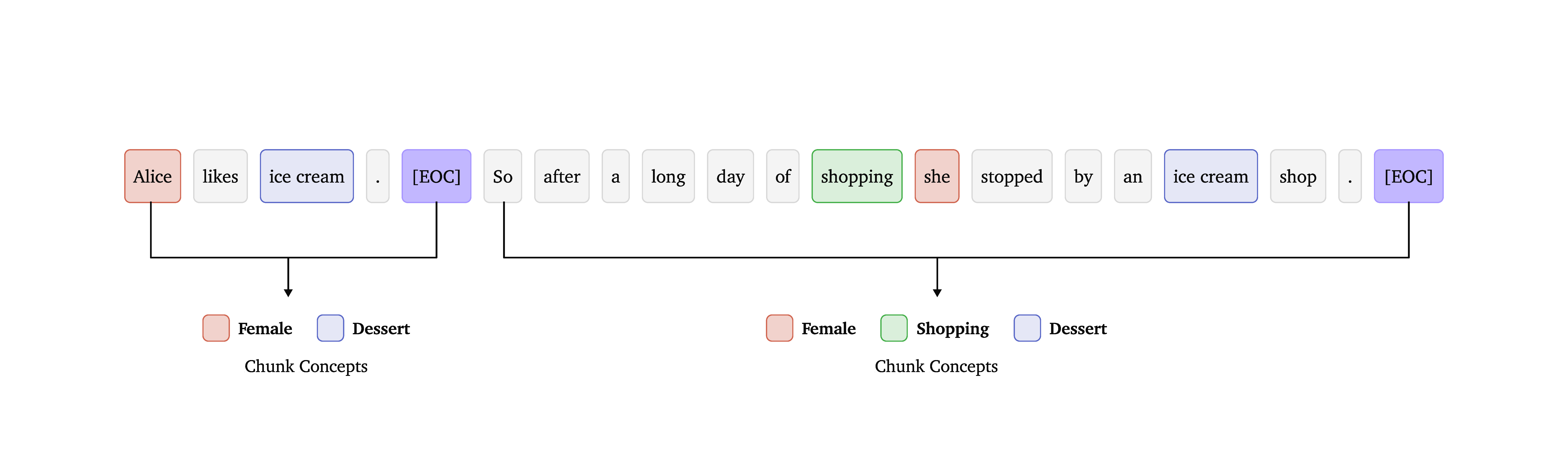}
  \caption{Token stream with chunk-level concept annotations. Each chunk is terminated by an \texttt{[EOC]} token and associated with a set of concept labels. Concept supervision is provided at the chunk level; the model learns to localize concepts to individual tokens via OR-aggregation during training.}
  \label{fig:chunk-concept-alignment}
\end{figure}
 
The annotated parquet files are tokenized into the token stream consumed by the model during training. We use the \texttt{cl100k\_base} encoding from tiktoken\footnote{\url{https://github.com/openai/tiktoken}}. We add three special tokens to represent our document and training structure -- padding token (\texttt{[PAD]}, ID $100{,}277$), end-of-chunk token within a document (\texttt{[EOC]}, ID $100{,}279$), and a mask token for diffusion training (\texttt{[MASK]}, ID $100{,}280$) -- and reuse the encoding's native \texttt{<|endoftext|>} token (ID $100{,}257$) as a document-boundary marker. Including the byte-pair tokens, the encoding's existing special tokens, and our additions, the full vocabulary contains $100{,}281$ tokens.

% \begin{center}
% \small
% \begin{tabular}{lll}
% \toprule
% \textbf{Token} & \textbf{ID} & \textbf{Role} \\
% \midrule
% % \texttt{<|endoftext|>}    & 100257 & End of document \\
% \texttt{[PAD]}          & 100277 & Padding \\
% % \texttt{[BOS]}          & 100278 & Beginning of sequence \\
% \texttt{[EOC]}   & 100279 & End of chunk (within a document) \\
% \texttt{[MASK]}         & 100280 & Masked position (for diffusion training) \\
% \bottomrule
% \end{tabular}
% \end{center}
 
\paragraph{Tokenization procedure.}\looseness=-1 Each chunk's text is tokenized without special tokens. An end-of-chunk (\texttt{EOC}) token is appended after every chunk, marking the boundary between consecutive chunks within a document. For the final chunk of each document (identified by the \texttt{doc\_end\_flag}), an \texttt{<|endoftext|>} token is additionally appended after the \texttt{EOC} token, signaling the document boundary. This structure preserves the document--chunk hierarchy in the token stream: the model can distinguish intra-document chunk boundaries within a document from boundaries between documents.
 
\paragraph{Chunk--concept alignment.} The concept annotations produced by the annotator are chunk-level labels: each chunk is associated with a list of concept IDs, with no token-level labels. The \texttt{[EOC]} token serves as the delimiter that aligns concept annotations with their corresponding token spans in the training data. During training, the concept loss (\cref{sec:architecture}) uses \emph{OR-aggregation} across all tokens within a chunk to bridge this chunk-level supervision with the model's token-level concept activations. \cref{fig:chunk-concept-alignment} illustrates this structure.

% a concept is treated as present for a chunk if it is active at \emph{any} token within the chunk, so the model's per-token activations are combined by a (soft) logical OR over the chunk's tokens and the aggregate is supervised against the chunk label. This multiple-instance formulation lets token-level localization emerge from chunk-level supervision. 

\subsection{Training data indexing for test-time attribution}
\label{sec:training-data-indexing}

The pipeline in \cref{sec:data-processing} produces two artifacts: approximately 11 billion text chunks annotated with concepts by Stage~3 of the Atlas pipeline (\cref{sec:data-stage3}), and, for each chunk, a 1024-dimensional embedding vector computed during the annotator's forward pass over the corpus. Because these embeddings are cached at annotation time, index construction reuses them directly and requires no additional annotator forward passes. These vectors form the index that supports test-time training data attribution.

\textbf{Vector database.} An exact (flat) index scores a query against every stored vector. While accurate, it imposes two costs that are each prohibitive at our scale: search latency that grows linearly in the number of vectors $N$, and memory that grows as $N \times d$ in full precision. We therefore use an approximate index---an Inverted File with Product Quantization (IVFPQ) index from the FAISS library \citep{johnson2019billion}---which addresses both costs: an inverted-file structure partitions the space so each query scans only a small fraction of the vectors, and product quantization compresses the stored vectors into compact codes. %We detail the index structure, hyperparameters, and the query-time distance computation in \cref{app:training-data-indexing}.

With our configuration, product quantization reduces the per-vector payload from $4{,}096$ bytes (a full-precision 1024-dimensional vector) to $64$ bytes, roughly a $64\times$ compression. We $\ell_2$-normalize all vectors at build and search time, so inner-product retrieval is equivalent to cosine similarity. Even so, the pretraining data index occupies approximately $808$~GB on disk, exceeding available RAM; we therefore store the inverted lists in a memory-mapped on-disk format, so only the lists touched by a query are paged into memory at search time, enabling search over the full corpus without loading the entire index into RAM.

\paragraph{Retrieval accuracy.}
Because IVFPQ is approximate, we assess how faithfully it reproduces exact search. On $1{,}000$ query chunks drawn from the index itself, we query the index using an $n_{\text{probe}}=16$, and measure recall@$k$ as the fraction of queries whose own source chunk---the known ground-truth nearest neighbor---appears among the top-k retrieved results. The index achieves a \textbf{recall@10 of 96.8\%}, indicating that the quantized search recovers the exact nearest neighbor in the large majority of cases while operating within the memory and latency budget imposed by a corpus of this scale.

%% file: sections/data/chunk_tag_example_webtext.tex
% Requires: \usepackage{tcolorbox}  \usepackage{array}  \usepackage{xcolor}  \usepackage{caption}

% Chunk colors
\definecolor{c0}{HTML}{4847e4}   % indigo
\definecolor{c1}{HTML}{735900}   % olive   (was mislabeled "teal")
\definecolor{c2}{HTML}{854F0B}   % brown   (was mislabeled "amber")
\definecolor{c3}{HTML}{534AB7}   % purple
\definecolor{background}{HTML}{f5f5f5}

% Tag key:value row   #1 = color   #2 = key   #3 = value
\newcommand{\tagrow}[3]{%
  {\color{#1}\footnotesize\textbf{#2:}~#3}\\[1.5pt]%
}

\begin{table}[t]
\centering
\begin{tcolorbox}[
  enhanced,
  arc=5pt,
  boxrule=0.5pt,
  colframe=gray!40,
  colback=background,
  left=8pt, right=8pt, top=4pt, bottom=4pt,
  fontupper=\sffamily\footnotesize,
]
\setlength{\tabcolsep}{5pt}
\begin{tabular}{@{}>{\raggedright\arraybackslash}p{0.56\linewidth}@{\hskip 10pt}l@{}}
% --- header (both cells centered over their column widths) ---
\multicolumn{1}{@{}>{\centering\arraybackslash}p{0.56\linewidth}@{\hskip 10pt}}{\footnotesize\textbf{Chunks}} &
\multicolumn{1}{>{\centering\arraybackslash}p{0.40\linewidth}@{}}{\footnotesize\textbf{Structured Tags}} \\[2pt]
% \noalign{{\color{gray!50}\hrule height 0.4pt}\vskip5pt}
% --- Chunk 1 ---
{\color{c0}\footnotesize%
The game increases the chances of finding a Shiny Pokemon by generating extra personality values in an attempt to find one that results in a Shiny Pokemon, with the number of attempts depending on the size of the current streak. For every Pokemon added to the streak up to 20 Pokemon, the game will make two extra attempts to find a Shiny personality value; i.e., the number of attempts at any given point in the streak is $1 + 2 *$ streak\_size, and caps at a maximum of 41 attempts when the streak is at least 20 Pokemon long.%
}
&
\begin{minipage}[t]{0.40\linewidth}\raggedright
    \tagrow{c0}{main}{gaming, mechanics, video-games, Pokemon, shiny-mechanics}
    \tagrow{c0}{purpose}{informational-guide, explanatory, how-to}
    \tagrow{c0}{tone}{neutral, informative, technical}
    \tagrow{c0}{minor}{gaming-progress, streaks, gaming-algorithms, random-generation}
\end{minipage}
\\[4pt]
% \noalign{\vskip2pt{\color{gray!40}\hrule height 0.4pt}\vskip4pt}
% --- Chunk 2 ---
{\color{c1}\footnotesize%
In Going for the Gold!, Ash and his friends met a fisherman named Rodman, who was trying to fish up a Shiny Magikarp with a Magikarp-shaped lure. Ash, Serena, and Clemont decided to try fishing too, with Ash giving the inexperienced Serena instructions of how to do it correctly. While fishing, Serena hooked up a Corsola, which she tried to battle with her Fennekin, but it simply hid itself behind Serena when Corsola tried to use Water Gun on it, causing the Coral Pokémon to get away.%
}
&
\begin{minipage}[t]{0.40\linewidth}\raggedright
    \tagrow{c1}{main}{fiction, media-franchise, anime, pokemon, adventure}
    \tagrow{c1}{purpose}{narrative, storytelling, character-development}
    \tagrow{c1}{tone}{casual, descriptive, lighthearted}
    \tagrow{c1}{minor}{fishing-techniques, lure-use, pokemon-battles, strategy-mistakes}
\end{minipage}
\end{tabular}
\end{tcolorbox}
\caption{A webtext document about \emph{Pokemon} split into two chunks, with its domain-specific, hierarchical structured tags.}
\label{tab:chunk-tag-example-webtext}
\end{table}

%% file: sections/data/topk_frequent_tags.tex
% Requires in preamble:
%   \usepackage{tcolorbox}
%   \tcbuselibrary{skins}
%   \usepackage{xcolor}
\definecolor{tagcream}{HTML}{ceb4fe}
\definecolor{tagframe}{HTML}{ceb4fe}
\definecolor{boxbg}{HTML}{f5f5f5}
\definecolor{boxframe}{HTML}{8d7cff}

\newcommand{\tagsingle}[1]{%
  \tcbox[
    on line,
    arc=3pt,
    outer arc=3pt,
    boxsep=1pt,
    left=4pt, right=4pt, top=2pt, bottom=2pt,
    boxrule=0.5pt,
    colframe=tagframe,
    colback=tagcream,
    fontupper=\footnotesize\sffamily
  ]{#1}\hspace{4pt}\vspace{1pt}%
}

\begin{tcolorbox}[
  enhanced,
  title={{Top-50 most frequent tags}},
  fonttitle=\small\bfseries,
  colback=boxbg,
  colframe=boxframe,
  colbacktitle=boxframe,
  coltitle=white,
  arc=5pt,
  boxrule=0.8pt,
  left=8pt, right=8pt, top=6pt, bottom=6pt
]
{\setlength{\baselineskip}{10\baselineskip}%
\tagsingle{academic writing}
\tagsingle{research reporting}
\tagsingle{science research}
\tagsingle{biology}
\tagsingle{physics}
\tagsingle{mathematics}
\tagsingle{computer science}
\tagsingle{medicine}
\tagsingle{data analysis}
\tagsingle{molecular biology}
\tagsingle{technology}
\tagsingle{research methodology}
\tagsingle{experimental results}
\tagsingle{scientific explanation}
\tagsingle{chemistry}
\tagsingle{statistics}
\tagsingle{real world context}
\tagsingle{theoretical physics}
\tagsingle{machine learning}
\tagsingle{arithmetic}
\tagsingle{educational}
\tagsingle{algorithms}
\tagsingle{computation}
\tagsingle{education}
\tagsingle{mathematical concepts}
\tagsingle{astrophysics}
\tagsingle{notation}
\tagsingle{analysis}
\tagsingle{fiction}
\tagsingle{programming}
\tagsingle{technical explanation}
\tagsingle{academic discussion}
\tagsingle{genetics}
\tagsingle{informational}
\tagsingle{software development}
\tagsingle{equations}
\tagsingle{documentation}
\tagsingle{engineering}
\tagsingle{comparison}
\tagsingle{information sharing}
\tagsingle{methodology}
\tagsingle{communication}
\tagsingle{narrative}
\tagsingle{materials science}
\tagsingle{functions}
\tagsingle{geography}
\tagsingle{comparative analysis}
\tagsingle{research methods}
\tagsingle{source code}
\tagsingle{economics}
\tagsingle{study findings}
\tagsingle{material science}
% \tagsingle{character development}
% \tagsingle{statistical analysis}
% \tagsingle{data structures}
% \tagsingle{biochemistry}
% \tagsingle{history}
% \tagsingle{geometry}
% \tagsingle{algebra}
% \tagsingle{literature}
% \tagsingle{question answering}
% \tagsingle{scientific reporting}
% \tagsingle{health medicine}
% \tagsingle{environmental science}
% \tagsingle{literature review}
% \tagsingle{methodology description}
% \tagsingle{astronomy}
% \tagsingle{demographics}
% \tagsingle{research summary}
% \tagsingle{optimization}
% \tagsingle{psychology}
% \tagsingle{personal narrative}
% \tagsingle{problem solving}
% \tagsingle{quantum mechanics}
% \tagsingle{research paper}
% \tagsingle{explanation}
% \tagsingle{scientific findings}
% \tagsingle{data interpretation}
% \tagsingle{implementation}
% \tagsingle{mathematical proof}
% \tagsingle{storytelling}
% \tagsingle{media}
% \tagsingle{results discussion}
% \tagsingle{functional}
% \tagsingle{public health}
% \tagsingle{verification}
% \tagsingle{programming languages}
% \tagsingle{definition}
% \tagsingle{graph theory}
% \tagsingle{entertainment}
% \tagsingle{experimental methods}
% \tagsingle{web development}
% \tagsingle{configuration}
% \tagsingle{news reporting}
% \tagsingle{linear algebra}
% \tagsingle{data collection}
% \tagsingle{data processing}
% \tagsingle{error analysis}
% \tagsingle{research}
% \tagsingle{technical reporting}%
}
\end{tcolorbox}

%% file: sections/architecture/main.tex
\section{Inherently interpretable architecture}
\label{sec:architecture}
In this section, we present the architectural and training choices to build a language model with inherently interpretable outputs, an instantiation of \cref{sec:recipe}. We depart from the autoregressive paradigm: the masking-based training objective and attention structure of diffusion models are better suited to the interpretability properties we want to enforce (\cref{subsec:beyond-ar}), and we adopt a causal block-attention pattern that retains efficient inference while preserving the diffusion training objective (\cref{subsec:causal-diffusion}). On top of this backbone, we introduce the concept module, a bottleneck inserted between the transformer and the language modeling head that routes every prediction through an explicit concept representation (\cref{subsec:concept-module}). We then describe the training procedure that ties these components together (\cref{subsec:training}).

\subsection{Beyond autoregressive models}
\label{subsec:beyond-ar}

Autoregressive (AR) models, built on a causal-attention transformer architecture, have been the standard choice for large language models, achieving impressive performance across a wide range of tasks~\citep{gpt4, claude, gemini, deepseek, llama3, qwen}. Beyond raw capability, the AR paradigm benefits from years of accumulated practical knowledge: training procedures, hyperparameter choices, and scaling laws have been studied extensively and documented in detail by the open-source community~\citep{olmo, llama3, gemma, deepseek}, making AR models a tempting default. However, their inductive biases do not align well with our objectives. We aim to build a model that supports faithful input attribution and concept-level control at inference time, and AR generation makes both properties harder to enforce. AR models predict one token at a time conditioned on a strictly causal context: concepts typically span multiple tokens rather than localizing to one, and attribution lacks a natural absence-of-information baseline. The same left-to-right factorization underlies well-documented failure modes, including the reversal curse~\citep{berglund2024reversal} and poor performance on tasks requiring graph backtracking~\citep{ye2025beyond}.

The limitations above are not incidental to AR models; they follow from the left-to-right factorization itself, so addressing them calls for a different generative paradigm. Diffusion language models offer a more natural fit. The masking objective sidesteps each of the failures identified above and provides three properties we exploit throughout this work. First, it gives the model an explicit, trained representation of ``no information at this position,'' the absence baseline that faithful input attribution requires and that strictly causal contexts cannot provide. Second, multiple tokens are predicted jointly at each denoising step,    giving us a natural interface for concept-level control over phrases rather than individual tokens, directly addressing the fact that concepts span multiple tokens rather than localizing to one. Third, diffusion models generate tokens in any order, so the model can choose where in the sequence to express an intervened concept rather than being forced to commit at the next position. We describe our specific instantiation, \causaldiff, in the next section.

\subsection{\causaldiff}
\label{subsec:causal-diffusion}
Our objectives place two demands on the architecture. First, we want the model to operate over groups of tokens rather than single positions, since the concepts we interpret and steer (\cref{subsec:concept-module}) typically span phrases rather than localizing to one token. Second, we want autoregressive-style inference efficiency: the KV caching and throughput that make AR models practical at scale. Masked diffusion gives us the first through its joint, any-order denoising, but, as we show below, its standard form sacrifices the second. The remainder of this section develops an attention structure that recovers both.

\begin{figure}[!htbp]
    \centering
\includegraphics[trim={3cm 3cm 3cm 3cm}, clip, width=\textwidth]
    {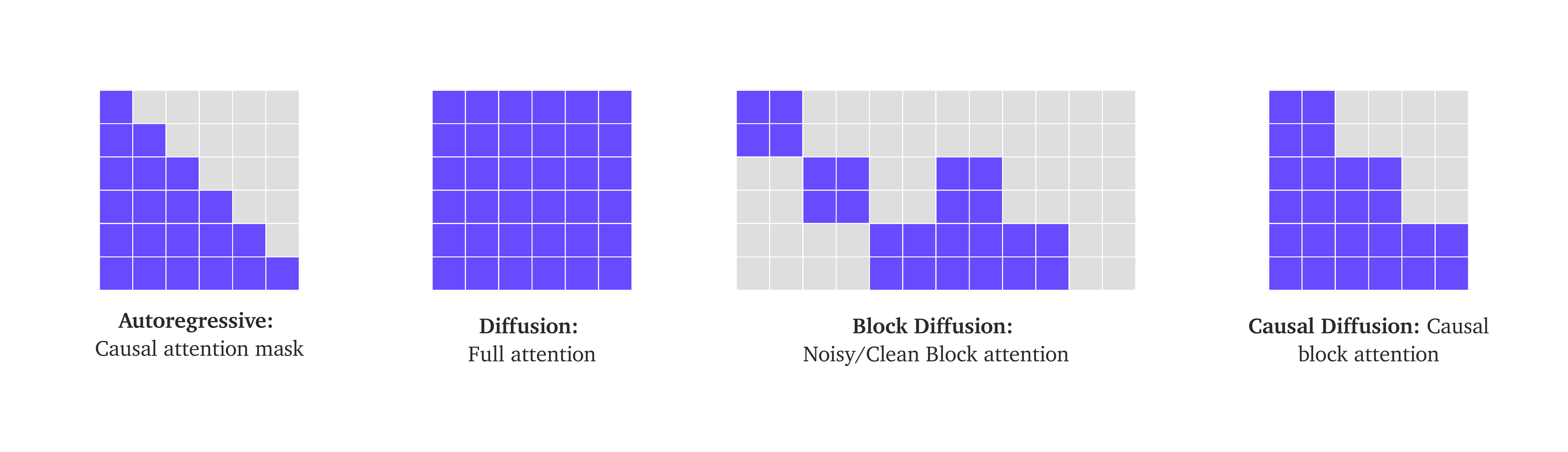}
    \caption{Attention patterns for Autoregressive, Diffusion, 
    \blockdiff, and \causaldiff models.}
    \label{fig:attention-patterns}
\end{figure}

\looseness-1 We build on masked diffusion models (MDMs; see \cref{sec:related-work} and \cref{sec:background}). Standard MDMs use full bidirectional attention: every token attends to every other token in the sequence at each denoising step (\cref{fig:attention-patterns}, panel b). Bidirectional context is what enables MDMs to denoise multiple tokens jointly in a single forward pass and to predict tokens in any order, since no position is privileged over another. The same property, however, prevents efficient inference. Because tokens that change between denoising steps are attended to by every other token, no representations can be cached across steps~\citep{israel2026enabling, blockdiffusion}. Each step recomputes the full attention over the full sequence, making inference substantially slower than autoregressive models of comparable size.

\looseness-1 Block diffusion~\citep{blockdiffusion} addresses this inference-time cost by modifying the attention pattern. The sequence is partitioned into blocks of fixed length $b$, and the attention mask is bidirectional within each block but causal across blocks (\cref{fig:attention-patterns}, panel c). Generation proceeds one block at a time, and KV caches built from previously generated blocks can be reused across denoising steps. The training algorithm, however, requires concatenating a noisy and a clean copy of the sequence as input: noisy blocks supply the diffusion loss, while clean previous blocks supply context. This roughly doubles the per-step memory and FLOPs relative to standard MDM training.

We propose \textbf{\causaldiff}, which retains \blockdiff's attention pattern but drops the clean copy of the sequence at training time (\cref{fig:attention-patterns}, panel d). The training objective is the standard masked diffusion loss from \cref{sec:bg-mdm}, applied to a single sequence of tokens with a block-causal attention mask: bidirectional within each block, causal across blocks. \citet{llada} showed that an MDM trained with full bidirectional attention can be sampled block-by-block at inference time with minimal quality degradation, suggesting that the masking objective does not require bidirectional context across the entire sequence. If block-causal attention is sufficient at inference, training under the same constraint costs nothing in expressive power: we obtain block-causal structure at half the training cost of \blockdiff, without the inference-time bottleneck of standard MDMs.
At inference, each new block is denoised by masked diffusion while conditioning on the keys and values cached from all previous blocks; once generated, its own keys and values are appended to the cache, exactly as in autoregressive decoding. The result is an MDM that retains diffusion's parallelism and any-order flexibility within each block while inheriting autoregressive-style KV caching across blocks.

\subsection{Concept module}
\label{subsec:concept-module}

One way to build an inherently interpretable language model is to decompose the hidden representation into a set of concepts and use those concepts, combined through a simple and interpretable function, for the model's predictions. Two properties follow directly from this decomposition: faithful concept attribution, since each concept's contribution to a given output can be computed directly, and concept steering, since we know how each concept is represented and can therefore bias the model towards or away from it. We achieve this with an additive bottleneck: the hidden representation is reconstructed as a sum of concept contributions before being passed to the language modeling head. We refer to this as our \emph{concept module}.

\begin{figure}[thbp!]
    \centering
\includegraphics[trim={2cm 8cm 2cm 8cm}, clip, width=\textwidth]{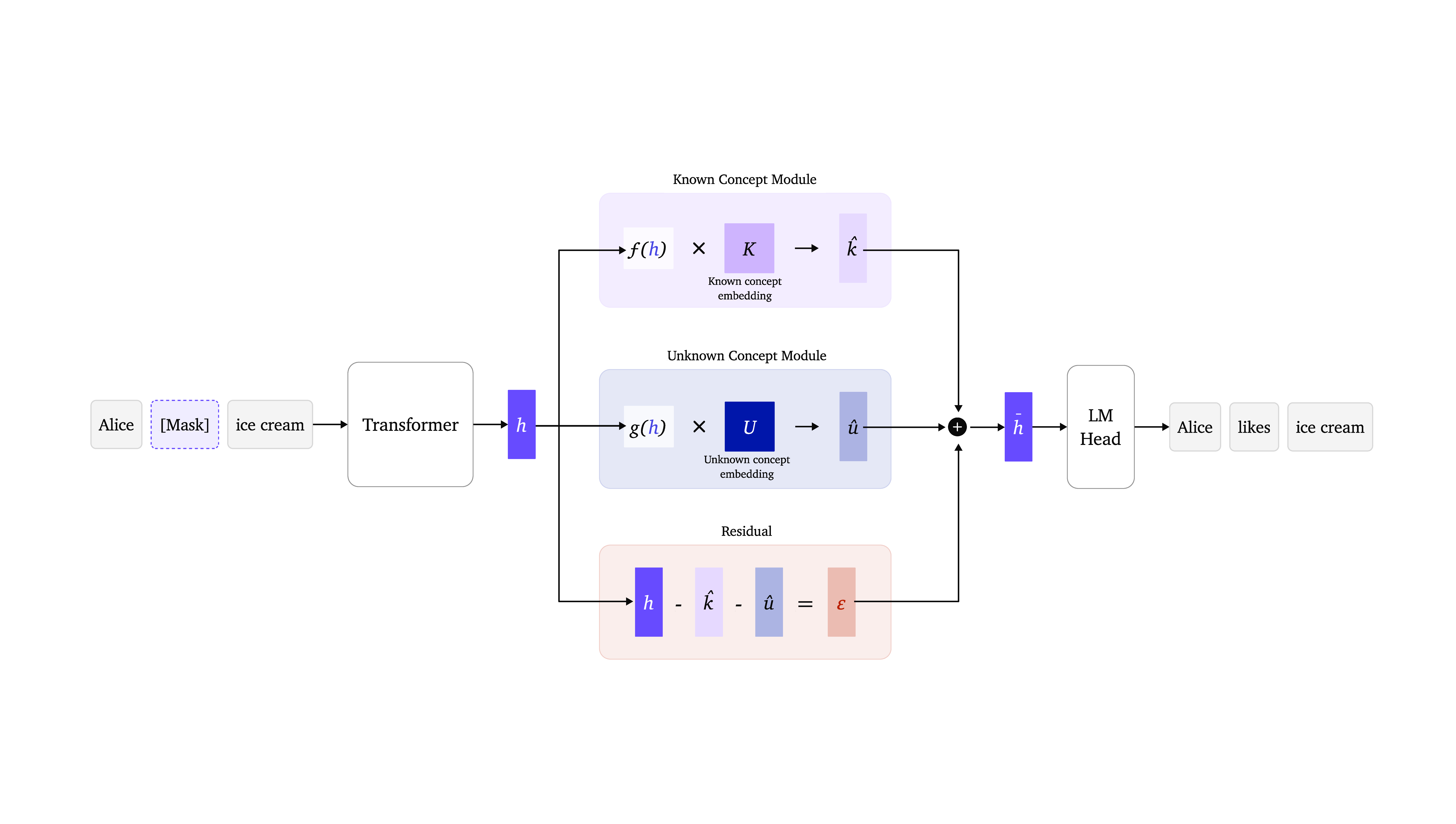}
\caption{The concept module bottleneck. The transformer hidden state $h$ is decomposed into known concept contributions $\hat{k}$, unknown concept contributions $\hat{u}$, and a residual $\varepsilon$.}
    \label{fig:concept-module-arch}
\end{figure}

The concept module sits between the transformer backbone and the language modeling head (\cref{fig:concept-module-arch}). For each input token, the transformer produces a hidden state $h \in \R^d$, which is normally passed straight to the language modeling head. The concept module instead decomposes $h$ into three additive components:
\begin{equation}
\bar{h} = \hat{k} + \hat{u} + \varepsilon,
\label{eq:bottleneck}
\end{equation}
where $\hat{k}$ is a weighted sum of \emph{known} concept embeddings, $\hat{u}$ is a weighted sum of \emph{unknown} concept embeddings, and $\varepsilon = h - \hat{k} - \hat{u}$ is the residual term. Only $\bar{h}$ is passed to the LM head, so every output logit becomes a linear function of concept activations. We apply dropout with rate $p_{\varepsilon}$ to $\varepsilon$ during training, to discourage the model from relying on the residual channel for prediction.

Both $\hat{k}$ and $\hat{u}$ are computed from $h$ via two small heads. The known head $f$ produces concept activation probabilities for the labeled concept set, and the unknown head $g$ produces activation probabilities for a set of unknown concepts:
\begin{equation}
k = \sigma(f(h)) \in \R^n, \qquad u = \sigma(g(h)) \in \R^m,
\label{eq:concept-activations}
\end{equation}
\looseness=-1 where $f$ and $g$ are small learnable networks, $\sigma$ denotes the elementwise sigmoid, $n$ is the number of known concepts, and $m \gg n$ is the number of unknown concepts. Each concept $i$ has a learned embedding $K_i \in \R^d$ (or $U_j$ for unknown), analogous to a token embedding, and the concept-weighted hidden states are
\begin{equation}
\hat{k} = \sum_{i=1}^{n} k_i K_i, \qquad \hat{u} = \sum_{j=1}^{m} u_j U_j.
\label{eq:concept-pooling}
\end{equation}
The full set of concept embeddings forms a vocabulary of concepts, in the same way that token embeddings form a vocabulary of tokens. Given the large number of unknown concepts, storing the unknown embedding matrix $U \in \R^{m \times d}$ directly would dominate the parameter count. We therefore factorize it as a low-rank product $U = AB$, with $A \in \R^{m \times r}$ and $B \in \R^{r \times d}$ for rank $r \ll d$, reducing the parameter count from $md$ to $r(m + d)$ while preserving capacity.

At inference time, the additive structure of $\bar{h}$ makes every prediction a transparent function of the concept activations. Because the LM head is linear, the logit for any output token decomposes as
\begin{equation}
\ell_y = \hat{k}^\top W_y + \hat{u}^\top W_y + \varepsilon^\top W_y,
\label{eq:logit-decomposition}
\end{equation}
where $W_y$ is the row of the LM head corresponding to token $y$. Each term is exact: there is no approximation involved in attributing a prediction to known concepts, unknown concepts, or the residual. This decomposition is what allows the model to support faithful concept attribution.

Taken together, these components make the concept module a self-contained, modular interface to the model's predictions:  known and unknown concepts, their embeddings, and the residual are separate, inspectable channels, and every output is an explicit, additive function of them. This organization is what makes the attribution and steering described in \cref{sec:capabilities} possible.

\subsection{Model training}
\label{subsec:training}
The architecture described in \cref{subsec:concept-module} forces every prediction to pass through an additive concept representation. In effect, it builds in the linear representation hypothesis~\citep{park2023linear}, the idea that high-level concepts are encoded as linear directions in representation space. Rather than hoping this property emerges, as it may or may not in a standard language model, our architecture enforces it by construction: every output is an explicit linear function of concept activations. This structural guarantee, however, does not by itself ensure that the concept module learns useful concepts, that the language modeling head produces fluent text, or that the known and unknown concepts encode complementary information. Each of these properties has to be optimized for. We describe the loss objectives that target them in \cref{subsubsec:loss-objectives}, and the training dynamics that control how those losses are applied over time in \cref{sec:training-dynamics}.

\subsubsection{Loss objectives}
\label{subsubsec:loss-objectives}

\looseness=-1\paragraph{Language modeling loss.} The primary objective is the masked diffusion loss $\mathcal{L}_{\text{MDM}}$ from \cref{sec:bg-mdm}, applied to the bottlenecked hidden state $\bar{h}$ rather than the raw transformer output $h$. Every token prediction is therefore evaluated as a function of the concept module's output. We denote this loss $\mathcal{L}_{\text{LM}}$.

% \paragraph{Concept loss.} The ground-truth concept labels are produced at the chunk level by the annotation pipeline of \cref{sec:data}. A chunk is a contiguous span of tokens terminated by an \texttt{[EOC]} token (\cref{fig:chunk-concept-alignment}). The labels are positive-only: a label tells us a concept appears somewhere in a chunk, but not at which token. We accommodate this weaker supervision signal with an OR-aggregation across the chunk. For known concept $c$, let $k_{c,t}$ denote its predicted activation at token $t$ (the $c$-th entry of $k$ from \cref{eq:concept-activations}). The probability that $c$ appears at least once in the chunk is

% \begin{equation}
% k^{\text{chunk}}_c = 1 - \prod_{t \in \text{chunk}} (1 - k_{c,t}).
% \label{eq:chunk-aggregation}
% \end{equation}

% Let $y_c \in \{0, 1\}$ denote the ground-truth chunk-level label for concept $c$. The concept loss is the binary cross-entropy between the aggregated probability and the chunk label, summed over all known concepts:
% \begin{equation}
% \mathcal{L}_{\text{concept}} = -\sum_{c=1}^{n} \Big[ y_c \log k^{\text{chunk}}_c + (1 - y_c) \log (1 - k^{\text{chunk}}_c) \Big].
% \label{eq:concept-loss}
% \end{equation}

% The OR-aggregation is satisfied as soon as the concept is predicted at any one token in the chunk, consistent with our chunk-level supervision.

\paragraph{Concept loss.} The ground-truth concept labels are produced at the chunk level by the annotation pipeline of \cref{sec:data}. A chunk is a contiguous span of tokens terminated by an \texttt{[EOC]} token (\cref{fig:chunk-concept-alignment}). The labels are positive-only: a label tells us a concept appears somewhere in a chunk, but not at which token. We accommodate this weaker supervision signal with an OR-aggregation across the chunk. Because the diffusion objective only supervises masked positions, the aggregation runs over the masked tokens of the chunk; let $\text{chunk}_{\mathcal{M}}$ denote this set, where $\mathcal{M}$ is the set of masked positions in the minibatch. For known concept $c$, let $k_{c,t}$ denote its predicted activation at token $t$ (the $c$-th entry of $k$ from \cref{eq:concept-activations}). The probability that $c$ appears at least once among the masked tokens of the chunk is
\begin{equation}
k^{\text{chunk}}_c = 1 - \prod_{t \in \text{chunk}_{\mathcal{M}}} (1 - k_{c,t}).
\label{eq:chunk-aggregation}
\end{equation}
Let $y_c \in \{0, 1\}$ denote the ground-truth chunk-level label for concept $c$. The concept loss is the binary cross-entropy between the aggregated probability and the chunk label, averaged over the known concepts and over chunks in the minibatch:
\begin{equation}
\mathcal{L}_{\text{concept}} = -\frac{1}{n}\sum_{c=1}^{n} \Big[ y_c \log k^{\text{chunk}}_c + (1 - y_c) \log (1 - k^{\text{chunk}}_c) \Big].
\label{eq:concept-loss}
\end{equation}
The OR-aggregation is satisfied as soon as the concept is predicted at any one masked token in the chunk, consistent with our chunk-level supervision.

\paragraph{Reconstruction loss.} The unknown head is trained to represent the part of the hidden state that is not captured by the known concepts. Given the ground-truth labels for known concepts, the ideal known representation and the corresponding target for the unknown head are
\begin{equation}
\hat{k}^{\text{GT}} = \sum_{i=1}^{n} k^{\text{GT}}_i K_i, \qquad \hat{u}^{\text{GT}} = h - \hat{k}^{\text{GT}},
\label{eq:gt-targets}
\end{equation}
where $\hat{u}^{\text{GT}}$ is the residual that remains after subtracting $\hat{k}^{\text{GT}}$ from the transformer hidden state. The unknown head is trained to match this target under a mean-squared error, averaged over masked positions in the minibatch:
\begin{equation}
\mathcal{L}_{\text{rec}} = \frac{1}{|\mathcal{M}|}\sum_{t \in \mathcal{M}} \| \hat{u}_t - \hat{u}_t^{\text{GT}} \|_2^2,
\label{eq:rec-loss}
\end{equation}

where $\mathcal{M}$ is the set of masked token positions in the minibatch. When $\hat{u} = \hat{u}^{\text{GT}}$, the residual term in the bottleneck satisfies $\varepsilon = 0$.

% \paragraph{Independence loss.} The reconstruction loss alone does not prevent the unknown head from encoding information that is already represented by the known concepts. To discourage such redundancy, we penalize the statistical dependence between the known and unknown representations using a normalized cross-covariance penalty in the spirit of the Hilbert-Schmidt Independence Criterion with a linear kernel~\citep{hsic_2009, hsic_2020}, following its use for concept decoupling in~\citet{sccbgm}. Over a minibatch of $B$ token representations, let
% \begin{equation}
% \Phi = H_k - \mathbf{1}\boldsymbol{\mu}_{\hat{k}}^\top, \qquad \Psi = H_u - \mathbf{1}\boldsymbol{\mu}_{\hat{u}}^\top,
% \label{eq:centered-features}
% \end{equation}
% where $H_k, H_u \in \R^{B \times d}$ stack the per-token $\hat{k}, \hat{u}$ components, $\boldsymbol{\mu}_{\hat{k}}, \boldsymbol{\mu}_{\hat{u}} \in \R^d$ are their column means, $\mathbf{1} \in \R^B$ denotes the all-ones vector, and $\| \cdot \|_F$ is the Frobenius norm. The independence loss is
% \begin{equation}
% \mathcal{L}_{\text{indep}} = \frac{1}{d^2 (B-1)} \, \big\| \Psi^\top \Phi \big\|_F^2.
% \label{eq:indep-loss}
% \end{equation}
% In practice, gradients flow only through the unknown representation $\Psi$; the known representation $\Phi$ is treated as a fixed input. Minimizing $\mathcal{L}_{\text{indep}}$ drives the cross-covariance between $\hat{k}$ and $\hat{u}$ toward zero, encouraging the two heads to encode complementary rather than redundant information of the hidden state.

\paragraph{Independence loss.} The reconstruction loss alone does not prevent the unknown head from encoding information that is already represented by the known concepts. To discourage such redundancy, we penalize the statistical dependence between the known and unknown representations using a normalized cross-covariance penalty in the spirit of the Hilbert-Schmidt Independence Criterion with a linear kernel~\citep{hsic_2009, hsic_2020}, following its use for concept decoupling in~\citet{sccbgm}. Over the $|\mathcal{M}|$ masked token positions in a minibatch, let
\begin{equation}
\Phi = H_k - \mathbf{1}\boldsymbol{\mu}_{\hat{k}}^\top, \qquad \Psi = H_u - \mathbf{1}\boldsymbol{\mu}_{\hat{u}}^\top,
\label{eq:centered-features}
\end{equation}
where $H_k, H_u \in \R^{|\mathcal{M}| \times d}$ stack the per-token $\hat{k}, \hat{u}$ components over the masked positions, $\boldsymbol{\mu}_{\hat{k}}, \boldsymbol{\mu}_{\hat{u}} \in \R^d$ are their column means, $\mathbf{1} \in \R^{|\mathcal{M}|}$ denotes the all-ones vector, and $\| \cdot \|_F$ is the Frobenius norm. The independence loss is
\begin{equation}
\mathcal{L}_{\text{indep}} = \frac{1}{d^2 (|\mathcal{M}|-1)} \, \big\| \Psi^\top \Phi \big\|_F^2.
\label{eq:indep-loss}
\end{equation}
In practice, gradients flow only through the unknown representation $\Psi$; the known representation $\Phi$ is treated as a fixed input. Minimizing $\mathcal{L}_{\text{indep}}$ drives the cross-covariance between $\hat{k}$ and $\hat{u}$ toward zero, encouraging the two heads to encode complementary rather than redundant information of the hidden state.

\paragraph{Final training objective.} The four losses are combined linearly with non-negative weights:
\begin{equation}
\mathcal{L} = \mathcal{L}_{\text{LM}} + \lambda_{\text{concept}} \mathcal{L}_{\text{concept}} + \lambda_{\text{rec}} \mathcal{L}_{\text{rec}} + \lambda_{\text{indep}} \mathcal{L}_{\text{indep}}.
\label{eq:combined-loss}
\end{equation}
The backbone and concept module are optimized jointly under this objective, where $\lambda_{\text{concept}}, \lambda_{\text{rec}}, \lambda_{\text{indep}}$ are hyperparameters. For both auxiliary losses $\mathcal{L}_{\text{rec}}$ and $\mathcal{L}_{\text{indep}}$, gradients are detached so that only the unknown head is updated.

\subsubsection{Training dynamics}
\label{sec:training-dynamics}

\begin{figure}[h]
    \centering
        \includegraphics[trim={3cm 3cm 3cm 3cm}, clip, width=\textwidth]{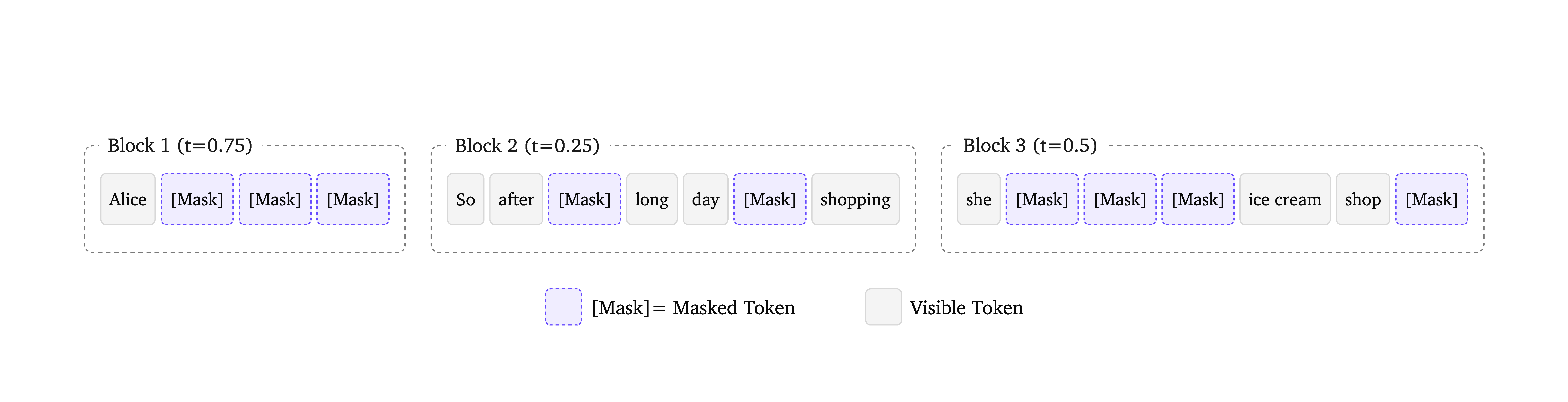}
\caption{Per-block masking schedule. Each block samples an independent noise level $t_b \sim \mathcal{U}(0,1)$, so blocks in the same sequence are denoised to different degrees within a single training step.}
    \label{fig:per-block-masking}
\end{figure}

\paragraph{Per-block masking schedule.} Standard masked diffusion models apply a single noise level $t \sim \mathcal{U}(0, 1)$ to the entire sequence at each training step. In the block-causal architecture (\cref{subsec:causal-diffusion}), we treat each block as an independent unit and sample a separate noise level $t_b \sim \mathcal{U}(0, 1)$ for each block $b$, so two blocks within the same context window can be at different stages of the denoising process simultaneously (\cref{fig:per-block-masking}). The model thus observes a richer distribution of partially-denoised contexts at every training step than under a single global noise level.

\paragraph{Concept teacher forcing schedule.} The known head's predictions are unreliable early in training, and even once they stabilize, routing $\bar{h}$ through the predicted activations $\hat{k}$ allows the language modeling loss to push those activations to encode information beyond the labeled concepts, a failure mode known as concept leakage~\citep{mahinpei2021promises}. \citet{cbm} address this by training the concept head independently of the downstream model and feeding ground-truth concepts forward at every step. We adopt the same substitution idea but apply it on a schedule: with probability $\alpha_{\text{known}}(s)$ at training step $s$, we replace the predicted known representation $\hat{k}$ with its ground-truth analogue $\hat{k}^{\text{GT}} = \sum_i k^{\text{GT}}_i K_i$ from \cref{eq:gt-targets} when forming $\bar{h}$, which we refer to as teacher forcing. We anneal $\alpha_{\text{known}}$ from $1$ at the start of training (full teacher forcing) to a smaller steady-state value, so the model relies progressively on its own predictions as the head becomes more accurate.

\paragraph{Unknown concept teacher forcing schedule.} The unknown head faces the same early-training instability as the known head, but no ground-truth labels exist to substitute for its predicted activations. Instead, we substitute the ideal target itself: from \cref{eq:gt-targets}, the unknown embedding should reconstruct $\hat{u}^{\text{GT}} = h - \hat{k}^{\text{GT}}$, which we can compute directly from the transformer hidden state and the labeled concepts. With probability $\alpha_{\text{unknown}}(s)$ at training step $s$, we replace the predicted $\hat{u}$ with $\hat{u}^{\text{GT}}$ when forming $\bar{h}$. As with $\alpha_{\text{known}}$, we anneal $\alpha_{\text{unknown}}$ from $1$ to a smaller steady-state value, so the language modeling head is shielded from a poorly-trained unknown head early on and learns to rely on the predicted unknown embedding as it becomes accurate.

A consolidated reference for the symbols introduced in this section is provided in \cref{appendix:architecture-notation}.

%% file: sections/attribution/main.tex
% \clearpage
\section{Interpretability capabilities}
\label{sec:capabilities}
The architecture of~\cref{sec:architecture} produces models that are both \emph{interpretable} and \emph{controllable}. Interpretability answers ``why a prediction was produced.'': through attribution, we identify the inputs, concepts, and training data associated with a prediction. Controllability is enabled through steering, which modifies the concept representation at inference time to control how selected concepts affect the model's behavior. We describe attribution and steering in the following sections.

\subsection{Attribution}
\label{sec:attribution}

Attribution asks why \steerling produced a given output. We approach it from three angles:

\begin{itemize}
    \item Input attribution identifies influential input tokens (\cref{sec:input-attribution}).
    \item Concept attribution traces influential internal concepts (\cref{sec:concept-attribution}).
    \item Training data attribution retrieves similar training examples (\cref{sec:training-data-attribution}).
\end{itemize}

\begin{figure}[!ht]
  \centering
  \includegraphics[trim={2cm 2cm 2cm 2cm}, clip, width=\textwidth]{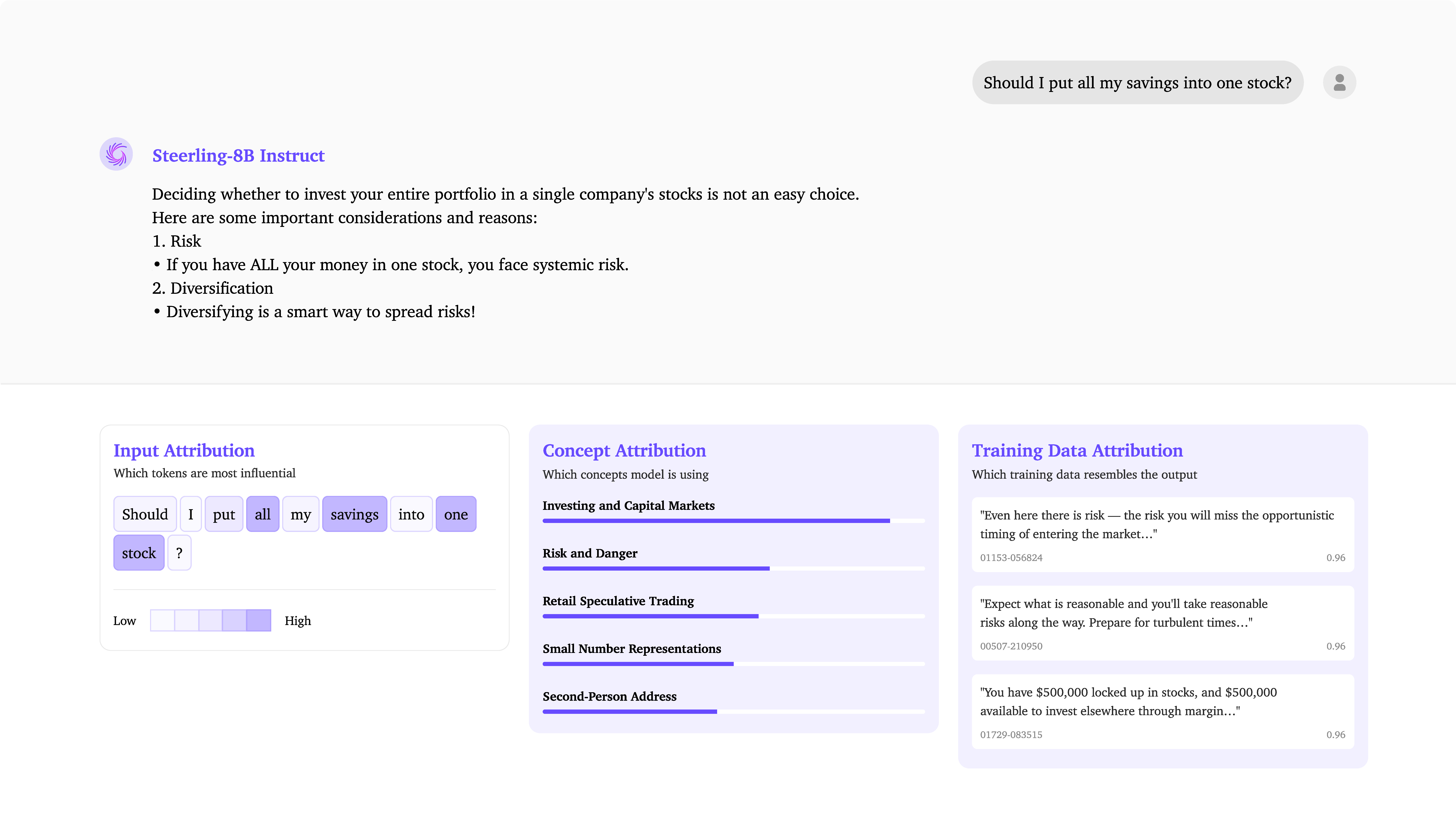}
  \caption{Three views of attribution for one output chunk. \textbf{Input attribution}
  (left): per-token Integrated Gradients scores over the prompt, with tokens that
  contribute most to the chunk shaded more intensely. \textbf{Concept attribution}
  (center): top contributing concepts, ranked by their chunk-level contribution.
  \textbf{Training data attribution} (right): retrieved training chunks, ranked by
  similarity to the chunk's representation and identified by chunk ID, which resolves
  to the source document for verification.}
  \label{fig:attribution}
\end{figure}

\cref{fig:attribution} shows the three 
methods applied to a single output chunk.

\subsubsection{Input attribution}
\label{sec:input-attribution}
Input attribution answers: \emph{which input tokens most influenced a given output?} We use Integrated Gradients (Equation~\ref{eq:integrated-gradients}) with the \texttt{[MASK]} embedding as the baseline. The baseline determines what the attribution \emph{measures}: how the output changes as the input moves from the baseline to its actual value. \texttt{[MASK]} makes that comparison meaningful for our model. The diffusion objective trains the model to predict masked tokens at every position, so \texttt{[MASK]} becomes a learned representation of ``no information at this position,'' and the integration runs from this learned ``absent'' state to the actual token along a path of states the model was trained on. \cref{fig:attribution}A shows the per-token scores for the highlighted chunk in this example.
Autoregressive models have no such trained baseline: zero embeddings and padding tokens are out-of-distribution states the model never learned to read as an absence of information.

We compute attribution using Integrated Gradients~\citep{integrated_gradient} on the token embeddings. For an input token $x^i$ with embedding $T_{x^i}$, we integrate the gradient of the output logit $\ell_y$ along the straight-line path from the baseline embedding $T_{\texttt{[MASK]}}$ to $T_{x^i}$, giving a score
\begin{equation}
  \phi(x^i, y)
  = \frac{1}{S} \sum_{s=1}^{S}
    \left(T_{x^i} - T_{\texttt{[MASK]}}\right)^\top
    \nabla_{T^{(s)}} \, \ell_y,
  \label{eq:integrated-gradients}
\end{equation}

\looseness=-1where $T^{(s)} = T_{\texttt{[MASK]}} + \tfrac{s}{S}(T_{x^i} - T_{\texttt{[MASK]}})$ is the $s$-th of $S$ interpolation points along that path. Replacing a token with \texttt{[MASK]} is an operation the model has performed countless times during training, so the Integrated Gradients path traces a well-defined direction through the model's learned representation space.

\subsubsection{Concept attribution}
\label{sec:concept-attribution}
Concept attribution answers: \emph{which internal concepts most influenced a given output?} The additive bottleneck of the concept module already decomposes the logit of any output token exactly into known-concept, unknown-concept, and residual terms (Equation~\ref{eq:logit-decomposition}). \cref{fig:attribution}B ranks the top contributing concepts for the highlighted chunk.

We report attribution at the chunk level, consistent with how concepts are supervised (\cref{subsec:training}). The contribution of a concept to a chunk is the sum of its per-token contributions over the tokens in that chunk:
\begin{equation}
  \Gamma^{\text{known}}_i = \sum_{t \in \text{chunk}} k_{i,t} \, K_i^\top W_{y_t},
  \qquad
  \Gamma^{\text{unknown}}_j = \sum_{t \in \text{chunk}} u_{j,t} \, U_j^\top W_{y_t},
  \label{eq:chunk-concept-attribution}
\end{equation}
where $k_{i,t}$ and $u_{j,t}$ are the activations of known concept $i$ and unknown concept $j$ at the token $y_t$ produced at position $t$, and an analogous sum gives the residual's contribution. Ranking concepts by these chunk-level contributions identifies which concepts drove the output, while the residual captures the part of the output that the concept inventory does not explain.

\subsubsection{Training data attribution}
\label{sec:training-data-attribution}

Training data attribution (TDA) answers: \emph{which training examples are most influential to a given output?} It enables applications such as alignment fine-tuning, factual provenance tracing, and auditing whether a model generalizes from proprietary fine-tuning data or falls back on pretraining knowledge. Unlike concept attribution, data attribution does not derive its scores from the model's own computation. We do not estimate the causal effect of removing a training example, as classical influence-function methods do~\citep{koh2017understanding}; at our scale the required Hessian inverse is intractable. Instead we frame attribution as \emph{approximate semantic-similarity retrieval}: given an output, we retrieve the training chunks whose representations are most similar to it.
A structural counterpart appears in PRISM~\citep{ley2026prototype}, which implements TDA as a lookup on a learned prototype layer rather than nearest-neighbor search in general representation space.
Our work is a scalable proxy for influence rather than a faithful or causal account. The retrieval view suits \steerling in particular: because its representations are aligned to human-interpretable concepts (\cref{sec:architecture}), semantically related chunks lie close together in representation space, so nearest-neighbor search tends to surface training chunks that share concepts with the output.

Because data attribution retrieves rather than computes, its central difficulty is representational. The query vector we extract from \steerling lives in the model's internal representation space, whereas the index stores corpus embeddings produced by a different model, so an output's internal representation cannot be matched against the index directly. We bridge them in two steps: forming a query representation from an output, then mapping it into the index's space. Given a model output, we segment it into chunks at the \texttt{[EOC]} token and forward each chunk through the language model, mean-pooling over its token positions to obtain a single vector that combines the known-head, unknown-head, and residual components of the hidden state. A learned \emph{projection}---a small MLP trained with a cosine-similarity objective---maps this internal vector into the corpus embedding space of the index built in \cref{sec:training-data-indexing}, placing an output and its related training chunks close together. The projected query is then matched against the index by approximate nearest-neighbor search, and the nearest chunks are returned as the attributed sources. \cref{fig:attribution}C shows the top-ranked retrieved chunks for the highlighted output chunk, ranked by cosine similarity; their source URLs can be inspected for verification. %We give details of the embedding model, vector dimensions, pooling, and projection architecture in \cref{app:training-data-indexing}.

\subsection{Steering}
\label{sec:model-control}

Steering is the ability to control model output without using prompts. This is enabled by the concept module and is not possible in other existing models. We steer generation at inference time, without updating any weights, by acting directly on the concept representations the model already uses. Steering can push generation toward a target concept (amplification) or away from it (suppression), and can be applied to both the known concepts supervised during training and the unknown concepts the model discovers on its own.

\subsubsection{Steering operation}
\label{sec:steering-operation}

Our model learns a direction (or embedding) for each concept and uses it internally to form predictions. We can steer the model toward a given concept by injecting the direction the model has already learned for it into the hidden representation. Injecting a direction into the hidden states to steer generation is a well-established technique~\citep{subramani2022extracting,  turner2023activation, rimsky2024steering, zou2023representation}, but these methods must first extract the direction post-hoc from a trained model's activations, so it is only an estimate of how a concept is represented and varies with the procedure used to recover it. Our architecture is better suited to steering because the direction is not estimated: each concept's embedding $K_c$ is a model parameter the model itself uses, so we inject a direction the model already relies on rather than one fit after the fact.

Concretely, we take the concept embedding $K_c$ and normalize it to unit length, giving a steering direction $e_c = K_c / \|K_c\|_2$. To steer toward several concepts at once, we sum their embeddings then normalize. We then add this direction, scaled by a steering strength $\gamma$, to the transformer hidden state at every masked position, at every layer from $L_{\mathrm{inj}}$ onward. The steering signal accumulates as the representation propagates toward the bottleneck:
\begin{equation}
  h^{(l)}_t \;\leftarrow\; h^{(l)}_t + \gamma \, e_c, \qquad l \geq L_{\mathrm{inj}}.
  \label{eq:concept-injection}
\end{equation}

\looseness=-1Because each concept aligns with the LM head differently, the effect of a given $\gamma$ is not comparable across concepts: the same strength shifts the output logits more for some concepts than others.  We therefore calibrate $\gamma$ per concept so that its largest effect on any output token equals a fixed target $\tau$:
\begin{equation}
\gamma = \frac{\tau}{\mathrm{peak}(e_c)},
\qquad
\mathrm{peak}(e_c) = \max_{y \in V} \, e_c^\top W_y,
\label{eq:gamma-calibration}
\end{equation}

where $\mathrm{peak}(e_c)$ is the largest logit shift the direction $e_c$ can induce over the vocabulary $V$. A single global $\tau$ then gives adaptive steering strength across concepts without per-concept tuning.

\subsubsection{Steering direction}
\label{sec:steering-direction}
The sign of the steering strength $\gamma$ sets the direction of the intervention. Amplification pushes generation toward the target concept, while suppression pushes it away.

\paragraph{Amplification.}
Amplification takes $\gamma > 0$, so the injection of \cref{eq:concept-injection} moves the hidden state along $e_c$, toward the concept $c$. The concept direction is added before the bottleneck, propagating the boosted activation through the concept module then the LM head to increase the logits of concept-expressing tokens.

\begin{figure}[htbp!]
\centering
\includegraphics[width=0.85\linewidth]{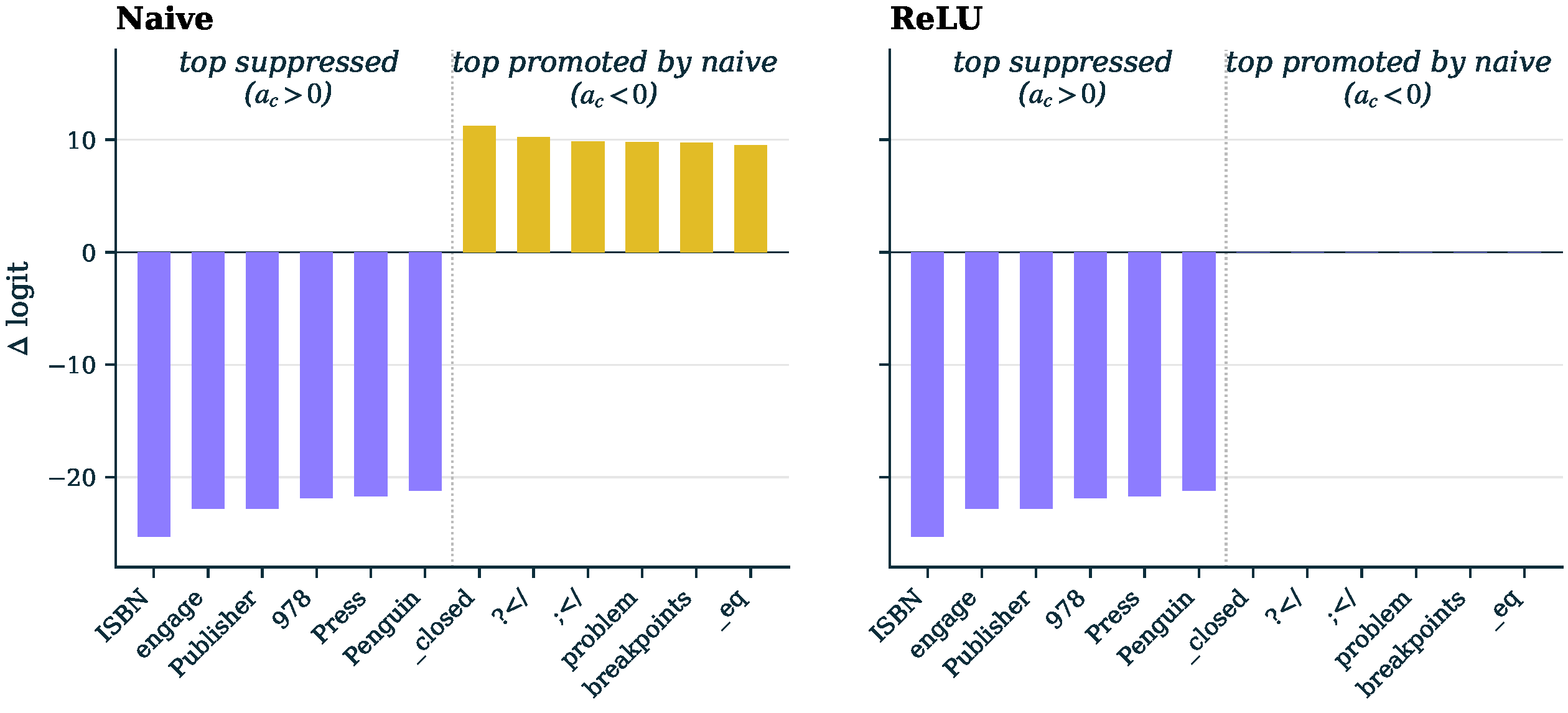}
\caption{
ReLU-gated concept suppression avoids unintended token promotion during negative steering.
Naive subtraction suppresses tokens positively aligned with the target concept, but promotes tokens with negative alignment.
For an \textit{academic and commercial publishing} concept, this boosts unrelated tokens such as \textit{problem} and \textit{breakpoints}.
The ReLU-gated update preserves suppression of publishing-related tokens while leaving anti-aligned tokens unchanged.
% \aya{colors and fonts need to be consistant with other}}
}
\label{fig:relu-logit-mask}
\end{figure}

\paragraph{Suppression.}
Suppression eliminates text associated with a target concept $c$. We combine two mechanisms to ensure the concept is properly suppressed: a negative hidden-state injection (\cref{eq:concept-injection} with $\gamma < 0$) that pushes the transformer representation away from $c$, and a ReLU-gated logit mask that subtracts $c$'s contribution from the LM head output. 
We explain why we use the ReLU gate below.
% The combination is more effective than either alone.

Since the LM head is a linear projection on top of the sum of concept embeddings, the contribution of concept $c$ to the output is its alignment with each vocabulary token:
\begin{equation}
  a_c = W e_c \in \R^{|V|},
  \qquad a_c[v] = W_v^\top e_c,
  \label{eq:concept-alignment}
\end{equation}
where $a_c[v]$ is the concept's contribution to the logit of token $v$. 
Subtracting this contribution directly ($\ell_v \to \ell_v - s \, a_c[v]$ with suppression strength $s > 0$) will suppress tokens with $a_c[v] > 0$ as intended, but simultaneously \emph{promotes} tokens with $a_c[v] < 0$, making unrelated tokens that happen to be anti-aligned with the concept (\cref{fig:relu-logit-mask}) dominate generation. 
We gate the subtraction with a ReLU so that only positively aligned tokens are affected:
\begin{equation}
  \ell_v \;\longrightarrow\; \ell_v - s \cdot \mathrm{ReLU}\big(a_c[v]\big).
  \label{eq:relu-logit-mask}
\end{equation}
This suppresses the tokens aligned with the concept while leaving anti-aligned tokens untouched.

%% file: sections/interpretability_metrics/main.tex
% \clearpage

\section{Interpretability Metrics}
\label{sec:interp_metrics}
This work is built on the claim that interpretability is best treated as a design constraint rather than a post-hoc analysis problem. If that claim is correct, then a model trained with interpretability built into its architecture and objectives should be more interpretable than one trained without, not in some abstract sense, but in ways one can measure directly. This section makes that claim measurable. The metrics defined below test the properties the concept module of~\cref{subsec:concept-module} was designed to deliver.

The concept module was trained to detect a set of concepts in context, via the concept loss. The model was trained to route its predictions through those concepts rather than through the residual, via the reconstruction loss. The known and unknown concepts were trained to be disentangled, so they do not encode the same information, via the independence loss. Each of these properties has a corresponding metric, and each metric tests a necessary condition for the architecture to be working as built. A fourth metric asks whether the learned concept embeddings  point at semantically related tokens; this property is not directly supervised by any training objective, and is included to test whether the architecture produces token-level interpretability beyond what was directly optimized for.

These metrics are familiar from the broader interpretability literature, where post-hoc methods such as sparse autoencoders, probes, and steering vectors are evaluated along similar axes of concept detection and disentanglement~\citep{wu2025axbench, bhalla2024towards}. These post-hoc methods exist because representations in standard transformers are in superposition~\citep{elhage2022superposition, bricken2023monosemanticity}:  more learned directions than dimensions, packed into overlapping subspaces that have to be recovered after training. The concept module sidesteps superposition by construction:  each concept has its own learned embedding, so the metrics do not have to ask whether a recovered direction behaves like a concept, but whether the concepts the model was trained on are doing the work the architecture assigned to them.
\begin{table}[htbp!]
\centering
\small
\begin{tabular}{@{}lcl@{}}
\toprule
\textbf{Metric} & \textbf{Direction} & \textbf{What it measures} \\
\midrule
Concept Loss               & $\downarrow$ & Does the concept module detect the right concepts? \\
Concept Independence Loss  & $\downarrow$ & Are known and unknown heads disentangled? \\
Concept Contribution       & $\uparrow$   & Do predictions route through the concept module? \\
Known Concept Alignment    & $\uparrow$   & Do concept embeddings point at related tokens? \\
\bottomrule
\end{tabular}
\caption{Interpretability metrics.  $\uparrow$ or $\downarrow$ indicates the direction of better performance.}
\label{tab:metrics-summary}
\end{table}

\paragraph{Concept Loss.} The concept loss (Equation~\ref{eq:concept-loss}) measures whether the concept module assigns the right concepts to a given chunk. The metric is the OR-aggregated binary cross-entropy between predicted concept presence (Equation~\ref{eq:chunk-aggregation}) and ground-truth chunk-level labels, computed over a held-out validation set of chunks annotated by the pipeline of \cref{sec:data}. A low value indicates that the concept module identifies concepts in unseen text in the same way it learned to identify them during training.

\paragraph{Concept Independence Loss.} The independence loss (Equation~\ref{eq:indep-loss}) measures the linear statistical dependence between the known and unknown concept representations. The metric is computed over per-token $\hat{k}$ and $\hat{u}$ on a held-out validation set. A low value indicates that the two representations carry linearly independent information. A linear kernel was chosen because the bottleneck composes its components additively, $\bar{h} = \hat{k} + \hat{u} + \varepsilon$, and the LM head is linear, so every output logit decomposes linearly into contributions from each pathway (Equation~\ref{eq:logit-decomposition}). Linear independence between $\hat{k}$ and $\hat{u}$ is therefore the exact property concept attribution requires: if the two representations are linearly independent in expectation, their contributions to any logit are additively separable, and the known and unknown pathways do not encode overlapping information that would inflate one attribution at the expense of the other.

\paragraph{Concept Contribution.} The concept contribution metric measures how much of each output prediction is explained by the concept module rather than by the residual. From the logit decomposition in \cref{eq:logit-decomposition}, the contribution of the concept module to the logit of any token $y$ is the sum of the known and unknown terms; the residual term $\varepsilon^\top W_y$ denotes what has not been captured by the concept module. The metric is the fraction of total logit magnitude attributable to the concept module:

\begin{equation}
\label{eq:concept-contribution}
\text{Concept Contribution} = \frac{|\hat{k}^\top W_y| + |\hat{u}^\top W_y|}{|\hat{k}^\top W_y| + |\hat{u}^\top W_y| + |\varepsilon^\top W_y|},
\end{equation}

averaged over predictions on a held-out validation set. A high value indicates that predictions can be mostly attributed to the concept module. The reconstruction loss (Equation~\ref{eq:rec-loss}) creates training pressure on this property by driving $\varepsilon \to 0$, but the metric is computed at the logit level rather than on $\|\varepsilon\|$ directly, so it tests the property that matters for attribution faithfulness: how much of the prediction are the concepts responsible for.

\paragraph{Known Concept Alignment.} Each known concept has a learned embedding $K_c \in \mathbb{R}^d$ that the LM head projects into vocabulary space, so the top-$k$ tokens for concept $c$ are $T_k(c) = \mathrm{TopK}(W K_c)$. The known concept alignment metric asks whether these top tokens semantically match the human-assigned label and description that the concept was originally annotated with by the pipeline of \cref{sec:data}. An LLM-judge is given the concept's label, its description, and the top tokens, and rates the alignment on a 1-5 scale, with 5 indicating that the top tokens unambiguously represent the named concept and 1 indicating no relationship. Details on the judge and the full prompt are given in \cref{app:judge-known-alignment}. The metric is the mean rating over a randomly sampled subset of concepts from the library.

Known concept alignment is not directly supervised by the training objective. The concept loss (Equation~\ref{eq:concept-loss}) supervises chunk-level concept presence using positive-only labels; it does not connect a concept embedding $K_c$ to the specific tokens that lexically express concept $c$. The language modelling loss supervises next-token prediction; it does not connect token predictions to the concept vocabulary. Any alignment between $K_c$ and semantically-related tokens therefore arises from the joint optimization of these two objectives, not from a direct training signal.

%% file: sections/scaling_laws/main.tex
% \clearpage
\section{Scaling laws}
\label{sec:scaling-laws}

A long-standing concern in the interpretability literature is that constraining a model's representations to be human-understandable necessarily costs capability~\citep{rudin2019stop, doshivelez2017rigorous, cbm}. A second concern, less studied, is that even when interpretability properties hold at small scale, there is no guarantee they survive at the scales where models actually get deployed~\citep{wei2022emergent}. This section tests both concerns empirically. Two model families, autoregressive (AR) and causal diffusion (CDLM), together with their interpretable counterparts, are trained across three orders of magnitude of compute. Scaling laws are fit to the resulting checkpoints for capability (compute-optimal loss) and for the interpretability metrics defined in \cref{sec:interp_metrics}, and each fit is then used to extrapolate to \steerling, a CDLM+Concept model trained at 8B parameters and 1.35T tokens. Specifically, this section addresses two research questions:

\begin{itemize}
\item \textbf{RQ1.} Do inherently interpretable architectures preserve compute-optimal scaling?
\item \textbf{RQ2.} Do the interpretability properties scale favorably with compute?
\end{itemize}

Across model families and three orders of magnitude of compute, adding the concept module to either backbone shifts the compute-optimal scaling exponents by \textbf{a small, fixed per-backbone offset}. The interpretability metrics defined in \cref{sec:interp_metrics} \textbf{improve favorably with compute on both backbones}, with all four metrics scaling in the expected direction. Extrapolating the small-scale fits to \steerling{} \textbf{predicts the deployed model's validation loss within $0.11$ nats} and bounds its interpretability metrics within their natural ranges.
A consolidated reference for the symbols introduced in this section is provided in \cref{appendix:scaling-notation}.

\subsection{Setup}

\label{subsec:scaling-setup}

\paragraph{Model families.} Four model families are compared. \textbf{AR} is a standard autoregressive transformer;  \textbf{CDLM} is the causal block-diffusion model defined in \cref{subsec:causal-diffusion};  and \textbf{AR+Concept} and \textbf{CDLM+Concept} are their inherently interpretable counterparts, obtained by inserting the concept module of \cref{subsec:concept-module} between the transformer hidden state and the LM head. The two backbones are held fixed across all experiments; only the presence of the concept module changes. The concept library used by the +Concept families is the one constructed in \cref{sec:data}, fixed across all model sizes.

\paragraph{IsoFLOP sweep.}
Each family is trained across four IsoFLOP slices, with slice targets reported in \cref{tab:isoflop-slices}. Each slice contains four to six model sizes whose token counts are adjusted so that their total compute lands within $\pm 15\%$ of the slice target. The slice targets differ between base and +Concept families: at the smallest backbones, the concept module's parameter overhead exceeds the backbone itself, making compute-optimal estimates unreliable in the lowest IsoFLOP slice. Slightly higher targets are therefore used for the +Concept families. Full architecture configurations and slice targets are reported in \cref{app:scaling_laws_arch_table}.

\paragraph{Training.}
All runs share the same data, optimizer, and schedule. Pretraining is on the real-data subset of Nemotron-CC-HQ~\citep{nemotron} at sequence length $N = 4096$. Optimization uses AdamW under a warmup-stable-decay (WSD) schedule~\citep{wsd}: a short linear warmup, an extended stable phase at peak learning rate, and a final $20\%$ linear decay to zero. Optimizer hyperparameters, tokenizer, and per-family configurations are reported in \cref{app:scaling_laws_arch_table}.

Each IsoFLOP checkpoint is annealed independently rather than sharing a single annealing tail. The $80\%$ stable-phase trajectory is treated as a reservoir of starting points; a $20\%$ linear decay is run from each one to its final loss. This is more expensive than estimating annealed losses from non-annealed checkpoints~\citep{scaling_behaviorDLM_von}, but it yields a dense compute-loss sweep without a separate full run per slice. The cheaper estimation procedure produced inconsistent results across model families and sizes (\cref{app:scaling_laws_annealing}).

\paragraph{Parameters and FLOPs calculation.}
Following~\citet{bi2024deepseek}, per-token forward+backward FLOPs for the AR and CDLM backbones are
\begin{equation}
\label{eq:flops-base}
M_{\text{base}} = 6P + 12 L d N,
\end{equation}
where $P$ is the non-embedding parameter count, $L$ the number of layers, $d$ the hidden dimension, and $N$ the sequence length. Total training FLOPs are $C = M \cdot D$ for $D$ training tokens.

The +Concept families add two heads acting on every token (\cref{subsec:concept-module}). The known head scores all $n$ concepts through a predictor of size $d \times n$, then composes the result through a separate embedding matrix of the same size via top-$k_{\text{known}}$ selection. The two matmuls give a per-token cost of $2 d n$. The unknown head is factorized through low-rank embeddings: a down-projection $d \to R$, a predictor over $m$ unknown concepts, top-$k_{\text{unknown}}$ selection, and composition through factorized embeddings, for a per-token cost of $2 d R + R m$. Combined:
\begin{equation}
\label{eq:flops-concept}
M_{\text{+Concept}} = \underbrace{6P + 12 L d N}_{\text{backbone}} + \underbrace{6\bigl(2 d n\bigr)}_{\text{known head}} + \underbrace{6\bigl(2 d R + R m\bigr)}_{\text{unknown head}}.
\end{equation}
Concept module hyperparameters and per-backbone parameter counts are reported in \cref{app:scaling_laws_arch_table}.

\begin{figure}[htbp!]
\centering
\includegraphics[width=0.45\linewidth]{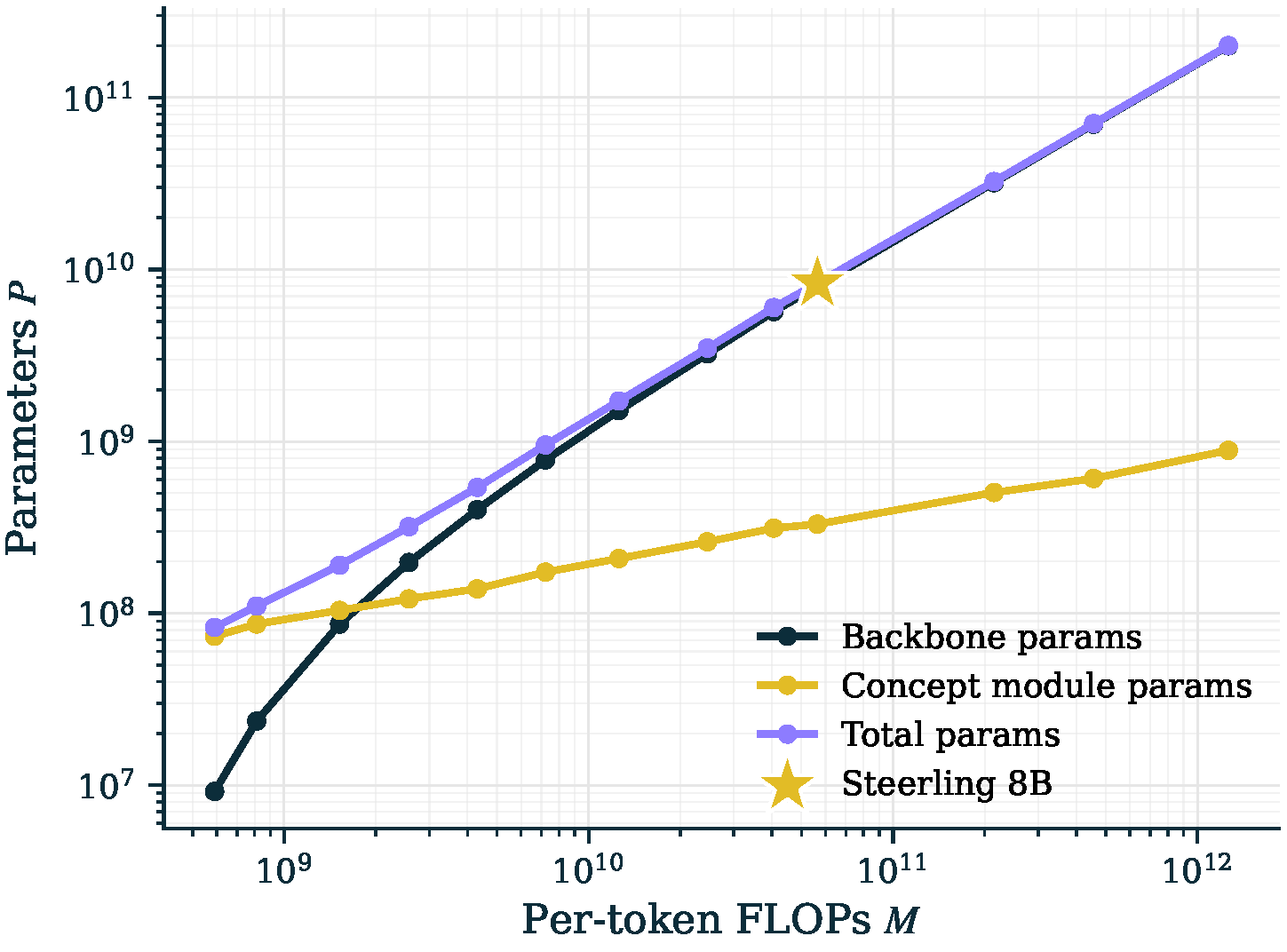}
\caption{Concept module overhead as a fraction of total parameters, vs.\ backbone size.}
% Module scales as $O(D)$, backbone as $O(D^2 L)$.}
\label{fig:cbm-overhead}
\end{figure}

\textbf{The parameter cost of the concept module shrinks rapidly with scale.} For a fixed concept library, the concept module adds a parameter overhead of $O(d)$, while backbone parameters scale as $O(d^2 L)$. The relative overhead therefore decays rapidly: the concept module accounts for $\sim 89\%$ of the total parameter count at 10M, $\sim 4\%$ at \steerling, and under $1\%$ at frontier scales (\cref{fig:cbm-overhead}).

\subsection{Compute-Optimal scaling}
\label{subsec:compute-optimal}

Compute-optimal scaling is fit per family on the checkpoints from \cref{subsec:scaling-setup}, yielding a set of $(P_i, D_i, \mathcal{L}_i)$ triples per family giving parameter count, training tokens, and validation loss, organized into four IsoFLOP slices. The exponents are then compared pair by pair across each backbone's interpretable and non-interpretable variant, allowing a direct test of \textbf{RQ1: do inherently interpretable architectures preserve compute-optimal scaling?}

\subsubsection{Methodology}
\label{subsubsec:co-methodology}

\paragraph{Validation losses.} For autoregressive models (AR, AR+Concept), $\mathcal{L}_i$ is validation cross-entropy on a held-out subset of the Nemotron pretraining corpus. For diffusion models (CDLM, CDLM+Concept), $\mathcal{L}_i$ is a Monte Carlo estimate of the masked diffusion model (MDLM) ELBO bound on the per-token negative log-likelihood~\citep{scaling_behaviorDLM_von, scaling_beyond_sahoo, mdlm}. Per batch, a single noise level $t \sim \mathcal{U}(10^{-3}, 1 - 10^{-3})$ is sampled and shared across the batch, each token position is independently masked with probability $t$, a forward pass is run, and the mean cross-entropy on the masked positions is computed. The validation loss is the average over batches. Note that this estimator integrates over $t \in (0, 1)$, while training samples $t \sim \mathcal{U}(0.05, 0.95)$; the two are therefore unbiased estimators of the same MDLM objective on slightly different intervals and will not match numerically. Details on the ELBO estimator and Monte Carlo sampling are given in \cref{app:scaling_laws_ELBO}.

\paragraph{Step 1: Per-slice parabola in $\log P$.}
Within each IsoFLOP slice (fixed target $C$), the per-size $(P_i, \mathcal{L}_i)$ pairs are fit to a parabola in $\log P$~\citep{hoffmann2022training}:
\begin{equation}
\mathcal{L}(\log P;\, C) = a(C)\,\bigl(\log P - \log P^*(C)\bigr)^2 + \mathcal{L}^*(C),
\label{eq:parabola}
\end{equation}
where $a(C)$, $\log P^*(C)$, and $\mathcal{L}^*(C)$ are free parameters, fit by nonlinear least squares (Levenberg-Marquardt). The fitted $\log P^*(C)$ locates the parameter count that minimizes loss at compute budget $C$, and $\mathcal{L}^*(C)$ is the corresponding loss. Doing this once per slice produces four $(C_i, P^*_i, \mathcal{L}^*_i)$ triples per family.

\looseness=-1\paragraph{Step 2: Power laws across compute budgets.} With the four per-slice triples, we fit two power laws,
\begin{equation}
P^*(C) = a_P\, C^{\alpha_P}, \qquad \mathcal{L}^*(C) = a_L\, C^{\alpha_L},
\label{eq:powerlaws}
\end{equation}
% each by ordinary least squares on log-transformed data: $\alpha_P$ and $\log a_P$ are the slope and intercept of $\log P^*$ regressed on $\log C$, and analogously for $\alpha_L$, $\log a_L$. Since per-checkpoint compute satisfies $C = M(P) \cdot D$ (Equation~\ref{eq:flops-base} and Equation~\ref{eq:flops-concept} ), the compute-optimal token count exponent is determined by $\alpha_D = 1 - \alpha_P$. The exponents $\alpha_P$, $\alpha_D$, and $\alpha_L$ characterize how the compute-optimal parameter count, training token budget, and loss scale with available compute. \nathaniel{so, $\alpha_D = 1 - \alpha_P$ becomes a good approximation eventually, but at the scales we have here, its anywhere from 20\% off to  ~170\% off. With the concept head, it ends up being ~500\% off I think? And then in step 3 we compute it without tying the 2 together at all, with both free, so I think we can just eliminate this or have a footnote mentioning that at our scales it isnt true, or something.}

each by ordinary least squares on log-transformed data: $\alpha_P$ and $\log a_P$ are the slope and intercept of $\log P^*$ regressed on $\log C$, and analogously for $\alpha_L$, $\log a_L$. The exponents $\alpha_P$ and $\alpha_L$ characterize how the compute-optimal parameter count and loss scale with available compute; the compute-optimal token exponent $\alpha_D$ is estimated from the joint fit of Step~3.

\paragraph{Step 3: Joint parametric fit for the irreducible loss.}
Steps 1 and 2 yield $\alpha_P$ and $\alpha_L$ but no irreducible-loss term. Following \citet{hoffmann2022training, quokka}, the joint loss surface
\begin{equation}
\mathcal{L}(P, D)= \mathcal{L}_\infty + \frac{A_P}{P^{\alpha}} + \frac{A_D}{D^{\beta}}
\label{eq:chinchilla}
\end{equation}
is fit on all $(P_i, D_i, \mathcal{L}_i)$ observations pooled across the four IsoFLOP slices, not just the slice minima. The five free parameters are $\mathcal{L}_\infty$, $A_P$, $A_D$, $\alpha$, $\beta$. The Huber loss on log-residuals is minimized,
\begin{equation}
\mathcal{L}_{\text{fit}}(\mathcal{L}_\infty, A_P, A_D, \alpha, \beta) = \sum_i \mathrm{Huber}_\delta\!\bigl(\log \mathcal{L}(P_i, D_i) - \log \mathcal{L}_i\bigr),
\end{equation}
\looseness=-1with $\delta = 10^{-3}$, using L-BFGS-B with grid-search initialization. The irreducible loss $\mathcal{L}_\infty$ and the data exponent $\alpha_D = \alpha / (\alpha + \beta)$ are reported from this fit. The IsoFLOP estimates of Step~2 remain the primary $\alpha_P$ and $\alpha_L$ values; the joint fit contributes the irreducible-loss term and the data exponent $\alpha_D$.

% \paragraph{Confidence intervals.}
% Confidence intervals on $\alpha_P$ and $\alpha_L$ come from parametric bootstrap with residual resampling \nathaniel{I think parametric is for when you have a noise model, and you did residual resampling. Drop word parametric.} on the per-slice parabola fits of Step~1: residuals are resampled with replacement and added to the fitted values, the parabolas are re-fit on each synthetic dataset, and the power laws are re-fit on the resulting slice minima. Confidence intervals on $\mathcal{L}_\infty$ and $\alpha_D$ come from parametric \nathaniel{and here} bootstrap with log-residual resampling on the joint fit of Step~3, with each iteration re-initialized from grid search. All confidence intervals are reported at the 90\% level over 10{,}000 bootstrap iterations.

\paragraph{Confidence intervals.}
Confidence intervals on $\alpha_P$ and $\alpha_L$ come from bootstrap with residual resampling on the per-slice parabola fits of Step~1: residuals are resampled with replacement and added to the fitted values, the parabolas are re-fit on each synthetic dataset, and the power laws are re-fit on the resulting slice minima. Confidence intervals on $\mathcal{L}_\infty$ and $\alpha_D$ come from bootstrap with log-residual resampling on the joint fit of Step~3, with each iteration re-initialized from grid search. All confidence intervals are reported at the 90\% level over 10{,}000 bootstrap iterations.

\subsubsection{Results}
\label{subsubsec:co-results}

\begin{figure}[htpb!]
\centering
\includegraphics[width=0.8\linewidth]{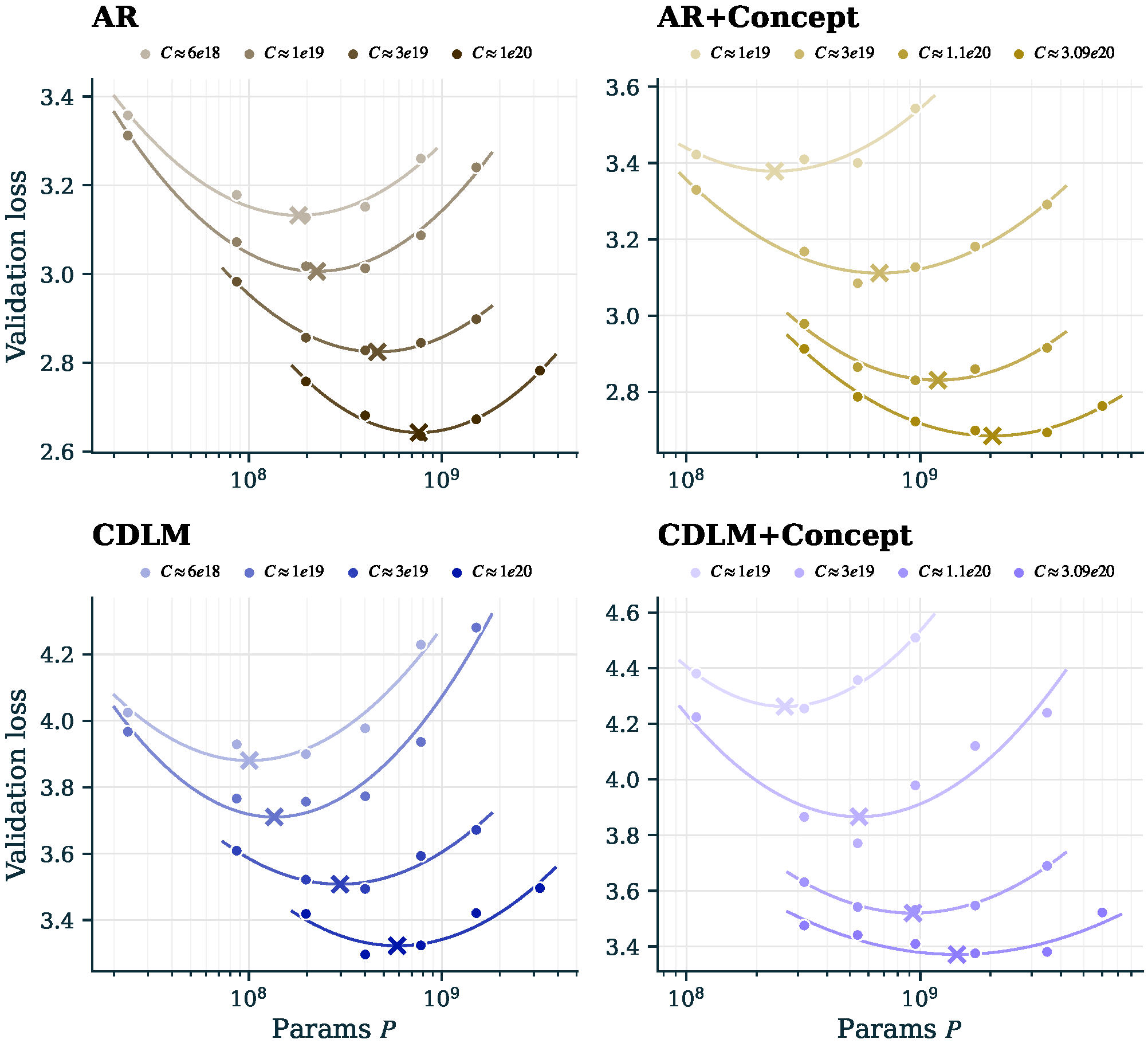}
\caption{IsoFLOP analysis for each family. Each curve fits a parabola to model sizes within a fixed compute target $C$; markers show measured validation losses; $\times$ markers locate the per-slice minima $P^*(C)$.}
\label{fig:isoflop-grid}
\end{figure}

\cref{fig:isoflop-grid} shows the per-slice parabolic fits for each of the four families. Markers show measured validation losses, and $\times$ markers locate the per-slice minima $P^*(C)$. The fits are well-behaved across all four families: the parabolas tighten with increasing compute, and the per-slice minima move smoothly toward larger $P$ as $C$ grows.

\begin{figure}[htbp!]
\centering
\includegraphics[width=0.95\linewidth]{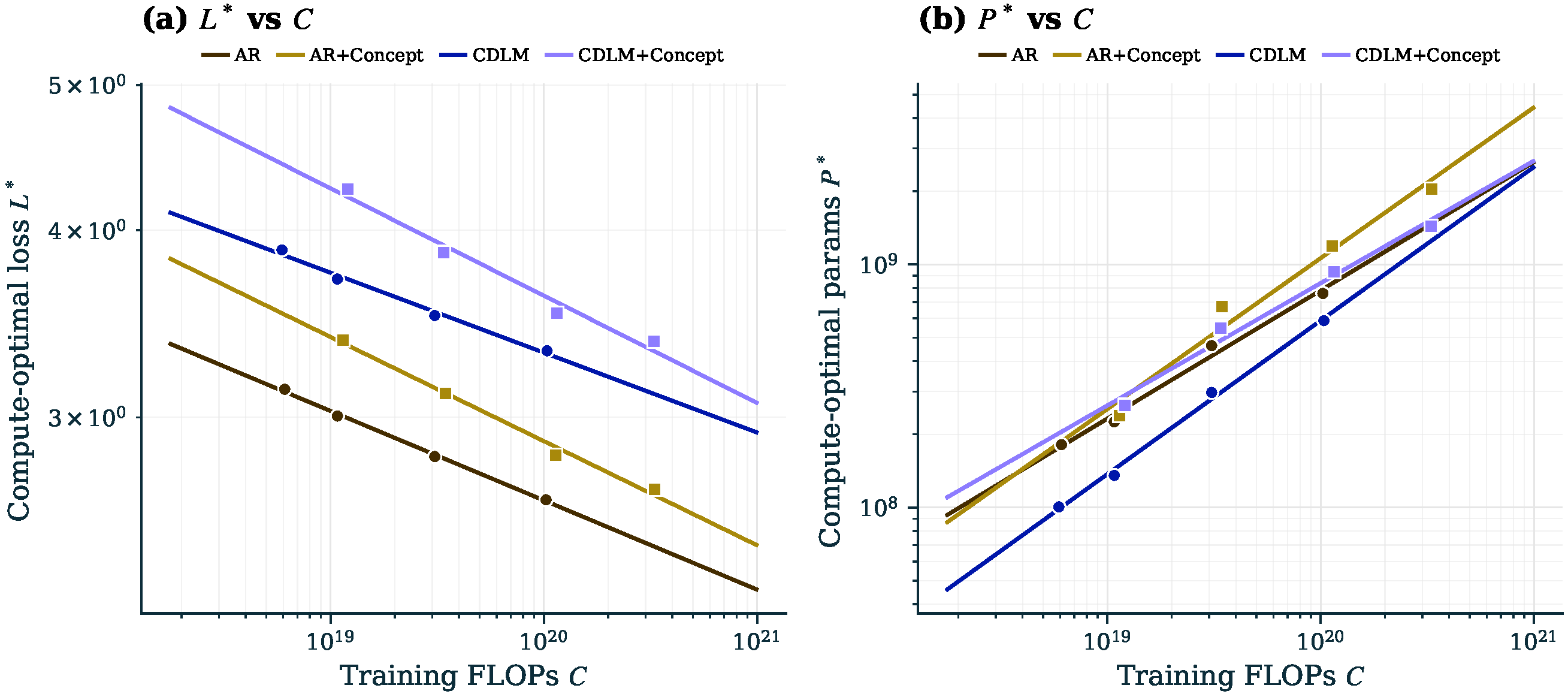}
\caption{Compute-optimal scaling laws for each family. Panel (a): $\mathcal{L}^*(C)$ from the Step~2 power-law fits. Panel (b): $P^*(C)$ from the Step~2 power-law fits. Markers show per-slice minima from \cref{fig:isoflop-grid}.}
\label{fig:scaling-laws}
\end{figure}

The resulting power-law fits across compute are shown in \cref{fig:scaling-laws}, with exponents and irreducible-loss asymptotes for all four families reported in \cref{tab:scaling-exponents}. The autoregressive baseline gives $\alpha_P = 0.528$, in line with prior estimates of $0.49$, $0.464$, and $0.524$ reported by \citet{hoffmann2022training}, \citet{shuai2024scaling}, and \citet{bi2024deepseek} respectively. The fit also yields $\mathcal{L}_\infty = 1.857$, comparable to Chinchilla's $1.69$. The diffusion baseline gives $\alpha_P = 0.632$, on the high end of prior masked-diffusion estimates of $0.514$, $0.566$, and $0.634$ reported by \citet{quokka}, \citet{scaling_behaviorDLM_von}, and \citet{mdlm} respectively. The higher value may result from the causal block attention mask (\cref{subsec:causal-diffusion}). Validation is reported in ELBO to follow prior work, although the choice of validation loss estimator can dramatically affect the fitted $\mathcal{L}^*$ values (\cref{app:scaling_laws_ELBO}). \citet{scaling_behaviorDLM_von} additionally reports $\alpha_P$ ranging from $0.535$ to $0.589$ across diffusion families and notes that the validation-loss ranking these exponents induce does not match the downstream performance ranking. The CDLM baseline yields $\mathcal{L}_\infty = 2.658$, modestly above Quokka's $2.41$ but within the variation attributable to the ELBO estimator choice. The goal of this section is not to rank diffusion families against each other, but to measure the effect of adding the concept module on top of a fixed backbone trained on the same data; the comparisons that follow are therefore within-pair rather than cross-family.

Adding the concept module shifts the compute-optimal scaling exponents by a small, fixed per-backbone offset. For both pairs, $\alpha_L$ becomes more negative when the concept module is added. This direction is consistent with the fixed-library bias: the smallest +Concept models are over-parameterized by the concept library and under-utilize their parameters, biasing their losses upward and steepening the apparent loss-vs-compute slope. The shifts on $\alpha_P$ are small in both pairs. The irreducible-loss asymptote $\mathcal{L}_\infty$ shifts downward in both pairs ($1.857 \to 1.193$ for AR, $2.658 \to 1.942$ for CDLM); within-pair $\mathcal{L}_\infty$ comparisons are valid since both members of each pair use the same loss methodology.

\keypoint{\textbf{Interpretability-by-design imposes a small per-backbone offset on compute-optimal scaling, not a scaling tax.}}

\input{sections/scaling_laws/tables/scaling_exponents.tex}

\begin{figure}[htbp!]
\centering
\includegraphics[width=0.7\linewidth]{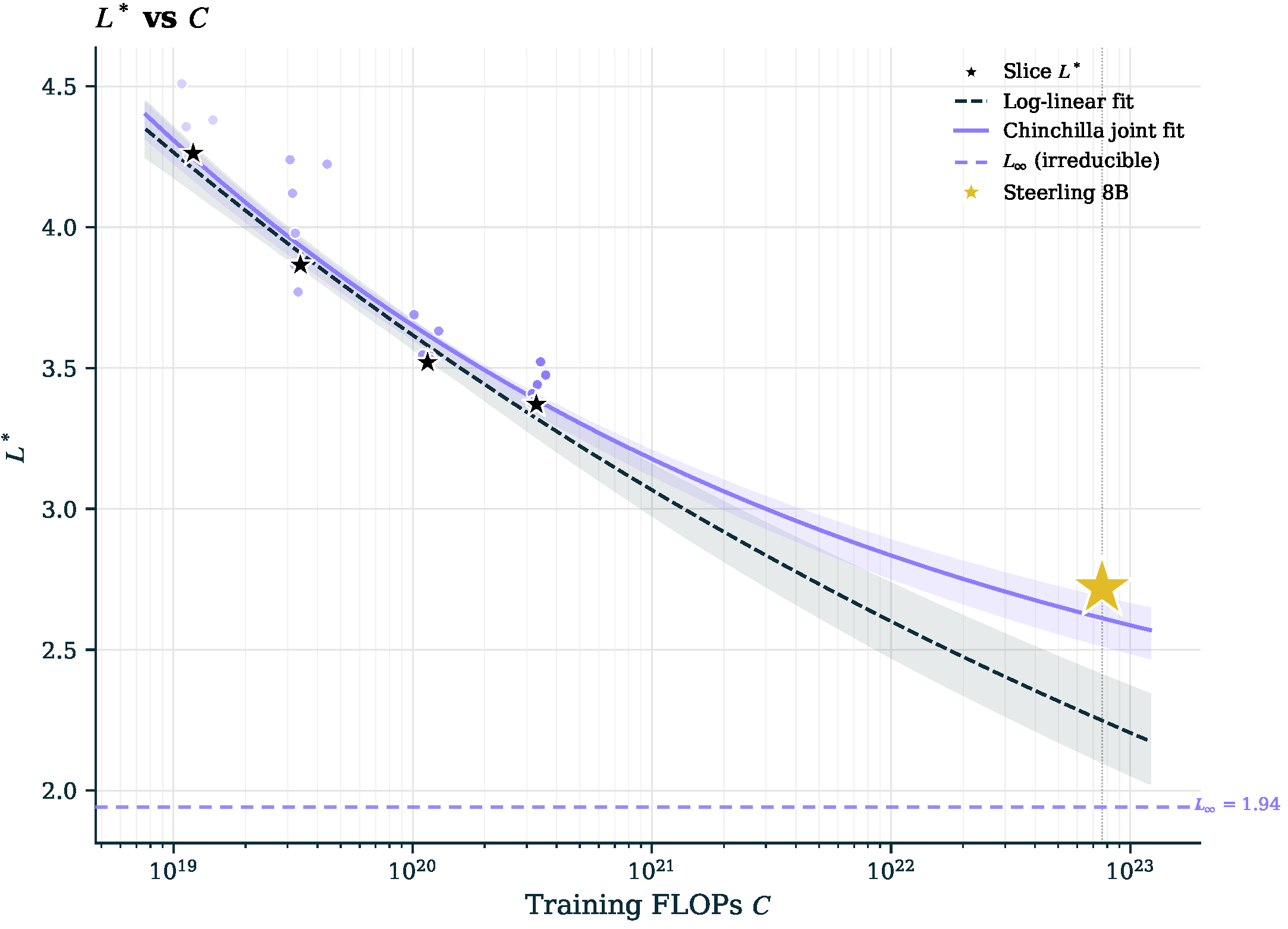}
\caption{Extrapolating CDLM+Concept to \steerling. Joint Chinchilla fit (solid, $\mathcal{L}_\infty$ dashed) and log-linear fit (dashed) on slice $\mathcal{L}^*$ minima. \steerling{} (filled star) lands within $0.11$ nats of the joint fit.}
\label{fig:8b-validation}
\end{figure}

\paragraph{Predicting \steerlingB from small models. }
The fitted exponents for the +Concept families are likely steeper than the asymptotic scaling rate, because the fixed concept library makes the smallest +Concept models artificially under-utilized and biases their losses upward. The joint Chinchilla form (Equation~\ref{eq:chinchilla}) accommodates this curvature by absorbing it into the irreducible loss $\mathcal{L}_\infty$; a naive log-linear extrapolation does not. \cref{fig:8b-validation} refits both forms on the small-scale checkpoints only (excluding \steerling), and extrapolates to $C = 7.6 \times 10^{22}$ FLOPs, the compute used to train \steerling{} at 8B parameters and 1.35T tokens. The \steerlingB model achieves $\mathcal{L}^* = 2.72$, while the joint Chinchilla fit predicts $\mathcal{L}^* = 2.61$, a gap of $0.11$ nats. A naive log-linear extrapolation predicts $\mathcal{L}^* \approx 2.25$, missing by $0.47$ nats. The joint fit's prediction error is roughly $4\times$ smaller than the log-linear extrapolation's.

\important{\textbf{Inherently interpretable architectures admit scaling-law extrapolation.} \steerlingB's validation loss is predicted from small-scale fits to within $0.11$ nats via the joint Chinchilla form.}

\subsection{Interpretability scaling}

\label{subsec:interp-scaling}

\cref{subsec:compute-optimal} established that adding the concept module preserves compute-optimal scaling: interpretability-by-design imposes at most a fixed per-backbone offset on validation loss. The question that remains is how interpretability itself behaves at scale. The metrics defined in \cref{sec:interp_metrics} are fit against training compute on the same checkpoints used in \cref{subsec:compute-optimal}, allowing a direct test of \textbf{RQ2: does interpretability scale favorably with compute?}

\subsubsection{Methodology}
\label{subsubsec:interp-methodology}
\paragraph{Metrics. } Four metrics from \cref{sec:interp_metrics} are fit against training compute: Concept Loss ($\downarrow$), Concept Independence Loss ($\downarrow$), Concept Contribution ($\uparrow$), and Known Concept Alignment ($\uparrow$).
\paragraph{Scaling-law fits.}
Unlike validation loss, interpretability metrics are not analyzed under a compute-optimal frontier. The metrics are modeled in two ways: against training compute alone, and jointly against parameters and tokens. In both cases the fit uses all checkpoints rather than per-slice minima. The metrics are bounded by construction (Concept Loss and Concept Independence Loss below by zero, Concept Contribution above by one, Known Concept Alignment above by five), and a log-linear fit ignores these bounds, predicting values outside them when extrapolated past the data range. The first form, following the irreducible-loss scaling laws of \citet{henighan2020scaling} as applied to sparse autoencoder scaling by \citet{gao2025scaling}, encodes the bound directly as a free parameter:
\begin{equation}
m(C) = e \pm A \, C^{-\beta},
\label{eq:metric-gao}
\end{equation}
with three free parameters $A, \beta, e$ fit jointly by nonlinear least squares, where $e$ is the irreducible loss the metric approaches at infinite compute. The sign is positive for $\downarrow$ metrics (approaching $e$ from above) and negative for $\uparrow$ metrics (approaching $e$ from below).

Compute alone is an incomplete description of the sweep: within an IsoFLOP slice, compute is fixed while model size and token count vary inversely, so \cref{eq:metric-gao} assigns every checkpoint in a slice the same prediction and treats their per-checkpoint variation as noise. The second form resolves each checkpoint at its own parameter count $P$ and token count $D$ by refitting the parametric decomposition of \cref{eq:chinchilla} to each metric, $m(P, D) = e \pm \left( A_P P^{-\alpha} + A_D D^{-\beta} \right)$, with $e$ playing the role of $\mathcal{L}_\infty$ and the sign following the convention of \cref{eq:metric-gao}. The five free parameters are fit by nonlinear least squares.

\paragraph{Confidence intervals.} Confidence intervals on the fitted parameters come from a residual bootstrap on both fits, matching the procedure of \cref{subsubsec:co-methodology}. 90\% intervals are reported over $10{,}000$ iterations, with intervals on the irreducible loss $e$ tabulated in \cref{tab:interp-scaling}, and the joint-fit parameters in \cref{tab:interp-joint}.

\

\subsubsection{Results}
\label{subsubsec:interp-results}
\begin{figure}[htbp!]
\centering
\includegraphics[width=\linewidth]{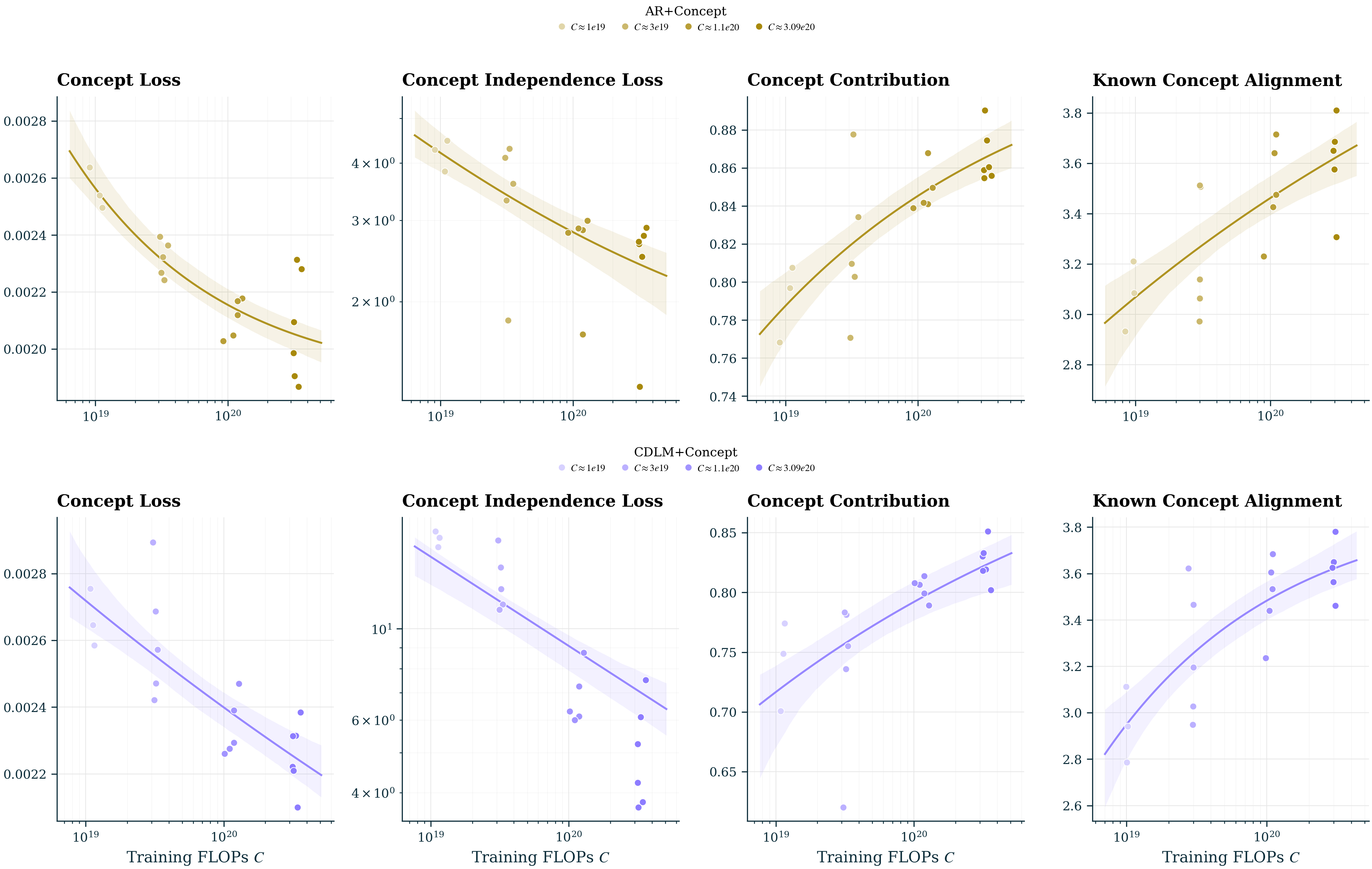}
\caption{Interpretability metrics scaling with compute, fit using the power-law-with-irreducible-loss form (Equation~\ref{eq:metric-gao}). Top row: AR+Concept. Bottom row: CDLM+Concept. Markers colored by IsoFLOP slice; shaded bands are 90\% bootstrap confidence intervals on the fitted curve.}
\label{fig:interp-scaling}
\end{figure}

\paragraph{Scaling with compute.}
\cref{fig:interp-scaling} shows the fitted curves for the four metrics on both +Concept families, with fitted parameters and bootstrap confidence intervals on the irreducible loss $e$ reported in \cref{tab:interp-scaling}. Every metric improves in the expected direction on both backbones: Concept Loss and Concept Independence Loss decrease with compute, while Concept Contribution and Known Concept Alignment increase. The fits explain a modest fraction of the per-checkpoint variance ($R^2$ between $0.49$ and $0.75$), reflecting the spread in evaluating interpretability properties on individual checkpoints. For several metrics the fitted asymptote is pinned to its natural bound, with bootstrap intervals concentrated near the boundary; this indicates that the small-scale data does not yet curve enough to identify $e$ from the data range alone, and the asymptote estimates should be read as upper or lower bounds on the true plateau rather than precise predictions.

\input{sections/scaling_laws/tables/interp_scaling_table.tex}
Across both backbones, all four metrics improve with compute under the asymptotic form. Concept Loss falls toward zero, indicating that the concept module continues to identify concepts more accurately as training scales. Concept Independence Loss falls similarly, indicating that the known and unknown heads become more disentangled. Concept Contribution rises toward one, indicating that predictions route through the concept module rather than the residual at increasing fractions. Known Concept Alignment rises toward saturating at the judge ceiling, indicating that concept embeddings increasingly point at semantically related tokens.

\begin{figure}[htbp!]
\centering
\includegraphics[width=\linewidth]{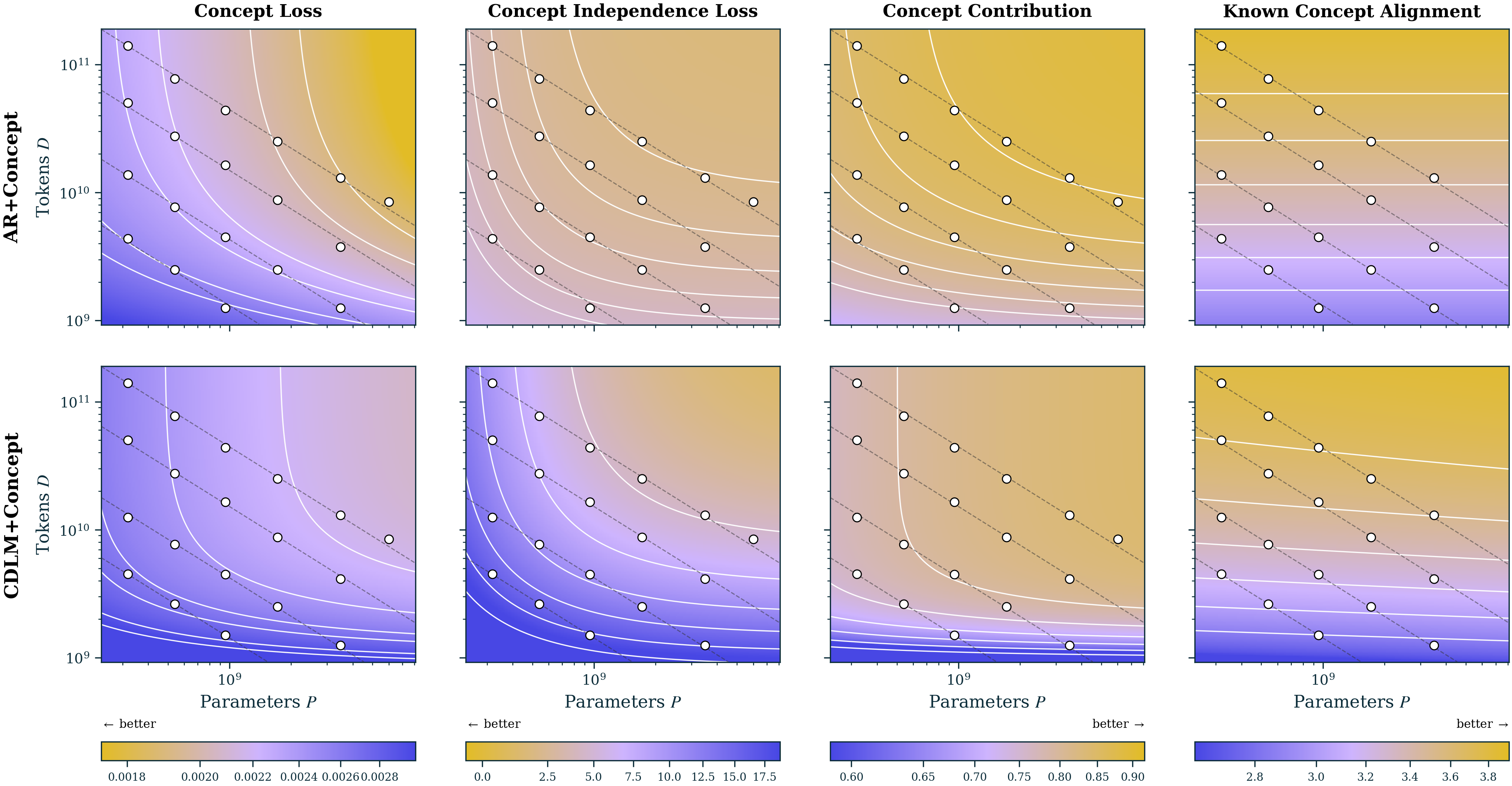}
\caption{Fitted interpretability surfaces $m(P, D)$ over parameters and tokens; gold indicates better values, and each metric column shares one color scale. White markers: checkpoints at their trained $(P, D)$; dashed diagonals: IsoFLOP slice budgets. Top row: AR+Concept. Bottom row: CDLM+Concept. Near-vertical contours indicate parameter-driven metrics, near-horizontal token-driven ones.}

\label{fig:interp-scaling-contours}
\end{figure}

\paragraph{Scaling in parameters and tokens.}
\cref{fig:interp-scaling-contours} shows the same metrics under the joint fit of \cref{subsubsec:interp-methodology}, plotted as surfaces over parameters and tokens, with fitted parameters reported in \cref{tab:interp-joint}. Contour lines connect $(P, D)$ configurations with equal metric value, and the dashed diagonals mark the IsoFLOP slice budgets. The joint form explains substantially more of the per-checkpoint variance than compute alone: across both families, $R^2$ rises from $0.49$ to $0.75$ under \cref{eq:metric-gao} to $0.62$ to $0.94$ under the joint fit. The variation within an IsoFLOP slice is therefore not evaluation noise but structure: at fixed compute, checkpoints differ systematically in how the budget is split between parameters and tokens, and the joint fit resolves this.

The contour orientations separate the metrics into two regimes. Concept Loss and Concept Independence Loss improve primarily with parameters:      their contours run near-vertical over the fitted range, so at fixed compute the larger models in a slice identify concepts more accurately and disentangle the known and unknown heads further. Known Concept Alignment improves primarily with tokens: its contours run near-horizontal, and for AR+Concept the fitted parameter term is negligible over the fitted range, making alignment mostly a function of data. Concept Contribution sits between the two regimes. A consequence is that no single split of a compute budget optimizes all four metrics at once: the token-heavy end of a slice favors alignment while the parameter-heavy end favors disentanglement, and the compute-optimal model of \cref{subsec:compute-optimal} is a compromise between them. As with the compute-only fits, several joint asymptotes pin to their natural bounds and should be read as bounds on the plateau rather than point estimates.

\input{sections/scaling_laws/tables/interp_joint_table.tex}
\important{\textbf{Interpretability scales favorably with compute across both autoregressive and diffusion,} with model size driving the concept and independence losses, and training data driving alignment.}

\subsubsection{Predicting \steerlingB from small models}
\label{subsubsec:interp-prediction}
\begin{figure}[htbp!]
\centering
\includegraphics[width=0.75\linewidth]{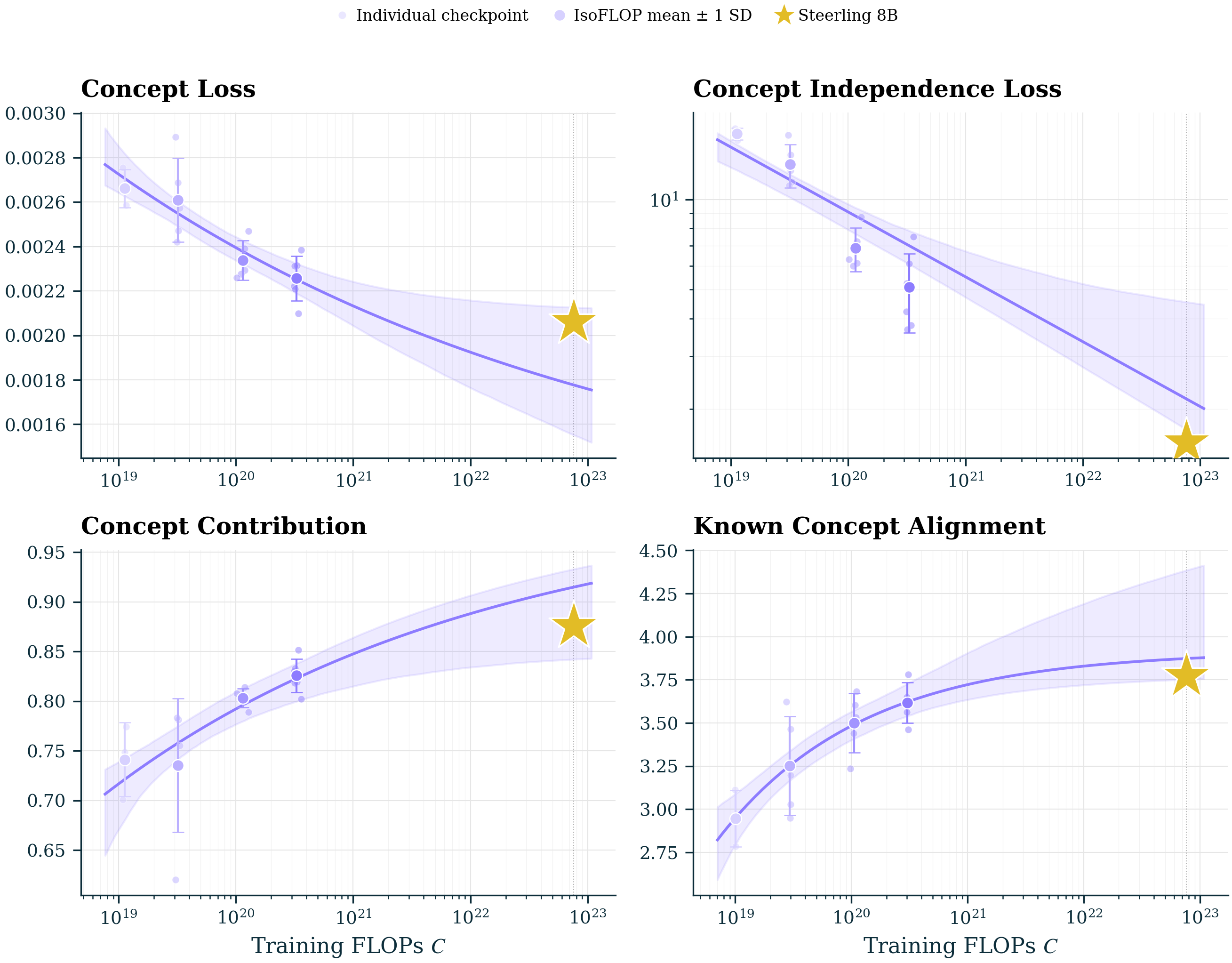}
\caption{Extrapolating CDLM+Concept interpretability metrics to \steerling. Asymptotic fits on small-scale checkpoints with 90\% bootstrap CI bands. Fits use all checkpoints; markers show per-slice means $\pm$ 1 SD, with individual checkpoints in the background. Filled stars: \steerling{} at 8B parameters and 1.35T~tokens.}
\label{fig:8b-interp-prediction}
\end{figure}

The fits of \cref{eq:metric-gao} are now refit on the small-scale checkpoints only and extrapolated to $C = 7.6 \times 10^{22}$ FLOPs, the compute used to train \steerling{} at 8B parameters and 1.35T tokens. \cref{fig:8b-interp-prediction} compares the extrapolations to the deployed model's measured metrics, and the rightmost columns of \cref{tab:interp-scaling} report the predicted and actual values per metric. Three of the four metrics land within tight bounds of the small-scale extrapolation: Concept Loss within $0.0004$ on a BCE scale, Concept Contribution within $0.04$ on a $[0, 1]$ scale, and Known Concept Alignment within $0.10$ on a $1$-$5$ scale. Concept Independence Loss exceeds the small-scale extrapolation in the favorable direction, with the deployed model achieving $1.55$ against a predicted $2.16$. The joint fit is refit and extrapolated the same way, resolving \steerling{} at its actual parameter and token counts rather than its compute; its predictions are reported in the rightmost columns of \cref{tab:interp-joint}.
\important{\textbf{Inherently interpretable architectures admit scaling-law extrapolation for interpretability properties.} Three of four metrics are predicted within error bounds from small-scale fits; the fourth exceeds its prediction in the favorable direction.}

%% file: sections/scaling_laws/tables/scaling_exponents.tex
\begin{table}[htbp!]
\centering
\small
\setlength{\tabcolsep}{4pt}
\begin{tabular}{@{}lcccc@{}}
\toprule
\textbf{Model} & $\alpha_P$ & $\alpha_D$ & $\alpha_L$ & $\mathcal{L}_\infty$ \\
\midrule
\multicolumn{5}{@{}l}{\textit{Ours (Autoregressive)}} \\
AR             & $0.528_{[-0.025, +0.023]}$ & $0.445_{[-0.036, +0.114]}$ & $-0.060_{[-0.002, +0.002]}$ & $1.857_{[-0.335, +0.071]}$ \\
AR+Concept     & $0.621_{[-0.041, +0.077]}$ & $0.524_{[-0.112, +0.108]}$ & $-0.070_{[-0.002, +0.003]}$ & $1.193_{[-0.703, +0.394]}$ \\
\midrule
\multicolumn{5}{@{}l}{\textit{Ours (\causaldiff)}} \\
CDLM           & $0.632_{[-0.091, +0.075]}$ & $0.481_{[-0.111, +0.162]}$ & $-0.053_{[-0.004, +0.004]}$ & $2.658_{[-0.708, +0.181]}$ \\
CDLM+Concept   & $0.503_{[-0.046, +0.051]}$ & $0.374_{[-0.170, +0.174]}$ & $-0.072_{[-0.003, +0.003]}$ & $1.942_{[-1.805, +0.529]}$ \\
\midrule
\multicolumn{5}{@{}l}{\textit{Masked diffusion (prior)}} \\
\citet{quokka}                        & $0.514$                    & $0.486$                    & ---                            & $2.41$ \\
\citet{scaling_behaviorDLM_von}       & $0.566_{[-0.022, +0.019]}$ & $0.434_{[-0.019, +0.020]}$ & $-0.0496_{[-0.0004, +0.0003]}$ & --- \\
\midrule
\multicolumn{5}{@{}l}{\textit{Autoregressive (prior)}} \\
\citet{hoffmann2022training}          & $0.490$ & $0.510$ & --- & $1.69$ \\
\citet{shuai2024scaling}              & $0.464$ & $0.536$ & --- & --- \\
\citet{bi2024deepseek}                & $0.524$ & $0.476$ & --- & --- \\
\bottomrule
\end{tabular}
\caption{Compute-optimal scaling exponents and irreducible-loss asymptotes. Subscripts are 90\% bootstrap confidence intervals.}
\label{tab:scaling-exponents}
\end{table}

%% file: sections/scaling_laws/tables/interp_scaling_table.tex
\begin{table}[htbp!]
\centering
\small
\setlength{\tabcolsep}{6pt}
\begin{tabular}{@{}lcccccc@{}}
\toprule
\textbf{Family} & $\beta$ & $e$ [90\% CI] & $R^2$ & 8B pred & 8B actual & $\Delta$ \\
\midrule
\multicolumn{7}{@{}l}{\textit{Concept Loss}} \\
\quad AR+Concept   & $0.385$ & $0.002_{[-0.002, +0.000]}$ & $0.704$ & $0.002$ & ---     & ---      \\
\quad CDLM+Concept & $0.054$ & $0.000_{[-0.000, +0.002]}$ & $0.648$ & $0.002$ & $0.002$ & $+0.000$ \\
\midrule
\multicolumn{7}{@{}l}{\textit{Concept Independence Loss}} \\
\quad AR+Concept   & $0.279$ & $1.310_{[-1.310, +0.000]}$ & $0.505$ & $1.550$ & ---     & ---      \\
\quad CDLM+Concept & $0.217$ & $0.000_{[-0.000, +3.687]}$ & $0.749$ & $2.157$ & $1.550$ & $-0.611$ \\
\midrule
\multicolumn{7}{@{}l}{\textit{Concept Contribution}} \\
\quad AR+Concept   & $0.212$ & $0.937_{[-0.047, +0.063]}$ & $0.630$ & $0.915$ & ---     & ---      \\
\quad CDLM+Concept & $0.134$ & $1.000_{[-0.149, +0.000]}$ & $0.496$ & $0.915$ & $0.876$ & $-0.039$ \\
\midrule
\multicolumn{7}{@{}l}{\textit{Known Concept Alignment}} \\
\quad AR+Concept   & $0.099$ & $5.000_{[-1.190, +0.000]}$ & $0.557$ & $4.200$ & ---     & ---      \\
\quad CDLM+Concept & $0.349$ & $3.920_{[-0.135, +1.080]}$ & $0.637$ & $3.870$ & $3.770$ & $-0.100$ \\
\bottomrule
\end{tabular}
\caption{Interpretability scaling fits. Subscripts on $e$ are 90\% bootstrap confidence intervals. The 8B columns compare the small-scale extrapolation against the actual \steerling{} values.}
\label{tab:interp-scaling}
\end{table}

%% file: sections/scaling_laws/tables/interp_joint_table.tex
\begin{table}[htbp!]
\centering
\small
\setlength{\tabcolsep}{5pt}
\begin{tabular}{@{}lccccccc@{}}
\toprule
\textbf{Family} & $\alpha$ & $\beta$ & $e$ [90\% CI] & $R^2$ & 8B pred & 8B actual & $\Delta$ \\
\midrule
\multicolumn{8}{@{}l}{\textit{Concept Loss}} \\
\quad AR+Concept   & $0.093$ & $0.767$ & $0.000_{[-0.000, +0.001]}$ & $0.945$ & ---     & ---     & ---      \\
\quad CDLM+Concept & $0.427$ & $1.255$ & $0.002_{[-0.000, +0.000]}$ & $0.893$ & $0.002$ & $0.002$ & $+0.000$ \\
\midrule
\multicolumn{8}{@{}l}{\textit{Concept Independence Loss}} \\
\quad AR+Concept   & $1.085$ & $0.529$ & $1.347_{[-1.347, +0.745]}$ & $0.624$ & ---     & ---     & ---      \\
\quad CDLM+Concept & $0.904$ & $0.659$ & $0.000_{[-0.000, +1.908]}$ & $0.937$ & $0.624$ & $1.550$ & $+0.921$ \\
\midrule
\multicolumn{8}{@{}l}{\textit{Concept Contribution}} \\
\quad AR+Concept   & $0.541$ & $0.739$ & $0.903_{[-0.028, +0.097]}$ & $0.784$ & ---     & ---     & ---      \\
\quad CDLM+Concept & $0.441$ & $2.077$ & $0.864_{[-0.012, +0.018]}$ & $0.883$ & $0.844$ & $0.876$ & $+0.032$ \\
\midrule
\multicolumn{8}{@{}l}{\textit{Known Concept Alignment}} \\
\quad AR+Concept   & $2.660$ & $0.154$ & $4.640_{[-0.363, +0.360]}$ & $0.926$ & ---     & ---     & ---      \\
\quad CDLM+Concept & $0.022$ & $0.346$ & $5.000_{[-1.125, +0.000]}$ & $0.933$ & $3.967$ & $3.770$ & $-0.194$ \\
\bottomrule
\end{tabular}
\caption{Joint interpretability scaling fits, $m(P, D) = e \pm \left( A_P P^{-\alpha} + A_D D^{-\beta} \right)$, with $\alpha$ the parameter exponent and $\beta$ the token exponent. Subscripts on $e$ are 90\% bootstrap confidence intervals. The 8B columns compare the small-scale extrapolation against the actual \steerling{} values.}
\label{tab:interp-joint}
\end{table}

%% file: sections/steerling_pretraining/main.tex
\section{\steerlingB: Pretraining}
\label{sec:pretraining}

With the recipe established in \cref{sec:recipe} and its scaling behavior characterized in \cref{sec:scaling-laws}, this section describes the full-scale pretraining of \steerlingB. We cover the concept-annotated 1.2T-token dataset (\cref{sec:pretraining-data}), the architectural and training choices that distinguish \steerlingB from a standard autoregressive or diffusion backbone (\cref{sec:pretraining-choices}), the pretraining run itself (\cref{sec:pretraining-training}), and the lessons we learned (\cref{sec:pretraining-lessons}).

% =====================================================================

% ---------------------------------------------------------------------
\subsection{Data}
\label{sec:pretraining-data}

We pretrain \steerlingB on 1.2T tokens drawn from a mixture of high-quality web, academic, mathematical, and code corpora. The bulk of the mix is Nemotron-CC-HQ~\citep{nemotron}, a quality-filtered slice of Nemotron-CC composed of both real webtext and synthetic question-and-answer rephrasings generated from the same documents. We choose Nemotron-CC-HQ because it yields the strongest downstream performance among public 1T pretraining datasets. The remainder of the corpus consists of peS2o~\citep{pes2o}, arXiv~\citep{weber2024redpajama}, OpenWebMath~\citep{openwebmath}, Algebraic Stack~\citep{llemma}, StarCoder~\citep{starcoder}, and Wikipedia and Wikibooks following the OLMo~2 mixture~\citep{olmocore}. We annotate the corpus with concepts at the chunk level using the \atlas pipeline (\cref{sec:data}). The full per-source token counts are reported in \cref{tab:pretraining-corpus}.

\input{sections/steerling_pretraining/tables/pretraining_corpus}

\subsection{Pretraining recipe}
\label{sec:pretraining-choices}

Training a standard language model requires choosing a handful of hyperparameters such as sequence length, batch size, learning rate, optimizer, and weight decay. The community has well-established defaults for each, and we adopt them without modification for \steerlingB. However, our interpretable causal-diffusion architecture (\cref{sec:architecture}) introduces design decisions without precedent in the autoregressive or masked-diffusion literature, for which no community defaults yet exist.

These fall into two groups: those that govern the diffusion process (\cref{subsec:causal-diffusion}) and those that govern the concept bottleneck (\cref{subsec:concept-module}). We describe each choice and the reasoning below; for every choice we ran a small ablation at the 1B scale, reported in full in \cref{app:pretraining-ablations}.

\paragraph{Diffusion process.}
\steerlingB uses the block-causal attention mask of \cref{subsec:causal-diffusion}, where tokens attend bidirectionally within a block of size $b$ and causally across blocks. The block size sets how many tokens are decoded in parallel at inference and the granularity of the key-value cache. We compare $b \in \{32, 64\}$ and find block size leaves the interpretability metrics unchanged while the larger value lowers validation loss, so we set $b = 64$. Diffusion training must also choose how the noise level $t$ is sampled at each step. We compare uniform sampling against a moving Gaussian curriculum that shifts from low to high masking over training~\citep{quokka}, and adopt the moving Gaussian for a slight edge on validation loss, consistent with prior work.\footnote{This schedule proved too aggressive over the full pretraining run (see~\cref{subsec:lesson-masking}).}

\paragraph{Concept bottleneck sizing.}
The concept module splits its representation into $n$ known concepts, fixed by the \atlas library, and $m$ unknown concepts learned during training. Here $m$ is a free hyperparameter. We find that raising it from $3n$ to $5n$ gives no measurable gain on any metric, so we keep the conservative $m = 3n$. The unknown embedding matrix $U \in \mathbb{R}^{m \times d}$ is the largest parameter the module adds; we factorize it as $U = AB$ with rank $R = 256$, which makes it roughly $15\times$ smaller and removes a heavy per-token matrix multiplication at no capability cost and only a small drop in concept contribution.

\paragraph{Concept bottleneck training dynamics.}
Both concept heads are trained from scratch, so their early predictions are unreliable. We therefore route ground-truth concepts to the LM head early and anneal toward the model's own predictions, holding the teacher forcing floor at $0.5$ for both heads, since decaying further hurts known concept alignment for no capability gain. The model can optionally carry a residual $\varepsilon = h - \hat{k} - \hat{u}$ in the bottleneck, an uninterpreted channel that absorbs whatever the two heads fail to reconstruct. Dropping it forces every dimension through the known and unknown heads, raising concept contribution to $1.0$ by construction, but in our comparison the capability gap is large enough that we keep $\varepsilon$.

% Finally, we detach the transformer hidden state at the unknown head's input so that its auxiliary losses do not reshape the backbone, which matches the direct variant on interpretability while helping downstream performance. 

\subsection{Pretraining run}
\label{sec:pretraining-training}

% The pretraining run executed the configuration of \cref{app:steerling8b-config} on $40$ A100 nodes over approximately three weeks of wall-clock time, completing its full 1.2T-token budget without divergence or manual intervention to the loss curve.

Pretraining used the configuration of \cref{app:steerling8b-config} on $320$ A100 GPUs ($40$ nodes, 8 GPUs per node) for approximately $21$ days, totaling $\sim$161K GPU-hours. The run completed its full 1.2T-token budget without divergence or manual intervention to the loss curve.

We monitored various metrics during the pretraining run. Validation loss was logged continuously, computed as a Monte Carlo estimate of the MDLM ELBO (\cref{subsubsec:co-methodology}). At every $50$B-token interval, we additionally saved a checkpoint and ran a broader evaluation suite of five downstream benchmarks from the language modelling harness: ARC-Challenge~\citep{clark2018think}, HellaSwag~\citep{zellers2019hellaswag}, PIQA~\citep{bisk2020piqa}, MMLU~\citep{hendrycks2020measuring}, and WinoGrande~\citep{sakaguchi2021winogrande}; together with the four interpretability metrics introduced in \cref{sec:interp_metrics}: concept loss, concept contribution, concept independence loss, and known concept alignment.

\begin{figure}[htbp!]
  \centering
  \includegraphics[width=\linewidth]{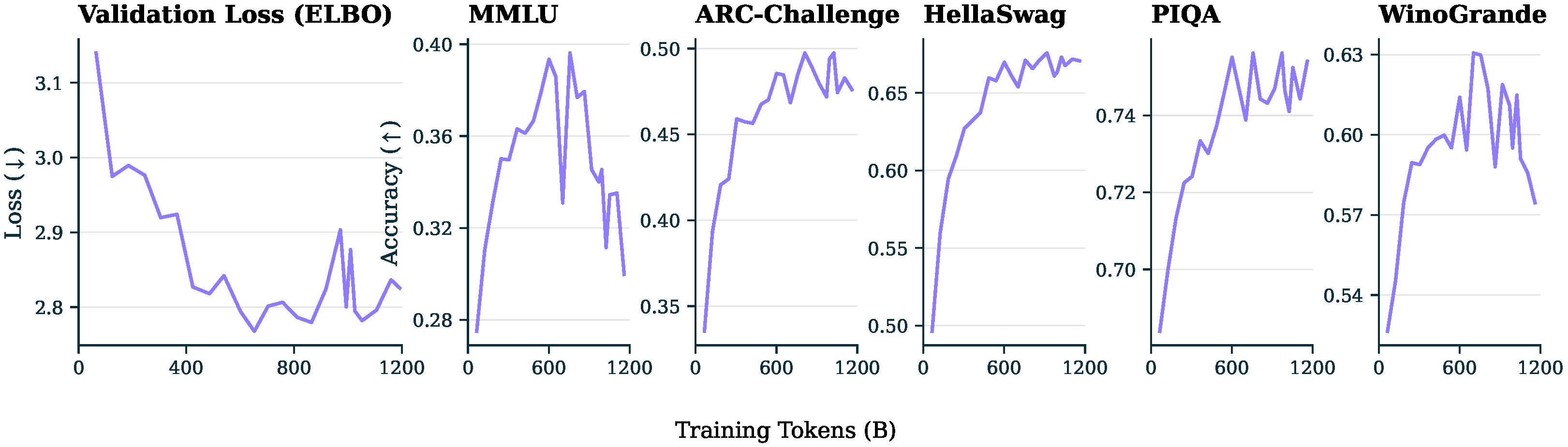}
  \caption{Validation loss and LM-Harness metrics for \steerlingB across the pretraining run.}
  \label{fig:pretraining-capability}
\end{figure}

\cref{fig:pretraining-capability} shows the capability metrics across the run. Validation loss descends rapidly through the first half and then plateaus, oscillating around a stable value. The harness benchmarks behave differently from one another. HellaSwag and PIQA rise smoothly and hold near their peaks to the end, and ARC-Challenge peaks late with only a small dip at the very end. MMLU and WinoGrande, by contrast, peak around the midpoint and then decline through the final third, with MMLU losing roughly a quarter of its peak value.

\begin{figure}[htbp!]
  \centering
  \includegraphics[width=\linewidth]{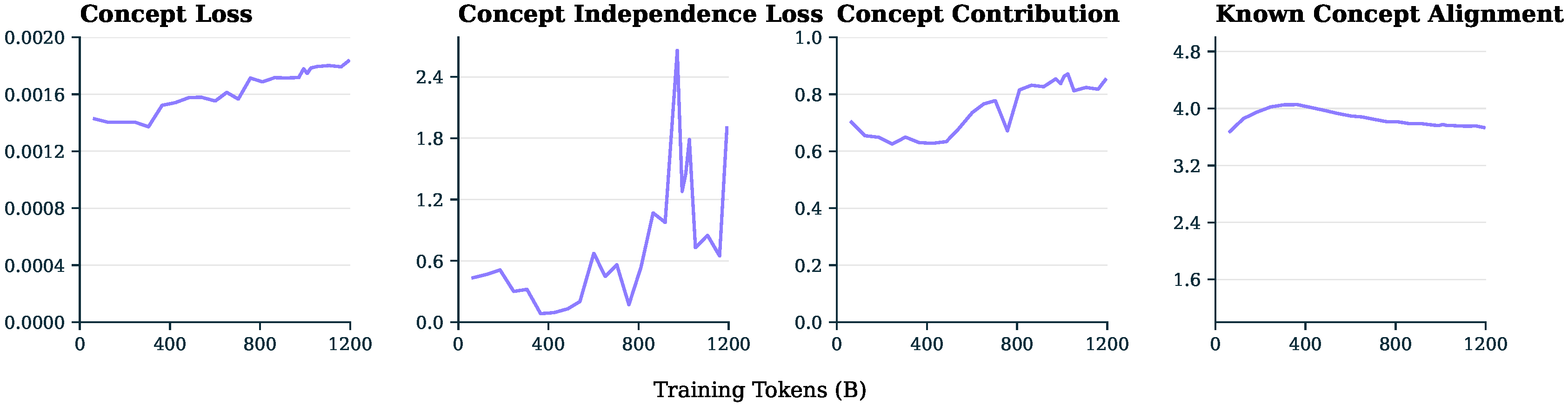}

  \caption{Interpretability metrics for \steerlingB across the pretraining run.}
  \label{fig:pretraining-interpretability}
\end{figure}

\cref{fig:pretraining-interpretability} shows the interpretability metrics across the run. Concept loss drops early and then climbs steadily, though the climb is small in absolute terms. Concept independence loss stays low through most of training, then rises sharply in the final third, peaking around $2.7$. Concept contribution dips slightly early before climbing from $0.62$ to $0.85$. Known concept alignment moves modestly on its $[1,5]$ scale, rising to a peak near $4$ in the first third before settling back.

\subsection{Pretraining lessons}

\label{sec:pretraining-lessons}

During the run, most metrics looked healthy. Validation loss kept descending or held steady, three of the four interpretability metrics drifted only modestly, and HellaSwag, PIQA, and ARC-Challenge plateaued near their peaks. However, four signals were less encouraging:
\begin{itemize}
    \item MMLU and WinoGrande began declining around the midpoint of training.
    \item Concept independence loss spiked in the final third.
    \item A late checkpoint showed weak math and code performance.
    \item The same checkpoint did not steer reliably through the concept module.
\end{itemize}
At the time we expected some of these to recover as the schedules stabilized, so we continued the run to its planned 1.2T token budget. Once the run completed, we examined the issues closely and traced them to four causes: an over-aggressive masking curriculum (\cref{subsec:lesson-masking}), late-training entanglement of the concept heads (\cref{subsec:lesson-entanglement}), insufficient math and code in the corpus (\cref{subsec:lesson-data}), and the absence of steering operations during training (\cref{subsec:lesson-steering}).

% \subsubsection{Over-aggressive masking}
% \label{subsec:lesson-masking}
% MMLU and WinoGrande peak around the midpoint of training and decline thereafter. This became clear once we overlaid the masking curriculum on the capability metrics (\cref{fig:appendix-pretraining-capability}), where both inflection points coincide with the curriculum passing center $\sim\!0.5$ on its way to its endpoint of $0.8$. The moving Gaussian was validated at small scale (\cref{app:abl-masking}), where it modestly outperformed uniform sampling. At pretraining scale, however, the curriculum spends a substantial fraction of training above $50\%$ masking, a regime in which most of the input is replaced with \texttt{[MASK]} tokens and the model has little context to leverage.

\subsubsection{Over-aggressive masking}
\label{subsec:lesson-masking}
MMLU and WinoGrande peak around the midpoint of training and decline thereafter. Overlaying the masking curriculum on the capability metrics (\cref{fig:appendix-pretraining-capability}) makes the cause visible: both inflection points coincide with the curriculum reaching $\sim\!0.5$ on its climb to its endpoint of $0.8$. The moving Gaussian curriculum was validated at small scale (\cref{app:abl-masking}), where it held a slight edge on validation loss. At pretraining scale, however, the curriculum spends a substantial fraction of training above $50\%$ masking, where most of the input is replaced with \texttt{[MASK]} tokens and the model has little context to leverage.

\subsubsection{Late-training entanglement of the concept heads}

\label{subsec:lesson-entanglement}

Concept independence loss stays low through most of the run and rises sharply in the final third. Overlaying the three schedules on the interpretability metrics shows why (\cref{fig:appendix-pretraining-interp}). Two of them move into adverse territory over the same window. The masking curriculum reaches its hard regime, so most input tokens are masked. At the same time the teacher forcing floors have been reached, with $\alpha_{\text{known}}$ at its floor and $\alpha_{\text{unknown}}$ decayed substantially, so the LM head leans increasingly on predicted concepts rather than ground-truth ones. With less supervision and more reliance on predicted contributions, the two heads come to rest on overlapping information, and disentangling their representations becomes harder. At ablation scale this effect was not visible.

\subsubsection{Limited math and code in the corpus}
\label{subsec:lesson-data}
Our corpus is dominated by Nemotron-CC-HQ, which carries little math or code. Together with the smaller dedicated sources in the mix (\cref{tab:pretraining-corpus}), the run saw slightly over $100$B math and code tokens out of $1.2$T total. \citet{scaling_behaviorDLM_von} report a similar shortfall on the same corpus. We caught this only at the end of pretraining, when GSM8K and HumanEval both came back low. With more math and code in the mixture, and these benchmarks included in the in-run evaluation sweep, the issue would have surfaced earlier and been addressable on the data side.

\subsubsection{The model does not respond to steering}
\label{subsec:lesson-steering}
A model built around an explicit concept module should be steerable through it, since injecting a concept's own direction (\cref{sec:model-control}) should bias generation toward that concept. We tested this on a late pretraining checkpoint following the evaluation protocol of \citet{wu2025axbench}. We elicit 128-token generations from a fixed prompt, sampling a continuation under the method being evaluated, and score each continuation with an LLM-judge on two axes. A \emph{concept} score (0--2) measures how strongly the target concept is present in the generated text, and a \emph{quality} score (0--2) measures the fluency and coherence of the output. Details on the judge and the full prompts are given in \cref{app:steering-judge}. Finally, we summarize the two scores by their harmonic mean:
\begin{equation}
    \text{Harmonic} = \frac{2}{\tfrac{1}{\text{concept}} + \tfrac{1}{\text{quality}}},
\end{equation}
which weights both axes equally and strongly penalizes a method that sacrifices one for the other. We report the mean over 72 randomly sampled concepts.

We study steering on three conditions: \emph{unsteered}, \emph{prompting} (via prepending the concept label and description to the prompt), and \emph{steered} generation (layer injection in \cref{sec:model-control}). 
In \cref{tab:steering_pretrained_method_comparison}, unsteered generation produces fluent text that does not surface the target concept, which is expected. Steering achieves the highest concept score of 1.072, showing that intervening through the model's own concept direction induces the target concept more strongly than though a text prompt. Yet, the cost of steering is quality. As a result, prompting yields the best harmonic mean of 1.156.

\input{sections/steerling_midtraining/tables/midtraining_steering_results}

From this experiment, we identify two limitations for steering. First, the quality drop under steering is substantial, suggesting \steerlingB does not gracefully integrate the injected direction into its forward pass. Second, layer injection does not generalize on less-frequent concepts, as roughly one third never activate at any injection strength $\gamma$, with the bottleneck activation $k_c$ staying near zero throughout generation. Both trace to the same cause, that \steerlingB never encounters concept injection during pretraining, so adding $\gamma \cdot K_c$ at inference is an out-of-distribution perturbation the model has no mechanism to respond to.

%% file: sections/steerling_pretraining/tables/pretraining_corpus.tex
\begin{table}[htbp!]
\centering

\begin{tabular}{lrrr}
\toprule
\textbf{Source} & \textbf{Documents} & \textbf{Chunks} & \textbf{Tokens} \\
\midrule
Nemotron-CC-HQ (real) & 740M & 5.1B & 547B \\
Nemotron-CC-HQ (synthetic) & 971M & 4.8B & 498B \\
peS2o & 38.8M & 565.4M & 59B \\
arXiv & 3.9M & 142.2M & 20.4B \\
Wikipedia \& Wikibooks & 6.1M & 36.8M & 3.8B \\
OpenWebMath & 2.9M & 76.8M & 12.1B \\
Algebraic Stack & 2.8M & 65.6M & 12.1B \\
StarCoder & 78.6M & 317M & 91.4B \\
\midrule
\textbf{Total} & \textbf{1.84B} & \textbf{11.1B} & \textbf{1.24T} \\
\bottomrule
\end{tabular}
\caption{\steerlingB pretraining corpus. Token counts are post-tokenization.}
\label{tab:pretraining-corpus}
\end{table}

%% file: sections/steerling_midtraining/tables/midtraining_steering_results.tex
\begin{table}[htbp!]
    \centering
    \begin{tabular}{lccc}
        \toprule
        Method     & Concept $\uparrow$ & Quality $\uparrow$ & Harmonic $\uparrow$ \\
        \midrule
        Unsteered  & 0.033          & 1.108          & 0.065 \\
        Prompting  & 0.908          & \textbf{1.588} & \textbf{1.156} \\
        Steered    & \textbf{1.072} & 0.972          & 1.020 \\
        \bottomrule
    \end{tabular}
    \caption{\looseness=-1 Steering results on random concepts of the pretrained \steerlingB. \emph{Steered} denotes layer injection. 
    % \aya{numbers need to be updates}
    }
    \label{tab:steering_pretrained_method_comparison}
\end{table}

%% file: sections/steerling_midtraining/main.tex
% \clearpage
\section{\steerlingB: Mid-training}
\label{sec:steerling-midtraining}
\label{sec:midtraining}

Mid-training is a short 150B-token run initialized from the final pretraining checkpoint. It has two aims. First, address the four weaknesses identified in pretraining (\cref{sec:pretraining-lessons}): aggressive masking, thin math and code coverage, unstable independence loss, and no exposure to steering during training. Second, tighten the model for downstream use, by reducing its reliance on teacher forcing and increasing its reliance on the concept heads. Here we describe the data mixture (\cref{sec:midtraining-data}), the recipe changes from pretraining including a dedicated steering phase (\cref{sec:midtraining-changes}, \cref{subsec:training-steering}), the resulting mid-trained model (\cref{sec:midtraining-results}), and its benchmark performance against open base models of comparable size (\cref{sec:base-model-results}).

\subsection{Data}
\label{sec:midtraining-data}

Our goal in mid-training is to improve \steerlingB on math and code, where the pretrained model lagged furthest behind, and to recover the reasoning and knowledge capabilities degraded by heavy masking late in pretraining. We follow the OLMo~2 mid-training recipe~\citep{olmocore}: start from a high-quality data mixture that lifts performance across the benchmark suite, then patch the specific capabilities the pretrained model is weakest on. For the natural-language portion of the mixture we again use Nemotron-CC-HQ, but restrict to real tokens; we benchmarked real, synthetic, and mixed against each other (\cref{app:nemotron-real-vs-synth}), and real tokens won.

\input{sections/steerling_midtraining/tables/midtraining_composition}

\looseness=-1To find the best midtraining mixture, we run a data ablation: starting from the final pretraining checkpoint of 1.2T tokens, we midtrain on 10B tokens for each of four candidate compositions and compare their downstream performance. The compositions, listed in \cref{tab:midtraining-composition}, are a math-heavy mixture, a balanced mixture following OLMo~2's Dolmino Mix, a code-augmented mixture that adds StarCoder to the balanced mixture, and a code-only mixture. The first three hold the Nemotron-real share roughly fixed and vary the math, code, and high-quality reference sources around it; the code-only mixture is an extreme that drops Nemotron entirely, isolating the effect of training on code alone.

\input{sections/steerling_midtraining/tables/midtraining_data_results}

The results are shown in \cref{tab:midtraining-results}. Every mixture except code-only \textbf{improves substantially} over the pretrained model, but each isolates a different lesson:

\begin{itemize}
    \item The math-heavy mixture delivers the largest math gain, lifting GSM8K from 0.140 to 0.441, and posts the best overall average, but this is driven by large gains on only two benchmarks while it lags on the rest.

    \item The balanced mixture drives the largest knowledge gain, recovering MMLU from 0.298 to 0.416, but with no dedicated code source, HumanEval again fails to improve.

    \item The code-only mixture pushes HumanEval the highest, as expected, but lags well behind every other composition elsewhere.

    \item The code-augmented mixture strikes a balance across reasoning, math, and code, and is the only composition to improve on every benchmark over the pretrained model, so we adopt it as the mid-training mixture.
\end{itemize}

The final midtraining mixture applies the code-augmented recipe over 150B tokens, with proportions given in \cref{tab:midtraining-mixture}.

\input{sections/steerling_midtraining/tables/midtraining_mixture}

\subsection{Mid-training recipe changes}
\label{sec:midtraining-changes}

% \paragraph{Masking schedule.} The first and most consequential change is to revert the masking schedule. Pretraining used a moving Gaussian curriculum whose center rose from 0.2 to 0.8; we traced \steerlingB's declining knowledge and reasoning scores to the high-masking regime this curriculum entered late in the run (\cref{sec:pretraining-lessons}). Mid-training instead samples the masking rate uniformly, following MDLM training~\citep{llada}, so the model sees a balanced spread of masking levels rather than a curriculum that drifts toward heavy masking. To confirm that uniform sampling is the right choice, we compare it against an 80\% Gaussian schedule (the endpoint of the pretraining curriculum) in a 10B-token ablation from the final pretraining checkpoint (\cref{tab:midtraining-masking}). The two schedules are close on most benchmarks, and the 80\% schedule is even slightly stronger on the math tasks. On MMLU, however, the heavy schedule falls to 0.280, below the pretrained model's 0.298, while uniform sampling lifts it to 0.416; the heavy-masking regime that hurt knowledge during pretraining hurts it again here, confirming our initial suspicion. We adopt the uniform schedule for mid-training.

\subsubsection{Schedules}
\label{sec:midtraining-schedules}

\paragraph{Masking schedule.} The first and most consequential change is to revert the masking schedule. Pretraining used a moving Gaussian curriculum whose center rose from 0.2 to 0.8, and \cref{sec:pretraining-lessons} traced \steerlingB's declining knowledge and reasoning scores to the high-masking regime this curriculum entered late in the run. Midtraining instead samples the masking rate uniformly, following MDLM training~\citep{llada}, so the model sees a balanced spread of masking levels rather than a curriculum that drifts toward heavy masking. To confirm that uniform sampling is the right choice at this stage, we compare it against an 80\% Gaussian schedule in a 10B-token ablation from the final pretraining checkpoint (\cref{tab:midtraining-masking}). The two schedules are close on most benchmarks, and the 80\% schedule is even slightly stronger on the math tasks. On MMLU, however, the heavy schedule falls to 0.280, below the pretrained model's 0.298, while uniform sampling lifts it to 0.416; the heavy-masking regime that hurt knowledge during pretraining hurts it again here, confirming our initial suspicion. We adopt a uniform schedule during midtraining.

\input{sections/steerling_midtraining/tables/midtraining_masking_results}

\paragraph{Teacher forcing.} During pretraining the model's predicted concepts were partly replaced with ground-truth ones: with probability $\alpha_{\text{known}}$ the known head's predicted activations were replaced by their labeled values, and the unknown head was also mixed with its residual at rate $\alpha_{\text{unknown}}$. Mid-training anneals $\alpha_{\text{known}}$ from its end-of-pretraining value of $0.5$ to $0$, and holds $\alpha_{\text{unknown}}$ at $0$ throughout. By the end of mid-training, every concept entering the bottleneck is model-predicted, matching the inference regime where labeled concepts are unavailable.

\paragraph{Learning rate.} Following standard mid-training practice~\citep{olmocore, grattafiori2024llama, deepseek}, the learning rate is decayed linearly to zero.

\subsubsection{Losses}
\label{sec:midtraining-losses}

\looseness=-1The second pretraining weakness was the independence loss (Equation~\ref{eq:indep-loss}), which destabilized late in the run as the model came to rely on its predicted concepts (\cref{sec:pretraining-lessons}). Midtraining makes the model rely on the predicted concepts entirely, which exposes a limitation of the single-term loss. The pretraining loss penalized the dependence between the known representation $\hat{k}$ and the unknown representation routed to the bottleneck, but with the unknown head now always supplying its own prediction, detached before the bottleneck (gradient detachment), the term applies no pressure on the transformer residual. We therefore replace it with two terms, $\mathcal{L}_{\text{indep}}(\hat{u}, \hat{k}) + \mathcal{L}_{\text{indep}}(\varepsilon, \hat{k})$: the first keeps the predicted unknown independent of the known concepts, the second keeps the residual $\varepsilon$ independent of them. Each term carries half the original weight, leaving the total magnitude unchanged.

\subsubsection{Architecture}
\label{sec:midtraining-architecture}
Two architectural changes tighten interpretability. First, the residual dropout rate $p_\varepsilon$ is raised from $0.1$ to $0.3$, applying more pressure on $\varepsilon$ to vanish and forcing the concept heads to carry a larger share of the prediction. Second, the concept heads are sparsified: the known head composes the bottleneck from its top $32$ concepts, and the unknown head from its top $128$. Sparsity makes attribution more interpretable, since each logit decomposes into at most $160$ concepts rather than the full bottleneck of thousands.

\subsubsection{Steering training}
\label{subsec:training-steering}

\begin{figure}[htbp!]
\centering
\includegraphics[trim={4cm 4cm 4cm 4cm}, clip, width=\textwidth]{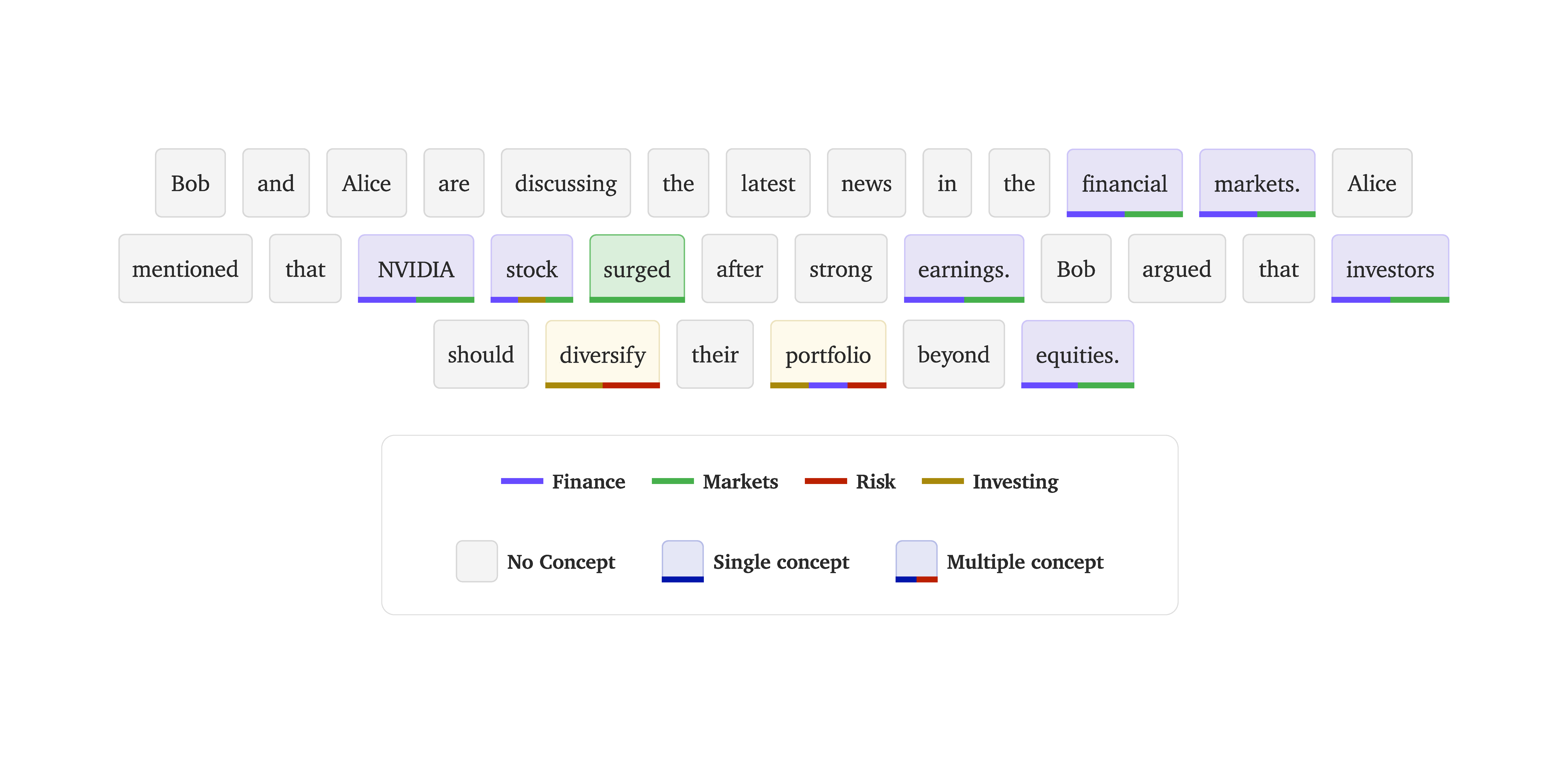}
\caption{Token-level concept annotations used for steering training. Each token is tagged only with the concepts attributed to it, rather than with all chunk-level concepts in pretraining.}
\label{fig:steering_midtraining_dataset}
\end{figure}

Pretraining labels mark that a concept appears in a chunk but not which tokens actually express it. Steering supervision needs that detail because injection happens at the token level: at each generation step, we add a concept embedding into the transformer's hidden state at each token position, so the loss must know which positions should actually respond. We therefore mid-train on a token-level dataset (\cref{fig:steering_midtraining_dataset}) of roughly $400$M tokens, where each token is tagged with its attributed concepts.

\paragraph{Steering losses.}
During a steering phase the target concept's direction $K_c$, scaled by $\gamma$, is injected at the masked positions attributed to $c$, exactly as steering is applied at inference. Let $\mathcal{I}$ denote the set of these injected positions. Two losses are added on top of the masked diffusion loss $\mathcal{L}_{\text{MDM}}$, one for each objective:

\begin{itemize}
\item \textbf{Respond objective.} At every injected position, the concept bottleneck should activate the injected concept. Its activation $k_{c,t}$ is driven toward $1$ by minimizing its negative log-likelihood:
\begin{equation}
    \mathcal{L}_{\text{respond}} = -\frac{1}{|\mathcal{I}|} \sum_{t \in \mathcal{I}} \log k_{c,t}.
    \label{eq:steer-respond}
\end{equation}

% \item \textbf{Express objective.} Activating the bottleneck is not sufficient if the generation remains dominated by unrelated tokens, so output mass is pushed onto the tokens that express $c$. These are the top entries of the vocabulary profile $W K_c$ from the logit decomposition (Equation~\ref{eq:logit-decomposition}), denoted $\mathcal{T}_c$. The total probability assigned to $\mathcal{T}_c$ is maximized, i.e. its negative log is minimized:
% \begin{equation}
% \mathcal{L}_{\text{express}} = -\frac{1}{|\mathcal{I}|} \sum_{t \in \mathcal{I}} \log \frac{\sum_{y \in \mathcal{T}_c} e^{\,z_{t,y}}}{\sum_{y \in V} e^{\,z_{t,y}}},
% \end{equation}
% where $z_{t,y}$ is the logit of token $y$ at position $t$, and the fraction is the share of output probability falling on the expressing tokens $\mathcal{T}_c$.
% \end{itemize}

\item \textbf{Express objective.} A high bottleneck activation does not by itself produce concept-expressing output: the concept direction may still place weight on unrelated tokens such as fillers. We therefore push the concept's output distribution at attributed positions onto the tokens that express $c$. 

These are the lifted set $\mathcal{T}_c$: vocabulary tokens with the highest \emph{lift}, defined as the ratio of token frequency in concept-tagged chunks to token frequency in the corpus. We maximize the total probability assigned to $\mathcal{T}_c$, \ie minimize its negative log:
\begin{equation}
\mathcal{L}_{\text{express}} = -\frac{1}{|\mathcal{I}|} \sum_{t \in \mathcal{I}} \log \frac{\sum_{y \in \mathcal{T}_c} e^{\,\ell_{t,y}}}{\sum_{y \in V} e^{\,\ell_{t,y}}},
\label{eq:express-loss}
\end{equation}
where $\ell_{t,y}$ is the logit of token $y$ at position $t$.

\end{itemize}

The masked diffusion loss $\mathcal{L}_{\text{MDM}}$ is kept active throughout so that generation stays coherent, while the pretraining interpretability losses (the concept, reconstruction, and independence terms) are disabled during steering phases. The steering-phase objective is
\begin{equation}
    \mathcal{L} = \mathcal{L}_{\text{MDM}} + \lambda_{\text{respond}}\, \mathcal{L}_{\text{respond}} + \lambda_{\text{express}}\, \mathcal{L}_{\text{express}},
    \label{eq:steer-total}
\end{equation}
where $\lambda_{\text{respond}}$ and $\lambda_{\text{express}}$ are both set to $1$.

\begin{figure}[!htt]
    \centering
    \includegraphics[width=0.89\linewidth]{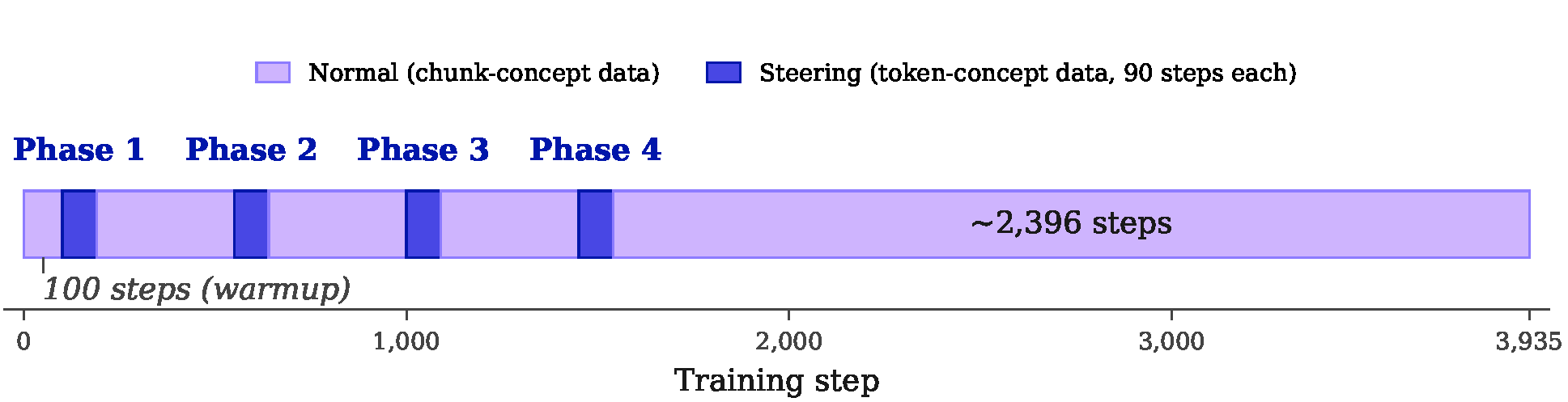}
    \caption{Interleaved mid-training schedule for \steerlingB. Four steering phases (90 steps each, shaded) are inserted within the standard run; the first normal block is a 100-step warmup. Normal phases use chunk-level concept data, steering phases use token-level concept data.}
    \label{fig:steering_midtraining_schedule}
\end{figure}

\paragraph{Steering training schedule.}
Steering is interleaved into the main midtraining run rather than added as a separate stage (\cref{fig:steering_midtraining_schedule}). 
We front-load it: four short steering phases come immediately after a 100-step warmup, and the remainder of mid-training is normal capability training. 
This placement keeps the steering objectives from interfering with the capabilities the rest of the run builds.
Each phase is one pass over the token-level dataset ($\sim$400M tokens); we use four phases because both $\mathcal{L}_{\text{respond}}$ and $\mathcal{L}_{\text{express}}$ converge by the fourth epoch. Normal phases use chunk-level data and the standard objective; steering phases use token-level data and the steering losses above.

\paragraph{Steering training works and does not cost capability.}
We evaluate steering as an ablation on the mid-training recipe: starting from the math-heavy composition tested in \cref{sec:midtraining-data}, the 10B-token run is repeated with the steering phases interleaved, so that the only difference from the recipe in \cref{tab:midtraining-results} is steering itself. Adding steering leaves LM Harness performance essentially unchanged across all five benchmarks (\cref{tab:steering-midtraining-ablation}), so teaching the model to respond to injection does not trade off against capability. It also achieves its intended effect: the steering benchmark scores all improve, and the model activates the target concept in a single injection step at $\gamma = 1$.

\input{sections/steerling_midtraining/tables/steering_midtraining_ablation}

\subsection{The mid-trained model}
\label{sec:midtraining-results}

The final mid-training run is 150B tokens on the code-augmented mixture (\cref{sec:midtraining-data}), starting from the final pretraining checkpoint, with the recipe changes of \cref{sec:midtraining-changes}: uniform masking, teacher forcing annealed to zero, the two-term independence loss, residual dropout raised to $0.3$, sparsified concept heads, and a learning rate decayed to zero. Steering is interleaved into this run rather than added as a separate stage, as four short phases over a token-level dataset of roughly $400$M tokens (\cref{subsec:training-steering}). The complete hyperparameter list is given in \cref{app:midtraining-config}.

% \paragraph{Mid-training rescues the capabilities lost in pretraining.}
% Midtraining addresses all three pretraining weaknesses of \cref{sec:pretraining-lessons} at once (\cref{tab:midtraining-rescue}). The knowledge lost under heavy masking comes back, with MMLU up \textbf{17 percentage points}. The two capabilities the pretraining corpus starved improve the most: math rises \textbf{30 percentage points} and code \textbf{7}. The benchmarks that were already healthy stay healthy, so nothing is traded away, and \textbf{every benchmark improves}, lifting the overall average by \textbf{10 percentage points}.

\paragraph{Mid-training rescues the capabilities lost in pretraining.}
Mid-training addresses the capability weaknesses of pretraining at once (\cref{tab:midtraining-rescue}). The knowledge lost under heavy masking comes back, with MMLU up \textbf{17 percentage points}. The two capabilities the pretraining corpus starved improve the most: math rises \textbf{30 percentage points} and code around \textbf{7} percentage points on average (HEval \& MBPP). The benchmarks that were already healthy stay healthy, so nothing is traded away. \textbf{Every benchmark improves}, lifting the overall average by \textbf{10 percentage points}.

\input{sections/steerling_midtraining/tables/midtraining_rescue}
\input{sections/steerling_midtraining/tables/midtraining_interp}

\paragraph{Mid-training improves interpretability.}
The effects of midtraining on interpretability metrics are shown in \cref{tab:midtraining-interp}. The late-pretraining rise in concept independence loss, the third weakness of \cref{sec:pretraining-lessons}, is mitigated after mid-training: the two-term penalty of \cref{sec:midtraining-changes} brings concept independence loss down by $19\%$, so the known and unknown representations are more cleanly separated than at the end of pretraining. Concept contribution rises, meaning the concept module accounts for a larger share of each prediction, and known concept alignment improves slightly.

% \paragraph{Mid-training improves Steering.}
% \aya{Giang please add this}

\paragraph{Mid-training improves steering.}
We evaluate the midtrained \steerlingB using the steering benchmark used for the pretraining checkpoint~\citep{wu2025axbench}. Midtraining improves every metric in~\cref{tab:steering_pretrained_vs_midtrained}: mean concept rises from 1.072 to 1.247, mean quality from 0.972 to 1.064, and their harmonic mean from 1.020 to 1.148. We additionally report \emph{mean sample harmonic} (the harmonic mean computed per sample, then averaged across samples). This rises from 0.843 to 0.963.

\input{sections/steerling_midtraining/tables/midtraining_steering_pre_vs_midtrained}

\subsection{Evaluation and results}
\label{sec:base-model-results}
\begin{figure}[htbp!]
    \centering
    \includegraphics[width=0.9\linewidth]{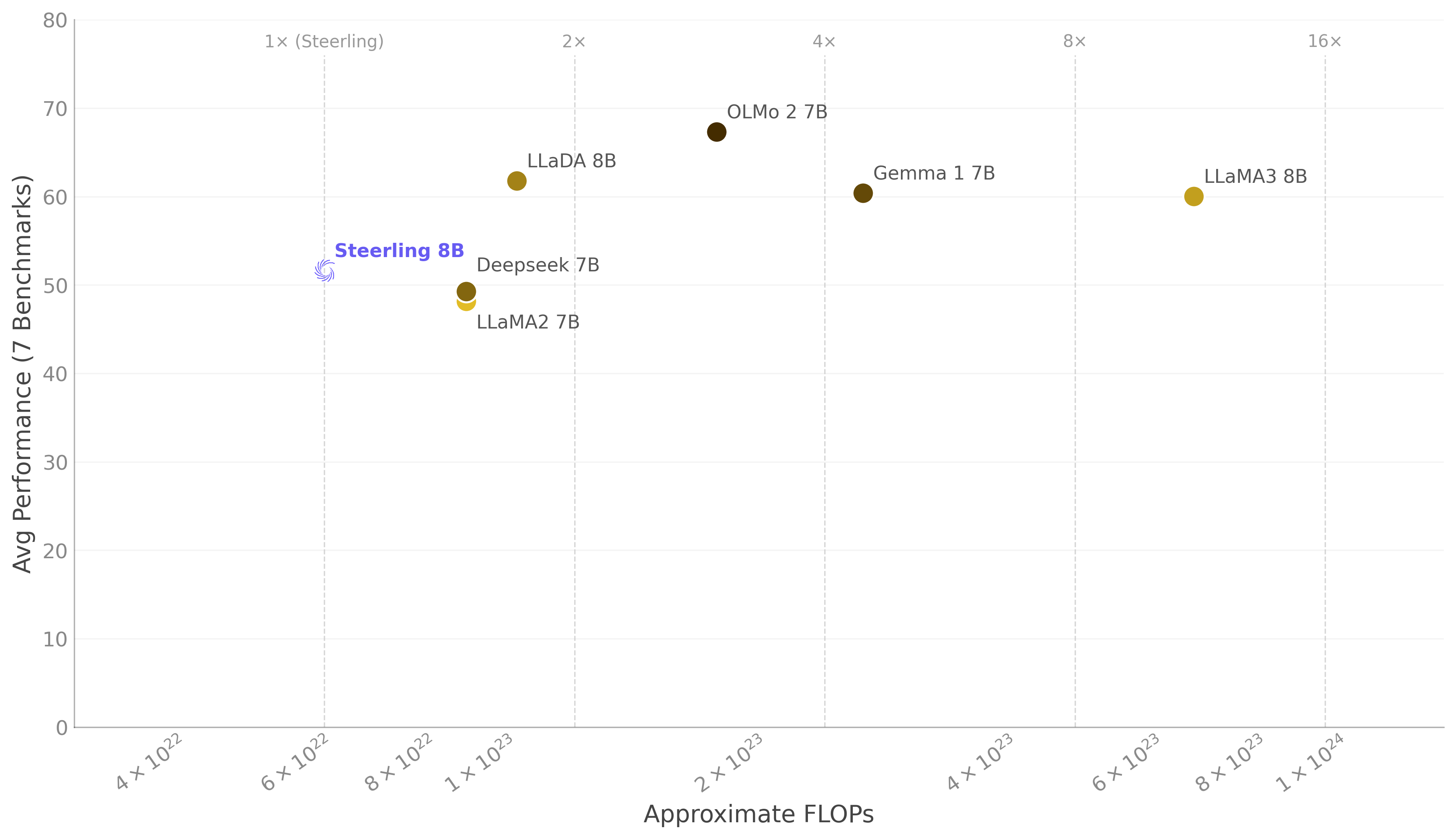}
    \caption{Average performance across the benchmark suite against approximate training FLOPs, with vertical lines at multiples of \steerlingB's compute.}
    \label{fig:flops}
\end{figure}

% \paragraph{\steerlingB base.}
\cref{tab:base-results} compares \steerlingB against open base models of comparable size on seven benchmarks, and \cref{fig:flops} places those scores on a compute axis. Every peer was trained on \textbf{2 to 16$\times$} more compute than \steerlingB, yet it lands within \textbf{$\sim$10\%} of them on average and ahead of the models trained at comparable budgets. Although \steerlingB is an inherently-interpretable architecture, we find this does not come at the cost of capability.
\keypoint{A model can be both interpretable and competitively performant.}

\input{sections/steerling_midtraining/tables/base_results}

\bigskip 
\vspace{2em}

%% file: sections/steerling_midtraining/tables/midtraining_composition.tex
% sections/steerling/tables/midtraining_composition.tex
\begin{table}[h!]
  \centering
\begin{tabular}{lcccc}
    \toprule
    Source & Math-heavy & Balanced & Code-augmented & Code-only \\
    \midrule
    Nemotron (real)        & 5.0B & 4.7B & 5.0B & --    \\
    \textit{Dolmino Math}  & 5.0B & 2.1B & 2.0B & --    \\
    StarCoder              & --   & --   & 1.0B & 10.0B \\
    FLAN                   & --   & 1.7B & 1.0B & --    \\
    peS2o                  & --   & 0.6B & 0.4B & --    \\
    Wikipedia \& Wikibooks & --   & 0.7B & 0.5B & --    \\
    Stack Exchange         & --   & 0.2B & 0.2B & --    \\
    \midrule
    Total                  & 10.0B & 10.0B & 10.0B & 10.0B \\
    \bottomrule
  \end{tabular}
  \caption{Data compositions compared in the mid-training ablation. Each arm is a 10B-token run from the final pretraining checkpoint of 1.2T tokens; entries are token counts in billions.}
  \label{tab:midtraining-composition}
\end{table}

%% file: sections/steerling_midtraining/tables/midtraining_data_results.tex
% sections/steerling/tables/midtraining_data_results.tex
\begin{table}[htbp!]
  \centering
  \setlength{\tabcolsep}{4pt}
  \begin{tabular}{lcccccccc}
    \toprule
    Composition & MMLU & GSM8K & ARC-C & HSwag & HEval & MBPP & WinoG & Avg. \\
    \midrule
    Pretrained model & 0.298 & 0.140 & 0.484 & 0.673 & 0.049 & 0.004 & 0.596 & 0.321 \\
    \midrule
    Math-heavy        & 0.376 & \textbf{0.441} & 0.492 & 0.681 & 0.037 & \textbf{0.012} & 0.616 & \textbf{0.379} \\
    Balanced     & \textbf{0.416} & 0.328 & 0.497 & \textbf{0.693} & 0.037 & 0.006 & 0.616 & 0.371 \\
    Code-augmented    & \textbf{0.416} & 0.328 & \textbf{0.498} & \textbf{0.693} & 0.055 & \textbf{0.012} & \textbf{0.628} & 0.376 \\
    Code-only         & 0.303 & 0.086 & 0.434 & 0.630 & \textbf{0.061} & \textbf{0.012} & 0.583 & 0.301 \\
    \bottomrule
  \end{tabular}
  \caption{Downstream performance of the four mid-training compositions, each a 10B-token run from the final pretraining checkpoint, against the base model. Best in each column in bold. HSwag: HellaSwag; HEval: HumanEval; WinoG: WinoGrande.}
  \label{tab:midtraining-results}
\end{table}

%% file: sections/steerling_midtraining/tables/midtraining_mixture.tex
% sections/steerling/tables/midtraining_mixture.tex
\begin{table}[htbp!]
  \centering
  \begin{tabular}{lcc}
    \toprule
    Source & Number of Tokens & Ratio (\%) \\
    \midrule
    Nemotron (real)        & 72.79B & 48.5 \\
    StarCoder              & 30.75B & 20.5 \\
    peS2o                  & 21.98B & 14.7 \\
    \textit{Dolmino Math} ($\sim$2$\times$) & 16.05B & 10.7 \\
    FLAN                   & 6.38B  & 4.3  \\
    Wikipedia \& Wikibooks & 1.50B  & 1.0  \\
    Stack Exchange         & 0.56B  & 0.4  \\
    \midrule
    Total                  & 150B   & 100.0 \\
    \bottomrule
  \end{tabular}
  \caption{The final \steerlingB midtraining mixture of 150B tokens. \textit{Dolmino Math} is upsampled roughly twofold. Proportions follow the code-augmented composition of~\cref{tab:midtraining-composition}.}
  \label{tab:midtraining-mixture}
\end{table}

%% file: sections/steerling_midtraining/tables/midtraining_masking_results.tex
% sections/steerling/tables/midtraining_masking_results.tex
\begin{table}[htbp!]
  \centering
  \setlength{\tabcolsep}{4pt}
  \begin{tabular}{lcccccccc}
    \toprule
    Masking & MMLU & GSM8K & ARC-C & HSwag & HEval & MBPP & WinoG & Avg. \\
    \midrule
    Pretrained model  & 0.298 & 0.140 & 0.484 & 0.673 & 0.049 & 0.004 & 0.596 & 0.321 \\
    50\% uniform       & \textbf{0.416} & 0.328 & 0.498 & \textbf{0.693} & \textbf{0.055} & \textbf{0.012} & \textbf{0.628} & \textbf{0.376} \\
    80\% Gaussian      & 0.280 & \textbf{0.355} & \textbf{0.500} & 0.686 & \textbf{0.055} & 0.008 & \textbf{0.628} & 0.359 \\
    \bottomrule
  \end{tabular}
  \caption{Masking schedule ablation, each a 10B-token run from the final pretraining checkpoint. Best in each column in bold. HSwag: HellaSwag; HEval: HumanEval; WinoG: WinoGrande.}
  \label{tab:midtraining-masking}
\end{table}

%% file: sections/steerling_midtraining/tables/steering_midtraining_ablation.tex
% sections/steerling/tables/steering_midtraining_ablation.tex
% \begin{table}[htbp!]
% \centering
% \small
% \setlength{\tabcolsep}{6pt}
% \begin{tabular}{@{}lcc@{}}
% \toprule
% & Math Heavy & \;+ steering \\
% \midrule
% \multicolumn{3}{@{}l}{LM Harness} \\
% \quad MMLU        & 0.370 & \textbf{0.384} \\
% \quad GSM8K       & \textbf{0.431} & 0.415 \\
% \quad ARC-C       & 0.499 & \textbf{0.505} \\
% \quad HellaSwag   & \textbf{0.682} & 0.681 \\
% \quad WinoGrande  & \textbf{0.618} & 0.611 \\
% \midrule
% \multicolumn{3}{@{}l}{Steering benchmark} \\
% \quad Concept $\uparrow$   & 1.208 & \textbf{1.244} \\
% \quad Quality $\uparrow$   & 0.989 & \textbf{1.139} \\
% \quad Harmonic $\uparrow$  & 1.088 & \textbf{1.189} \\
% \bottomrule
% \end{tabular}
% \caption{Steering ablation: Adding the steering phases improves every metric on the steering benchmark~\citep{wu2025axbench}, while LM Harness performance stays in range on every benchmark.}
% \label{tab:steering-midtraining-ablation}
% \end{table}

% sections/steerling/tables/steering_midtraining_ablation.tex
\begin{table}[htbp!]
\centering
\small
\setlength{\tabcolsep}{4pt}
\begin{tabular}{@{}lcccccccc@{}}
\toprule
& \multicolumn{5}{c}{LM Harness $\uparrow$} & \multicolumn{3}{c}{Steering benchmark $\uparrow$} \\
\cmidrule(lr){2-6} \cmidrule(lr){7-9}
& MMLU & GSM8K & ARC-C & HSwag & WinoG & Concept & Quality & Harmonic \\
\midrule
Math-heavy   & 0.370 & \textbf{0.431} & 0.499 & \textbf{0.682} & \textbf{0.618} & 1.208 & 0.989 & 1.088 \\
\;+ steering & \textbf{0.384} & 0.415 & \textbf{0.505} & 0.681 & 0.611 & \textbf{1.244} & \textbf{1.139} & \textbf{1.189} \\
\bottomrule
\end{tabular}
\caption{Steering ablation: Adding the steering phases improves every metric on the steering benchmark~\citep{wu2025axbench}, while LM Harness performance stays mostly unchanged.}
\label{tab:steering-midtraining-ablation}
\end{table}

%% file: sections/steerling_midtraining/tables/midtraining_rescue.tex
% sections/steerling/tables/midtraining_rescue.tex
\begin{table}[htbp!]
  \centering
  \setlength{\tabcolsep}{6pt}
  \begin{tabular}{lcccccccc}
    \toprule
    & MMLU & GSM8K & ARC-C & HSwag & HEval & MBPP & WinoG & Avg. \\
    \midrule
    Pretrained        & 29.8 & 14.0 & 48.4 & 67.3 & 4.9 & 0.4 & 59.6 & 32.1 \\
    Mid-trained       & \textbf{46.4} & \textbf{44.4} & \textbf{52.3} & \textbf{70.3} & \textbf{8.5} & \textbf{11.0} & \textbf{64.2} & \textbf{42.4} \\
    \bottomrule
  \end{tabular}
  \caption{\steerlingB before and after mid-training across the LM Harness suite (accuracy, \%). HSwag: HellaSwag; HEval: HumanEval; WinoG: WinoGrande.}
  \label{tab:midtraining-rescue}
\end{table}

%% file: sections/steerling_midtraining/tables/midtraining_interp.tex
% sections/steerling/tables/midtraining_interp.tex
\begin{table}[htbp!]
\centering
\small
\setlength{\tabcolsep}{8pt}
\begin{tabular}{@{}lcc@{}}
\toprule
& Pretrained & Mid-trained \\
\midrule
\quad Concept Loss $\downarrow$              & 0.002 & 0.002 \\
\quad Concept Independence Loss $\downarrow$ & 1.907 & \textbf{1.546} \\
\quad Concept Contribution $\uparrow$        & 0.851 & \textbf{0.876} \\
\quad Known Concept Alignment $\uparrow$     & 3.730 & \textbf{3.770} \\
\bottomrule
\end{tabular}
\caption{Interpretability metrics for \steerlingB before and after mid-training.}
\label{tab:midtraining-interp}
\end{table}

%% file: sections/steerling_midtraining/tables/midtraining_steering_pre_vs_midtrained.tex
% sections/steerling/tables/steering_pretrained_vs_midtrained.tex
\begin{table}[htbp!]
\centering
\small
\setlength{\tabcolsep}{8pt}
\begin{tabular}{@{}lcccc@{}}
\toprule
Checkpoint & Concept $\uparrow$ & Quality $\uparrow$ & Harmonic $\uparrow$ & Sample harmonic $\uparrow$ \\
\midrule
Pretrained (1.2T)   & 1.072 & 0.972 & 1.020 & 0.843 \\
Mid-trained (1.35T) & \textbf{1.247} & \textbf{1.064} & \textbf{1.148} & \textbf{0.963} \\
\bottomrule
\end{tabular}
\caption{Steering benchmark scores for the pretrained and mid-trained \steerlingB checkpoints. Mid-training improves every steering metric.}
\label{tab:steering_pretrained_vs_midtrained}
\end{table}

%% file: sections/steerling_midtraining/tables/base_results.tex
% sections/steerling/tables/base_results.tex
\begin{table}[htbp!]
  \centering
  \small
  \setlength{\tabcolsep}{6pt}
  \begin{tabular}{lccccc|cc|c}
    \toprule
    Model & HSwag & WinoG & PIQA & MMLU & ARC-C & GSM8K & Math & Avg. \\
    \midrule
    \steerlingB & 70.3 & 64.2 & 75.9 & 46.4 & 52.3 & 44.4 & 8.0 & 51.6 \\
    \midrule
    LLaMA2 7B   & 76.0 & 72.5 & 79.1 & 45.9 & 46.3 & 13.1 & 4.3 & 48.2 \\
    DeepSeek 7B & 75.4 & 70.5 & 79.2 & 48.2 & 48.1 & 17.4 & 6.0 & 49.3 \\
    Gemma 1 7B  & 81.2 & 72.3 & 81.2 & 64.3 & 53.2 & 46.4 & 24.3 & 60.4 \\
    LLaDA 8B    & 70.5 & 74.8 & 73.6 & 65.9 & 45.9 & 70.3 & 31.4 & 61.8 \\
    LLaMA3 8B   & 79.1 & 77.3 & 80.6 & 65.4 & 53.1 & 48.7 & 16.0 & 60.0 \\
    OLMo 2 7B   & 83.8 & 77.2 & 80.1$^{*}$ & 63.7 & 79.8 & 67.5 & 19.1$^{*}$ & 67.3 \\
    \bottomrule
  \end{tabular}
  \caption{\steerlingB base model against open base models of comparable size. HSwag: HellaSwag; WinoG: WinoGrande. Values marked $^{*}$ are taken from the OLMo~3 report~\citep{olmo}; all other peer numbers are from the respective model reports.}
  \label{tab:base-results}
\end{table}

%% file: sections/steerling_posttraining/main.tex
% \clearpage
\section{\steerlingB: Post-training}
\label{sec:posttraining}

Our post-training is an instruction supervised fine-tuning (ISFT) performed on the mid-trained model described in \cref{sec:midtraining}. There are two goals here. The first goal is the typical instruction-tuning goal, where the model's behavior is changed from in-filling \texttt{[MASK]} tokens to match the context distribution, to in-filling \texttt{[MASK]} tokens such that they answer the user-provided instruction. The second goal is to maintain the interpretability capabilities from pretraining and mid-training.

\subsection{Data}

\paragraph{ISFT Dataset.} For the ISFT dataset we use the \textit{Dolci Instruct SFT} dataset, which was originally used to train OLMo 3 \citep{olmo}. Specifically, we use the variant without tools\footnote{Available on Hugging Face as \href{https://huggingface.co/datasets/allenai/Dolci-Instruct-SFT-No-Tools}{\texttt{allenai/Dolci-Instruct-SFT-No-Tools}}.}. This dataset has then been annotated with concepts. Each dialog-turn is treated as a separate document, and chunked using the same approach as for the pretraining dataset (\cref{sec:data-stage1}), and then the entire dataset is annotated using the same concept annotator (\cref{sec:data-processing}).

\paragraph{Dialog template with chunks.} To convert multi-turn dialogs into token sequences, we apply the Llama-3 chat template \citep{grattafiori2024llama}, omitting the beginning-of-sequence (\texttt{[BOS]}) token since \steerling, as a diffusion model, does not require one. This follows LLaDA \citep{llada}, which uses the same template. The template has the following structure:
\begin{lstlisting}[breaklines=true, breakatwhitespace=false, basicstyle=\ttfamily\small]
<|start_header_id|>system<|end_header_id|>\n\n{system}<|eot_id|>
<|start_header_id|>user<|end_header_id|>\n\n{user}<|eot_id|>
<|start_header_id|>assistant<|end_header_id|>\n\n{assistant}<|eot_id|>
\end{lstlisting}
\noindent where \verb!<|start_header_id|>!, \verb!<|end_header_id|>!, and \verb!<|eot_id|>! are special tokens.

To integrate the notion of chunks and concepts, we treat framing tokens (e.g., \texttt{\mbox{<|start\_header\_id|>}\allowbreak\mbox{user}\allowbreak\mbox{<|end\_header\_id|>}\allowbreak\mbox{\textbackslash n\textbackslash n}} and \verb!<|eot_id|>!) as separate chunks without concepts. The system, user, and assistant messages are themselves segmented into chunks and an \texttt{[EOC]} token is appended after each chunk. Because the loss functions require knowing which chunk and concept each token is aligned to, and which tokens are framing tokens, we implement a custom version of the template that provides these annotations in addition to the token IDs.

\paragraph{Unfolded multi-turn dialog.} Following LLaDA \citep{llada}, we unfold each multi-turn dialogue into its prefixes: an observation $\{\textbf{u}_1, \textbf{a}_1, \dots, \textbf{u}_I, \textbf{a}_I\}$ of user ($\textbf{u}_i$) and assistant ($\textbf{a}_i$) messages becomes $I$ observations $\{\textbf{u}_1, \textbf{a}_1\}$, $\{\textbf{u}_1, \textbf{a}_1, \textbf{u}_2, \textbf{a}_2\}$, \dots, $\{\textbf{u}_1, \textbf{a}_1, \dots, \textbf{u}_I, \textbf{a}_I\}$. In a fully causal model this would be unnecessary, since $\textbf{a}_i$ never attends to later messages. \steerling, however, attends bidirectionally within each block of 64 tokens. Because messages are concatenated without padding, a message can end anywhere within a block, so the final block of $\textbf{a}_i$ typically also contains the start of $\textbf{u}_{i+1}$. Without unfolding, every non-final assistant turn would appear with part of $\textbf{u}_{i+1}$ visible in the same block, whereas at inference $\textbf{a}_i$ is generated before $\textbf{u}_{i+1}$ exists. Unfolding ensures that every assistant turn appears as the final message in some observation, matching the inference setting.

\subsection{Post-training recipe}

The post-training recipe is inspired by LLaDA \citep{llada}, but adds additional considerations in order to satisfy the interpretability requirements.

\paragraph{Padding.} Unlike LLaDA \citep{llada}, which pads sequences with end-of-text tokens (termed \texttt{<|endoftext|>}) to make them equal in length, we pad only the last diffusion-block that contains content with \texttt{<|endoftext|>} tokens. Tokens after, do not contribute to the gradient. This is because \steerling is block-causal, thus we have no need to learn how to generate blocks after the last diffusion-block.

\paragraph{Masking.} Similar to mid-training (\cref{sec:steerling-midtraining}) and LLaDA \citep{llada}, we use a uniform masking with masking ratio $t \sim \mathcal{U}(0, 1)$ sampled per observation. However, unlike LLaDA, we apply masking noise to all input tokens, not just the final assistant response ($\textbf{a}_I$) and \texttt{<|endoftext|>} tokens. This ensures that masking remains in-distribution for all tokens, which is one of our interpretability criteria (\cref{sec:inherent-interp}) as it enables faithful input-attribution.

\paragraph{Losses.} We only accumulate the $\mathcal{L}_{LM}$ loss on the masked tokens of the final assistant output and \texttt{<|endoftext|>} tokens. This is because only these tokens need to be predicted in a generation setting. 

Regarding interpretability losses, the same losses used mid-training are used (\cref{subsubsec:loss-objectives} and \cref{sec:midtraining-losses}), except for the steering losses (\cref{subsec:training-steering}) which are not included. Additionally, these losses are again only applied to chunks and tokens of the final assistant response ($\textbf{a}_I$) and the \texttt{[EOS]} tokens from the finishing block.

Specifically, the concept loss ($\mathcal{L}_{concept}$) in \eqref{eq:concept-loss} is applied only to chunks of the final assistant response. For the reconstruction loss ($\mathcal{L}_{rec}$) in \eqref{eq:rec-loss} and independence loss ($\mathcal{L}_{indep}$) in \eqref{eq:indep-loss}, the set of mask tokens $\mathcal{M}$ is the mask tokens from the final assistant output and the \texttt{[EOS]} in the finishing block.

\paragraph{Hyperparameters.} We use the same linear-warmup-and-cooldown schedule for the learning-rate, the same learning-rate, and the same batch-size, as LLaDA \citep{llada}. The batch-size is 256 observations, the linear-warmup takes 50 steps from $0$ to $2.5 \times 10^{-5}$, and the linear-cooldown is on the last 10\% to $2.5\times 10^{-6}$.

Regarding interpretability hyperparameters, post-training starts where mid-training ends. Teacher forcing is entirely disabled for all of post-training (i.e. $\alpha_{\mathrm{known}} = \alpha_{\mathrm{unknown}} = 0$). The same residual dropout is used (i.e. $p_\eps = 0.3$), and the same top-k concepts from the known and unknown heads are used, 32 and 128 respectively (\cref{sec:midtraining-architecture}). Also similar to mid-training, $\mathcal{L}_{LM}$, $\mathcal{L}_{concept}$, $\mathcal{L}_{rec}$, and $\mathcal{L}_{indep}$ are all weighted equally.

\subsection{Generation}

For benchmarking, we generate text blockwise. Each block is initialized as 64 \texttt{[MASK]} tokens and decoded iteratively: at each step, we perform one forward pass and unmask the masked position with the highest confidence, repeating until the block contains no masked tokens. We then initialize the next 64-token block and proceed in the same way. This procedure follows LLaDA \citep{llada}, except that we use a fixed block size of 64 tokens.

For the post-trained model, we observe a degenerate mode in which the model first unmasks the last token of the block as \texttt{[EOS]}, then the second-to-last token as \texttt{[EOS]}, and so on, yielding a block consisting entirely of \texttt{[EOS]} tokens. This behavior is plausible given the training data: \texttt{[EOS]} frequently occupies the final position of a block, and \texttt{[EOS]} tokens typically appear adjacent to one another, making these predictions high-confidence. We did not observe this behavior when generating with LLaDA \citep{llada} under a comparable procedure. The issue is resolved by setting the logit of \texttt{[EOS]} to $-\infty$ during generation. The \verb! <|eot_id|>! token still marks the end of the generation: once \verb! <|eot_id|>! is unmasked, all subsequent tokens in the block are set to \texttt{[EOS]}.

When a maximum generation length is specified, all positions beyond it are likewise set to \texttt{[EOS]}. We find that this helps the model terminate the generation naturally, since it is informed of where the generation must end. Note that this intervention has no analogue in autoregressive models, which cannot condition on future positions.

\subsection{Evaluation and results}

For pretraining and mid-training, ELBO-based likelihood estimates (\cref{app:scaling_laws_ELBO}) are often used for benchmarking. However, for the post-trained model we follow OLMo~3 and LLaDA and exclusively use generation for benchmarking \citep{olmo, llada}.

\begin{figure}[htbp!]
    \centering
    \includegraphics[width=0.9\linewidth]{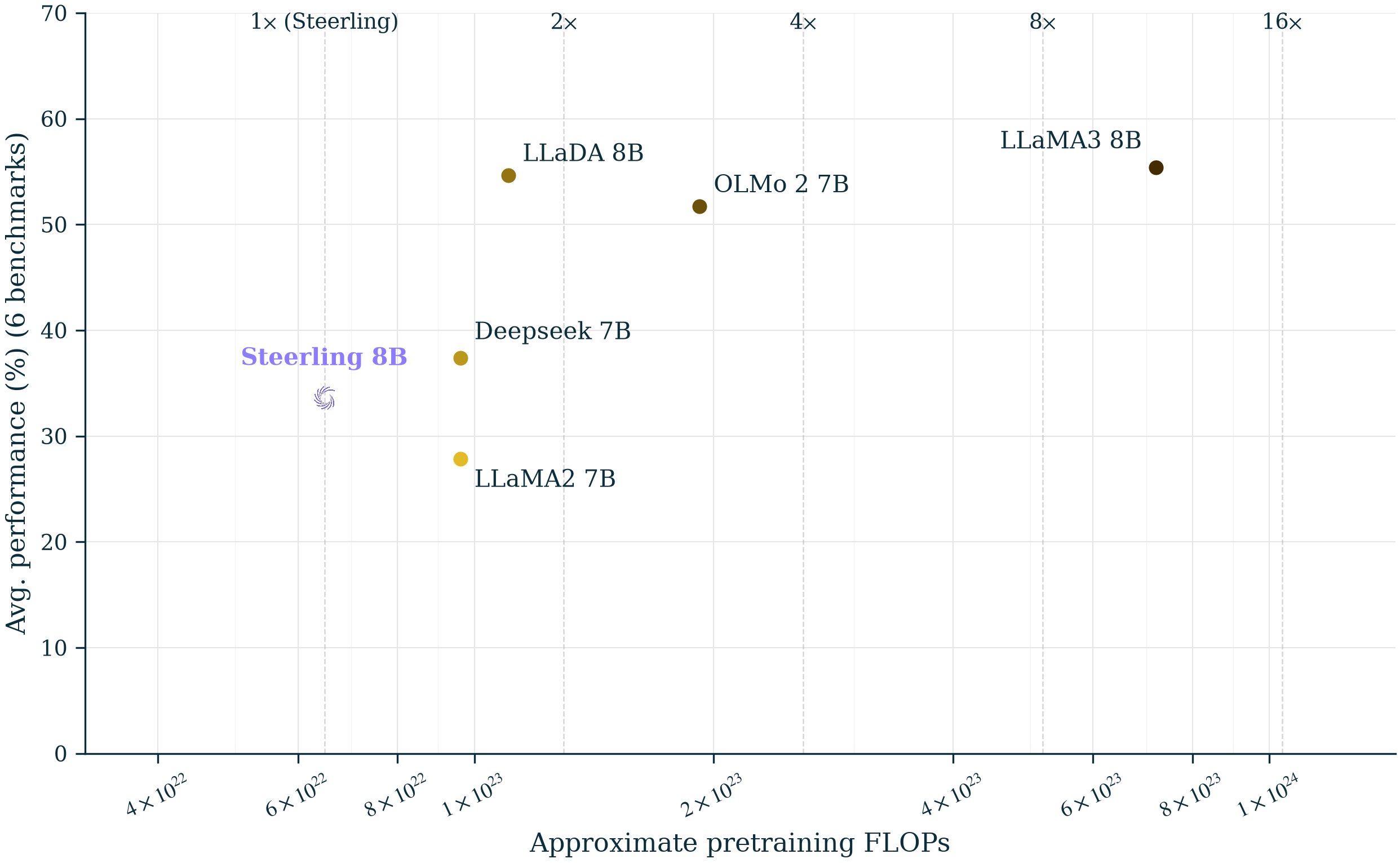}
    \caption{Average performance across the benchmark suite against approximate training FLOPs, with vertical lines at multiples of \steerlingB's compute.}
    \label{fig:posttraining-flops}
\end{figure}

\input{sections/steerling_posttraining/tables/gen_benchmark_results}

\cref{tab:sft-results} and \cref{fig:posttraining-flops} compares the SFT version of \steerlingB against open base models of comparable size. We generally find that the post-training results follow the mid-training results (\cref{sec:steerling-midtraining}), that \steerlingB's performance is competitive given the pretraining FLOPs while also maintaining the interpretability capabilities.

\paragraph{Post-training maintains interpretability requirements.}

\input{sections/steerling_posttraining/tables/posttraining_interp}

Interpretability metrics after post-training are shown in \cref{tab:posttraining-interp}, which shows that both concept contribution and known concept alignment improves, meaning the concepts accounts for a larger share of the output, while also being of a higher quality, a difficult combination to achieve. However, the concept independence loss does decrease, indicating there is more overlap between known and unknown concepts.

Post-training in it's optimization specifications just seeks to maintain the interpretability criteria, so it's unclear why such differences would be observed. One hypothesis, is that concept expression differs significantly in an instruction context compared to a pre-training context. The concepts of the assistant chunks, might be significantly different from the user chunks, which is unlike pre-training where the concepts of adjacent chunks are likely to be similar, this can make concept prediction more challenging in post-training. On the other hand, assistant answers tends to carry a more direct message, so the concept expression can be more clearly expressed in the output.

%% file: sections/steerling_posttraining/tables/gen_benchmark_results.tex
% Requires \usepackage{booktabs} and \citep support (e.g., natbib).
\begin{table}[htbp!]
  \centering
  \small
  \setlength{\tabcolsep}{6pt}
  \begin{tabular}{lcccc|cc|c}
    \toprule
    Model & MMLU & MMLU-Pro & ARC-C & GPQA & GSM8K & Math & Avg. \\
    \midrule
    Steerling 8B & 44.6 & 17.4 & 50.5 & 29.5 & 49.2 & 10.4 & 33.6 \\
    \midrule
    LLaMA2 7B    & 44.1 & 4.6 & 57.3 & 28.4 & 29.0 & 3.8 & 27.9 \\
    DeepSeek 7B  & 48.2 & 21.33 & 49.4 & 26.59 & 63.0 & 15.8 & 37.4 \\
    LLaDA 8B     & 65.5 & 37.0 & 88.5 & 31.8 & 78.6 & 26.6 & 54.7 \\
    LLaMA3 8B    & 68.4 & 41.9 & 82.4 & 31.9 & 78.3 & 29.6 & 55.4 \\
    OLMo 2 7B    & 61.1 & 33.2 & 79.7 & 36.4 & 74.6 & 25.3 & 51.7 \\
    \bottomrule
  \end{tabular}
  \caption{Steerling-8B-Instruct against open SFT models of comparable size. Scores are percentages; Avg. is the unweighted mean across all six benchmarks, rounded to one decimal place. Scores for LLaDA, LLaMA2, LLaMA3, and DeepSeek are from the LLaDA paper~\citep{llada}, except DeepSeek's MMLU-Pro and GPQA scores, which are from the Open LLM Leaderboard~\citep{deepseek-open-llm-leaderboard}. OLMo~2's scores are from its report~\citep{olmocore}, with ARC-C from Table~8 of \citet{lin-etal-2025-efficient}.}
  \label{tab:sft-results}
\end{table}

%% file: sections/steerling_posttraining/tables/posttraining_interp.tex
% sections/steerling_posttraining/tables/posttraining_interp.tex
\begin{table}[htbp!]
\centering
\small
\setlength{\tabcolsep}{8pt}
\begin{tabular}{@{}lccc@{}}
\toprule
& Pretrained & Mid-trained & Post-training \\
\midrule
\quad Concept Loss $\downarrow$              & 0.002 & 0.002 & 0.002 \\
\quad Concept Independence Loss $\downarrow$ & 1.907 & \textbf{1.546} & 2.863 \\
\quad Concept Contribution $\uparrow$        & 0.851 & 0.876 & \textbf{0.893} \\
\quad Known Concept Alignment $\uparrow$     & 3.730 & 3.770 & \textbf{3.810} \\
\bottomrule
\end{tabular}
\caption{Interpretability metrics for \steerlingB for pre-, mid-, and post-training.}
\label{tab:posttraining-interp}
\end{table}

%% file: sections/related_works/main.tex
\section{Related work}
\label{sec:related-work}

\subsection{Underspecification and the Rashomon effect}
\label{sec:related-underspecification}

\textbf{The phenomenon.} For overparameterized neural networks, many distinct 
models achieve equivalent performance on the training objective while differing 
in their internal 
mechanisms~\citep{d2022underspecification,breiman2001statistical, black2022model, rudin2024amazing}. 
\citet{fisher2018all} formalized this as the \emph{Rashomon set}, the collection 
of models within a small tolerance of the optimum, and showed that feature 
importance varies substantially across its members. 
\citet{semenova2022existence} demonstrated that these sets are large in practice, 
with structurally diverse models routinely coexisting at equivalent performance.

\textbf{The problem for post-hoc interpretability.} Even a perfectly faithful 
explanation is an account of one arbitrary member of the Rashomon set; a different 
training run could yield an equally valid model with an entirely different 
internal decomposition. Worse,
\citet{brunet2022implications} show that models with nearly identical accuracy 
can produce contradicting explanations, e.g. opposite-sign attributions for the 
same feature, with no diagnostic to predict when this occurs. \citet{pawelczyk2020counterfactual} 
show that counterfactual recommendations derived from one model's decision 
boundary can be invalid for an equally valid alternative.

\looseness=-1
\textbf{How inherent interpretability addresses this?} Our approach constrains the Rashomon set during training. The 
concept library is fixed before training begins, the concept module forces 
every trained model to decompose its output through the same concept variables, 
and the masking objective ensures a shared absence baseline. Different runs still yield 
different parameters, but the attribution interface: which concepts exist, how 
attribution is computed, and what ``absent'' means, is fixed.

 \subsection{Large language models} 
\label{sec:related-lm} 

Most large language models are autoregressive (AR): they generate a sequence left to right, one token at a time, each conditioned on all preceding tokens \citep{radford2018improving}. Trained at scale with a next-token prediction objective, these models attain strong performance across a wide range of tasks, with training recipes and open model weights now widely documented by the community \citep{llama3, olmo, deepseek}. Most deployed production models belong to this family \citep{gemini, claude, gpt4}, making the AR pipeline a mature and well-understood baseline.

Diffusion language models generate by iteratively denoising a corrupted sequence, producing many tokens in parallel and in arbitrary order. Discrete diffusion comes in several families that trade off quality and efficiency: masked diffusion attains the strongest perplexity \citep{mdlm, llada}, uniform-state diffusion yields higher-quality samples in the few-step regime and is well suited to guidance \citep{austin2021structured,schiff2025simple,sahoo2025diffusion}, and interpolating diffusion supports KV caching for faster decoding \citep{sahoo2025esoteric}. \blockdiff bridges the two paradigms by factorizing autoregressively over blocks while denoising within each block, recovering KV caching at the cost of full block-wise attention \citep{blockdiffusion}. Recent work has further shown that masked diffusion models scale to billions of parameters \citep{llada} and now serve production workloads \citep{seed-diffusion, mercury}. Masked diffusion models are also a natural substrate for interpretability: the \texttt{[MASK]} token gives a learned baseline for attribution, parallel generation supports concept-level control, and any-order generation enables clean interventions. \steerling builds on this with \causaldiff, a block-causal formulation that attains the benefits of \blockdiff at roughly half the cost, and adds the concept module on top.

\subsection{Scaling laws}
\label{sec:related-scaling} 

Scaling laws characterize how model performance improves with compute, parameters, and data, and have become the standard tool for planning large training runs. For autoregressive language models, this line of work established power-law relationships between loss and scale and the compute-optimal allocation of parameters and tokens \citep{kaplan2020scaling, hoffmann2022training, bi2024deepseek}. More recently, the same methodology has been extended to diffusion language models, both for masked diffusion \citep{mdlm, scaling_behaviorDLM_von} and through compute-optimal studies tailored to the diffusion objective \citep{quokka, scaling_beyond_sahoo}. We follow this methodology and fit compute-optimal scaling laws to inherently interpretable models, measuring the effect of the concept module on both the autoregressive and diffusion families.

\looseness=-1
A separate line of work asks whether interpretability itself scales. Sparse autoencoders trained on frozen model activations recover more and finer features as the autoencoder grows, with reconstruction quality and feature-quality metrics following clean power laws \citep{gao2025scaling, templeton2024scaling}, and recurring neuron populations become more selective and monosemantic as the base model grows \citep{dravid2026neuron}. These works ask whether a post-hoc probe gets better as the probe is scaled, on a fixed, uninterpretable model. They do not ask how a model's own interpretability scales with its training compute. Adapting the irreducible-loss scaling form of \citet{gao2025scaling}, we test whether interpretability-by-design preserves compute-optimal scaling and whether the model's interpretability metrics improve predictably with compute across several distinct measures.

\subsection{Interpretable-by-design architectures}
\label{sec:related-cbm}
A range of architectures build interpretability into the model rather than recovering it post hoc. Among these, Concept Bottleneck Models (CBMs) \citep{cbm} route predictions through a layer of human-understandable concepts. Concept Embedding Models relax the bottleneck to recover accuracy \citep{espinosa2022concept}. Originally developed for classification, the approach was later extended to generative models, where intervening on the concept layer enables interpretable and controllable generation \citep{cbgm}. It has since reached protein language modeling \citep{cbplm} and single-cell counterfactuals generation \citep{sccbgm}. Backpack language models pursue a similar goal through a different mechanism, attaching interpretable sense vectors to each token in place of named concepts \citep{hewitt2023backpack}. 
A parallel line pursues interpretability through prototypes rather than concepts. PRISM~\citep{ley2026prototype} forms each prediction from a sparse, non-negative mixture of learned prototypes, where each prototype anchors to a coherent neighborhood of training examples. 
This yields structural training-data attribution at scale without post-hoc estimation.
All of these operate at a substantially smaller scale, and with far fewer concepts, than \steerling. We use an additive CBM to scale the concept vocabulary to over a \emph{hundred thousand concepts}, with the goal of reaching millions.

% \subsection{Training and finetuning interpretability constraints}
% We are not the first to follow this paradigm, but we scale it. 
% \aya{Do we really need this?}

\subsection{Attribution methods}

\paragraph{Input attribution.}
% Integrated Gradients \citep{integrated_gradient} integrate the gradient along a straight-line path from a baseline to the input, with axiomatic guarantees on sensitivity and implementation invariance. Standard implementations use a zero or padding embedding as the baseline. We use \texttt{[MASK]} instead. The diffusion training objective makes \texttt{[MASK]} a learned representation of ``no information at this position,'' so the integration path stays inside the model's training distribution. Autoregressive models offer no comparable learned baseline through their training objective.

We use Integrated Gradients \citep{integrated_gradient}, but the core of our input attribution is not the algorithm: it is the choice of baseline. Prior Integrated Gradient implementations use a zero or padding embedding as the baseline~\citep{kokhlikyan2020captum,nguyen2021effectiveness}; the model never learned to interpret either as absence of information. We use \texttt{[MASK]} instead. The diffusion training objective makes \texttt{[MASK]} a learned representation of ``no information at this position,'' so the integration path stays inside the model's training distribution. Autoregressive models offer no comparable learned baseline through their training objective.

\paragraph{Concept attribution.}
Methods that attribute predictions to human-interpretable concepts divide into post-hoc and architectural approaches. Post-hoc methods include linear probes \citep{alain2016understanding} and TCAV \citep{kim2018interpretability}, which detect concepts in activations or test their influence on predictions, and sparse autoencoders \citep{huben2024sparse, gao2025scaling}, which factorize activations into interpretable features. All are approximations of an internal representation that was not designed to be decomposed. Architectural methods embed concept supervision in the model itself, and we cover this family in Section~\ref{sec:related-cbm}. \steerling sits in the architectural family and produces an exact additive decomposition of every output logit, so concept attribution reads off the forward pass rather than estimating it.

\paragraph{Training data attribution.}
Influence functions estimate the effect of perturbing a training point on a model's predictions \citep{koh2017understanding}. Scaling them to large language models requires approximations to the Hessian inverse \citep{grosse2023studying}, and the resulting estimates can be brittle in non-convex regimes \citep{bae2022if}. TracIn takes a different approach, tracing influence along the optimization trajectory \citep{pruthi2020estimating}. 
A complementary line avoids estimation altogether: OLMoTrace \citep{liu2025olmotrace} retrieves verbatim substring matches between an output and the training corpus, surfacing documents the model could have seen.
A separate line makes TDA a first-class output of the architecture rather than a post-hoc estimate, achieving retrieval roughly 500× faster than influence-function baselines at matched memory~\citep{ley2026prototype}.
We adopt the same retrieval framing but match in latent representation space rather than by surface form: given an output, we return the training chunks most similar to it in that space. We do not claim our method estimates causal influence at trillion-token scale.

\subsection{Model steering}
\looseness=-1Methods for steering model behavior at inference time fall into two families. Both add a learned direction to hidden activations, with the sign of the injection determining whether the target behavior is amplified or suppressed. They differ in how the direction is obtained. The first family derives the direction from positive and negative examples of the target behavior, generating a new direction per task: representation engineering~\citep{zou2023representation}, steering vectors~\citep{turner2023activation}, and contrastive activation addition~\citep{rimsky2024steering}. The second family selects directions from a pre-built feature dictionary that is independent of any specific behavior we desire to steer. Parsimonious concept engineering~\citep{luo2024pace} and sparse autoencoder features~\citep{huben2024sparse} decompose activations into a fixed set of interpretable features, any of which can serve as a steering direction. Both families obtain the direction \emph{post-hoc}, from activations the model has already produced. Our model instead has its steering directions built into the architecture: each concept contributes an exact additive term to every output logit, so positive or negative steering is a closed-form edit on that term, requiring neither contrastive derivation nor post-hoc dictionary construction.

% \newpage

%% file: sections/conclusion.tex
\section{Conclusion}
\label{sec:conclusion}
Interpretability is often treated as a cost paid against capability.
In this work, we tested whether that cost, indeed, grows with scale. However, we demonstrate concrete settings where this is not the case. 
Concretely, across three orders of magnitude of compute, on both autoregressive and causal-diffusion language models, training a model with interpretable structure shifts compute-optimal scaling by a small, fixed offset rather than a penalty that compounds with scale. 
More surprisingly, we find that all the interpretability constraints we train for improve with compute: larger models have more independent, and more semantically aligned use of human-understandable concepts. 
Under the metrics we measure, the model does not become harder to understand as it becomes more capable; it becomes easier.

The central change is where interpretability enters the modeling process. 
Standard pipelines train opaque predictors and then ask whether
post-hoc methods can recover faithful explanations. Instead, we ask which conditions training must enforce for explanations to be faithful, and build those conditions into the data, architecture, objective, and losses. 
The resulting recipe is not a collection of interpretability add-ons: each component exists because removing it breaks a specific condition required for faithful attribution. Because the same concept variables support both attribution and intervention, a user can decompose an output, inspect the relevant concepts and similar training data, edit the responsible concept direction, and verify the immediate logit-level effect, all without retraining.

The choices made in this work are first instantiations, not settled
directions. We fixed the concept library before training, to topics
mostly describing the content of training document segments. We chose
one bottleneck design, the additive module, among many possible
alternatives. We supervised concepts at the chunk level, and our
training stops at supervised finetuning. Each of these choices can be
improved upon, and we expect future work to do so substantially.

A few prospects seem tantalizing. The first is post-training:
reward objectives can be expressed over concepts, so that
training targets not only what a model says, but which concepts it uses to decide. 
In agentic settings, the same structure lets an agent's decisions be decomposed into inspectable concepts that are monitored and corrected mid-trajectory. 
Concept libraries can grow to match, becoming hierarchical and adaptive rather than fixed. 
Furthermore, individuals should be able to interactively define, audit, and extend the vocabularies through which models explain themselves. 
Second, the attribution interfaces can become more legible in turn: today they return structured artifacts, ranked concept contributions, token-level scores, and retrieved training chunks; future work can translate these into natural-language explanations that remain grounded in the underlying decomposition. 

Third, our scaling laws suggest that a frontier-grade interpretable model is feasible, including agentic systems whose every action can be decomposed, audited, and steered. We therefore view \steerlingB less as a finished system, but as evidence that this research program is viable.

Taken together, the results suggest a different scaling paradigm for capable AI systems. 
Interpretability, steerability, and other reliability requirements need not be retrofitted after training, nor treated as a tax against capability. 
They can be specified as a contract, optimized as part of the training process, and measured as models scale. 
Our results point to another possibility: if interpretability can be specified, trained, and scaled like any other capability, then the opacity of today's most capable systems is not a law of nature.

%% file: sections/contributions.tex
\section*{Authorship and Credit Attribution}
\label{sec:contributions}

\noindent
The results outlined in this manuscript were a collaborative effort that spanned data annotation, architecture, training infrastructure, post-training, and product. No component was built in isolation; each underwent iterative feedback from all members of the team. The statements of work below describe key primary responsibilities rather than any exclusive ownership.

\paragraph{Writing.} All members of the team contributed to writing the manuscript.

\paragraph{Data.} Nathaniel Monson, Saqib Azim, and Julius Adebayo built the Atlas data annotation system. Nathaniel Monson led the work on the LLM annotation infrastructure, designed the concept library, and the human validation study. Saqib Azim led and implemented the pipeline for Stage 2 of the Atlas system, covering clustering, de-duplication, and labeling clusters of tags as concepts; he also built the baseline k-nearest-neighbor embedding annotators and led the implementation of the distributed index of the training data. Julius Adebayo designed and trained the multi-head concept annotator, and then used it to annotate the entire pre-training corpus.

\paragraph{Model Architecture, Training Recipe, Scaling Laws, and Training.}
Aya Abdelsalam Ismail led the design of the model architecture, training recipe, and concept module. She devised, along with  Andreas Madsen, the causal diffusion attention formulation. Giang Nguyen devised the mid-training steering formulation and led its design, empirical validation, and production implementation. He also contributed to training recipe design, and implementation of both input feature and concept attribution. Aya Abdelsalam Ismail designed and executed the scaling law formulation.

\paragraph{Infrastructure.} Andreas Madsen developed the new dataloader and distributed multi-node training framework used throughout model training. Using the infrastructure, Julius Adebayo then wrote the pre-training package for model training.  Andreas Madsen and Julius Adebayo managed the internal SLURM cluster used for research and production. 

\paragraph{Open Source.} Aya Abdelsalam Ismail and Giang Nguyen led the release of Steerling base and instruct models along with the necessary artifacts to enable attribution and steering.

\paragraph{Model Post-Training.} Andreas Madsen led post-training for Steerling. He reproduced the LLaDA-8B instruction supervised fine-tuning setup, adapted the recipe to block-diffusion and interpretability training, trained the supervised fine-tuned model, and identified and fixed correctness issues across the training and evaluation stack.

\paragraph{Model Post-Processing.} Nathaniel Monson led the work on concept-naming for the trained model with feedback from Giang Nguyen and Aya Abdelsalam Ismail.

\paragraph{Clarity Platform.} Zhichen Guo and Isaac Plant designed the Clarity product that serves \steerlingB to users. Andreas Madsen led platform engineering: he designed and implemented the microservice architecture and gRPC interfaces on which the serving, attribution, and product backend systems are built. Zhichen Guo implemented the frontend and Clarity API. Muawiz Chaudhary and Andreas Madsen designed the model serving engine for the new causal diffusion model. Muawiz Chaudhary led the implementation of the serving engine with feedback from Andreas Madsen. Saqib Azim implemented the production training data attribution pipeline.

\paragraph{Supervision.} Julius Adebayo supervised the project.

\section*{Acknowledgments}

\noindent
The authors would like to thank Pietro Barbiero for his helpful feedback on this manuscript.

%% file: sections/appendix/architecture.tex
\clearpage

\part{Architecture}
\label{app:architecture}

\section{Symbol reference}
\label{appendix:architecture-notation}

\input{sections/appendix/tables/arch_notation}

%% file: sections/appendix/tables/arch_notation.tex
\begin{table}[htbp!]
\centering
\small
\setlength{\tabcolsep}{6pt}
\renewcommand{\arraystretch}{1.0}
\begin{tabular}{@{}lll@{}}
\toprule
Symbol & Type & Meaning \\
\midrule
\multicolumn{3}{l}{\textit{Hidden states}} \\
$h$ & vector & Transformer hidden state \\
$\bar{h}$ & vector & Bottlenecked state passed to the LM head \\
$\bar{h}=\hat{k}+\hat{u}+\varepsilon$ & equation & Concept module decomposition \\
\midrule
\multicolumn{3}{l}{\textit{Concept module heads}} \\
$f$ & function & Known head \\
$g$ & function & Unknown head \\
$k=\sigma(f(h))$ & vector & Per-concept activation probabilities, known \\
$u=\sigma(g(h))$ & vector & Per-concept activation probabilities, unknown \\
$k_{\text{known}}$ & scalar & Top-$k$ count for the known head \\
$k_{\text{unknown}}$ & scalar & Top-$k$ count for the unknown head \\
\midrule
\multicolumn{3}{l}{\textit{Concept embeddings}} \\
$K$ & matrix & Known concept embedding matrix \\
$U$ & matrix & Unknown concept embedding matrix \\
$K_i$ & vector & Embedding of known concept $i$ \\
$U_j$ & vector & Embedding of unknown concept $j$ \\
$n$ & scalar & Number of known concepts \\
$m$ & scalar & Number of unknown concepts ($m\gg n$) \\
$R$ & scalar & Factorization rank of unknown embedding matrix \\
\midrule
\multicolumn{3}{l}{\textit{Concept contributions}} \\
$\hat{k}=\sum_i k_i K_i$ & vector & Known concept contribution \\
$\hat{u}=\sum_j u_j U_j$ & vector & Unknown concept contribution \\
$\varepsilon=h-\hat{k}-\hat{u}$ & vector & Residual term \\
\midrule
\multicolumn{3}{l}{\textit{Logit decomposition}} \\
$W_y$ & vector & Row of the LM head for output token $y$ \\
$\ell_y$ & scalar & Output logit for token $y$ \\
\bottomrule
\end{tabular}
\caption{Concept module notation, grouped by role.}
\label{tab:arch-notation-1}
\end{table}

\begin{table}[htbp!]
\centering
\small
\setlength{\tabcolsep}{6pt}
\renewcommand{\arraystretch}{1.0}
\begin{tabular}{@{}lll@{}}
\toprule
Symbol & Type & Meaning \\
\midrule
\multicolumn{3}{l}{\textit{Losses}} \\
$\mathcal{L}_{\text{LM}}$ & loss & Language modeling loss ($\mathcal{L}_{\text{MDM}}$ on $\bar{h}$) \\
$\mathcal{L}_{\text{concept}}$ & loss & Concept loss (chunk-level BCE) \\
$\mathcal{L}_{\text{rec}}$ & loss & Reconstruction loss for the unknown head \\
$\mathcal{L}_{\text{indep}}$ & loss & Independence loss between $\hat{k}$ and $\hat{u}$ \\
$\mathcal{L}$ & loss & Combined training objective \\
$\lambda_{\text{concept}},\lambda_{\text{rec}},\lambda_{\text{indep}}$ & scalars & Loss weights \\
\midrule
\multicolumn{3}{l}{\textit{Supervision and targets}} \\
$y_c$ & scalar & Ground-truth chunk label, known concept $c$ \\
$k^{\text{chunk}}_c$ & scalar & OR-aggregated chunk-level activation \\
$k^{\text{GT}}_i$ & scalar & Ground-truth activation of known concept $i$ \\
$\hat{k}^{\text{GT}}$ & vector & Ground-truth known concept contribution \\
$\hat{u}^{\text{GT}}=h-\hat{k}^{\text{GT}}$ & vector & Target for the unknown head \\
\midrule
\multicolumn{3}{l}{\textit{Independence loss}} \\
$H_k, H_u$ & matrices & Stacked per-token $\hat{k}$, $\hat{u}$ over a minibatch \\
$\boldsymbol{\mu}_{\hat{k}},\boldsymbol{\mu}_{\hat{u}}$ & vectors & Column means of $H_k$, $H_u$ \\
$\Phi$ & matrix & Centered known features \\
$\Psi$ & matrix & Centered unknown features \\
\midrule
\multicolumn{3}{l}{\textit{Training dynamics}} \\
$\mathcal{M}$ & set & Masked token positions in the minibatch \\
$t_b$ & scalar & Per-block noise level (block $b$) \\
$\alpha_{\text{known}}(s)$ & scalar & Teacher forcing prob., known, step $s$ \\
$\alpha_{\text{unknown}}(s)$ & scalar & Teacher forcing prob., unknown, step $s$ \\
$p_{\text{cfg}}$ & scalar & Dropout rate for the known head \\
$p_\varepsilon$ & scalar & Dropout rate for the residual $\varepsilon$ \\
$B$ & scalar & Minibatch size \\
$b$ & scalar & Block size in causal block-diffusion \\
\bottomrule
\end{tabular}
\caption{Concept module notation, grouped by role.}
\label{tab:arch-notation-2}
\end{table}

%% file: sections/appendix/capabilities.tex
\clearpage
\part{Interpretability capabilities}
\label{app:capabilities}
\section{Attribution details}

\subsection{Training data attribution}
\label{appendix:tda}
The training data attribution pipeline of~\cref{sec:training-data-attribution} indexes the training corpus offline and retrieves against it at inference time. We describe each component here.

\paragraph{Corpus index.}
We index the full training corpus, approximately 11 billion chunks, with a fine-tuned embedding model, \texttt{FT-Qwen3-Embedding-0.6B}, derived from \texttt{Qwen3-Embedding-0.6B} (the annotator model of section~\ref{sec:data}), encoding each chunk into a 1024-dimensional vector. To make nearest-neighbor search tractable at this scale, we build a FAISS index with an inverted-file structure and product quantization (IVFPQ): the inverted file partitions the corpus into coarse clusters so a query is compared only against the closest few, and product quantization compresses each stored vector to keep the index in memory. Retrieval performs approximate nearest-neighbor search over the top-$n$ coarse clusters.

\paragraph{Chunk representation.}
At inference time, each chunk of the model response is tokenized and forwarded through the language model to obtain per-token hidden states. These are mean-pooled over the real (non-padding) positions into a single 4096-dimensional representation, formed from the known-head, unknown-head, and residual components of the hidden state, so it encodes the meaning of the chunk as the model represents it.

\paragraph{Transducer.}
The chunk representation is 4096-dimensional while the corpus index is 1024-dimensional, so we bridge the two spaces with the transducer: a two-hidden-layer MLP with roughly 15 million parameters that maps $\mathbb{R}^{4096} \to \mathbb{R}^{1024}$, trained to preserve semantic content under a cosine-similarity objective,
\begin{equation}
  \mathcal{L}_{\text{trans}} = 1 - \cos\!\left(\hat{\mathbf{e}}, \mathbf{e}^*\right),
  \label{eq:transducer-loss}
\end{equation}
where $\hat{\mathbf{e}}$ is the transducer's predicted embedding and $\mathbf{e}^*$ is the target embedding from \texttt{FT-Qwen3-Embedding-0.6B} for the same chunk.

\paragraph{Pipeline.}
End to end, each response chunk is encoded and mean-pooled into a single representation, transduced into the corpus embedding space, and used to query the IVFPQ index by approximate nearest-neighbor search over the top-$n$ coarse clusters, returning the most similar training chunks as the attributed sources.

%% file: sections/appendix/data.tex
\clearpage
\part{Data}
\label{app:data}
\section{Atlas: From documents to concepts}

% \begin{figure}[H]
% \begin{lstlisting}[basicstyle=\ttfamily\footnotesize,frame=single]
% SYSTEM:
% You are an expert in evaluating thematic coherence. Your task is to evaluate how well a group of short-tags belong together in a cluster based on their semantic similarity and thematic unity. Use a 1-5 scale with these specific criteria:

% Scoring Criteria:
% - 5 (Highly Coherent): All tags share a clear, specific theme or domain. They would naturally appear together in the same context or document section.
% - 4 (Mostly Coherent): Most tags clearly relate to a common theme with 1-2 minor outliers or broader variations.
% - 3 (Moderately Coherent): Tags share a general domain but with notable diversity or multiple sub-themes present.
% - 2 (Weakly Coherent): Only loose connections exist; tags might share abstract similarities but come from different contexts.
% - 1 (Incoherent): Tags are unrelated or from completely different domains with no meaningful connection.

% USER:
% Evaluate the coherence of the following set of short-tags sampled from a cluster:
% {tags}

% Respond with only a single number from 1 to 5 as your rating. Do not include any explanation.
% \end{lstlisting}
% \caption{Prompt for cluster coherence evaluation}
% \label{fig:prompt-cluster-coherence-eval}
% \end{figure}

\paragraph{Text-assigned concept evaluation.}
For the tag, concept, and predicted-concept relevance evaluations described in Section~\ref{sec:data-atlas}, we use an LLM judge to rate whether candidate concepts are present in a text chunk. The judge is \texttt{Mistral-Small-3.1-24B-Instruct} run at temperature 0. It receives a text chunk together with one or more candidate concepts, each represented by a label and one-sentence description, and returns a JSON object assigning each candidate concept a relevance score from 1 to 5. The full prompt is shown in Figure~\ref{fig:prompt-text-assigned-concept-eval}.

\begin{figure}[H]
\begin{lstlisting}[basicstyle=\ttfamily\footnotesize,frame=single]
SYSTEM:
You are an expert at evaluating whether concepts are genuinely present in a passage of text. For each concept, you need to score based on how well the concept is exhibited (either explicitly or in a subtle manner) by the text passage. Use the following 1-5 scoring criteria:
- 5: Very strongly present/central to the text - concept is explicitly discussed and/or forms a core theme
- 4: Strongly present/well-demonstrated - concept is clearly evident with substantial supporting content
- 3: Moderately present/clearly related - concept is reasonably connected with some supporting evidence
- 2: Weakly present/tangentially related - minimal connection or only surface-level mention
- 1: Not present/completely inappropriate - concept is absent, irrelevant, or text merely uses related vocabulary without actually being about that concept

Be precise in your evaluation - consider both explicit mentions and substantive demonstration of the concept.

Respond with a single-line JSON object where keys are the concept letters (A, B, C, etc.) and values are scores (1-5). **DO NOT include any explanations or additional text.**

USER:
Evaluate how well each concept below is exhibited in the given text chunk using the 1-5 scale.

=== TEXT CHUNK ===
{text_chunk}

=== CONCEPTS TO EVALUATE ===
{concepts_with_descriptions}
\end{lstlisting}
\caption{Prompt for text-assigned concepts evaluation}
\label{fig:prompt-text-assigned-concept-eval}
\end{figure}

\section{Additional details on the human interpretability study}
\label{app:human-eval-details}

This appendix gives additional methodological and statistical details for the human interpretability study in Section~\ref{sec:data-human-eval}. The main text reports the study design and headline results; here we report agreement statistics, robustness checks, and model-based analyses.

\subsection{Sampling and annotation protocol}

We sampled $100$ concepts stratified by top-level taxonomy branch: ten concepts from each of the nine largest branches and ten from the aggregated remainder. For each sampled concept, we constructed a lifted-word list by ranking lemmatized words according to
\[
\operatorname{lift}(w,c) = \frac{P(w \mid c)}{P(w)} ,
\]
subject to a minimum-support filter. Annotators saw only these lifted words, not the pipeline concept name, concept description, taxonomy position, or source documents.

In Phase~1, annotators wrote a name or short phrase for the concept and rated whether the lifted words formed a recognizable concept on a $1$--$5$ scale. Phase~1 collected $303$ named-concept responses from $9$ annotators, with a median of $3$ annotators per concept. Human-written names averaged $4.1$ words.

In Phase~2, annotators rated candidate names for the same lifted-word lists. Each candidate set contained the pipeline's LLM-generated label, two human labels from Phase~1, an embedding-neighbor distractor, and a taxonomy-neighbor distractor. A small number of cases also included a low-effort filler label as a floor control when one of the other candidate types was unavailable. Candidate order was randomized, annotators were blind to candidate provenance, and no annotator scored a label they had written. Phase~2 collected $205$ scoring records over $34$ concepts from $8$ annotators, for $1{,}025$ individual candidate-name ratings. The $20$ most-rated concepts were scored by all eight Phase~2 annotators.

\subsection{Phase~1 agreement and coherence}

Phase~1 was designed to test whether lifted-word evidence contains recoverable semantic structure before any pipeline label is shown. The mean coherence score was $3.52$. Annotators judged $55\%$ of responses to form a recognizable concept ($\geq 4$), rated $27\%$ as borderline ($=3$), and flagged $17\%$ as incoherent or noisy ($\leq 2$).

Agreement was moderate rather than perfect, as expected for a task involving short word lists and concepts drawn from many technical domains. ICC$(1)$ was $0.43$. The within-concept standard deviation was $0.65$, compared with total scale standard deviation $1.05$, and $53\%$ of concepts were unanimous on the coherent-vs-not split. These agreement statistics support the main-text conclusion: lifted-word evidence is often meaningful, but not uniformly so.

\subsection{Ordinal mixed-model analysis}

The primary Phase~2 outcome is an ordinal fit rating $y_i \in \{1,2,3,4,5\}$ for a candidate label. Because ratings are ordinal and clustered by rater, concept, label, and scoring context, we fit a Bayesian cumulative-link mixed model with crossed random effects. Human labels are the reference category. The model has the form
\[
\Pr(y_i \leq k)
=
\operatorname{logit}^{-1}(\tau_k - \eta_i),
\]
where $k \in \{1,2,3,4\}$ indexes the ordinal thresholds and
\[
\eta_i =
\beta_{\mathrm{LLM}}\mathbf{1}_{\mathrm{LLM},i}
+ \beta_{\mathrm{emb}}\mathbf{1}_{\mathrm{embedding},i}
+ \beta_{\mathrm{tax}}\mathbf{1}_{\mathrm{taxonomy},i}
+ \beta_{\mathrm{fill}}\mathbf{1}_{\mathrm{filler},i}
+ u^{\mathrm{rater}}_{r[i]}
+ u^{\mathrm{concept}}_{c[i]}
+ u^{\mathrm{context}}_{s[i]}
+ u^{\mathrm{label}}_{\ell[i]} .
\]
Positive coefficients indicate higher expected fit scores. The main contrast is $\beta_{\mathrm{LLM}}$, comparing the pipeline label to human-written labels.

The LLM--human proportional-odds ratio was $2.38$ ($95\%$ CrI $[1.23,4.01]$), with posterior probability $0.99$ that the pipeline label receives higher fit ratings than a human label. On the response scale, pipeline labels received a top-two rating ($\geq 4$) $79\%$ of the time, compared with $63\%$ for human labels.

\subsection{Assumption-light robustness checks}

We also ran simpler checks that make fewer modeling assumptions. In paired comparisons, the pipeline label outscores a human label with probability $0.62$ (cluster-bootstrap $95\%$ CI $[0.58,0.66]$, $402$ pairs). A Gaussian mixed model on the raw $1$--$5$ ratings estimates a $+0.47$ point advantage for pipeline labels over human labels ($95\%$ CI $[0.28,0.65]$). These checks agree with the ordinal mixed model: pipeline labels are not merely competitive with human labels, but are rated higher on average under blind evaluation.

\begin{table}[t]
\centering
\small
\begin{tabular}{ll}
\toprule
Analysis & Result \\
\midrule
Mean Phase~2 fit score & LLM $3.98$ vs.\ human $3.50$ \\
Top-two rating rate ($\geq 4$) & LLM $79\%$ vs.\ human $63\%$ \\
Bayesian cumulative-link model & OR $2.38$, $95\%$ CrI $[1.23,4.01]$ \\
Posterior probability of LLM advantage & $0.99$ \\
Paired comparison probability & $0.62$, bootstrap $95\%$ CI $[0.58,0.66]$ \\
Gaussian mixed model & $+0.47$ points, $95\%$ CI $[0.28,0.65]$ \\
\bottomrule
\end{tabular}
\caption{Robustness checks for the Phase~2 comparison between pipeline labels and independently generated human labels.}
\label{tab:human-eval-robustness}
\end{table}

\subsection{Dependence on lifted-word coherence}

If the LLM were hallucinating plausible names uniformly, its advantage should not depend strongly on whether the lifted-word evidence itself is coherent. We therefore examined how Phase~2 label fit varies with Phase~1 coherence.

Concepts with higher Phase~1 coherence receive higher Phase~2 fit scores for both human and pipeline labels. The correlation between Phase~1 coherence and Phase~2 fit is $+0.42$ for human labels and $+0.46$ for pipeline labels. The pipeline advantage is positive across coherence bins, but larger when the underlying word evidence is clearer: the LLM--human gap is $+0.22$ for low-coherence concepts, $+0.54$ for mid-coherence concepts, and $+0.52$ for high-coherence concepts. This pattern is consistent with the LLM naming real statistical structure rather than assigning plausible labels independently of the evidence.

\subsection{Power analysis}

Finally, we checked whether the realized Phase~2 sample size was sufficient for the observed LLM--human contrast. A simulation under the fitted clustered generative model gives power above $0.99$ for the observed contrast at roughly $20$ fully-rated concepts and essentially $1.00$ at the realized sample size. This analysis should not be interpreted as certifying every individual concept in the full library. Rather, it shows that the stratified pilot is well-powered for the aggregate comparison between pipeline labels and human-written labels.

\subsection{Limitations}

The human study validates the labeling operation used by Atlas, not every concept individually. A minority of lifted-word lists are ambiguous or noisy, and some concepts require domain expertise that may not be uniformly available across annotators. The Phase~2 comparison also evaluates names relative to lifted-word evidence, rather than full source-document evidence. These limitations make the result conservative in one respect and incomplete in another: humans often recover and endorse the same semantic structure from sparse evidence alone, but the study does not eliminate the need for additional per-domain or per-concept audits in downstream use.

%% file: sections/appendix/interpretability.tex
\clearpage

\part{Interpretability metrics}
\label{app:interpretability-metrics}

\section{Known concept alignment judge}
\label{app:judge-known-alignment}

\looseness=-1 The Known Concept Alignment metric from section~\ref{sec:interp_metrics} relies on an LLM-judge to rate concepts. The judge uses \texttt{Mistral-Small-3.1-24B-Instruct} at temperature 0. It receives a concept's human-assigned label, its one-sentence description, and the top-$K$ tokens scored by the concept embedding through the LM head. It returns a single integer rating from 1 to 5. The full prompt is shown in Figure~\ref{fig:prompt-known-alignment}.

\begin{figure}[H]
\begin{lstlisting}[basicstyle=\ttfamily\footnotesize,frame=single]
SYSTEM:
You are an expert in neural network interpretability. You are evaluating whether
a concept head in a language model has learned to represent a specific named
concept.

You will be given:
1. The concept's human-assigned label and description
2. The top activated tokens from the concept head

Rate how well the top tokens represent the named concept on a 1-5 scale:
- 5: Tokens strongly and clearly represent the concept. Most tokens are
     directly related.
- 4: Tokens mostly represent the concept with minor noise or tangential items.
- 3: Tokens partially represent the concept. Some relevant tokens but also
     significant off-topic items.
- 2: Tokens weakly relate to the concept. Only a few tokens connect; most are
     unrelated.
- 1: Tokens do not represent the concept at all.

IMPORTANT: A token does not need to be an exact word from the label or
description to count as relevant. Proper nouns, abbreviations, sub-words, and
semantically related terms all count. For example, if the concept is 'academic
publishers', then 'Penguin', 'Wiley', 'ISBN', 'paperback', and 'imprint' are all
relevant even though none appear in the label.

Respond with ONLY one line:
ALIGNMENT: <integer 1-5>

USER:
Concept label: {concept_label}
Concept description: {concept_description}

=== TOP ACTIVATED TOKENS ===
{top_tokens}
\end{lstlisting}
\caption{Prompt for the Known Concept Alignment judge.}
\label{fig:prompt-known-alignment}
\end{figure}

%% file: sections/appendix/scaling_laws.tex
\clearpage
\part{Scaling laws}
\label{app:scaling-laws}
\section{Symbol and notations}
\label{appendix:scaling-notation}
\input{sections/appendix/tables/scaling_notation}

\section{Architectures, IsoFLOP slices, and hyperparameters}
\label{app:scaling_laws_arch_table}

\paragraph{Backbones.}
Table~\ref{tab:arch-configs} lists the configuration of each backbone size used in the scaling sweep. All four families share the same backbone architecture; +Concept variants add the concept module on top, configured per~\cref{tab:concept-config-stable}~and~\cref{tab:concept-config-anneal}.

\begin{table}[htbp!]
\centering
\small
\begin{tabular}{@{}lrrrr@{}}
\toprule
\textbf{Size} & \textbf{Layers $L$} & \textbf{Hidden $d$} & \textbf{Backbone params} & \textbf{+Concept total params} \\
\midrule
10M   & 6  & 320  & 9.2M       & 82.8M      \\
25M   & 6  & 512  & 23.6M      & 110.2M     \\
85M   & 10  & 768  & 86.5M      & 190.5M     \\
200M  & 13  & 1024  & 197.7M     & 319.1M     \\
400M  & 17  & 1280  & 401.1M     & 540.0M     \\
800M  & 17  & 1792  & 779.9M     & 953.6M     \\
1.5B  & 20  & 2304  & 1{,}510M   & 1{,}718M   \\
3B    & 24  & 3072  & 3{,}228M   & 3{,}489M   \\
5B    & 24  & 3840  & 5{,}694M   & 6{,}007M   \\
\bottomrule
\end{tabular}
\caption{Backbone architectures used across all four families. Backbone parameter counts exclude embeddings; +Concept totals include the concept module heads (concept classifier, and factorized unknown head concept embeddings are excluded). Sequence length is 4096 throughout.}
\label{tab:arch-configs}
\end{table}

\paragraph{IsoFLOP slices.}
Table~\ref{tab:isoflop-slices} reports the four IsoFLOP target compute budgets per family. The +Concept families start at higher targets because the concept module's per-token FLOPs are non-negligible at small backbone sizes. Each slice contains four to six model sizes whose per-checkpoint compute lands within $\pm 15\%$ of the target.

\begin{table}[htbp!]
\centering
\small
\begin{tabular}{@{}lcccc@{}}
\toprule
\textbf{Family} & \textbf{Slice 1} & \textbf{Slice 2} & \textbf{Slice 3} & \textbf{Slice 4} \\
\midrule
AR             & $6{\times}10^{18}$ & $10^{19}$    & $3{\times}10^{19}$    & $10^{20}$           \\
CDLM           & $6{\times}10^{18}$ & $10^{19}$    & $3{\times}10^{19}$    & $10^{20}$           \\
AR+Concept     & $10^{19}$          & $3{\times}10^{19}$ & $1.1{\times}10^{20}$ & $3.09{\times}10^{20}$ \\
CDLM+Concept   & $10^{19}$          & $3{\times}10^{19}$ & $1.1{\times}10^{20}$ & $3.09{\times}10^{20}$ \\
\bottomrule
\end{tabular}
\caption{IsoFLOP target compute budgets per family.}
\label{tab:isoflop-slices}
\end{table}

\paragraph{Shared hyperparameters.}
All scaling-law runs share optimizer, learning rate, batch size, warmup, and architectural defaults. The optimizer is AdamW with $\beta_1 = 0.9$, $\beta_2 = 0.95$, $\varepsilon = 10^{-8}$, weight decay $0.1$ (excluding embeddings), and gradient clipping at $1.0$. Peak learning rate is fixed at $4 \times 10^{-4}$ across all model sizes and all families. The schedule is warmup-stable-decay (WSD) with an $80/20$ stable/decay split and decay to zero. Warmup is the minimum of $2000$ steps or $2\%$ of total training steps, to accommodate runs with smaller token budgets. The total batch size is $524{,}288$ tokens (128 sequences $\times$ 4096 tokens) across every run. Backbone architectural defaults (post-norm RMSNorm, QK-norm, RoPE base $5 \times 10^{5}$, SwiGLU MLP with ratio $4$, no biases, $\text{clip\_qkv}{=}10$, dropout $0$) are held fixed across all backbones. Per-(size, slice) token counts are determined by the slice target $C$ and the FLOP equation (\cref{eq:flops-base}); intermediate stable-phase checkpoints are also included in IsoFLOP fits where they fall within $\pm 15\%$ of a slice target.

\paragraph{CDLM-specific settings.}
The two diffusion families (CDLM and CDLM+Concept) share an additional set of settings governing the masking process: causal block size $64$, with the noise level sampled uniformly at training time, $t \sim \mathcal{U}(0.05, 0.95)$. These are held fixed across all CDLM model sizes.

\paragraph{Concept module settings.}
The two +Concept families (AR+Concept, CDLM+Concept) share the concept module configuration. Settings differ between the stable phase (\cref{tab:concept-config-stable}) and the anneal phase (\cref{tab:concept-config-anneal}), with the anneal phase additionally applying top-$k$ sparsification to both heads. The parameters $\alpha_{\text{known}}(t)$ and $\alpha_{\text{unknown}}(t)$ are the probabilities of using ground-truth components in place of predicted ones during training; see~\cref{sec:training-dynamics} for the motivation and definition.

\begin{table}[htbp!]
\centering
\small
\begin{tabular}{@{}lll@{}}
\toprule
\textbf{Setting} & \textbf{Value} & \textbf{Notes} \\
\midrule
$n$ (known concepts)                & $33{,}732$ & shared across all sizes \\
Unknown ratio                         & $3\times$  & $\Rightarrow m = 101{,}196$ \\
Factorization rank $R$                & 256        & for unknown head \\
Top-$k_{\text{known}}$                     & 16         & predictor + compose \\
$\alpha_{\text{known}}(t)$            & $1.0 \to 0.5$ by step $0.10\,T_{\max}$ & cosine \\
$\alpha_{\text{unknown}}(t)$          & $0.0 \to 0.5$ by step $0.10\,T_{\max}$ & linear \\
$\lambda_{\text{concept}}$            & 1.0        & concept loss weight \\
$\lambda_{\text{rec}}$                & 1.0        & reconstruction loss weight \\
$\lambda_{\text{indep}}$              & 1.0        & independence loss weight \\
$p_{\text{cfg}}$                      & 0.1        & known head dropout rate \\
$p_\varepsilon$                       & 0.3        & residual dropout rate \\
\bottomrule
\end{tabular}
\caption{Concept module settings during the stable phase. Both $\alpha$ schedules ramp from initial value to $0.5$ by step $0.10\,T_{\max}$, then hold constant for the remainder of the stable phase. $T_{\max}$ denotes \texttt{max\_steps}.}
\label{tab:concept-config-stable}
\end{table}

\begin{table}[htbp!]
\centering
\small
\begin{tabular}{@{}lll@{}}
\toprule
\textbf{Setting} & \textbf{Value} & \textbf{Notes} \\
\midrule
$\alpha_{\text{known}}(t)$            & $0.5 \to 0.0$ over anneal & linear \\
$\alpha_{\text{unknown}}(t)$          & constant at $1.0$         & \\
Top-$k_{\text{known}}$                & 32        & \texttt{topk\_known} \\
Top-$k_{\text{unknown}}$              & 128       & factorized compose, \texttt{apply\_topk\_to\_unknown} \\
\bottomrule
\end{tabular}
\caption{Concept module settings during the anneal phase. The anneal phase resumes from the final stable-phase checkpoint and runs a linear LR decay to zero. Inherited settings ($n$, unknown ratio, factorization rank, loss weights, dropouts) match~\cref{tab:concept-config-stable}.}
\label{tab:concept-config-anneal}
\end{table}

\section{ELBO estimation for validation loss}

\label{app:scaling_laws_ELBO}

Autoregressive models report exact negative log-likelihood (cross-entropy on the next token); however, in diffusion models, the per-token NLL cannot be computed exactly and must instead be estimated via Monte Carlo on an Evidence Lower Bound (ELBO). Different MC schemes give materially different absolute loss values, which propagates into reported $\mathcal{L}_\infty$ values and, when the bias is non-uniform across model sizes, into reported scaling exponents.  Here we compare four estimators on our CDLM checkpoints.  We find that the compute-optimal parameter count $P^*(C)$ and the exponent $\alpha_P$ are robust to estimator choice, while absolute loss values $\mathcal{L}^*$ and the irreducible-loss asymptote $\mathcal{L}_\infty$ vary by up to a nat between estimators.

\subsection{Estimators}
Let $\mathbf{x}_0 = (x_0^1, \ldots, x_0^N)$ denote a sequence of length $N$ drawn from the validation distribution $\mathcal{D}$. At noise level $t \in [0, 1]$, the forward process replaces each token independently with \texttt{[MASK]} with probability $t$ and leaves it unchanged with probability $1 - t$:
\begin{equation}
x_t^i = \begin{cases} \texttt{[MASK]} & \text{with probability } t, \\ x_0^i & \text{with probability } 1 - t. \end{cases}
\label{eq:elbo-forward-process}
\end{equation}
We denote by $\mathcal{M}_t = \{i : x_t^i = \texttt{[MASK]}\}$ the set of masked positions in $\mathbf{x}_t$, and by $p_\theta(x_0^i \mid \mathbf{x}_t)$ the model's predicted distribution over the token at position $i$ given the corrupted sequence $\mathbf{x}_t$. All estimators below average over $\sim$100M tokens of validation data.

\paragraph{Fixed-rate mask loss.}
The first estimator computes per-token cross-entropy at a single fixed mask rate $\tau$, with no integration over the noise schedule:
\begin{equation}
\hat{\mathcal{L}}_{\text{fixed}}(\tau) = \E_{\mathbf{x}_0, \mathbf{x}_\tau} \left[\frac{1}{|\mathcal{M}_\tau|} \sum_{i \in \mathcal{M}_\tau} -\log p_\theta(x_0^i \mid \mathbf{x}_\tau)\right].
\label{eq:elbo-fixed}
\end{equation}
We set $\tau = 0.5$. This estimator computes per-token cross-entropy at one fixed corruption level rather than an ELBO bound on $-\log p_\theta(\mathbf{x}_0)$, but is the closest validation analogue of the training objective, which integrates over $[0.05, 0.95]$.

\paragraph{Fixed-grid discretized ELBO.}
The second estimator approximates the ELBO integral over $t$ via $K$ fixed bins $\{t_1, \ldots, t_K\}$ averaged uniformly:
\begin{equation}
\hat{\mathcal{L}}_{\text{grid}} = \frac{1}{K} \sum_{k=1}^{K} \E_{\mathbf{x}_0, \mathbf{x}_{t_k}} \left[\frac{1}{|\mathcal{M}_{t_k}|} \sum_{i \in \mathcal{M}_{t_k}} -\log p_\theta(x_0^i \mid \mathbf{x}_{t_k})\right].
\label{eq:elbo-grid}
\end{equation}
% We use $K = 9$ with $t_k = k/10$ for $k = 1, \ldots, 9$. This is a discrete trapezoid approximation to $\int_0^1 \mathcal{L}(t) \, dt$ on $[0.1, 0.9]$. \nathaniel{to be excessively nitpicky, this isn't trapezoidal, right? }

We use $K = 9$ with $t_k = k/10$ for $k = 1, \ldots, 9$. With uniform weights this is a midpoint-rule approximation to the ELBO integral over $t \in [0.05, 0.95]$, matching the interval sampled during training.

\paragraph{MDLM ELBO.}

The third estimator follows the low-variance form of the MDLM ELBO from equation~\ref{eq:mdm-loss}. Under the linear schedule $\alpha_t = 1 - t$, the canonical NELBO bound on $-\log p_\theta(\mathbf{x}_0)$ is
\begin{equation}
-\log p_\theta(\mathbf{x}_0) \leq \E_{t \sim \mathcal{U}(0, 1)} \left[\frac{1}{t} \sum_{i \in \mathcal{M}_t} -\log p_\theta(x_0^i \mid \mathbf{x}_t)\right].
\label{eq:elbo-canonical}
\end{equation}
The $1/t$ prefactor cancels in expectation with $\E[|\mathcal{M}_t|/N \mid t] = t$ when $t \sim \mathcal{U}(0, 1)$, giving the per-token form
\begin{equation}
\hat{\mathcal{L}}_{\text{MDLM}} = \E_{t \sim \mathcal{U}(0, 1)} \, \E_{\mathbf{x}_0, \mathbf{x}_t} \left[\frac{1}{|\mathcal{M}_t|} \sum_{i \in \mathcal{M}_t} -\log p_\theta(x_0^i \mid \mathbf{x}_t)\right].
\label{eq:elbo-mdlm}
\end{equation}

We sample one $t$ per batch and average over batches. This estimator is unbiased for the per-token NELBO bound and matches the formulations used in \citet{sahoo2024, mdlm, quokka, scaling_behaviorDLM_von, scaling_beyond_sahoo}; it is also what we report in section~\ref{subsec:compute-optimal}. Note that the CDLM training loss is itself an unbiased estimator of the same quantity, except that training clips $t$ to $[\beta, \omega] = [0.05, 0.95]$ to avoid degenerate extremes. The two estimators are therefore numerically close but not identical: $\hat{\mathcal{L}}_{\text{MDLM}}$ integrates over the full $[0, 1]$ interval, while the training loss integrates over $[0.05, 0.95]$.

\paragraph{Per-block ELBO.}
The fourth estimator matches the training distribution of our causal block-diffusion model (~\cref{subsec:causal-diffusion}), where each block has its own independently sampled noise level. We partition each sequence into $N/b$ blocks of size $b$. For each block, we sample an independent noise level $t_b \sim \mathcal{U}(0, 1)$ and mask positions within that block at rate $t_b$. We then compute mean cross-entropy on all masked positions:
\begin{equation}
\hat{\mathcal{L}}_{\text{block}} = \E_{\{t_b\} \overset{\text{iid}}{\sim} \mathcal{U}(0, 1)} \, \E_{\mathbf{x}_0, \mathbf{x}_t} \left[\frac{1}{|\mathcal{M}_t|} \sum_{i \in \mathcal{M}_t} -\log p_\theta(x_0^i \mid \mathbf{x}_t)\right].
\label{eq:elbo-block}
\end{equation}
We use $b = 64$, matching the training configuration, so each sequence contributes 64 independent noise levels per validation step. The cancellation argument from~\cref{eq:elbo-canonical} applies per-block, so this estimator is also unbiased for the per-token NELBO bound, with lower Monte Carlo variance than $\hat{\mathcal{L}}_{\text{MDLM}}$ because each batch covers a wider distribution of $t$ values.

\subsection{Results}

\begin{figure}[htbp!]
\centering
\includegraphics[width=\linewidth]{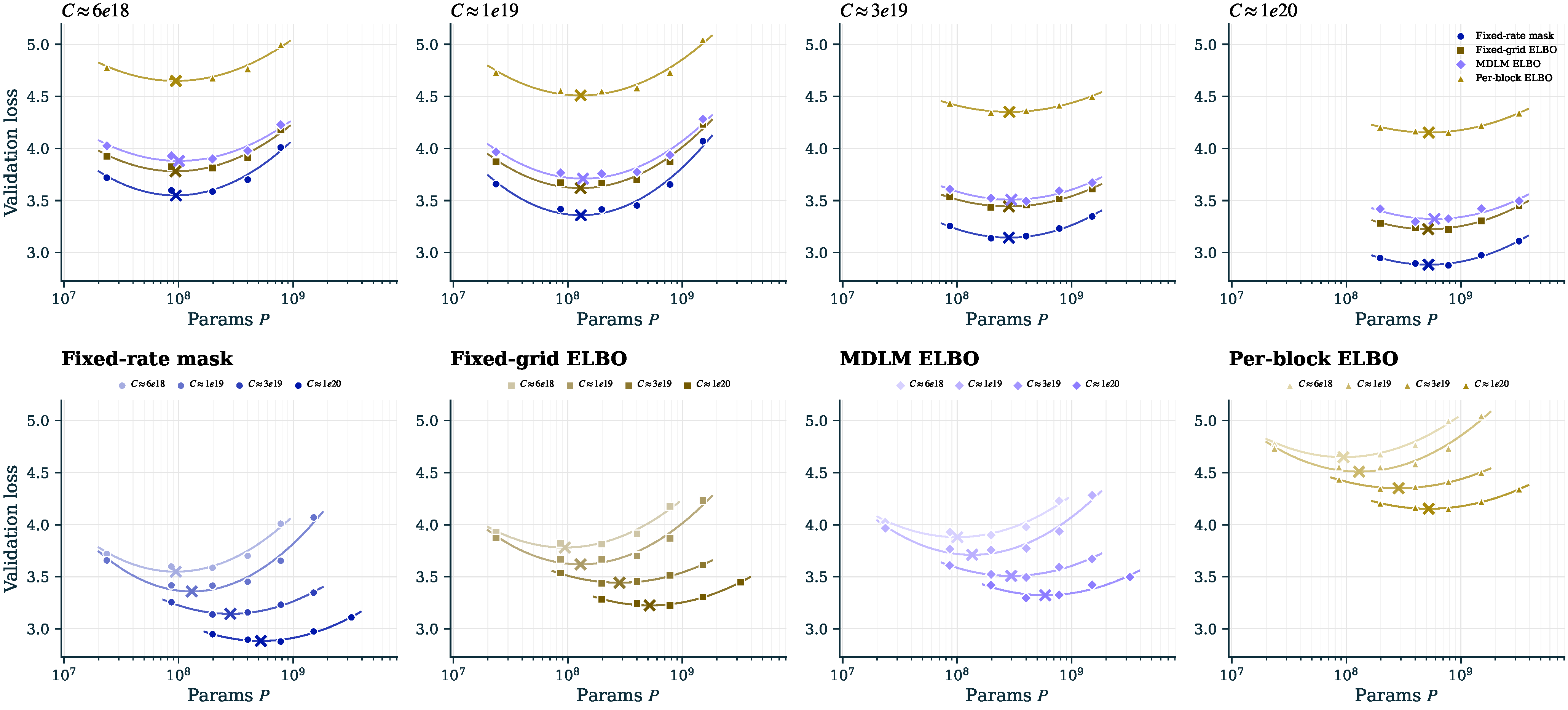}
\caption{IsoFLOP comparison of the four ELBO estimators on CDLM. Top row: per-slice parabolic fits with all four estimators overlaid. Bottom row: per-estimator IsoFLOP grids with all four slices overlaid.}
\label{fig:elbo-isoflops}
\end{figure}

\paragraph{IsoFLOP shapes and absolute losses.}
Figure~\ref{fig:elbo-isoflops} compares the four estimators across the four CDLM IsoFLOP slices. The per-slice parabolas (top row) have similar shapes and per-slice minima at comparable parameter counts $P^*(C)$ across all estimators, indicating that the location of the compute-optimal model size is robust to estimator choice. The absolute loss values, however, differ substantially: at any fixed slice, the four estimators are vertically offset by up to a nat, with uniform mask consistently lowest and per-block ELBO highest. The bottom row shows each estimator's full IsoFLOP grid; the parabolas tighten with compute under all four estimators, and the slice minima move smoothly toward larger $P$ as $C$ grows.

\begin{figure}[htbp!]
\centering
\includegraphics[width=0.9\linewidth]{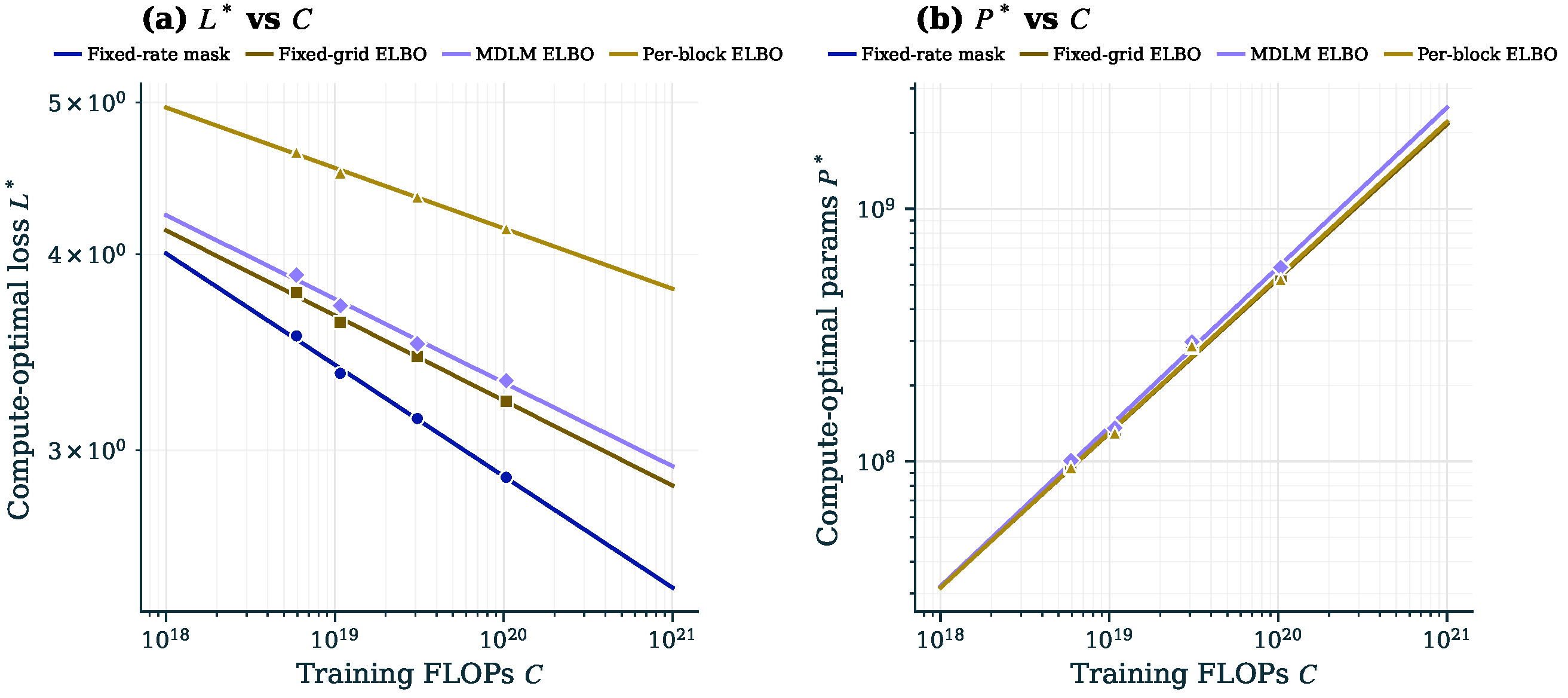}
\caption{Compute-optimal scaling laws under the four ELBO estimators. (a) $\mathcal{L}^*(C)$ versus compute. (b) $P^*(C)$ versus compute.}
\label{fig:elbo-powerlaw}
\end{figure}

\paragraph{Effect on power laws.}
Figure~\ref{fig:elbo-powerlaw} plots the compute-optimal scaling laws $\mathcal{L}^*(C)$ and $P^*(C)$ for each estimator. The $P^*(C)$ fits (panel b) are nearly indistinguishable across estimators, with all four lines overlapping within marker width and $\alpha_P$ values clustering tightly between $0.602$ and $0.632$. The $\mathcal{L}^*(C)$ fits (panel a) show parallel-ish slopes shifted vertically by an estimator-dependent offset; the loss exponent $\alpha_L$ varies modestly between $-0.071$ (uniform mask) and $-0.039$ (per-block ELBO). Table~\ref{tab:elbo-comparison} summarizes the exponents with $90\%$ bootstrap confidence intervals.

\input{sections/appendix/tables/elbo_comparison.tex}

\subsection{Effect of mask rate on scaling exponents}
\label{subsec:elbo-mask-rate}

\begin{figure}[htbp!]
\centering
\includegraphics[width=\linewidth]{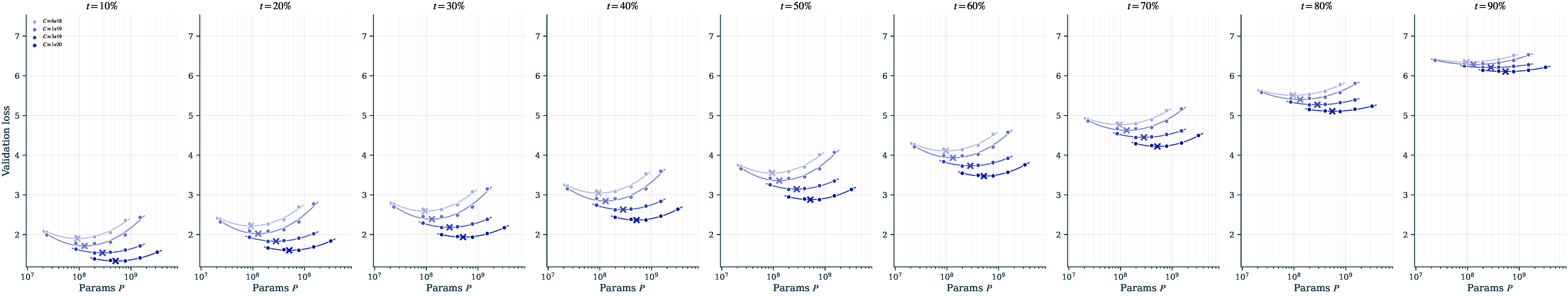}
\caption{IsoFLOP analysis at nine fixed mask rates $t \in \{0.1, 0.2, \ldots, 0.9\}$ on CDLM. $\times$ markers locate the per-slice minima $P^*(C)$.}
\label{fig:elbo-mask-rate-isoflops}
\end{figure}

\begin{figure}[htbp!]
\centering
\includegraphics[width=\linewidth]{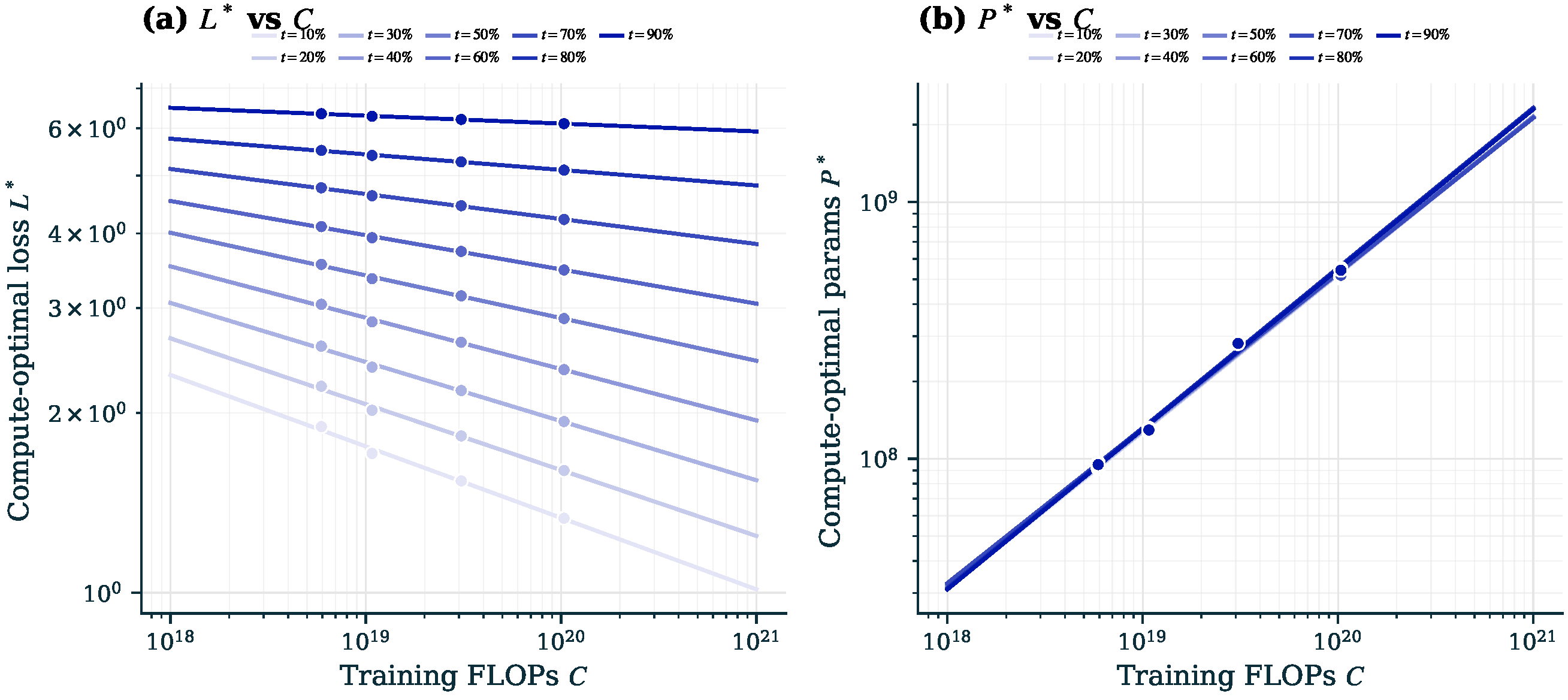}
\caption{Compute-optimal scaling laws at nine fixed mask rates on CDLM. (a) $\mathcal{L}^*(C)$ versus compute. (b) $P^*(C)$ versus compute. The $\mathcal{L}^*(C)$ slopes flatten monotonically as $t$ grows; $\alpha_P$ falls into two groups, $\approx 0.60$ for $t \leq 0.7$ and $\approx 0.62$ for $t \geq 0.8$.}
\label{fig:elbo-mask-rate-powerlaw}
\end{figure}

The four estimators above differ in how they aggregate across the noise schedule, mixing the effect of \emph{which} mask rates are sampled with the effect of \emph{how} they are integrated. To isolate the role of the mask rate alone, the IsoFLOP fit is re-run at nine fixed mask rates $t \in \{0.1, 0.2, \ldots, 0.9\}$ on the same CDLM checkpoints. Figure~\ref{fig:elbo-mask-rate-isoflops} shows the per-slice parabolas at each mask rate, Figure~\ref{fig:elbo-mask-rate-powerlaw} shows the resulting $P^*(C)$ and $\mathcal{L}^*(C)$ fits, and~\cref{tab:mask-rate-exponents} reports the exponents.s
The parameter exponent $\alpha_P$ separates into two groups: $\alpha_P \approx 0.60$ for $t \leq 0.7$ and $\alpha_P \approx 0.62$ for $t \geq 0.8$. The loss exponent $\alpha_L$ flattens monotonically and substantially, from $-0.120$ at $t = 0.1$ to $-0.013$ at $t = 0.9$: at low mask rates the loss has substantial room to fall with compute, while at high mask rates it approaches the random-token floor where additional compute buys little. This decoupling explains the four-estimator pattern documented above: any aggregation across $t$ inherits a similar $\alpha_P$ from the per-rate fits but produces an $\alpha_L$ determined by which mask rates carry weight.

\input{sections/appendix/tables/mask_rate_exponents.tex}

\subsection{Discussion}

\paragraph{Implications for $\mathcal{L}_\infty$.}
The IsoFLOP exponents $\alpha_P$ and $\alpha_L$ are robust to estimator choice, but the irreducible-loss asymptote $\mathcal{L}_\infty$ from the joint Chinchilla fit is not: across the four estimators, $\mathcal{L}_\infty$ spans roughly $1.9$ to $3.4$, a $1.5$-nat range that exceeds the entire $\mathcal{L}_\infty$ spread reported across diffusion scaling papers. This indicates that absolute $\mathcal{L}_\infty$ values are not directly comparable across diffusion methodologies that use different ELBO estimators, even on the same model and data. Within-methodology comparisons remain valid, as both members of each $\pm$Concept pair in the main results use the same estimator.

The CDLM model is trained with $t \sim \mathcal{U}(0.05, 0.95)$. The fixed-rate estimator at $\tau = 0.5$ matches the middle of this training interval and yields the lowest $\mathcal{L}_\infty$ across the four estimators,  with $\alpha_P$ in line with prior masked-diffusion estimates~\citep{quokka, scaling_behaviorDLM_von, mdlm}. The other three estimators integrate over a wider $t$ range, including the boundary regions $t \in [0, 0.05]$ and $t \in [0.95, 1]$ that the model was not directly trained on. The main paper reports MDLM ELBO to follow the convention adopted by prior diffusion scaling papers~\citep{mdlm, quokka, scaling_behaviorDLM_von, scaling_beyond_sahoo}, which makes the reported $\alpha_P$ directly comparable to theirs;  the higher absolute $\mathcal{L}_\infty$ relative to e.g.~Quokka's falls inside the estimator-induced spread documented here.

\section{Annealing each IsoFLOP checkpoint}
\label{app:scaling_laws_annealing}

The IsoFLOP analysis of~\cref{sec:scaling-laws} relies on annealed checkpoints, since the warmup-stable-decay schedule's final $20\%$ decay reduces validation loss by a non-trivial amount. Annealing each checkpoint independently is more expensive than estimating annealed losses from raw stable-phase checkpoints with a constant correction, as proposed by \citet{scaling_behaviorDLM_von}.

For the +Concept families, annealing is not solely an LR-decay procedure: the interpretable-component schedules also shift during the anneal phase (Appendix~\ref{app:scaling_laws_arch_table},~\cref{tab:concept-config-stable} and~\cref{tab:concept-config-anneal}). $\alpha_{\text{known}}$ ramps from $0.5$ to $0.0$, transitioning the model from teacher-forced ground-truth concept representations to its own learned known-head predictions. $\alpha_{\text{unknown}}$ shifts to $1.0$, transitioning the model to fully rely on the learned unknown head. Top-$k$ sparsification activates on both heads, restricting predictions to a small subset of the $\sim$135K concepts so that attribution remains interpretable. The anneal phase therefore transitions the model from its training-time configuration to the deployed inference-time configuration.

\looseness=-1 CDLM+Concept checkpoints in both states are compared on validation loss and the four interpretability metrics of section~\ref{sec:interp_metrics}. The gap is substantial in two places: compute-optimal allocation ($\alpha_P$ shifts by $0.11$) and Concept Independence Loss (which decreases $3\text{-}10\times$ under annealed evaluation).

\subsection{Validation loss}
\label{subsubsec:annealing-validation}

\begin{figure}[htbp!]
\centering
\includegraphics[width=\linewidth]{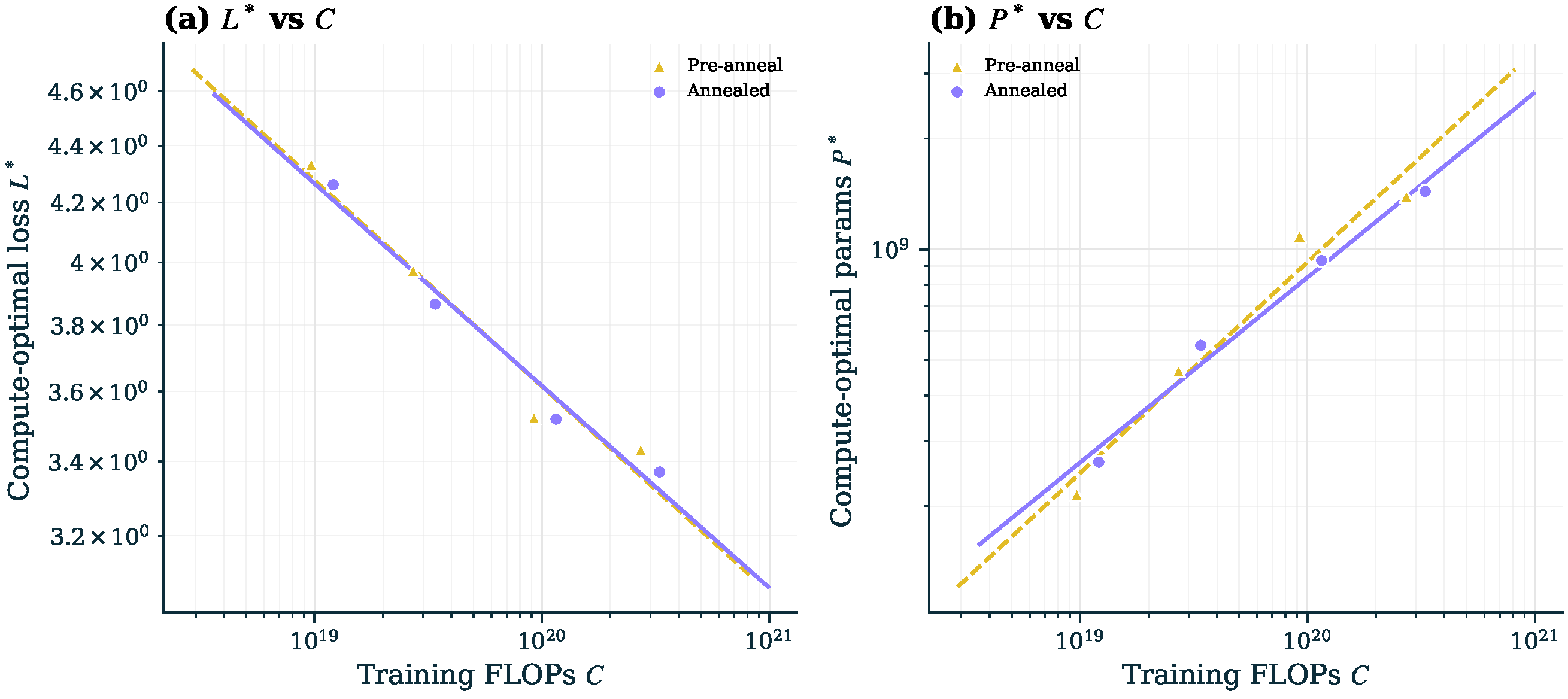}
\caption{CDLM+Concept validation loss, pre-anneal versus annealed. Top row: per-slice parabolas in each state. Bottom row: $\mathcal{L}^*(C)$ and $P^*(C)$ power-law fits.}
\label{fig:pre-anneal-elbo}
\end{figure}

Figure~\ref{fig:pre-anneal-elbo} shows IsoFLOP fits to CDLM+Concept checkpoints evaluated in both states, with the resulting power-law exponents reported in~\cref{tab:pre-anneal-validation}. The parameter exponent $\alpha_P$ shifts from $0.574$ pre-anneal to $0.686$ post-anneal, a $0.11$ absolute increase. The loss exponent $\alpha_L$ is essentially unchanged, and the slice-wise $\mathcal{L}^*$ values shift by at most $0.11$ nats.

The $\alpha_P$ shift is not a uniform "more tokens" effect of the anneal phase: a uniform downward shift in $\mathcal{L}^*$ across slices would change the intercept of $\log \mathcal{L}^*$ versus $\log C$ but leave the per-slice $\log P^*$ minima fixed, preserving $\alpha_P$ exactly (the slice-wise parabolas would shift vertically without translating along the parameter axis). The observed shift in $\alpha_P$ requires that annealing improves loss \emph{asymmetrically} across model sizes within each slice, moving the parabola minimum along $\log P$. The result is that the compute-optimal model size $P^*(C)$ is systematically larger under annealed evaluation than under pre-anneal evaluation.

\input{sections/appendix/tables/pre_anneal_validation.tex}

 \subsection{Interpretability}
\label{subsubsec:annealing-interpretability}

\begin{figure}[htbp!]
\centering
\includegraphics[width=0.85\linewidth]{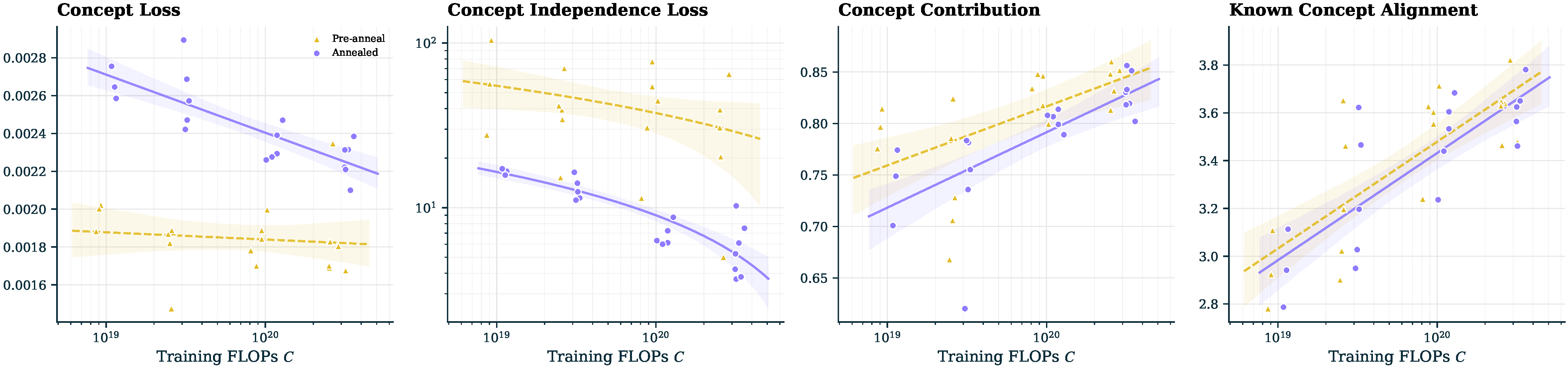}
\caption{CDLM+Concept interpretability metrics, pre-anneal versus annealed. Three metrics are robust; Concept Independence Loss decreases $3$-$10\times$ under annealed evaluation.}
\label{fig:pre-anneal-interp}
\end{figure}

Figure~\ref{fig:pre-anneal-interp} shows the four interpretability metrics evaluated in both states across the same CDLM+Concept checkpoints, with the resulting power-law fits in~\cref{tab:pre-anneal-interp}. Three of the four metrics are robust to annealing: Concept Loss and Concept Contribution shift by at most $0.02$ on their respective scales, with near-identical slopes; Known Concept Alignment is essentially unchanged ($\beta = 0.447$ pre-anneal vs $0.437$ annealed, $R^2$ within $0.01$). Concept Independence Loss is the outlier: pre-anneal HSIC sits in the $30$-$80$ range with a noisy trend ($R^2 = 0.14$), while annealed HSIC drops to the $5$-$20$ range with a clean trend ($R^2 = 0.79$), a $3$-$10\times$ reduction across checkpoints.

The pattern is consistent with which inference-time quantities each metric depends on. Known Concept Alignment is computed from the concept embeddings $K_c$ projected to vocabulary space and is invariant to the inference-time mixing of the known and unknown heads, so the schedule changes during anneal do not affect it. Concept Loss and Concept Contribution depend on the relative contributions of known, unknown, and residual pathways at inference; these shift modestly with $\alpha_{\text{known}}$ and $\alpha_{\text{unknown}}$. Concept Independence Loss is much more sensitive: multiple schedule changes (top-$k$ sparsification, the shift to fully model-predicted heads) act on the representations $\hat{k}$ and $\hat{u}$ during anneal, and the metric drops $3$-$10\times$.

\input{sections/appendix/tables/pre_anneal_interp.tex}

%% file: sections/appendix/tables/scaling_notation.tex
\begin{table}[h]
\centering
\begin{tabular}{lll}
\toprule
Symbol & Type & Meaning \\
\midrule
\multicolumn{3}{l}{\textit{Compute and resources}} \\
$C$ & scalar & Total training FLOPs \\
$M$ & scalar & Per-token FLOPs (forward + backward) \\
$P$ & scalar & Non-embedding parameter count \\
$D$ & scalar & Number of training tokens (total) \\
$D_i$ & scalar & Training tokens for the $i$-th checkpoint \\
\midrule
\multicolumn{3}{l}{\textit{Validation losses}} \\
$\mathcal{L}_i$ & scalar & Measured validation loss for checkpoint $i$ \\
$\mathcal{L}^*(C)$ & function & Compute-optimal validation loss at budget $C$ \\
$\mathcal{L}_\infty$ & scalar & Irreducible validation loss \\
$\mathcal{L}(P, D)$ & function & Joint Chinchilla loss surface \\
$\mathcal{L}_{\text{fit}}$ & function & Huber loss minimized in the joint fit \\
\midrule
\multicolumn{3}{l}{\textit{Power-law parameters}} \\
$P^*(C)$ & function & Compute-optimal parameter count at budget $C$ \\
$a_P, a_L$ & scalars & Power-law coefficients (parameter and loss) \\
$\alpha_P$ & scalar & Exponent on compute for parameter scaling \\
$\alpha_L$ & scalar & Exponent on compute for loss scaling \\
$\alpha_D$ & scalar & Exponent on compute for training-token scaling \\
$A_P, A_D$ & scalars & Chinchilla coefficients (parameter and data sides) \\
$\alpha, \beta$ & scalars & Chinchilla exponents (parameter and data sides) \\
\midrule
\multicolumn{3}{l}{\textit{Interpretability scaling}} \\
$m(C)$ & function & A metric as a function of compute \\
$e$ & scalar & Irreducible value of a metric \\
$A, \beta$ & scalars & Coefficient and exponent for metric scaling \\
\bottomrule
\end{tabular}
\caption{Notation introduced in the scaling-law analysis, grouped by role: compute and resources, validation losses, power-law parameters, and interpretability scaling.}
\label{tab:scaling-notation}
\end{table}

%% file: sections/appendix/tables/elbo_comparison.tex
\begin{table}[t]
\centering
\small
\setlength{\tabcolsep}{4pt}
\begin{tabular}{@{}lcccc@{}}
\toprule
\textbf{Estimator} & $\alpha_P$ & $\alpha_L$ & $\alpha_D$ & $\mathcal{L}_\infty$ \\
\midrule
Fixed-rate mask & $0.612\;[0.564,\,0.661]$ & $-0.071\;[-0.075,\,-0.067]$ & $0.374\;[0.317,\,0.525]$ & $1.914\;[1.263,\,2.070]$ \\
Fixed-grid ELBO & $0.611\;[0.568,\,0.658]$ & $-0.054\;[-0.057,\,-0.051]$ & $0.368\;[0.309,\,0.555]$ & $2.416\;[1.786,\,2.560]$ \\
MDLM ELBO & $0.632\;[0.541,\,0.707]$ & $-0.053\;[-0.058,\,-0.049]$ & $0.481\;[0.370,\,0.643]$ & $2.658\;[1.950,\,2.839]$ \\
Per-block ELBO & $0.615\;[0.572,\,0.660]$ & $-0.039\;[-0.041,\,-0.036]$ & $0.340\;[0.281,\,0.558]$ & $3.353\;[2.685,\,3.526]$ \\
\bottomrule
\end{tabular}
\caption{IsoFLOP power-law exponents and irreducible-loss asymptotes under four ELBO estimators on CDLM. Subscripts are 90\% bootstrap confidence intervals.}
\label{tab:elbo-comparison}
\end{table}

%% file: sections/appendix/tables/mask_rate_exponents.tex
\begin{table}[htbp!]
\centering
\small
\setlength{\tabcolsep}{8pt}
\begin{tabular}{@{}lcccc@{}}
\toprule
\textbf{Mask rate $t$} & $\alpha_P$ & $\alpha_L$ & $R^2(P)$ & $R^2(\mathcal{L})$ \\
\midrule
$0.1$ & $0.612$ & $-0.120$ & $0.991$ & $0.992$ \\
$0.2$ & $0.612$ & $-0.110$ & $0.991$ & $0.994$ \\
$0.3$ & $0.612$ & $-0.099$ & $0.992$ & $0.995$ \\
$0.4$ & $0.609$ & $-0.086$ & $0.992$ & $0.996$ \\
$0.5$ & $0.609$ & $-0.071$ & $0.993$ & $0.997$ \\
$0.6$ & $0.605$ & $-0.057$ & $0.993$ & $0.997$ \\
$0.7$ & $0.604$ & $-0.042$ & $0.993$ & $0.998$ \\
$0.8$ & $0.624$ & $-0.026$ & $0.996$ & $0.998$ \\
$0.9$ & $0.623$ & $-0.013$ & $0.995$ & $0.997$ \\
\bottomrule
\end{tabular}
\caption{IsoFLOP power-law exponents at each fixed mask rate $t$ (CDLM family).}
\label{tab:mask-rate-exponents}
\end{table}

%% file: sections/appendix/tables/pre_anneal_validation.tex
\begin{table}[h]
\centering
\small
\setlength{\tabcolsep}{8pt}
\begin{tabular}{@{}lcccc@{}}
\toprule
\textbf{Condition} & $\alpha_P$ & $\alpha_L$ & $R^2(P)$ & $R^2(L)$ \\
\midrule
Pre-anneal & $0.574$ & $-0.073$ & $0.959$ & $0.956$ \\
Annealed   & $0.503$ & $-0.072$ & $0.983$ & $0.976$ \\
\bottomrule
\end{tabular}
\caption{Compute-optimal scaling-law exponents for CDLM+Concept, evaluated pre- and post-anneal.}
\label{tab:pre-anneal-validation}
\end{table}

%% file: sections/appendix/tables/pre_anneal_interp.tex
\begin{table}[h]
\centering
\small
\setlength{\tabcolsep}{8pt}
\begin{tabular}{@{}llcc@{}}
\toprule
\textbf{Metric} & \textbf{Condition} & $\beta$ & $R^2$ \\
\midrule
Concept Loss                & Pre-anneal & $-0.000$ & $0.014$ \\
                            & Annealed   & $-0.000$ & $0.648$ \\
\midrule
Concept Independence Loss   & Pre-anneal & $-17.425$ & $0.141$ \\
                            & Annealed   & $-7.497$ & $0.801$ \\
\midrule
Concept Contribution        & Pre-anneal & $+0.057$ & $0.350$ \\
                            & Annealed   & $+0.073$ & $0.536$ \\
\midrule
Known Concept Alignment     & Pre-anneal & $+0.447$ & $0.609$ \\
                            & Annealed   & $+0.449$ & $0.645$ \\
\bottomrule
\end{tabular}
\caption{Interpretability metric trends versus compute for CDLM+Concept, evaluated pre- and post-anneal. $\beta$ is the log-linear slope of the metric against $\log_{10}(C)$.}
\label{tab:pre-anneal-interp}
\end{table}

%% file: sections/appendix/steerling_pretraining.tex
\clearpage
\part{ \steerlingB pretraining details}
\label{app:steerling-pretraining}

\section{Pretraining recipe ablations}
\label{app:pretraining-ablations}

The interpretable causal-diffusion architecture introduces design choices with no established defaults in the autoregressive or masked diffusion literature. For each choice in the recipe (\cref{sec:pretraining-choices}) we run a small ablation that sweeps one variable from a fixed baseline, a 1B model trained on 30B sampled from Nemotron-CC-HQ, with the full default configuration in \cref{tab:ablation-baseline}.

\input{sections/appendix/tables/ablation_baseline}

We track five metrics at every saved checkpoint. Three measure capability: validation loss; MMLU soft score, the log-probability the model assigns to the correct answer letter on MMLU~\citep{hendrycks2020measuring}, taken relative to the four choices~\citep{openathena2026delphi}, that is $\log P(\text{correct}) - \log \sum_{c \in \{A,B,C,D\}} P(c)$, which is bounded above by $0$ and varies continuously with capability while hard accuracy is still pinned near chance; and HellaSwag accuracy~\citep{zellers2019hellaswag}. Validation loss and MMLU soft score are continuous and retain signal at the 1B-parameter scale, where standard accuracy is near random and unable to discriminate between configurations~\citep{olmo, openathena2026delphi}. The remaining two measure interpretability: concept contribution (Equation~\ref{eq:concept-contribution}) and known concept alignment (\cref{sec:interp_metrics}).

\subsection{Diffusion block size}
\label{subsubsec:diff_block_size}

\begin{figure}[htbp!]
  \centering
  \includegraphics[width=\linewidth]{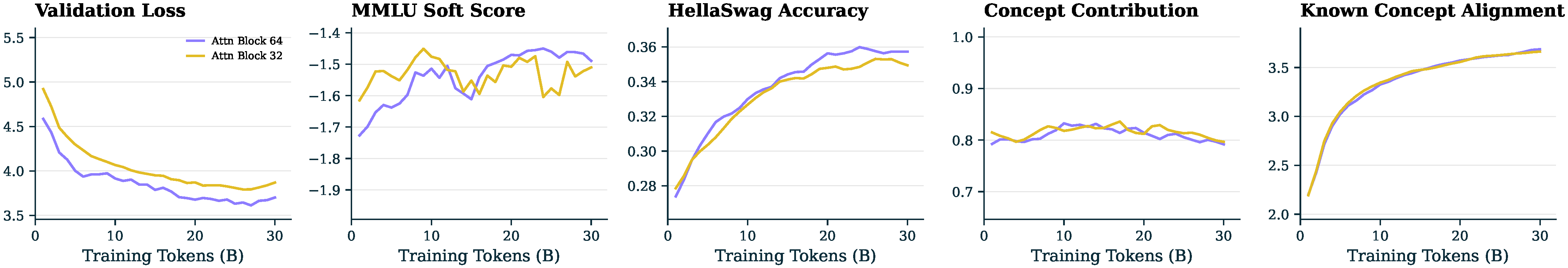}
  \caption{Block size ablation, $b \in \{32, 64\}$.}
  \label{fig:abl-block-size}
\end{figure}

\steerling is a causal diffusion model with the block-causal attention mask of \cref{fig:attention-patterns}: tokens within a block attend to each other and to all previous blocks, so within a single block the attention is effectively bidirectional. The block size $b$ controls how many tokens see each other bidirectionally, the maximum number of tokens that can be decoded in parallel at inference, and the granularity at which keys and values are cached across blocks. We compare $b \in \{32, 64\}$ on the five metrics introduced above, with results shown in \cref{fig:abl-block-size}.

The two settings are indistinguishable on both interpretability metrics: concept contribution and known concept alignment track each other within the noise band over the whole run. On capability the picture tilts toward the larger block. MMLU soft score is noisy and overlapping, with neither value holding a consistent lead, but validation loss separates cleanly, with $b = 64$ sitting below $b = 32$ for essentially the entire run, and HellaSwag shows the same tilt, with $b = 64$ pulling slightly ahead over the back half. Since the larger block is at least as good everywhere, better on validation loss and HellaSwag, and decodes more tokens in parallel at inference, we adopt it.

\keypoint{Block size leaves interpretability unchanged; $b = 64$ improves validation loss and HellaSwag, so we adopt it.}

% --- Choice 2 -------------------------------------------------------

\subsection{Diffusion masking schedule}
\label{app:abl-masking}

\begin{figure}[htbp!]
  \centering
  \includegraphics[width=0.4\linewidth]{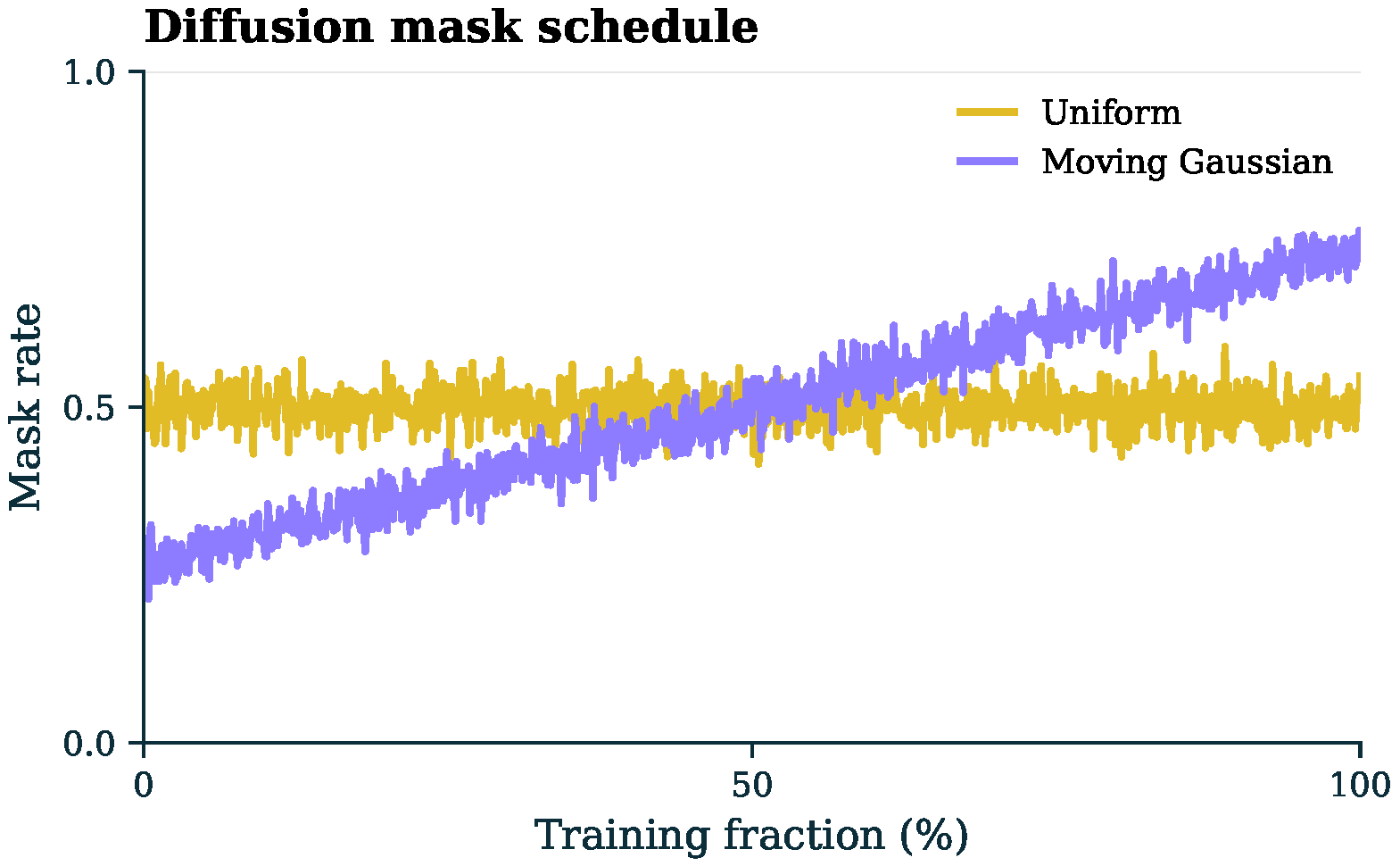}
  \caption{Mean mask rate per training step under the two sampling schedules.}
  \label{fig:masking-schedules}
\end{figure}

\begin{figure}[htbp!]
  \centering
  \includegraphics[width=\linewidth]{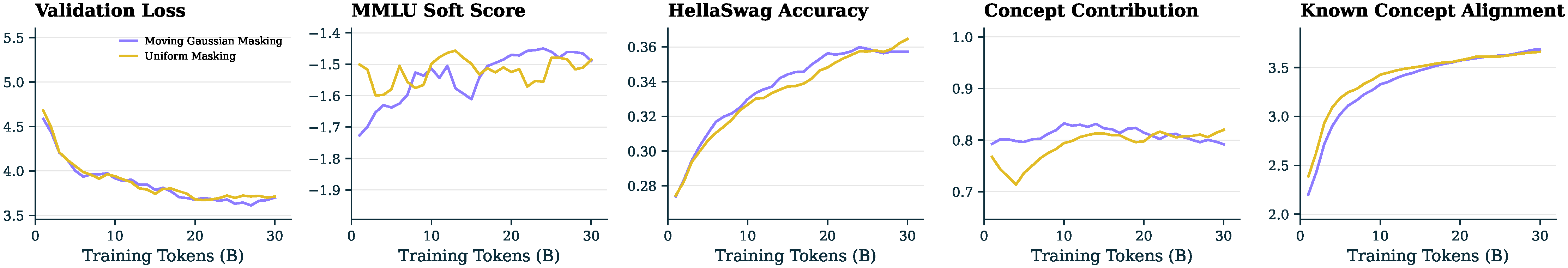}
  \caption{Masking schedule ablation, uniform vs.\ moving Gaussian.}
  \label{fig:abl-masking}
\end{figure}

Unlike autoregressive models, where the training objective is fixed at next-token cross-entropy, diffusion language models must additionally choose how the noise level $t$ is drawn at each step. \citet{llada} sample $t \sim \mathcal{U}(0,1)$ uniformly. \citet{quokka} report a small but consistent gain from a moving Gaussian curriculum, in which $t$ is drawn from a Gaussian window whose center shifts from low to high noise over training, exposing the model to easier (less masked) sequences early and harder ones late. The block-causal attention of \steerling differs from the full-attention setting in which both were measured, so we re-examine the choice here. We compare uniform $t \sim \mathcal{U}(0.05, 0.95)$~\citep{mdlm, llada, sahoo2024} against a moving Gaussian curriculum with center increasing linearly from $0.2$ to $0.8$ and $\sigma = 0.3$; the schedules are shown in \cref{fig:masking-schedules} and results in \cref{fig:abl-masking}.

At this scale the two schedules are very close. The moving Gaussian holds a very slight edge on validation loss over the back half of the run, in line with what \citet{quokka} report, while MMLU soft score and HellaSwag overlap throughout, with uniform finishing marginally ahead on HellaSwag. The interpretability metrics track each other after an early transient on concept contribution. With no setting clearly ahead, we adopt the moving Gaussian on the validation-loss edge and consistency with prior work.\footnote{This schedule proved too aggressive over the full pretraining run; see \cref{subsec:lesson-masking}.}

\keypoint{The two masking schedules are very close at ablation scale, with a slight edge to the moving Gaussian on validation loss; we adopt it.}

% --- Choice 3 -------------------------------------------------------
\subsection{Unknown concept capacity}
\begin{figure}[htbp!]
  \centering
  \includegraphics[width=\linewidth]{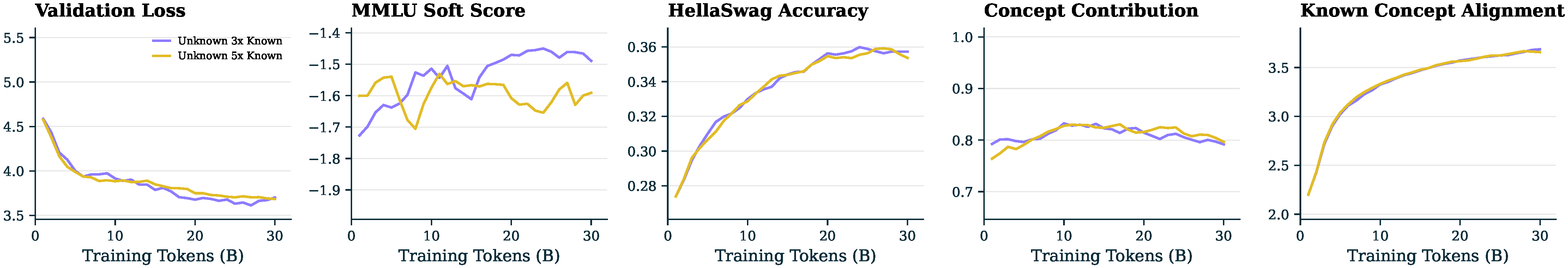}
  \caption{Unknown concept capacity ablation, $m \in \{3n, 5n\}$.}
  \label{fig:abl-unknown-capacity}
\end{figure}

\steerling's concept module splits its representation into known concepts, given by the data pipeline, and unknown concepts, learned during training. The number of known concepts $n$ is fixed by the \atlas concept library, but the number of unknown concepts $m$ is free. A larger $m$ gives the model more room to discover recurring patterns the known library does not cover, at the cost of parameters, compute, inference latency, and memory that could otherwise serve the language modeling objective. We compare $m = 5n$ against $m = 3n$; results are shown in \cref{fig:abl-unknown-capacity}.

The two settings are indistinguishable on the interpretability metrics. On capability they are also close, and where they separate it slightly favours the smaller setting: $m = 3n$ holds a mild edge on MMLU soft score over the back half of the run, while validation loss and HellaSwag overlap throughout. The extra capacity of $m = 5n$ buys no measurable improvement, so we adopt the more conservative setting.

\keypoint{Raising unknown capacity from $3n$ to $5n$ yields no measurable gain; we keep the conservative $m = 3n$.}

% --- Choice 4 -------------------------------------------------------

\subsection{Unknown embedding factorization}
\begin{figure}[htbp!]
  \centering
  \includegraphics[width=\linewidth]{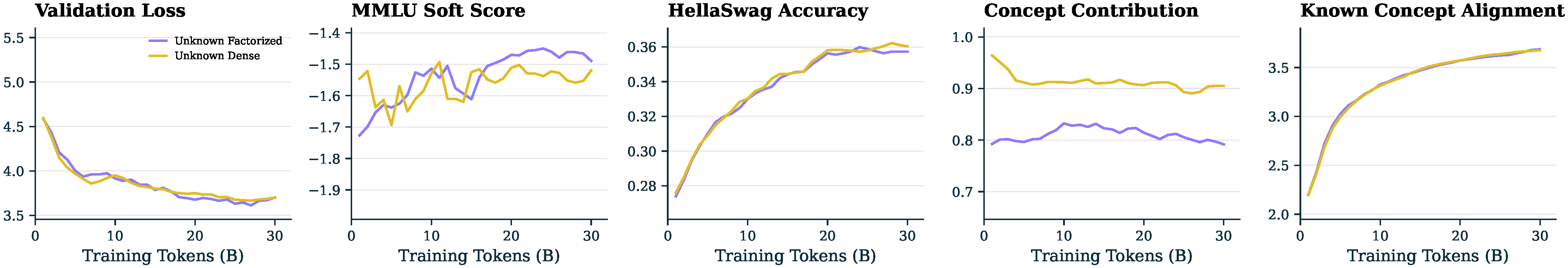}
  \caption{Unknown embedding factorization ablation: dense vs.\ factorized ($R = 256$).}
  \label{fig:abl-factorization}
\end{figure}

The unknown head's embedding matrix $U \in \mathbb{R}^{m \times d}$ is the largest single parameter the concept module adds: at $m \gg n$ it dominates the parameters \steerling carries over a standard backbone of the same size, and computing $\hat{u} = u^\top U$ at every token is a heavy matmul in the forward pass. We therefore factorize $U = AB$ with $A \in \mathbb{R}^{m \times R}$, $B \in \mathbb{R}^{R \times d}$, and rank $R = 256 \ll d$, cutting both the parameter count and the per-step compute of the unknown pathway. We compare the dense $U$ against the factorized form; results are in \cref{fig:abl-factorization}.

Factorization is close to free on capability: validation loss, MMLU soft score, and HellaSwag all overlap the dense baseline throughout, and known concept alignment is identical. The one cost is concept contribution, where the dense head holds a steady $\sim\!0.90$ against $\sim\!0.80$ for the factorized head, as the low-rank bottleneck forces the model to rely more on the residual. However, the saving is large: at the \steerlingB scale the dense $U$ is roughly $15\times$ larger than its factorized form, and factorization also removes the corresponding matrix multiplication from the per-token forward pass. We judge the small contribution cost well worth this and adopt $R = 256$.

\keypoint{Factorizing $U$ at $R = 256$ makes the unknown embedding roughly $15\times$ smaller with no capability cost and a small drop in concept contribution; we adopt it.}

\subsection{Use of the residual term}
\begin{figure}[htbp!]
  \centering
  \includegraphics[width=\linewidth]{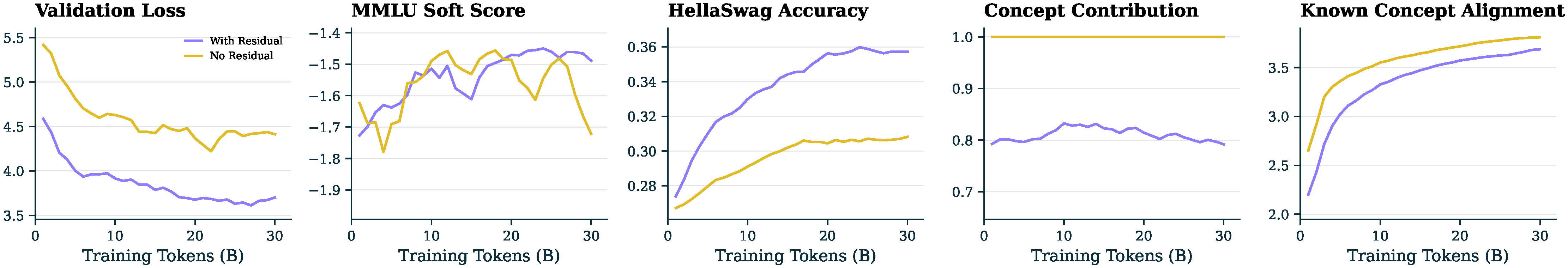}
  \caption{Residual term ablation: with vs.\ without $\varepsilon$.}
  \label{fig:abl-residual}
\end{figure}

The bottlenecked hidden state passed to the LM head, $\bar{h} = \hat{k} + \hat{u} + \varepsilon$, carries a residual $\varepsilon = h - \hat{k} - \hat{u}$ that absorbs whatever the two heads fail to reconstruct, so that $\bar{h} = h$ identically. Dropping it sets $\bar{h} = \hat{k} + \hat{u}$ and forces every dimension of the bottleneck to be the sum of a known and an unknown concept contribution, leaving the model no uninterpreted channel. We compare the two formulations; results are in \cref{fig:abl-residual}.

Removing the residual sharply harms the capability metrics: validation loss separates from the first tokens and never recovers, ending well above the residual baseline; HellaSwag plateaus around $0.31$ against $0.36$ for the baseline; and MMLU soft score is noisier and weaker over the back half. The perfect concept contribution here is true by construction, not a sign of a better decomposition: with no $\varepsilon$ channel the hidden state has nowhere else to fall. It does not justify the capability degradation, and given the uncertainty over how that cost would compound at larger scale, we keep the residual.

\keypoint{Dropping $\varepsilon$ forces concept contribution to $1.0$ by construction but inflicts a large, persistent capability penalty; we keep the residual.}

\begin{figure}[htbp!]
  \centering
  \includegraphics[width=0.85\linewidth]{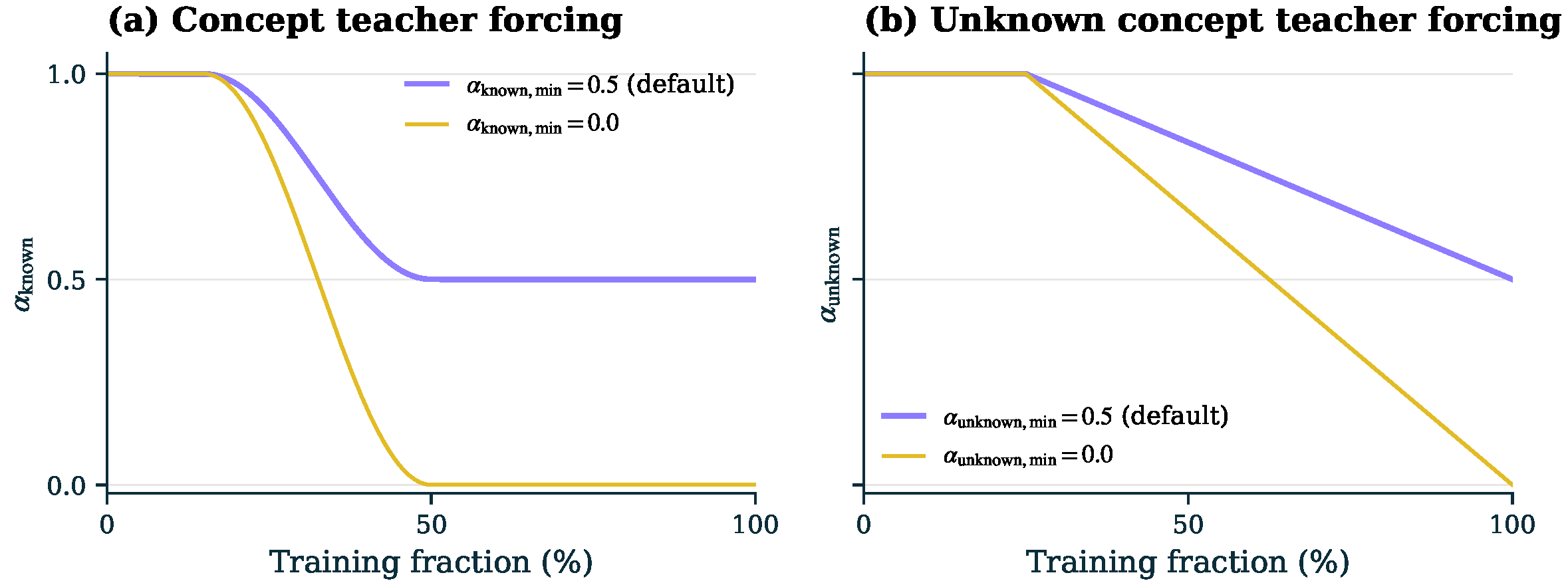}
  \caption{Teacher forcing schedules for the two ablations. (a) $\alpha_{\text{known}}(s)$ with cosine decay. (b) $\alpha_{\text{unknown}}(s)$ with linear decay. Defaults in bold.}
  \label{fig:tf-schedules}
\end{figure}

\subsection{Concept teacher forcing schedule}
\begin{figure}[htbp!]
  \centering
  \includegraphics[width=\linewidth]{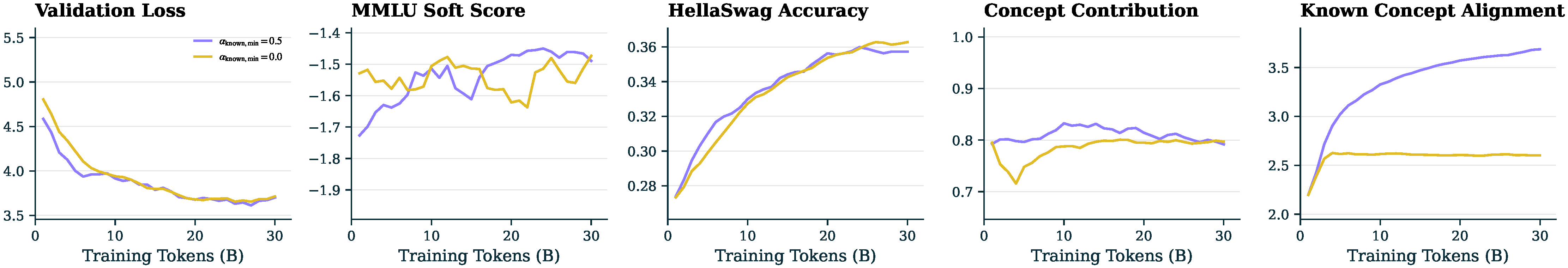}
  \caption{Concept teacher forcing floor ablation, $\alpha_{\text{known,min}} \in \{0.5, 0.0\}$.}
  \label{fig:abl-tf-known}
\end{figure}

The known head is trained from scratch alongside the transformer, so early on its predicted activations $\hat{k}$ are unreliable, and routing $\bar{h}$ through them can drive concept leakage into the language modeling loss~\citep{mahinpei2021promises}. The standard mitigation is to feed the ground-truth $\hat{k}^{\text{GT}}$ to the LM head instead, known as independent training~\citep{cbm}, but at our scale a model that never sees its own predictions would be unprepared for inference, where no ground truth exists. We therefore use the schedule of \cref{subsubsec:loss-objectives}, where $\alpha_{\text{known}}(s)$ starts at full teacher forcing, decays via cosine annealing during the early phase, and holds at a floor afterwards. Shape, warmup, and starting value are fixed at the defaults of \cref{tab:ablation-baseline}, and we ablate the floor over $\alpha_{\text{known,min}} \in \{0.5, 0.0\}$. The schedules are shown in \cref{fig:tf-schedules}(a) and results in \cref{fig:abl-tf-known}.

Capability is unaffected by the floor, with validation loss, MMLU soft score, and HellaSwag overlapping across the run. The difference is on the interpretability side, and it is large. Holding $\alpha_{\text{known}}$ at $0.5$ keeps known concept alignment climbing to $\sim\!3.7$, whereas decaying to $0.0$ stalls it near $2.6$ from early in training. We therefore hold the floor at $0.5$.

\keypoint{Decaying $\alpha_{\text{known}}$ to $0.0$ during pretraining leaves capability intact but collapses known concept alignment; we hold the floor at $0.5$.}

% --- Choice 8 -------------------------------------------------------

\subsection{Unknown concept teacher forcing schedule}
\begin{figure}[htbp!]
  \centering
  \includegraphics[width=\linewidth]{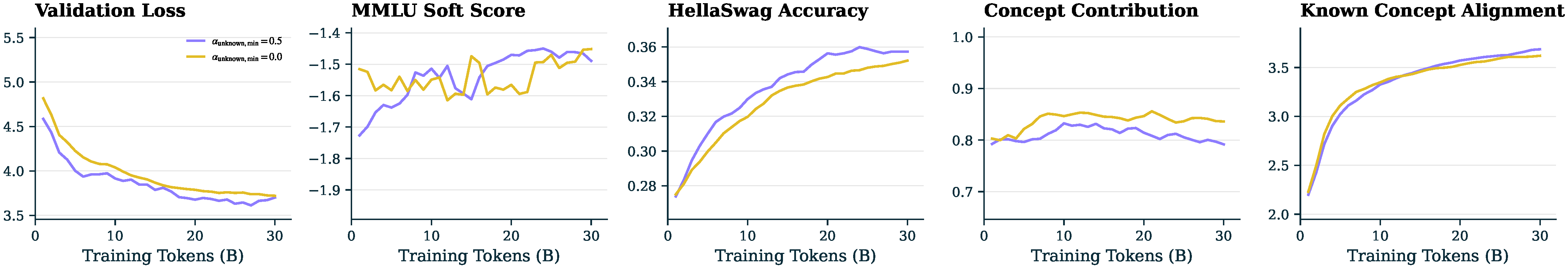}
  \caption{Unknown concept teacher forcing floor ablation, $\alpha_{\text{unknown,min}} \in \{0.5, 0.0\}$.}
  \label{fig:abl-tf-unknown}
\end{figure}

The unknown head faces the same early-training instability as the known head, but no ground-truth labels exist to substitute for its predicted activations $\hat{u}$. The natural target is instead the analytical residual $\hat{u}^{\text{GT}} = h - \hat{k}^{\text{GT}}$ from Equation~\ref{eq:gt-targets}, computed from the transformer hidden state and the labeled known concepts. We use the schedule of \cref{subsubsec:loss-objectives}, where $\alpha_{\text{unknown}}(s)$ starts at full teacher forcing, decays linearly, and holds at a floor afterwards. As with the known head, shape, warmup, and starting value are fixed at the defaults of \cref{tab:ablation-baseline}, and we ablate the floor over $\alpha_{\text{unknown,min}} \in \{0.5, 0.0\}$. The schedules are shown in \cref{fig:tf-schedules}(b) and results in \cref{fig:abl-tf-unknown}.

Unlike the known head, the unknown floor has only modest effects, but where the metrics separate they favour the higher floor: validation loss is slightly lower and HellaSwag consistently higher at $\alpha_{\text{unknown,min}} = 0.5$, while concept contribution is slightly higher at $\alpha_{\text{unknown,min}} = 0.0$, and the rest overlap. We adopt $0.5$.

\keypoint{The unknown teacher forcing floor has modest effects that favour $0.5$ on capability; we adopt it.}

\input{sections/steerling_pretraining/tables/ablation_summary}

\clearpage

\section{Final pretraining \steerlingB configuration}
\label{app:steerling8b-config}
\input{sections/appendix/tables/steerling8b_config}
\clearpage
\section{Pretraining diagnostic}
\label{app:pretraining-diagnostic}

Here we look into the issues raised in the pretraining lessons (\cref{sec:pretraining-lessons}). We plot each metric alongside the training schedule that most plausibly drives it, on a secondary axis, allowing the alignment between schedule transitions and metric inflections to be read directly.

\begin{figure}[htbp!]
  \centering
  \includegraphics[width=\linewidth]{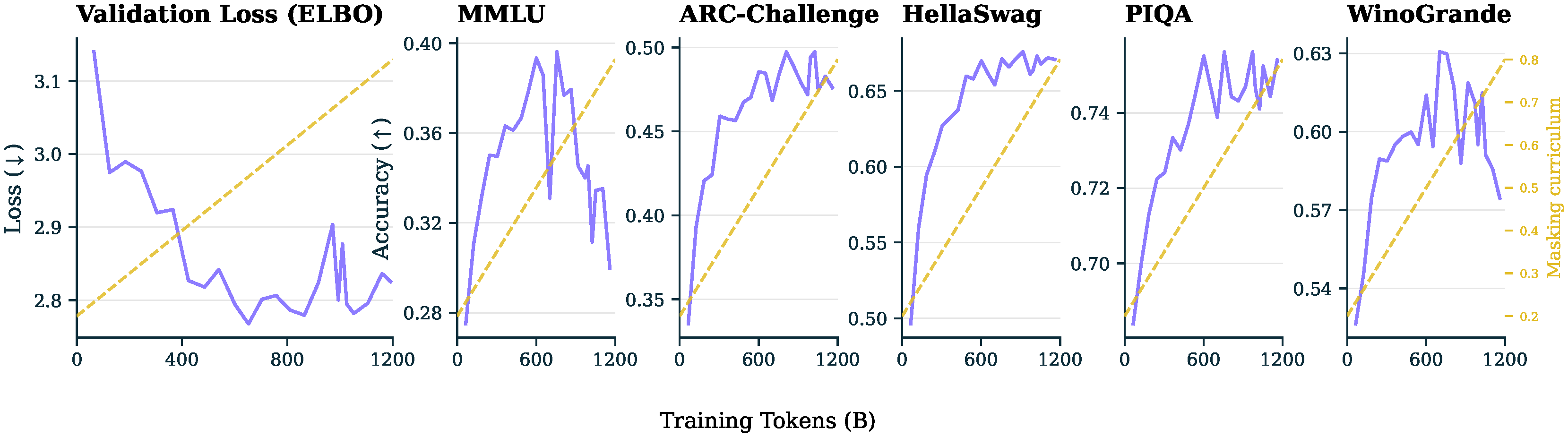}
  \caption{Capability metrics during \steerlingB pretraining with the masking curriculum overlaid (right axis, blue dashed). The curriculum passes center $0.5$ around the midpoint of training.}
  \label{fig:appendix-pretraining-capability}
\end{figure}

Figure~\ref{fig:appendix-pretraining-capability} overlays the masking curriculum on validation loss and the five downstream benchmarks. The schedule begins at center $0.2$ and rises linearly to $0.8$ over the course of pretraining, passing $0.5$ around the midpoint. MMLU and WinoGrande peak shortly before the schedule crosses $0.5$ and decline as it continues to steepen. ARC-Challenge peaks slightly later with a smaller decline; HellaSwag and PIQA hold near their peak values; validation loss plateaus rather than continuing to descend. The decline on the harder reasoning benchmarks tracks the masking curriculum directly.

\begin{figure}[htbp!]
  \centering
  \includegraphics[width=\linewidth]{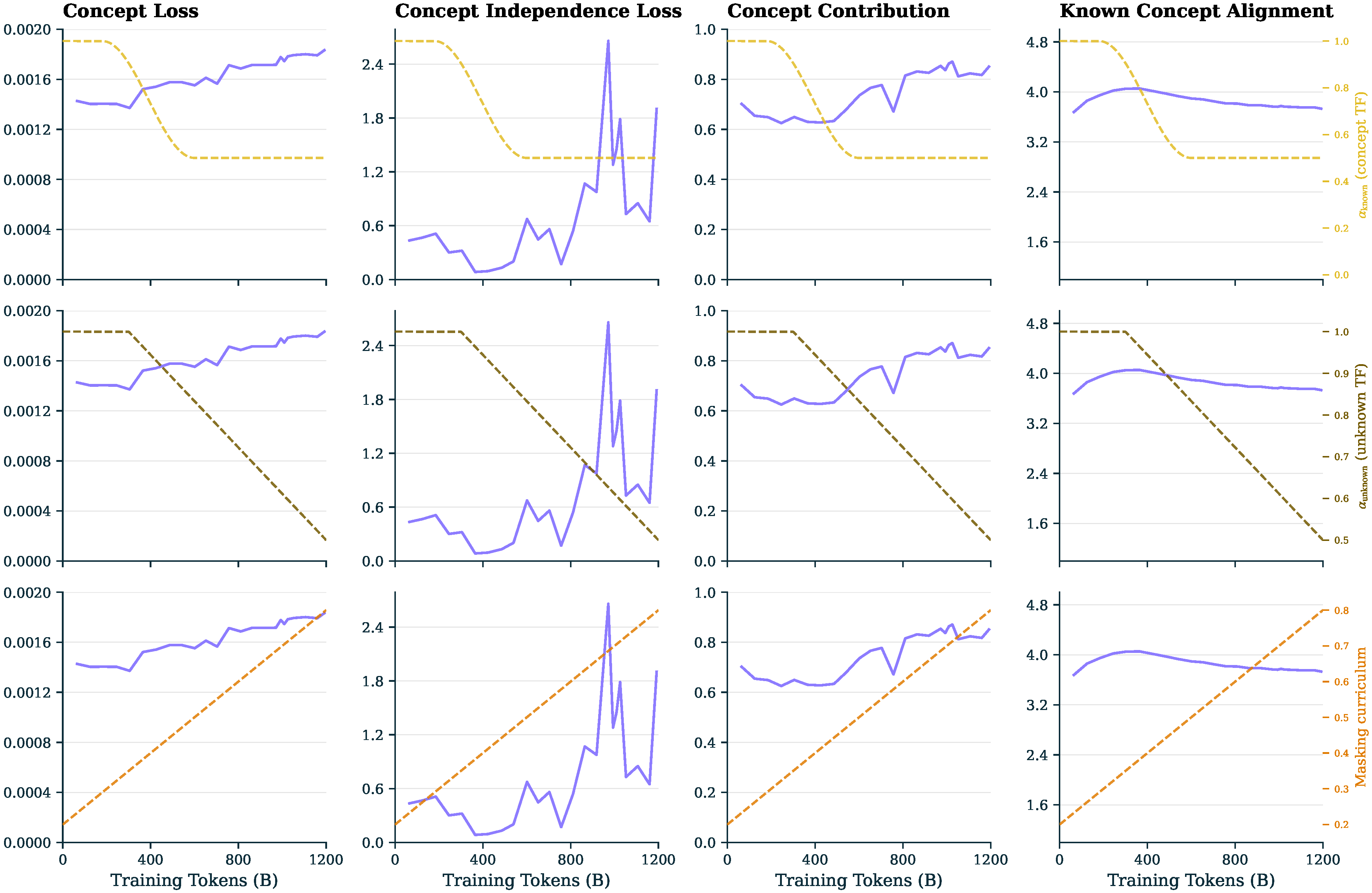}
  \caption{Interpretability metrics during \steerlingB pretraining with the three training schedules overlaid:  $\alpha_{\text{known}}$, $\alpha_{\text{unknown}}$, and masking curriculum.}
  \label{fig:appendix-pretraining-interp}
\end{figure}

Figure~\ref{fig:appendix-pretraining-interp} overlays the masking, concept teacher forcing, and unknown concept teacher forcing schedules on the four interpretability metrics. The teacher forcing schedules decay from $1.0$ to their floor of $0.5$: $\alpha_{\text{known}}$ via cosine annealing over the first $50\%$ of training, $\alpha_{\text{unknown}}$ linearly across the full run. Concept independence loss is near zero through the first two thirds of training and rises sharply in the final third, coinciding with the masking curriculum reaching its hard regime and the two teacher forcing schedules having moved substantially away from their starting values. Concept contribution climbs over the same window as the LM head depends increasingly on predicted concepts. Known concept alignment and concept loss change only modestly throughout, on small absolute scales.

\clearpage

\section{Steering judge}
\label{app:steering-judge}

The steering evaluation in ~\cref{tab:steering_pretrained_method_comparison} relies on an LLM-judge to score generated continuations. 
We use Mistral-Small-3.1-24B-Instruct at temperature 0. For each continuation we call the judge twice: once with the prompt in~\cref{fig:judge-concept-prompt} to produce a \emph{concept score} from 0 to 2 measuring how strongly the target concept appears in the continuation, and once with the prompt in~\cref{fig:judge-quality-prompt} to produce a \emph{quality score} from 0 to 2 measuring fluency and coherence. Each prompt is sent as a single user message through the model's chat template, with no separate system instruction. Both prompts instruct the judge to score only the continuation, not the prefix it was generated from.

\begin{figure}[htbp!]
\begin{lstlisting}[basicstyle=\ttfamily\footnotesize,frame=single]
You are an impartial evaluator.

Task: Rate how strongly the ANSWER expresses the TARGET CONCEPT, based on
meaning.
- Score the ANSWER text only.
- The concept may be expressed without using the exact label words; judge by
  meaning.

Examples:
Rating: [[2]] - if the answer clearly matches the concept description with
                multiple explicit supporting details.
Rating: [[1]] - if the answer is loosely related but provides only minimal or
                indirect evidence.
Rating: [[0]] - if the answer is unrelated to the concept.

Score:
2 = concept clearly present with strong evidence
1 = concept present but weak/indirect/minimal
0 = concept absent

Output ONLY one line in exactly this format: Rating: [[0]] or Rating: [[1]]
or Rating: [[2]]

[TARGET CONCEPT LABEL]
{concept_label}

[TARGET CONCEPT DESCRIPTION]
{concept_desc}

[ANSWER]
{answer}

Now score the ANSWER's concept expression. Answer with 'Rating: [[0]]' or
'Rating: [[1]]' or 'Rating: [[2]]' only.
\end{lstlisting}
\caption{Prompt for the steering concept-score judge.}
\label{fig:judge-concept-prompt}
\end{figure}

\begin{figure}[htbp!]
\begin{lstlisting}[basicstyle=\ttfamily\footnotesize,frame=single]
You are an impartial evaluator.

Task: Rate the text quality of the CONTINUATION text ONLY, considering
fluency, coherence, and readability.
IMPORTANT:
- Do NOT score the PREFIX quality; score ONLY the continuation's text quality.

Examples:
Rating: [[2]] - The continuation is fluent and coherent, and it follows
                naturally from the prefix.
Rating: [[1]] - The continuation is mostly understandable but has noticeable
                issues (awkwardness, jumps, mild repetition).
Rating: [[0]] - The continuation is hard to read (incoherent, severe
                repetition) or does not connect to the prefix.

Score:
0 = very poor (incoherent, severe repetition, hard to read)
1 = understandable but with issues (awkward phrasing, jumps, mild repetition)
2 = fluent, coherent, easy to read

Output ONLY one line in exactly this format: Rating: [[0]] or Rating: [[1]]
or Rating: [[2]]

[PREFIX - for context only, do not score]
{prompt_text}

[CONTINUATION - SCORE THIS ONLY]
{answer}

Now score the CONTINUATION's text quality. Answer with 'Rating: [[0]]' or
'Rating: [[1]]' or 'Rating: [[2]]' only.
\end{lstlisting}
\caption{Prompt for the steering quality-score judge.}
\label{fig:judge-quality-prompt}
\end{figure}

\clearpage

%% file: sections/appendix/tables/ablation_baseline.tex
\begin{table}[htbp!]
\centering
\small
\setlength{\tabcolsep}{4pt}
\renewcommand{\arraystretch}{0.95}
\begin{tabular}{@{}ll@{}}
\toprule
\textbf{Setting} & \textbf{Value} \\
\midrule
\multicolumn{2}{l}{\textit{Architecture}} \\
\midrule
Backbone size & 1.5B (non-embedding) \\
Hidden dimension $d$ & 2304 \\
Layers $L$ & 20 \\
Attention heads & 18 \\
Sequence length $N$ & 4096 \\
Block size $b$ & 64 \\
Unknown concept capacity & $m = 3n$ \\
Unknown factorization rank $R$ & 256 \\
Unknown decomposition & MLP \\
Gradient flow to unknown head & detached \\
\midrule
\multicolumn{2}{l}{\textit{Optimization}} \\
\midrule
Optimizer & AdamW ($\beta_1 = 0.9, \beta_2 = 0.95$) \\
Weight decay & 0.1 (excluding embeddings) \\
Peak learning rate & $4 \times 10^{-4}$ \\
LR schedule & constant at peak, 2\% warmup, no decay \\
Batch size & $524{,}288$ tokens \\
Training tokens & 20B \\
\midrule
\multicolumn{2}{l}{\textit{Concept module schedules}} \\
\midrule
Masking & Gaussian curriculum, center $0.2 \to 0.8$, $\sigma = 0.3$ \\
$\alpha_{\text{known}}(s)$ & $1.0 \to 0.5$, cosine \\
$\alpha_{\text{unknown}}(s)$ & $1.0 \to 0.5$, linear \\
$p_\varepsilon$ & 0.3 \\
$p_{\text{cfg}}$ & 0.1 \\
\midrule
\multicolumn{2}{l}{\textit{Loss weights}} \\
\midrule
$\lambda_{\text{concept}}$ & 1.0 \\
$\lambda_{\text{rec}}$ & 1.0 \\
$\lambda_{\text{indep}}$ & 1.0 \\
\midrule
\multicolumn{2}{l}{\textit{Data}} \\
\midrule
Source & Nemotron-CC-HQ \citep{nemotron} \\
\bottomrule
\end{tabular}
\caption{\looseness-1 Default configuration for pretraining ablations. Each ablation swaps one setting from this baseline.}
\label{tab:ablation-baseline}
\end{table}

%% file: sections/steerling_pretraining/tables/ablation_summary.tex
\begin{table}[htbp!]
\centering
\small
\setlength{\tabcolsep}{6pt}
\renewcommand{\arraystretch}{1.15}
\begin{tabular}{@{}llll@{}}
\toprule
Choice & Values & Default & Finding \\
\midrule
Block attention size $b$        & $\{32, 64\}$         & $64$      & No interp.\ effect; lower loss \\
Masking schedule                & uniform vs.\ Gaussian & Gaussian  & Very close; slight edge \\
Unknown capacity $m$            & $\{3n, 5n\}$         & $3n$      & No gain from more capacity \\
Unknown factorization rank $R$  & dense vs.\ $256$     & $256$     & $\sim\!15\times$ smaller, no cost \\
Residual term $\varepsilon$     & with vs.\ without    & with      & Removing it costs capability \\
$\alpha_{\text{known}}$ floor   & $\{0.5, 0.0\}$       & $0.5$     & $0.0$ collapses alignment \\
$\alpha_{\text{unknown}}$ floor & $\{0.5, 0.0\}$       & $0.5$     & Little effect \\
\bottomrule
\end{tabular}
\caption{Summary of the architecture and training ablations in \cref{sec:pretraining-choices}.}
\label{tab:ablation-summary}
\end{table}

%% file: sections/appendix/tables/steerling8b_config.tex
\begin{table}[htbp!]
\centering
\small
\setlength{\tabcolsep}{5pt}
\renewcommand{\arraystretch}{}
\begin{tabular}{@{}lll@{}}
\toprule
\textbf{Setting} & \textbf{Value} & \textbf{Notes} \\
\midrule
\multicolumn{3}{l}{\textit{Backbone}} \\
Layers $L$                         & $32$         & \\
Hidden dimension $d$               & $4096$       & \\
Attention heads                    & $32$         & GQA with $4$ KV heads \\
Sequence length $N$                & $4096$       & \\
MLP                                & SwiGLU       & ratio $4$, no biases \\
Normalisation                      & RMSNorm      & post-norm, QK-norm \\
Position encoding                  & RoPE         & base $5 \times 10^5$ \\
Weight tying                       & yes          & \\
\midrule
\multicolumn{3}{l}{\textit{Diffusion}} \\
Block size $b$                     & $64$         & \\
Masking schedule                   & moving Gaussian & center $0.2 \to 0.8$, $\sigma = 0.3$ \\
$t$ range                          & $[0.05, 0.95]$ & clipped at sampling \\
\midrule
\multicolumn{3}{l}{\textit{Concept module}} \\
Known concepts $n$                 & $33{,}732$   & from \atlas \\
Unknown concepts $m$               & $101{,}196$  & $m = 3n$ \\
Unknown factorization rank $R$     & $256$        & \\
Top-$k_{\text{known}}$             & $16$         & \\
Unknown decomposition              & MLP          & \\
$\alpha_{\text{known}}(s)$         & $1.0 \to 0.5$, cosine & warmup $15\%$, decay until $50\%$ \\
$\alpha_{\text{unknown}}(s)$       & $1.0 \to 0.5$, linear & warmup $25\%$, decay until $100\%$ \\
$p_{\text{cfg}}$                   & $0.1$        & known head dropout \\
$p_\varepsilon$                    & $0.1$        & residual dropout \\
$\lambda_{\text{concept}}$         & $1.0$        & \\
$\lambda_{\text{rec}}$             & $1.0$        & \\
$\lambda_{\text{indep}}$           & $1.0$        & \\
Gradient flow to unknown head      & detached     & \\
\midrule
\multicolumn{3}{l}{\textit{Optimizer}} \\
Optimizer                          & AdamW        & $\beta_1 = 0.9$, $\beta_2 = 0.95$, $\varepsilon = 10^{-8}$ \\
Peak learning rate                 & $4 \times 10^{-4}$ & \\
Schedule                           & WSD          & $100\%$ stable; decay deferred to mid-training \\
Warmup                             & $2000$ steps & \\
Weight decay                       & $0.1$        & excluding embeddings \\
Gradient clipping                  & $1.0$        & \\
\midrule
\multicolumn{3}{l}{\textit{Run}} \\
Batch size                         & $5{,}242{,}880$ tokens/step & $1280$ sequences $\times$ $4096$ tokens \\
Token budget                       & $1.2$T       & \\
Hardware                           & $40$ nodes   & A100s \\
Precision                          & bf16         & \\
\bottomrule
\end{tabular}
\caption{Final pretraining configuration of \steerlingB.}
\label{tab:steerling8b-config}
\end{table}

%% file: sections/appendix/steerling_midtraining.tex
\clearpage
\part{\steerlingB mid-training details}
\label{app:steerling8b-midtraining}

\section{Mid-training recipe}

\subsection{Nemotron: real, synthetic, and mixed}

\label{app:nemotron-real-vs-synth}

Nemotron-CC-HQ contains two kinds of natural-language data: real webtext, and synthetic question-and-answer rephrasings generated from the same documents. To decide which to use for the midtraining mixture, we compare three natural-language sources, each a 10B-token run from the final pretraining checkpoint: real only, synthetic only, and a mixture of the two (\cref{tab:nemotron-real-vs-synth}).

\input{sections/appendix/tables/nemotron_real_vs_synth}

\looseness=-1 All three improve substantially over the base model and track closely on most benchmarks, with near-identical averages. Real tokens give the strongest knowledge and math scores (MMLU and GSM8K), the capabilities mid-training most needs to recover, so we use real tokens for the natural-language portion.

\subsection{Final mid-training \steerlingB configuration}
\label{app:midtraining-config}
Table~\ref{tab:midtraining-config} lists the midtraining configuration alongside the pretraining values for comparison. Hyperparameters not listed are unchanged from pretraining (Appendix~\ref{app:steerling8b-config}).
\input{sections/appendix/tables/midtraining_config}

%% file: sections/appendix/tables/nemotron_real_vs_synth.tex
% sections/steerling/tables/nemotron_real_vs_synth.tex
\begin{table}[htbp!]
  \centering
  \setlength{\tabcolsep}{5pt}
  \begin{tabular}{lccccccc}
    \toprule
    Source & MMLU & GSM8K & ARC-C & HSwag & HEval & WinoG & Avg. \\
    \midrule
    Base model & 0.298 & 0.140 & 0.484 & 0.673 & 0.049 & 0.596 & 0.373 \\
    Real              & \textbf{0.376} & \textbf{0.441} & 0.492 & 0.681 & 0.037 & \textbf{0.616} & \textbf{0.440} \\
    Synthetic         & 0.370 & 0.426 & \textbf{0.503} & 0.677 & \textbf{0.055} & 0.615 & 0.441 \\
    Mixed             & 0.370 & 0.431 & 0.499 & \textbf{0.682} & 0.031 & \textbf{0.618} & 0.439 \\
    \bottomrule
  \end{tabular}
  \caption{Nemotron natural-language source comparison, each a 10B-token run on a fixed math base from the final pretraining checkpoint. HSwag: HellaSwag; HEval: HumanEval; WinoG: WinoGrande.}
  \label{tab:nemotron-real-vs-synth}
\end{table}

%% file: sections/appendix/tables/midtraining_config.tex
% sections/steerling/tables/midtraining_config.tex
\begin{table}[htbp!]
\centering
\small
\setlength{\tabcolsep}{8pt}
\begin{tabular}{@{}lcc@{}}
\toprule
& Pretraining & Mid-training \\
\midrule
\multicolumn{3}{@{}l}{Data} \\
\quad Tokens                          & 1.2T & 150B \\
\quad Mixture                         & Nemotron-CC-HQ & reasoning + Code \\
\quad Natural-language source         & real + synthetic & real \\
\midrule
\multicolumn{3}{@{}l}{Masking} \\
\quad Schedule                        & moving Gaussian & uniform \\
\quad Range / center                  & $0.2 \to 0.8$ & $\mathcal{U}(0.05, 0.95)$ \\
\midrule
\multicolumn{3}{@{}l}{Concept module} \\
\quad $\alpha_{\text{known}}$ floor   & 0.5 & $0$ (annealed) \\
\quad $\alpha_{\text{unknown}}$ floor & 0.5 & $0$ (fixed to prediction) \\
\quad Independence loss               & single term & two terms \\
\quad Residual dropout $p_\varepsilon$ & 0.1 & 0.3 \\
\quad Known head top-$k$              & dense & 32 \\
\quad Unknown head top-$k$            & dense & 128 \\
\midrule
\multicolumn{3}{@{}l}{Optimization} \\
\quad Learning rate                   & constant & decayed to $0$ \\
\midrule
\multicolumn{3}{@{}l}{Steering phases} \\
\quad Phases                          & --- & 4 (interleaved) \\
\quad Steering data                   & --- & token-level, ${\sim}400$M tokens \\
\quad Injection strength $\gamma$     & --- & 1.0 \\
\quad Injection layers                & --- & $\ell \geq L_{\mathrm{inj}}$ \\
\quad $\lambda_{\text{respond}}, \lambda_{\text{express}}$ & --- & 1.0, 1.0 \\
\bottomrule
\end{tabular}
\caption{Mid-training configuration for \steerlingB, with pretraining values for comparison.}
\label{tab:midtraining-config}
\end{table}